\documentclass{article}
\usepackage[utf8]{inputenc}
\usepackage[margin=1in]{geometry}

\usepackage{microtype}
\usepackage{graphicx}
\usepackage{subfigure}
\usepackage{booktabs} 
\usepackage{natbib}
\setcitestyle{open={(},close={)}}

\usepackage[colorlinks=true,citecolor=blue]{hyperref}

\usepackage{amsmath,amsthm,amsfonts,amssymb,mathdots,array,mathrsfs,bm,bbm,stmaryrd,graphicx,subfigure,xcolor}
\usepackage{algpseudocode,algorithm,algorithmicx}
\usepackage{breakcites}

\usepackage[T1]{fontenc}
\usepackage{enumerate}
\usepackage{inputenc}

\usepackage{subfigure}

\usepackage{booktabs,balance}
\usepackage{rotating}
\usepackage{boldline}
\usepackage{makecell}
\usepackage{multirow}
\usepackage{balance}
\usepackage{epstopdf}

\newtheorem{theorem}{Theorem}

\title{\bf GCWSNet: Generalized Consistent Weighted Sampling for Scalable and Accurate Training of Neural Networks\vspace{0.5in}}

\author{\textbf{Ping Li} \  and \ \textbf{Weijie Zhao} \\\\
Cognitive Computing Lab\\
Baidu Research\\
10900 NE 8th St. Bellevue, WA 98004, USA\\
  \texttt{\{liping11, weijiezhao\}@baidu.com}
  }
\date{\vspace{0.5in}}

\begin{document}
\maketitle

\begin{abstract}
\vspace{0.3in}

\noindent We develop the ``generalized consistent weighted sampling'' (GCWS) for hashing the ``powered-GMM'' (pGMM) kernel (with a tuning parameter $p$). It turns out that GCWS provides a numerically stable scheme for applying power transformation on the original data, regardless of the magnitude of $p$ and the data. The power transformation is often effective for boosting the performance, in many cases considerably so. We feed the hashed data to neural networks on a variety of public classification datasets and name our method ``GCWSNet''. Our extensive experiments show that GCWSNet often improves the classification accuracy. Furthermore, it is evident from the experiments that GCWSNet converges substantially faster. In fact, GCWS often reaches a reasonable accuracy with merely (less than) one epoch of the training process. This property is much desired because many applications, such as advertisement click-through rate (CTR) prediction models, or data streams (i.e., data seen only once), often train just one epoch. Another beneficial side effect is that the computations of the first layer of the neural networks become additions instead of multiplications because the input data become binary (and highly sparse).\\

\noindent Empirical comparisons with (normalized) random Fourier features (NRFF) are provided. We also propose to reduce the model size of GCWSNet by count-sketch and develop the theory for analyzing the impact of using count-sketch  on the accuracy of GCWS. Our analysis shows that an ``8-bit'' strategy should work well in that we can always apply an 8-bit count-sketch hashing on the output of GCWS hashing without hurting the accuracy much. Note that the outputs of count-sketch on top of GCWS hashing are also integers meaning that a lot of multiplications in training neural nets can still be avoided. \\

\noindent There are many other ways to take advantage of GCWS when training deep neural networks. For example,  one can apply GCWS on the outputs of the last layer to boost the accuracy of trained deep neural networks. In our view, GCWS and variants are a gem which has not been exploited much in the machine learning community. We hope this work would generate more interest in pGMM, GCWS, and their variants, for research and industrial practice.  GCWSNet has been implemented with the PaddlePaddle \url{https://www.paddlepaddle.org.cn} deep learning platform.

\end{abstract}

\newpage

\section{Introduction}

There has been a surge of interest in speeding up the training process of large-scale machine learning algorithms. For example, \citet{Proc:Weinberger_ICML2009} applied count-sketch type of randomized algorithms~\citep{Article:Charikar_2004} to approximate large-scale linear classifiers.  \cite{Proc:Li_NIPS11_Learning} applied (b-bit) minwise hashing~\citep{Proc:Broder_WWW97,Proc:Broder_STOC98,Proc:Li_Church_EMNLP05,Proc:Li_Konig_WWW10} to approximate the resemblance (Jaccard) kernel for high-dimensional binary (0/1) data, using highly efficient linear classifiers. In this paper, we first propose the ``pGMM'' kernel and then present the idea of ``generalized consistent weighted sampling'' (GCWS) to approximate the pGMM kernel, in the context of training neural~networks. 

The so-called ``generalized min-max'' (GMM) kernel was proposed in~\cite{Proc:Li_KDD17}. For defining the GMM kernel, the first step is a simple transformation on the original data. Consider, for example, the original data vector $u_i$, $i=1$ to $D$. The following transformation, depending on whether an entry $u_i$ is positive or negative,
\begin{align}\label{eqn:transform}
 \left\{\begin{array}{cc}
\hspace{-.11in}\tilde{u}_{2i-1} = u_i,\hspace{0.1in} \tilde{u}_{2i} = 0&\text{if } \ u_i >0\\
\tilde{u}_{2i-1} = 0,\hspace{0.1in} \tilde{u}_{2i} =  -u_i &\text{if } \ u_i \leq 0
\end{array}\right.
\end{align}
converts general data types to non-negative data only. For example, when $D=2$ and $u = [-3\ \ 17]$, the transformed data vector becomes $\tilde{u} = [0\ \ 3\ \ 17\ \ 0]$. The GMM kernel is then defined as follows:
\begin{align}\label{eqn:GMM}
{\textit{GMM}}(u,v) = \frac{\sum_{i=1}^{2D}\min\{\tilde{u}_i,\tilde{v}_i\}}{\sum_{i=1}^{2D} \max\{\tilde{u}_i,\tilde{v}_i\}}.
\end{align}
\cite{Proc:Li_Zhang_WWW17} developed some (rather limited) theories for the GMM kernel. For example, under certain distributional assumptions on the data, the GMM similarity converges to the following interesting limit: 
\begin{align}
\textit{GMM}(u,v) \rightarrow \frac{1-\sqrt{(1-\rho)/2}}{1+\sqrt{(1-\rho)/2}},
\end{align}
where $\rho$ is the true correlation between $u$ and $v$. More precisely, $\rho$ is the correlation parameter in the bi-variate distribution from which $u$ and $v$ are generated. 

Note that, the GMM kernel defined in~\eqref{eqn:GMM} still has no tuning parameter, unlike (e.g.,) the popular Gaussian (RBF) kernel. An extremely simple strategy to introduce tuning parameters is our proposed ``pGMM'' kernel: 
\begin{align}\label{eqn:pGMM}
&{\textit{pGMM}}(u,v;p) =  \frac{\sum_{i=1}^{2D}\left(\min\{\tilde{u}_i,\tilde{v}_i\}\right)^p}{\sum_{i=1}^{2D}\left( \max\{\tilde{u}_i,\tilde{v}_i\}\right)^p},
\end{align}
where $p\in\mathbb{R}$ is a tuning parameter. 
Immediately, readers would notice that this is mathematically equivalent to first applying a power transformation on the data ($\tilde{u}, \tilde{v}$) before computing the GMM kernel. Readers will soon see that, combined with GCWS hashing, our proposal of pGMM provides a convenient and also numerically highly stable scheme for applying the power transformation, regardless of the magnitude of $p$.

\subsection{Kernel SVM Experiments}

While the main focus of this paper is on training neural networks, we nevertheless provide a set of experimental studies on kernel SVMs for evaluating the pGMM kernel as formulated in~\eqref{eqn:pGMM}, in comparison with the linear kernel and the (best-tuned) RBF (Gaussian) kernel. The results are reported in Table~\ref{tab:SVM}. One reason for presenting this set of experiments is for ensuring reproducibility, as we (and probably many practitioners too) have noticed that tools like LIBSVM/LIBLINEAR are easy to use with essentially deterministic predictions.

\vspace{0.1in}

For example, for the SEMG dataset, the accuracies for the linear kernel and the (best-tuned) RBF kernel are very low, which are $19.3\%$ and $29.0\%$, respectively. Perhaps surprisingly, the GMM kernel reaches $54\%$ (with no tuning parameter) and the pGMM kernel achieves an accuracy of $56.1\%$ when $p=2$. For the M-Noise1 dataset, the best power parameter is $p = 80$ with an accuracy of $85.2\%$. Readers can check out the deep learning experiments reported in~\citet{Proc:Larochelle_ICML07} and confirm that they are inferior to the pGMM kernel, for the M-Noise1 dataset. 

\newpage

\begin{table}[t]
\caption{\textbf{Public classification datasets and $l_2$-regularized kernel SVM results}. We report the test classification accuracies for the linear kernel, the best-tuned RBF kernel, the original (tuning-free) GMM kernel, and the best-tuned pGMM kernel, at their individually-best SVM regularization $C$ values. The results on linear kernels were conducted using LIBLINEAR~\citep{Article:Fan_JMLR08} and the kernel SVM experiments were conducted using LIBSVM and pre-computed kernel matrices (hence it is difficult to train on datasets of more than 30,000 data points). All datasets are from the UCI repository except for M-Noise1 and M-Image, which were used by~\citet{Proc:Larochelle_ICML07,Proc:ABC_UAI10} for testing deep learning algorithms and tree methods.
}
\begin{center}{
{\begin{tabular}{l r r r r c c c l  }
\hline \hline
Dataset     &\# train  &\# test  &\# dim & \# class & linear  &RBF  &GMM  &pGMM (p)   \\
\hline
SEMG & 1800 & 1,800 & 2,500 & 6 & 19.3 & 29.0 & 54.0 & 56.1 (2) \\
DailySports & 4,560 & 4,560 & 5,625 & 19 & 77.7 & 97.6 & 99.6 & 99.6 (0.6) \\
M-Noise1 & 10,000 & 4,000 & 784 & 10 & 60.3 & 66.8 & 71.4 & 85.2 (80) \\
M-Image & 12,000 & 50,000 & 784 & 10 & 70.7 & 77.8 & 80.9 & 89.5 (50) \\
PAMAP101 & 188,209 & 188,208 & 51 & 20 & 75.3 & --- &  --- & ---\\
Covtype & 290,506 & 290,506 & 54 & 7 & 71.5 & --- &  --- & ---\\
\hline\hline
\end{tabular}}
}
\end{center}\label{tab:SVM}
\end{table}

It is not at all our intention to debate which type of classifiers work the best. We believe the performance highly depends on the datasets. Nevertheless, we hope that it is clear from Table~\ref{tab:SVM} that the pGMM kernel is able to achieve comparable or better accuracy than the RBF kernel on a wide range of datasets. Note that, there are more than 10 datasets used in~\citet{Proc:Larochelle_ICML07,Proc:ABC_UAI10}, such as M-Noiose2, ..., M-Noise6, etc which exhibit similar behaviors as M-Noise1. To avoid boring the readers, we therefore only present the experimental results for M-Noise1 and M-Image. 

\subsection{Linearizing pGMM Kernel via Generalized Consistent Weighted Sampling}

When using LIBSVM pre-computed kernel functionality, we have found it is already rather difficult if the number of training examples exceeds merely 30,000. It has been a well-known challenging task to scale up kernel learning for large datasets~\citep{Book:Bottou_07}. This has motivated many studies for developing hashing methods to approximately linearize nonlinear kernels. In this paper, we propose using the ``generalized consistent weighted sampling'' (GCWS) to approximate the pGMM kernel, in the context of training neural networks for any tuning parameter $p$, as illustrated in Algorithm~\ref{alg:GCWS}.

\begin{algorithm}{

\textbf{Input:} Data vector $u_i$  ($i=1$ to $D$)

\vspace{0.08in}

Generate vector $\tilde{u}$ in $2D$-dim by (\ref{eqn:transform}).

\vspace{0.08in}

For $i$ from 1 to $2D$

\hspace{0.25in}$r_i\sim \textit{Gamma}(2, 1)$, $c_i\sim \textit{Gamma}(2, 1)$,  $\beta_i\sim \textit{Uniform}(0, 1)$

\hspace{0.2in} $t_i\leftarrow \lfloor p\frac{\log \tilde{u}_i }{r_i}+\beta_i\rfloor$, $a_i\leftarrow \log(c_i)- r_i(t_i+1-\beta_i)$

End For

\vspace{0.08in}

\textbf{Output:} $i^* \leftarrow \textit{arg}\min_i \ a_i$,\hspace{0.3in}  $t^* \leftarrow t_{i^*}$
}\caption{Generalized consistent weighted sampling (GCWS) for hashing the pGMM kernel.}
\label{alg:GCWS}
\end{algorithm}

Given another data vector $v$, we feed it to GCWS using the same set of random numbers: $r_i$, $c_i$, $\beta_i$. To differentiate the hash samples, we name them, respectively,
$(i^*_{u}, t^*_{u})$ and $(i^*_{v}, t^*_{v})$. We first present the basic probability result as the following theorem. 
\begin{theorem}\label{thm:pGMM}
\begin{align}\label{eqn:GCWS_Prob}
    P[(i_u^*,t_u^*)=(i_v^*,t_v^*)]=\textit{pGMM}(u,v).
\end{align}
\end{theorem}

\vspace{0.1in}

The proof of Theorem~\ref{thm:pGMM} directly follows from the basic theory of consistent weighted sampling (CWS)~\citep{Report:Manasse_CWS10,Proc:Ioffe_ICDM10,Proc:Li_WWW21}. Although the original CWS algorithm is designed (and proved) only for non-negative data, we can see that the two transformations, i.e., converting general data types to non-negative data by (\ref{eqn:transform}) and applying the power on the converted data as in (\ref{eqn:pGMM}), are only pre-processing steps and the same proof for CWS will go through for GCWS. \\

Note that, in Algorithm~\ref{alg:GCWS}, while the value of the output $i^*$ is upper bounded by $2D$, the other integer output $t^*$ is actually unbounded. This makes it less convenient for the implementation. The next Theorem provides the basis for the (b-bit) implementation of GCWS. The proof is also straightforward. 
\begin{theorem}\label{thm:pGMM_b}
Assume that we can map $(i_u^*, t_u^*)$ uniformly to a space of $b$ bits denoted by $(i_u^*, t_u^*)_b$. Similarly, we have $(i_v^*, t_v^*)_b$. Then 
\begin{align}\label{eqn:pGMM_b} 
    P[(i_u^*,t_u^*)_b=(i_v^*,t_v^*)_b]=\textit{pGMM}(u,v) + \frac{1}{2^b}\left\{1-\textit{pGMM}(u,v)\right\}.
\end{align}
\end{theorem}

\subsection{Practical Implementation of GCWS}

For each input data vector, we need to generate $k$ hash values, for example, $(i^*_{u,j}, t^*_{u,j})$, $j = 1$ to $k$, for data vector $u$. The mapping of $(i^*_{u,j}, t^*_{u,j})$ uniformly to $(i^*_{u,j}, t^*_{u,j})_b$ is actually not a trivial task, in part because $t^*$ is unbounded. Based on the intensive experimental results in~\cite{Proc:Li_KDD15} (for the original CWS algorithm), we will take advantage of following approximation: 
\begin{align}\label{eqn:GCWS_Prob_approx} 
    P[i_u^*=i_v^*]\approx P[(i_u^*,t_u^*)=(i_v^*,t_v^*)] =\textit{pGMM}(u,v).
\end{align}
and will only keep the lowest $b$ bits of $i^*$ (unless we specify otherwise). 

Suppose that, for data vector $u$, we have obtained $k=3$ hash values which after we keep only the lowest $b =2$ bits, become $(3,0,1)$. We then concatenate their ``one-hot'' representations to obtain $[1\ 0\ 0\ 0 \ \ 0\ 0\ 0\ 1 \ \ 0\ 0\ 1\ 0]$, which is fed to subsequent algorithms for classification, regression, or clustering. In other words, with $k$ hashes and $b$ bits, for each input data vector we obtain a binary vector of length $2^b \times k$ with exactly $k$ 1's. 

\subsection{The History of Consistent Weighted Sampling}

For binary (0/1) data, the pGMM kernel becomes the resemblance (Jaccard) similarity and GCWS is essentially equivalent to the celebrated minwise hashing algorithm~\citep{Proc:Broder_WWW97,Proc:Broder_STOC98,Proc:Li_Church_EMNLP05,Proc:Li_Konig_WWW10}, with numerous practical applications~\citep{Proc:Fetterly_WWW03,Proc:Nitin_WSDM08,Proc:Buehrer_WSDM08,Article:Urvoy08,Article:Dourisboure09,Article:Forman09,Proc:Pandey_WWW09,Proc:Cherkasova_KDD09,Proc:Chierichetti_KDD09,Proc:Gollapudi_WWW09,Proc:Najork_WSDM09,Proc:Bendersky_WSDM09,Proc:Li_NIPS11_Learning,Proc:Shrivastava_ECML12,Proc:Schubert_KDD14,Proc:Fu_FAST15,Proc:Pewny_SP15,Proc:Manzoor_KDD16,Proc:Raff_SIGMOD17,Proc:Tymoshenko_EMNLP18,Proc:Zhu_SIGMOD19,Proc:Lei_SIGMOD20,Proc:Thomas_ECCV20}. Note that minwise hashing can also be used to estimate 3-way and multi-way resemblances~\citep{Proc:Li_Konig_NIPS10}, not limited to only pairwise similarity. 

\vspace{0.1in}

For general non-binary and non-negative data, the development of consistent weighted sampling algorithm in its current form was due to~\cite{Report:Manasse_CWS10,Proc:Ioffe_ICDM10}, as well as the earlier versions such as~\cite{Proc:Gollapudi_CIKM06}. \cite{Proc:Li_KDD15} made the observation about the ``0-bit'' CWS, i.e.,~\eqref{eqn:GCWS_Prob_approx} and applied it to approximate large-scale kernel machines using linear algorithms. \cite{Proc:Li_KDD17} generalized CWS to general data types which can have negative entries and demonstrated the considerable advantage of CWS over random Fourier features~\citep{Proc:Rahimi_NIPS07,Proc:Li_AISTATS21}. From the computational perspective, CWS is efficient in sparse data. For dense data, algorithms based on rejection sampling~\citep{Proc:Kleinberg_FOCS99,Proc:Charikar_STOC02,Proc:Shrivastava_NIPS16,Proc:Li_Li_AAAAI21} can be much more efficient than CWS. For relatively high-dimensional datasets, the method of ``bin-wise CWS'' (BCWS)~\citep{Proc:Li_NIPS19_BCWS} would always be recommended. Simply speaking, the basic idea of BCWS is to divide the data matrix into bins and then apply CWS (for sparse data) or rejection sampling (for dense data) in each bin. Finally, we should add that the mathematical explanation of the ``0-bit'' approximation~\eqref{eqn:GCWS_Prob_approx}, empirically observed by~\cite{Proc:Li_KDD15}, remains an open problem. The recent work by~\cite{Proc:Li_WWW21} developed a related algorithm based on extremal processes and mathematically proved the effectiveness of the ``0-bit'' approximation for that algorithm, which is closely related to but not CWS.

\newpage

\subsection{Our Contributions}

In this paper, we develop GCWS (generalized CWS) for hashing the pGMM kernel and then apply GCWS for efficiently training neural networks. We name our procedure GCWSNet and show that GCWSNet can often achieve more accurate results compared with the standard neural networks. We also have another beneficial observation, that the convergent speed of GCWSNet can be substantially faster than the standard neural networks. In fact, we notice that GCWS typically achieves a reasonable accuracy in (less than) one epoch. This characteristic of GCWSNet might be a huge advantage. For example, in many applications such as data streams, one of the 10 challenges in data mining~\citep{yang200610}, each sample vector is only seen once and hence only one epoch is allowed in the training process. In commercial search engines, when training click-through rate (CTR) deep learning models, typically only one epoch is used~\citep{Proc:Fan_KDD19,Proc:Zhao_CIKM19,Proc:Fei_SIGIR21,Proc:Xu_SIGMOD21} because the training data size is  on the petabyte scale and new observations (i.e., user click information) keep arriving at a very fast rate. Additionally, there is another side benefit of GCWSNet, because the computations at the first layer become additions instead of multiplications.

\section{GCWSNet}

For each input data vector, we first generate $k$ hash values $(i^*, t^*)$ using GCWS as in Algorithm~\ref{alg:GCWS}. We adopt the ``0-bit'' approximation~\eqref{eqn:GCWS_Prob_approx} and keep only the lowest $b$ bits of $i^*$. For example, suppose $k=3$,  $b =2$ bits, and the hash values become $(3,0,1)$. We concatenate their ``one-hot'' representations to obtain $[1\ 0\ 0\ 0 \ \ 0\ 0\ 0\ 1 \ \ 0\ 0\ 1\ 0]$, which is the input data fed to neural networks. In short, with $k$ hashes and $b$ bits, for each input data vector we obtain a binary vector of length $2^b \times k$ with exactly $k$ 1's.

The experiments were conducted on the  PaddlePaddle  \url{https://www.paddlepaddle.org.cn} deep learning platform. We adopted the Adam optimizer~\citep{Proc:Kingma_ICLR15} and set the initial learning rate to be $0.001$. We use ``ReLU'' for the activation function; other learning rates gave either similar or worse results in our experiments. The batch size was set to be 32 for all datasets except Covtype for which we use 128 as the batch size.  We report experiments for neural nets with no hidden units (denoted by ``$L=1$''), one layer of $H$ hidden units (denoted by ``$L = 2$'') and two layers (denoted by ``$L=3$'') of hidden units ($H$ units in the first hidden layer and $H/2$ units in the second hidden layer). We have also experimented with more layers.  

\vspace{0.1in}

\begin{figure}[h]
\begin{center}
\mbox{
\includegraphics[width=2.2in]{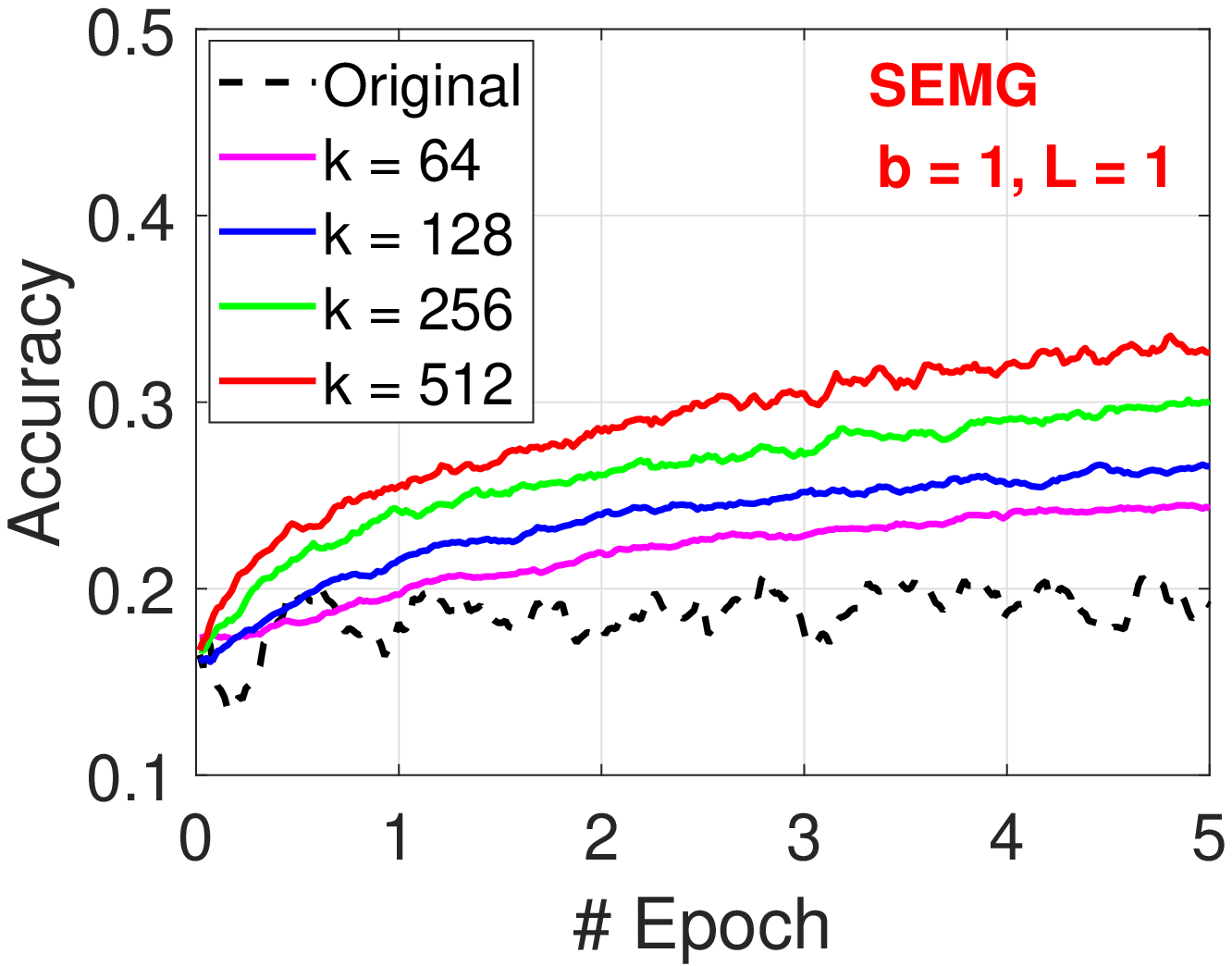}
\includegraphics[width=2.2in]{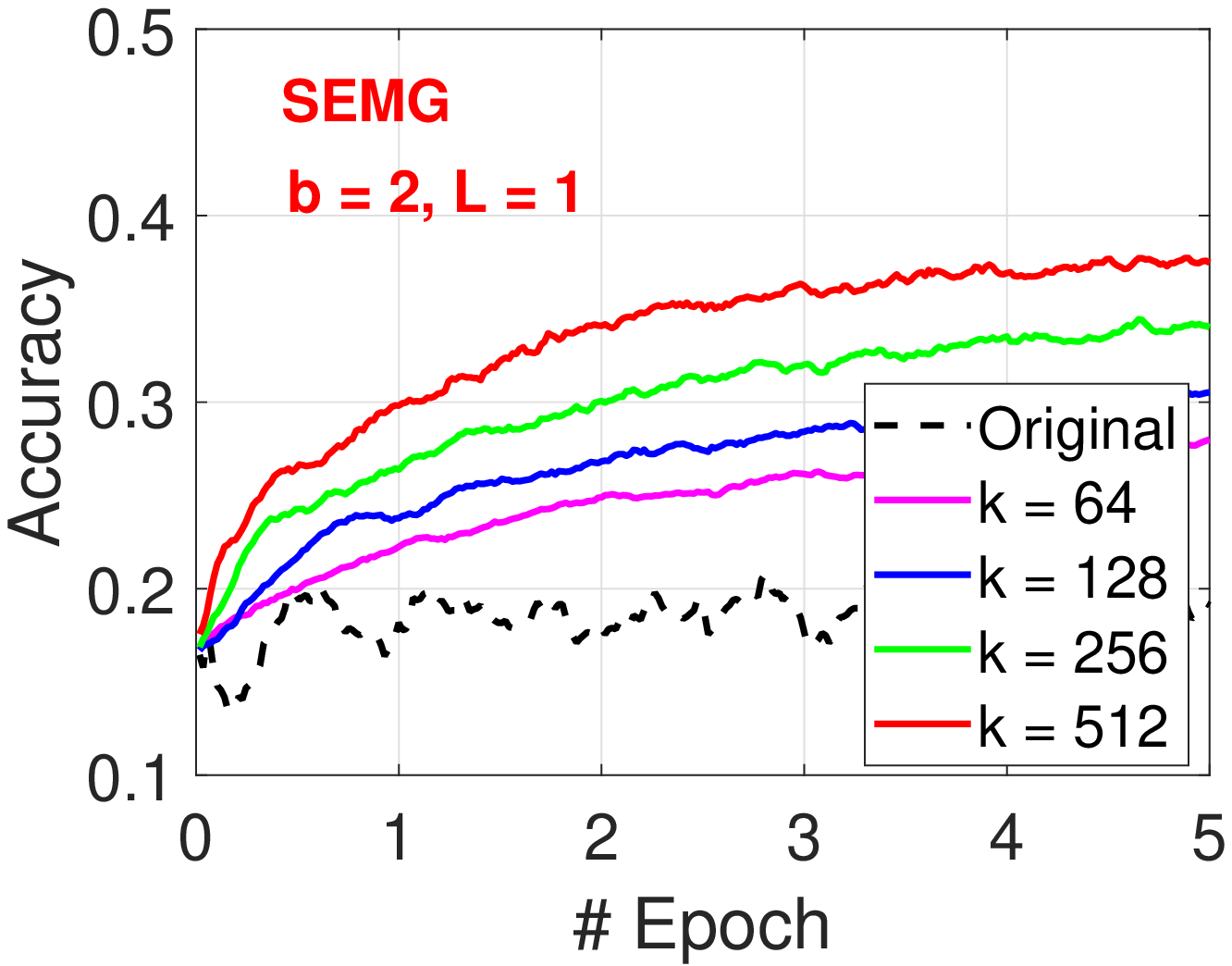}
\includegraphics[width=2.2in]{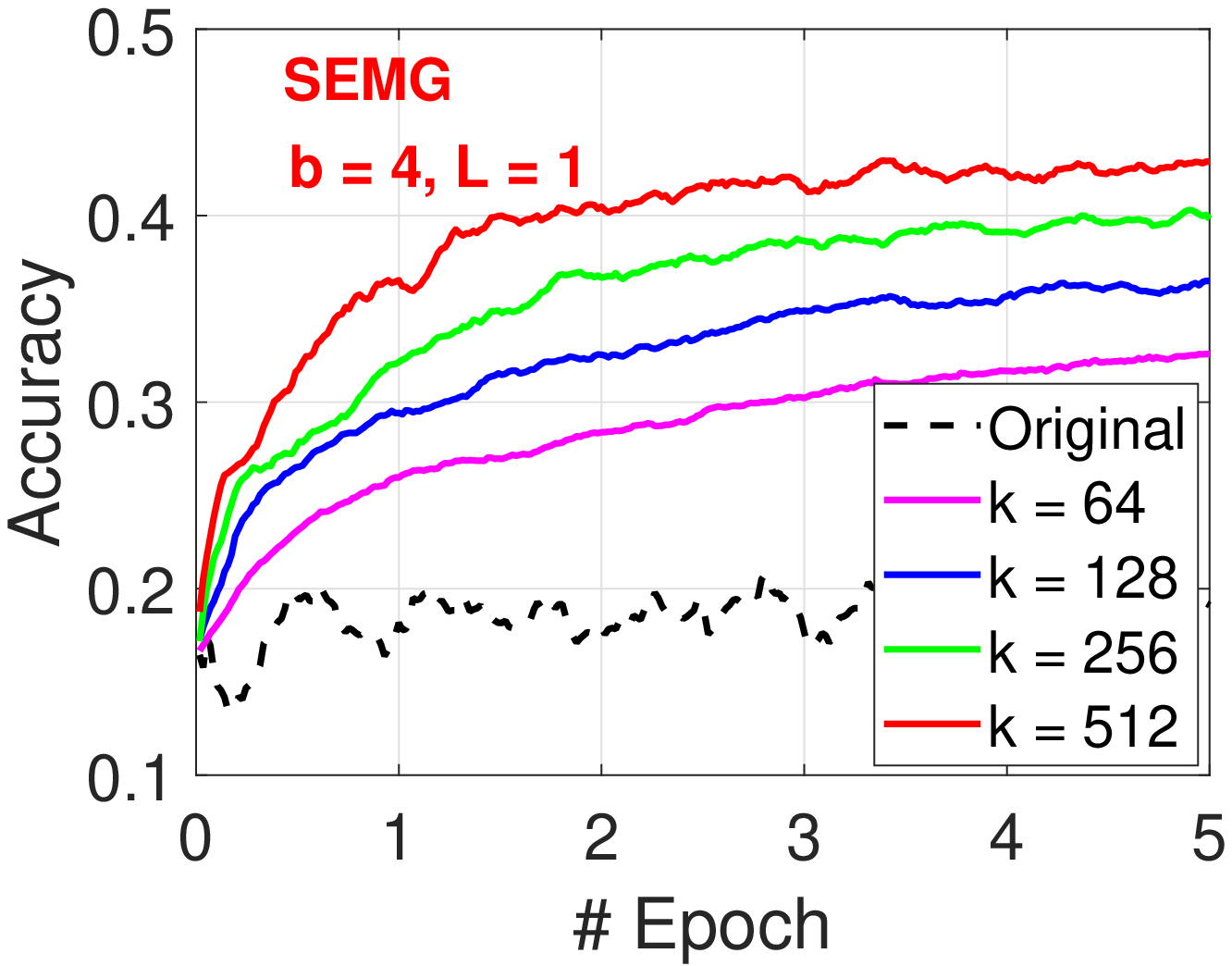}
}

\mbox{
\includegraphics[width=2.2in]{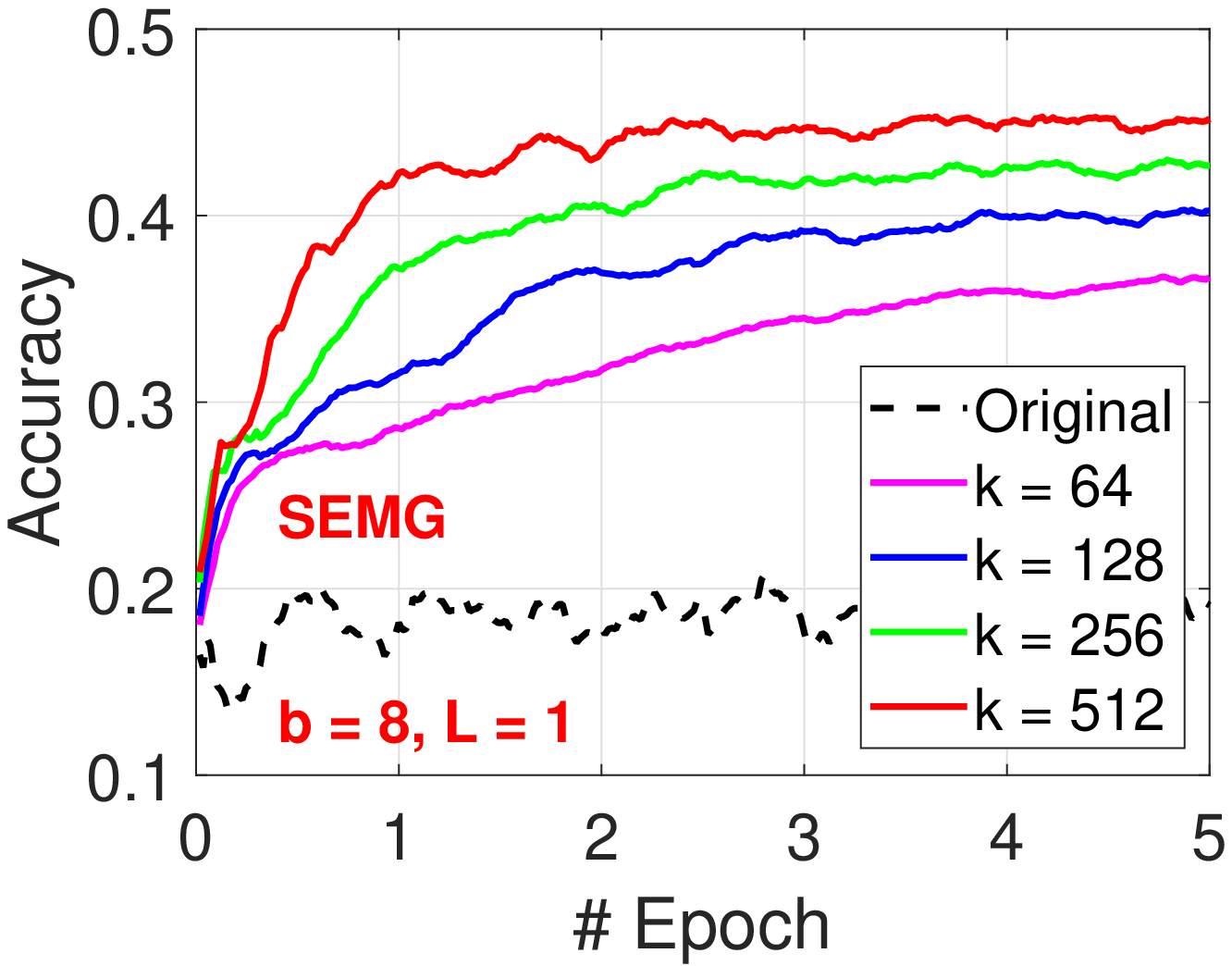}
\includegraphics[width=2.2in]{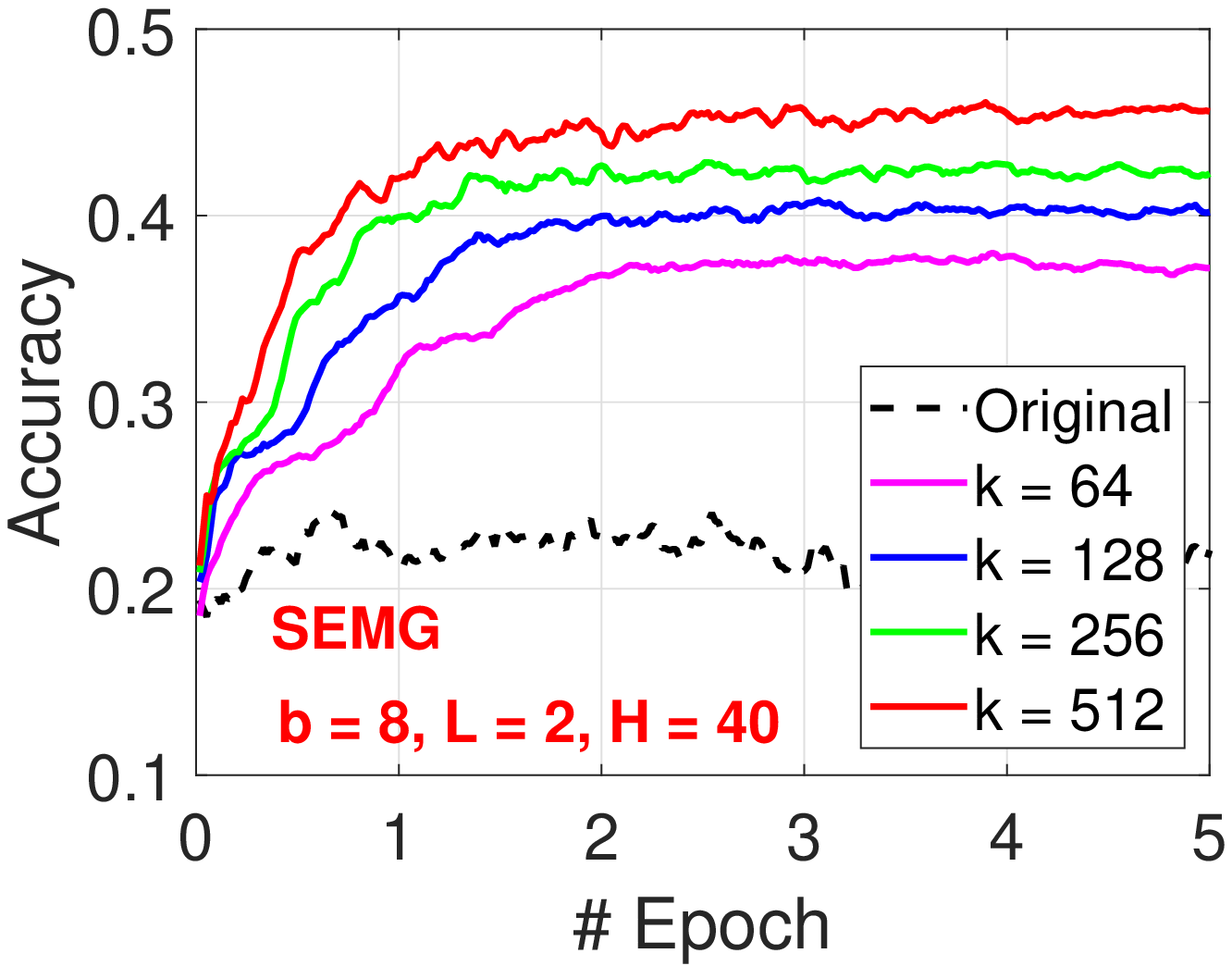}
\includegraphics[width=2.2in]{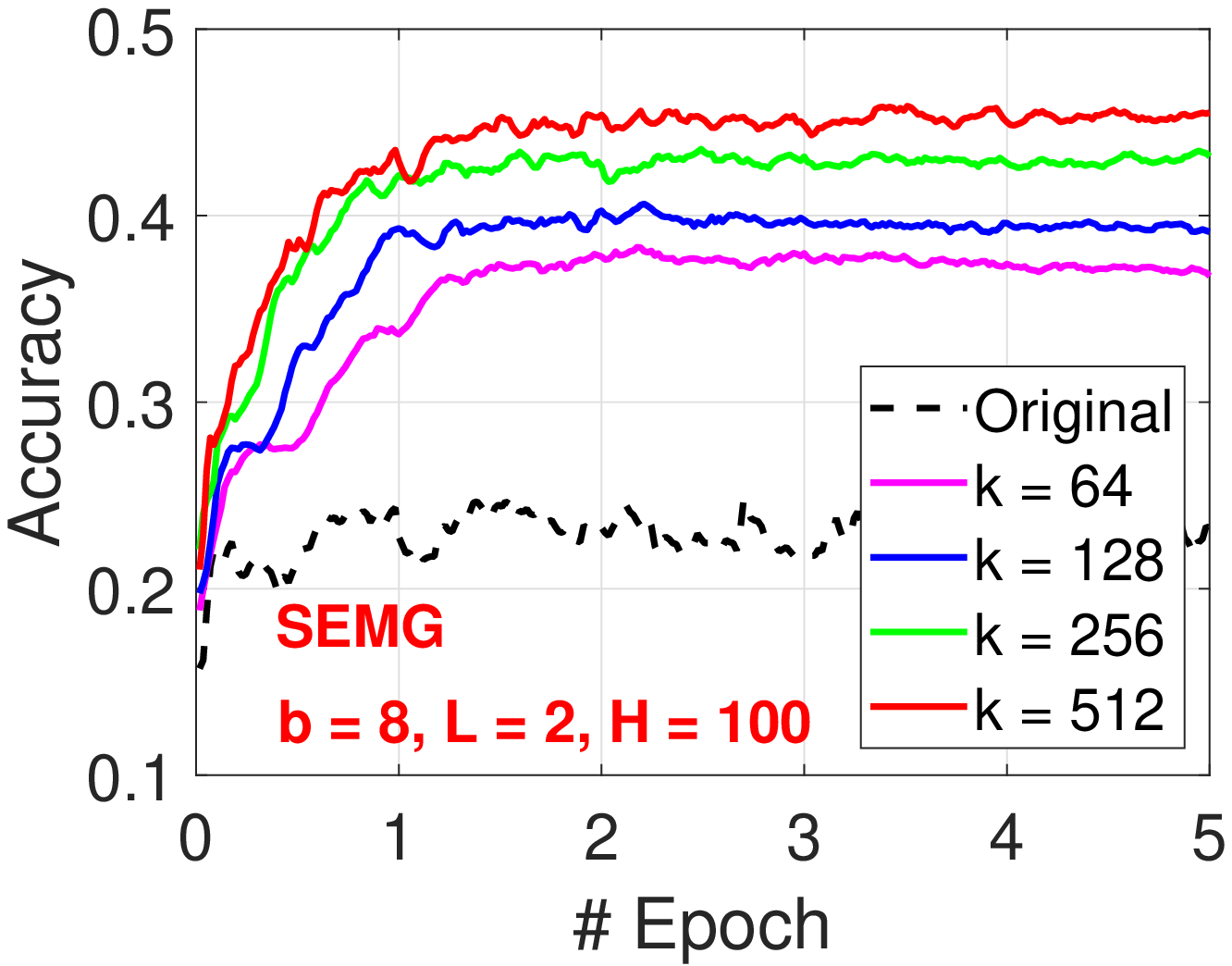}
}
\mbox{
\includegraphics[width=2.2in]{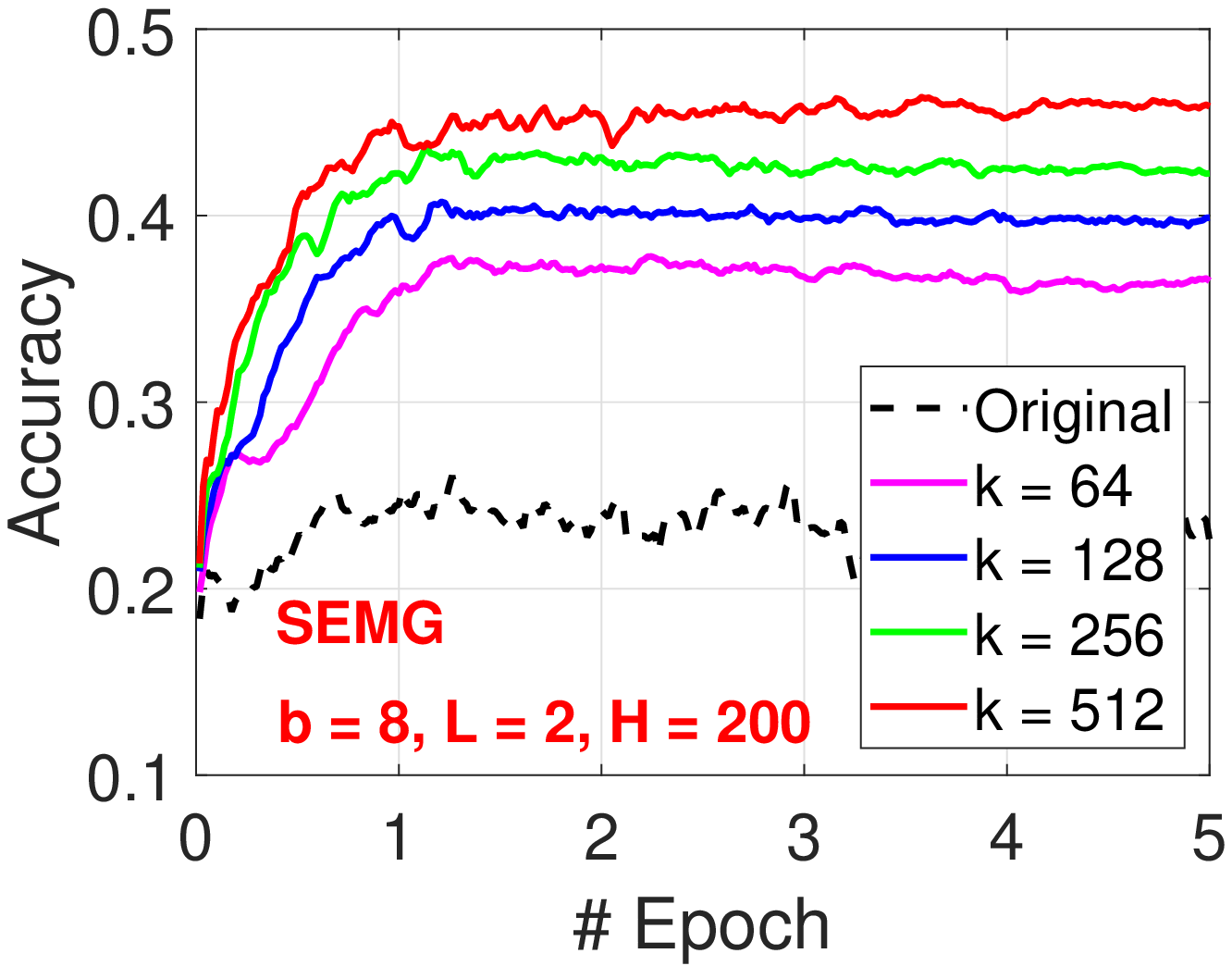}
\includegraphics[width=2.2in]{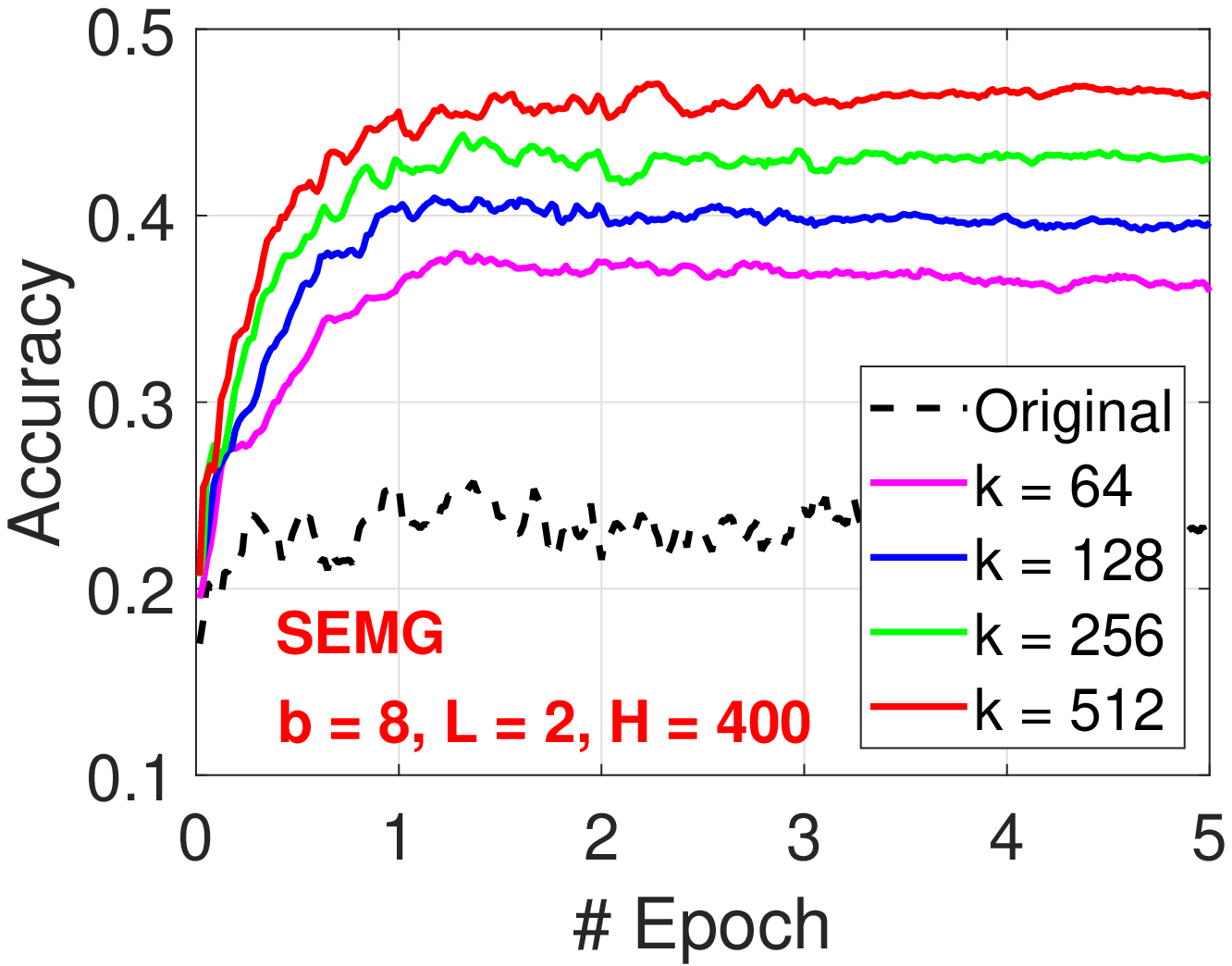}
\includegraphics[width=2.2in]{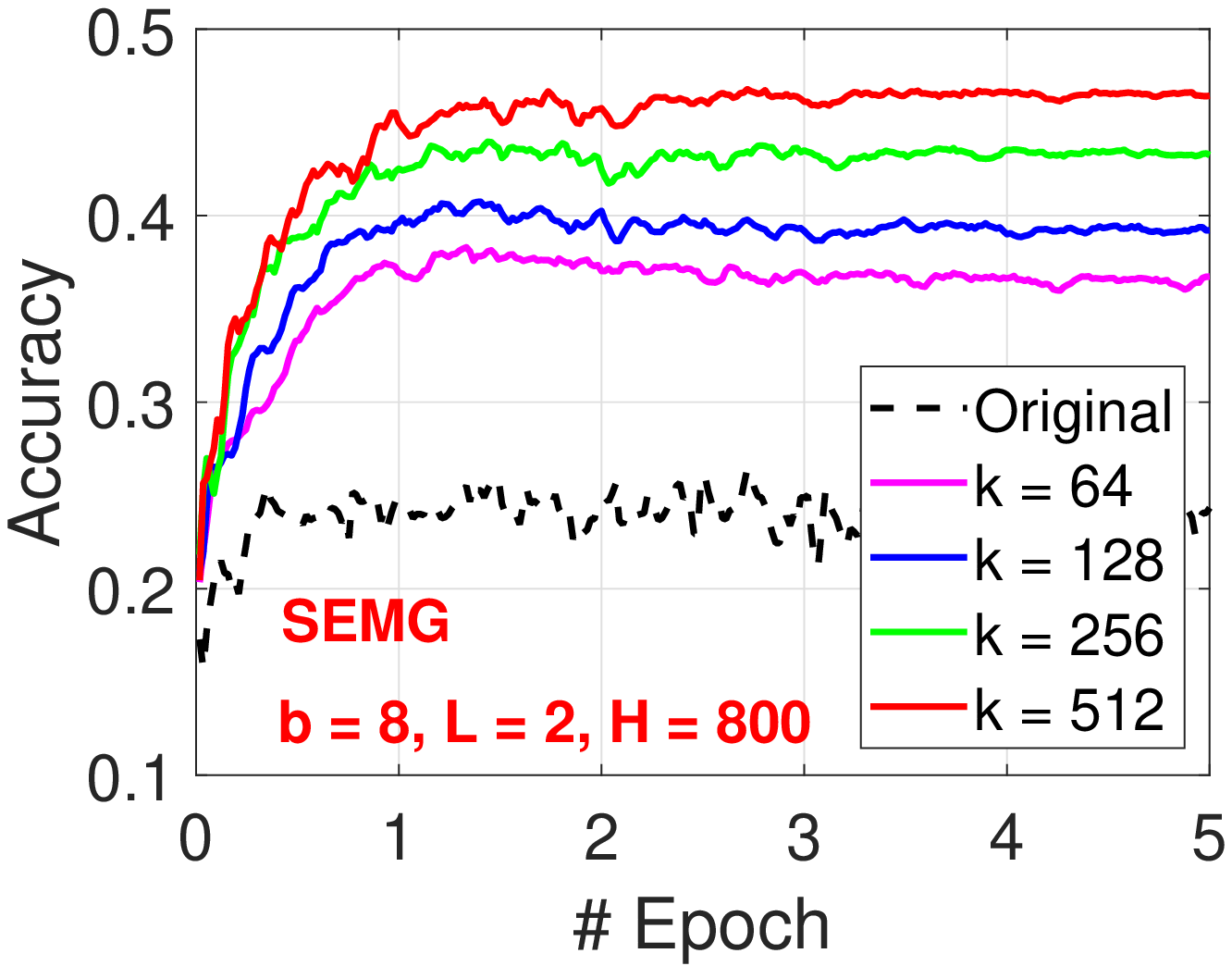}
}

\mbox{
\includegraphics[width=2.2in]{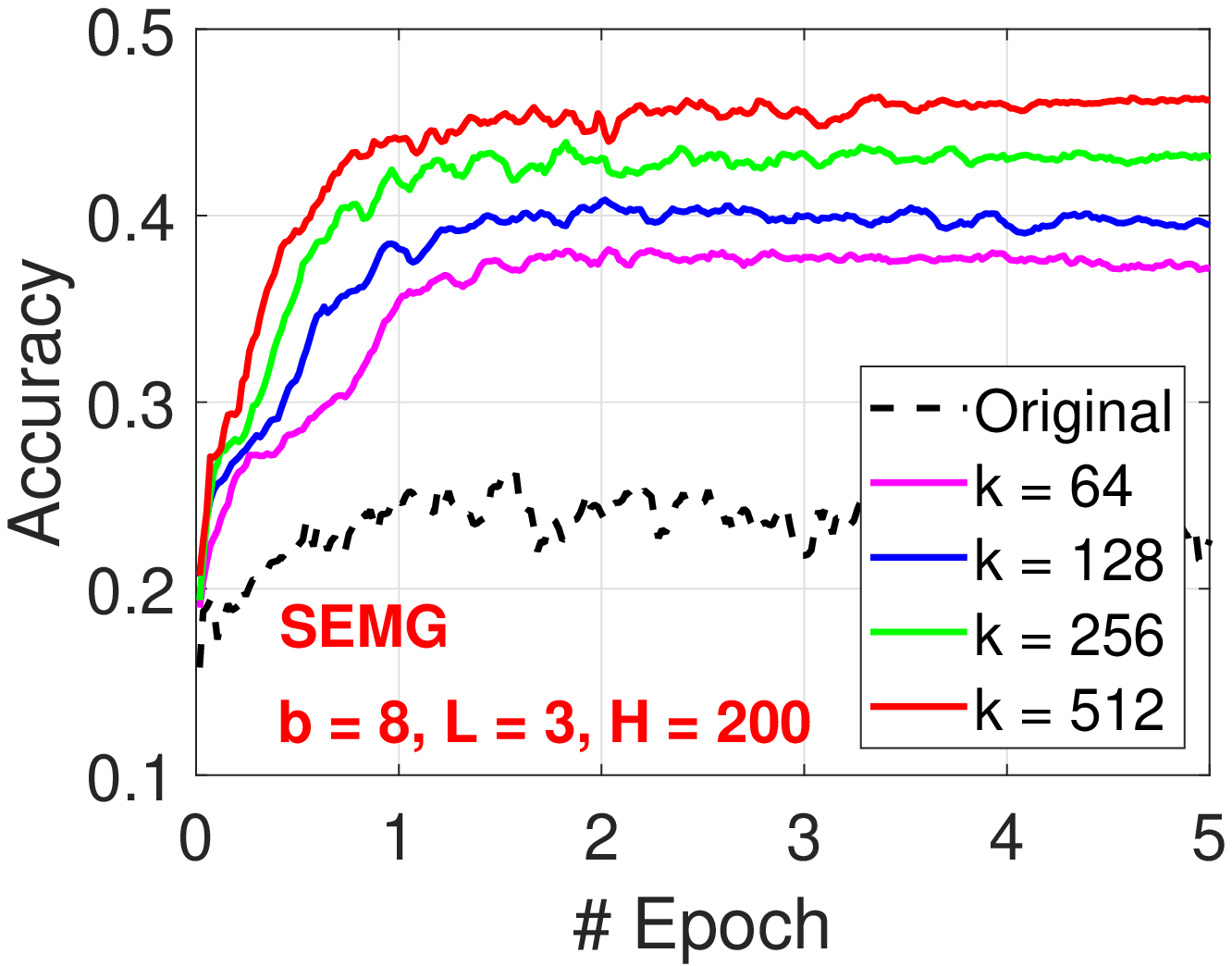}
\includegraphics[width=2.2in]{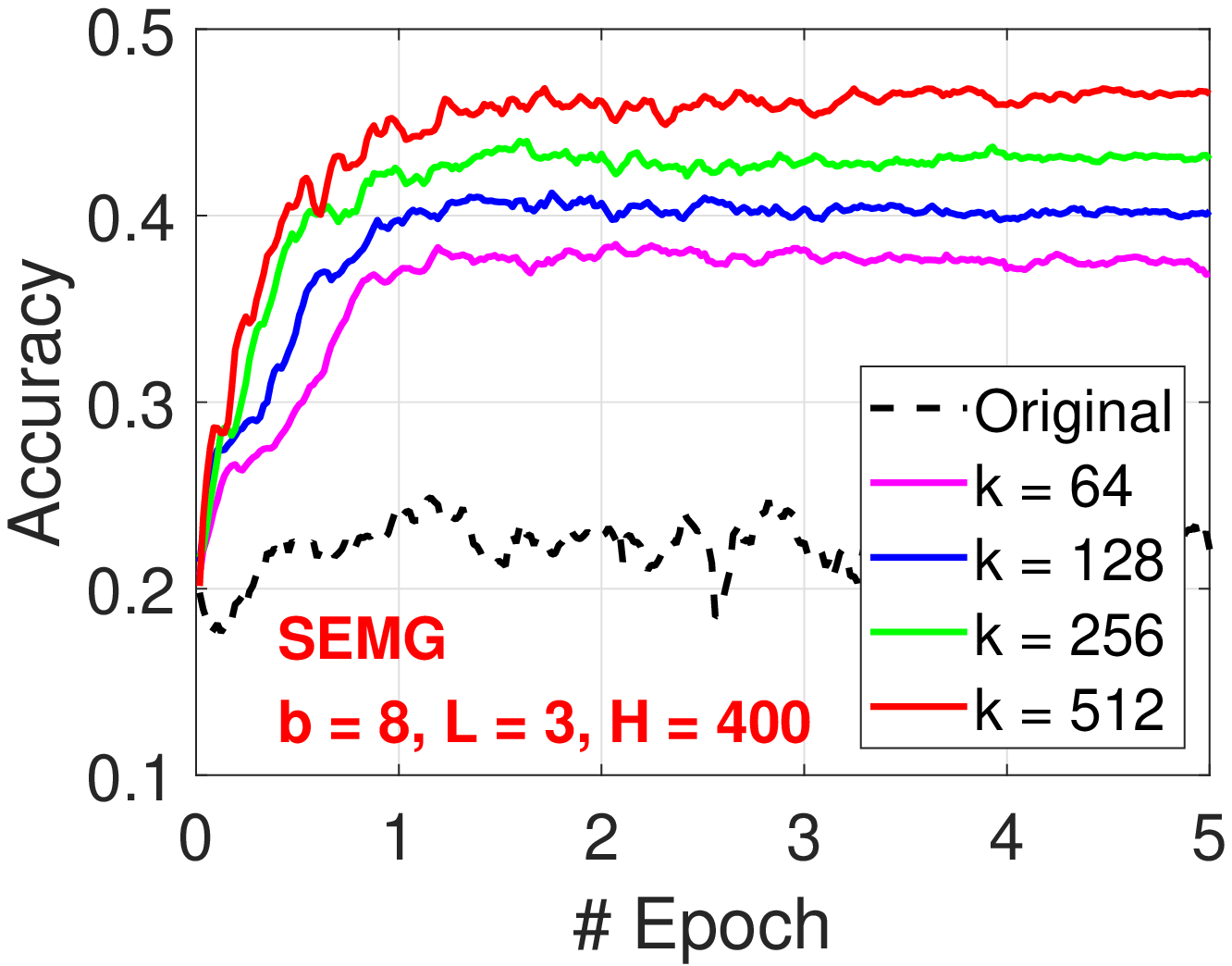}
\includegraphics[width=2.2in]{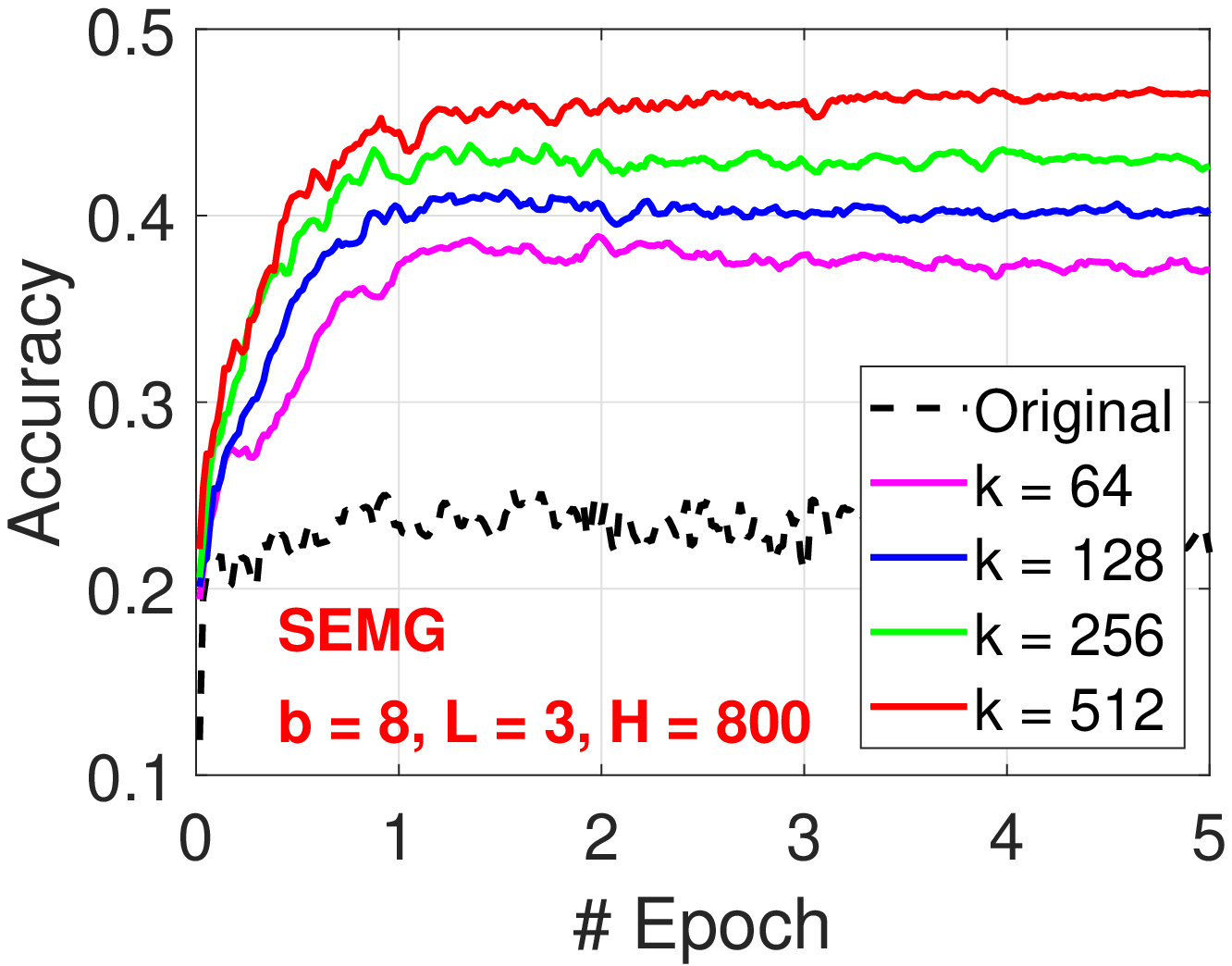}
}

\end{center}

\caption{GCWSNet with $p=1$, for $b\in\{1,2,4,8\}$ and $k\in\{64,128,256,512\}$, on the SEMG dataset (see Table~\ref{tab:SVM} for the details). We report the test accuracy for 5 epochs. Here ``one epoch'' means all the training data points have been used exactly once.   ``$L = 1$'' means no hidden units, i.e., just logistic regression. ``$L=2$'' means one hidden layer of $H$ units. ``$L=3$'' means two hidden layers with $H$ units in the first hidden layer and $H/2$ units in the second hidden layer. The results for the original data are reported as dashed (black) curves, which are substantially worse than the results of GCWSNet, as one would expect from Table~\ref{tab:SVM}. One important observation is that GCWSNet converges much faster, (for this dataset) reaching  the close-to-the-best accuracy after finishing merely one epoch of training.}\label{fig:SEMG2_p1}
\end{figure}

Figure~\ref{fig:SEMG2_p1} presents the experimental results on the SEMG dataset using GCWS (with $p=1$) along with the results  on the original data (dashed black curves). This is a small dataset with only 1,800 training samples and 1,800 testing samples. It is relatively high-dimensional with 2,500 features. Recall that in Table~\ref{tab:SVM}, on this dataset, we obtain an accuracy of $19.3\%$ using linear classifier and $29\%$ using the (best-tuned) RBF kernel. In Figure~\ref{fig:SEMG2_p1}, the dashed (black) curves for $L=1$ (i.e., neural nets with no hidden units) are consistent with the results in Table~\ref{tab:SVM}, i.e., an accuracy of $<20\%$ for using linear classifier. Using one layer of $H$ hidden units (i.e., $L=2$) improves the accuracy quite noticeably (to about $25\%$). However, using two layers of hidden units (i.e., $L=3$) further improves the accuracy only very little (if any). 

\vspace{0.1in}

For the SEMG dataset, GCWS (even with $p=1$) seems to be  highly effective. We report the experiments for $b\in\{1,2,4,8\}$ and $K\in\{64,128,256,512\}$. The cost of training SGD is largely determined by the number of nonzero entries. Note that even with $k=512$, the number of nonzero entries in each input training data vector is only $k=512$ while the original SEMG dataset has 2,500 nonzero entries in each data vector. 

\vspace{0.1in}

We recommend using a large $b$ value as long as the model is still small enough to be stored. With $b=8$ bits, we notice that, regardless of $H$ and $K$ (and the number of hidden layers), GCWSNet converges  fast and typically  reaches a reasonable accuracy at the end of one epoch. Here ``one epoch'' means all the training samples have been used exactly once. This property can be highly beneficial in practice. In data stream applications, one of the 10 challenges in data mining~\citep{yang200610}, each training example is seen only once and the process of model training is by nature one-epoch.

\newpage\clearpage

There are other important applications which would also prefer one-epoch training. For example, in commercial search engines such as \url{www.baidu.com} and \url{www.google.com}, a crucial component is the click-through rate (CTR) prediction task, which routinely uses very large distributed deep learning models. The training data size is  on the petabyte scale and new observations (i.e., user clicks) keep arriving at an extremely fast rate. As far as we know from our own experience, the training of CTR model is typically just one epoch~\citep{Proc:Fan_KDD19,Proc:Zhao_CIKM19,Proc:Fei_SIGIR21,Proc:Xu_SIGMOD21}.

\vspace{0.3in}

\begin{figure}[h!]
\begin{center}

\mbox{
\includegraphics[width=2.2in]{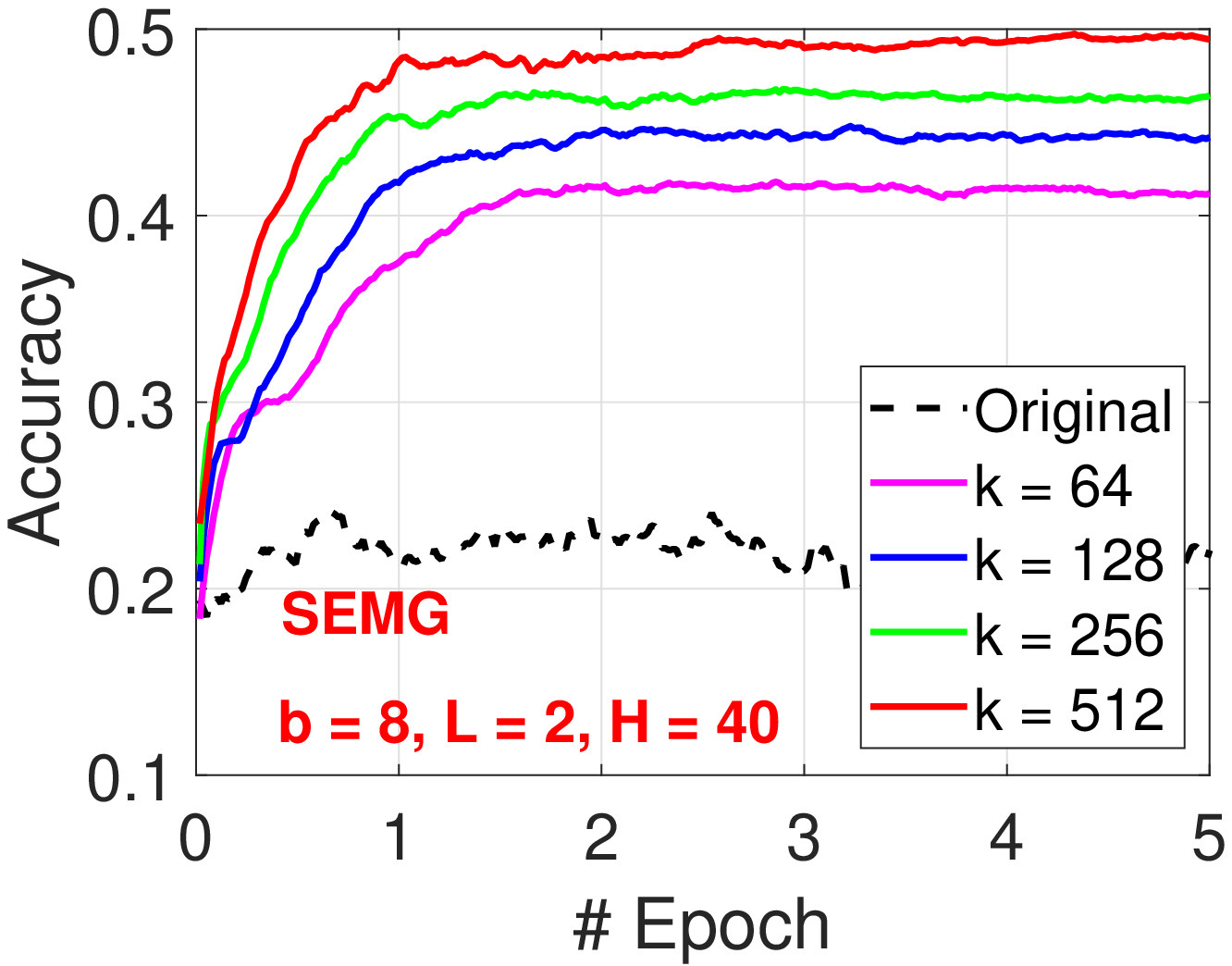}
\includegraphics[width=2.2in]{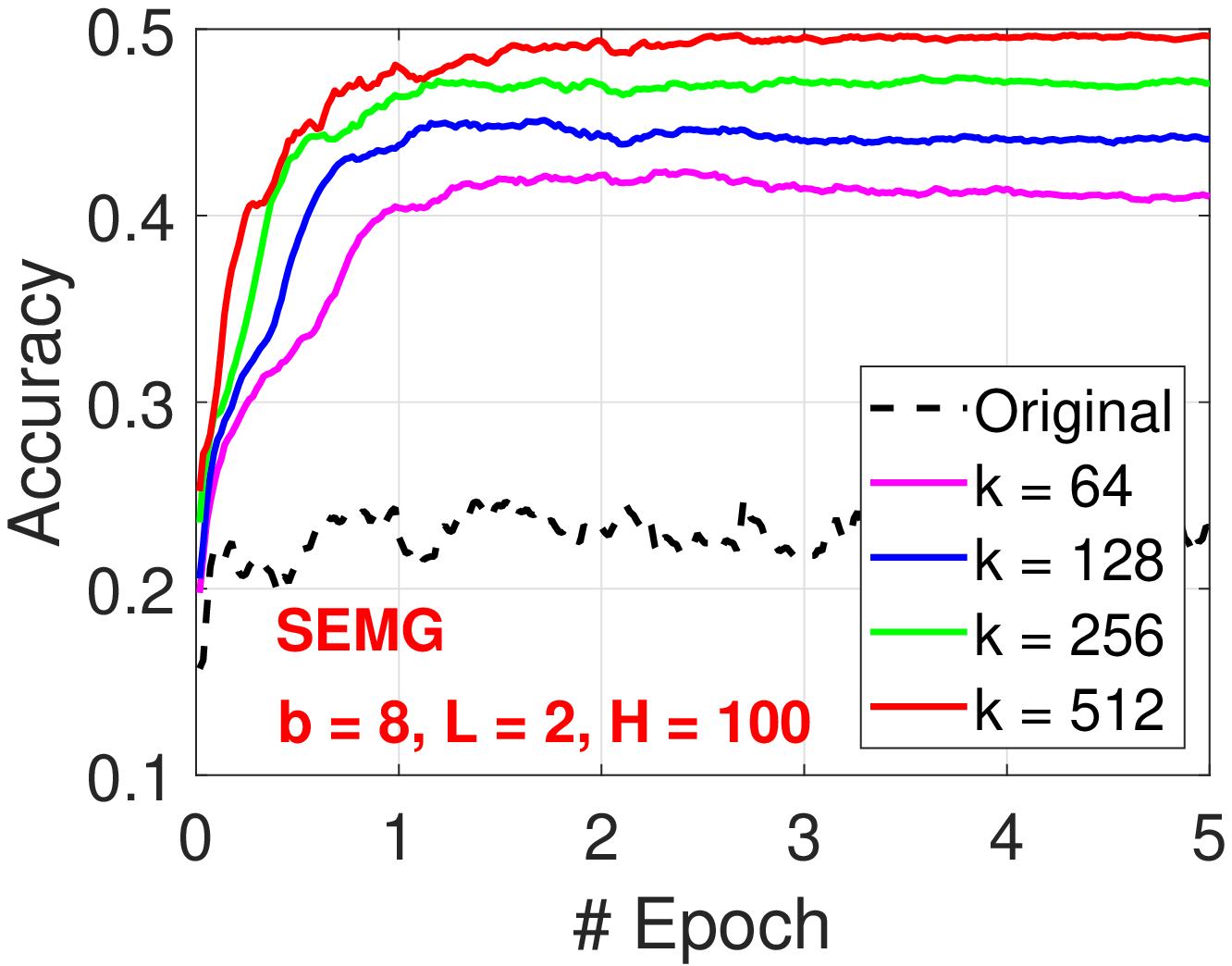}
\includegraphics[width=2.2in]{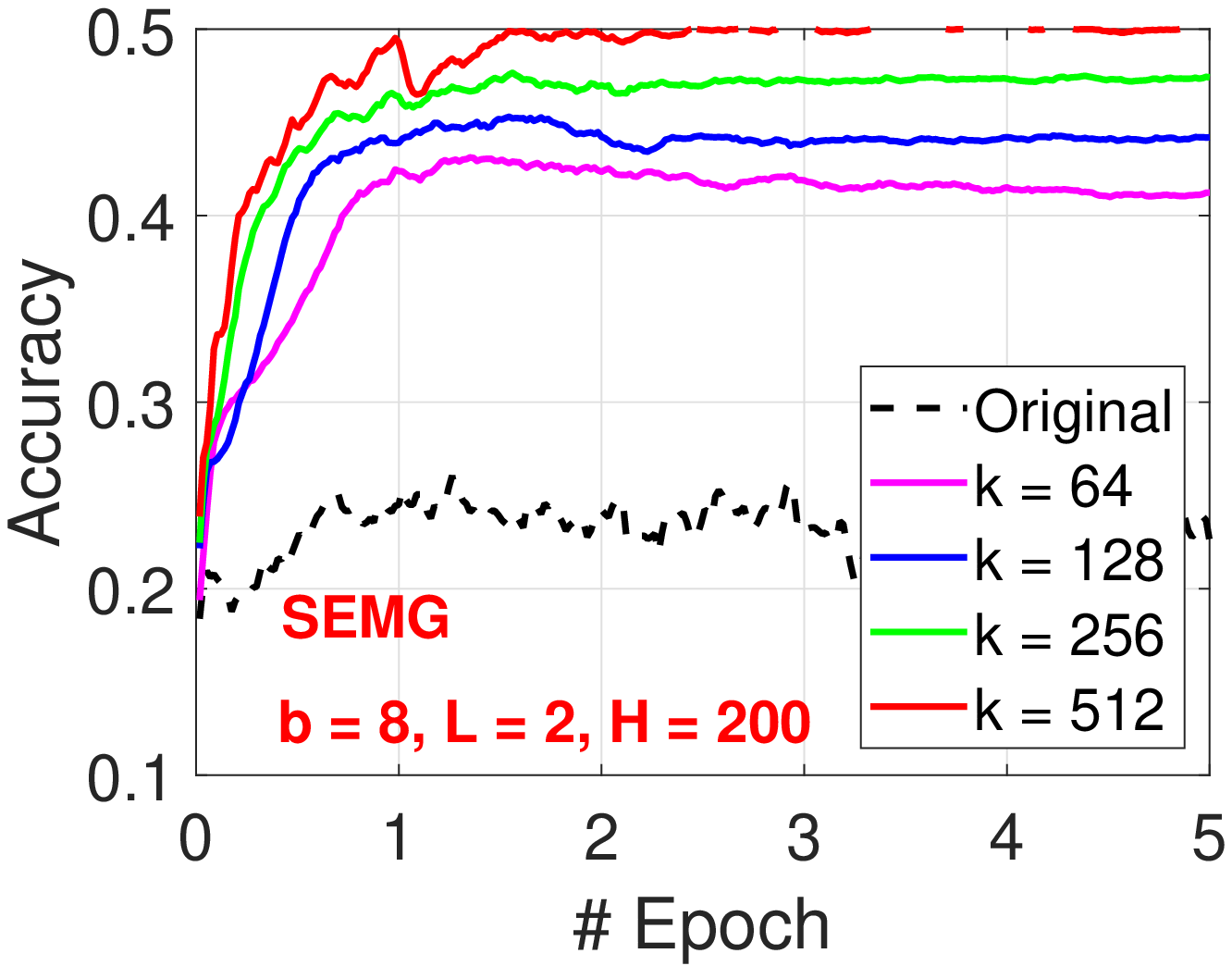}
}

\end{center}

\caption{GCWSNet with $p=2$ on the SEMG dataset. Note that for the original data (dashed black curves), we still report the original results without applying the power transformation. The results on the original data with power transformation using $p=2$ are actually not too much different from using $p=1$.  }\label{fig:SEMG2_p2}\vspace{0.3in}
\end{figure}

Table~\ref{tab:SVM} shows that, for the SEMG dataset, the pGMM kernel with $p=2$ improves the classification accuracy. Thus, we also report the results of GCWSNet for $p=2$ in Figure~\ref{fig:SEMG2_p2}. Compared with the corresponding plots/curves in Figure~\ref{fig:SEMG2_p1}, the improvements are quite obvious. Figure~\ref{fig:SEMG2_p2} also  again confirms that training just one epoch can already reach good accuracies.

\vspace{0.1in}

\begin{figure}[h]

\begin{center}

\mbox{
\includegraphics[width=2.2in]{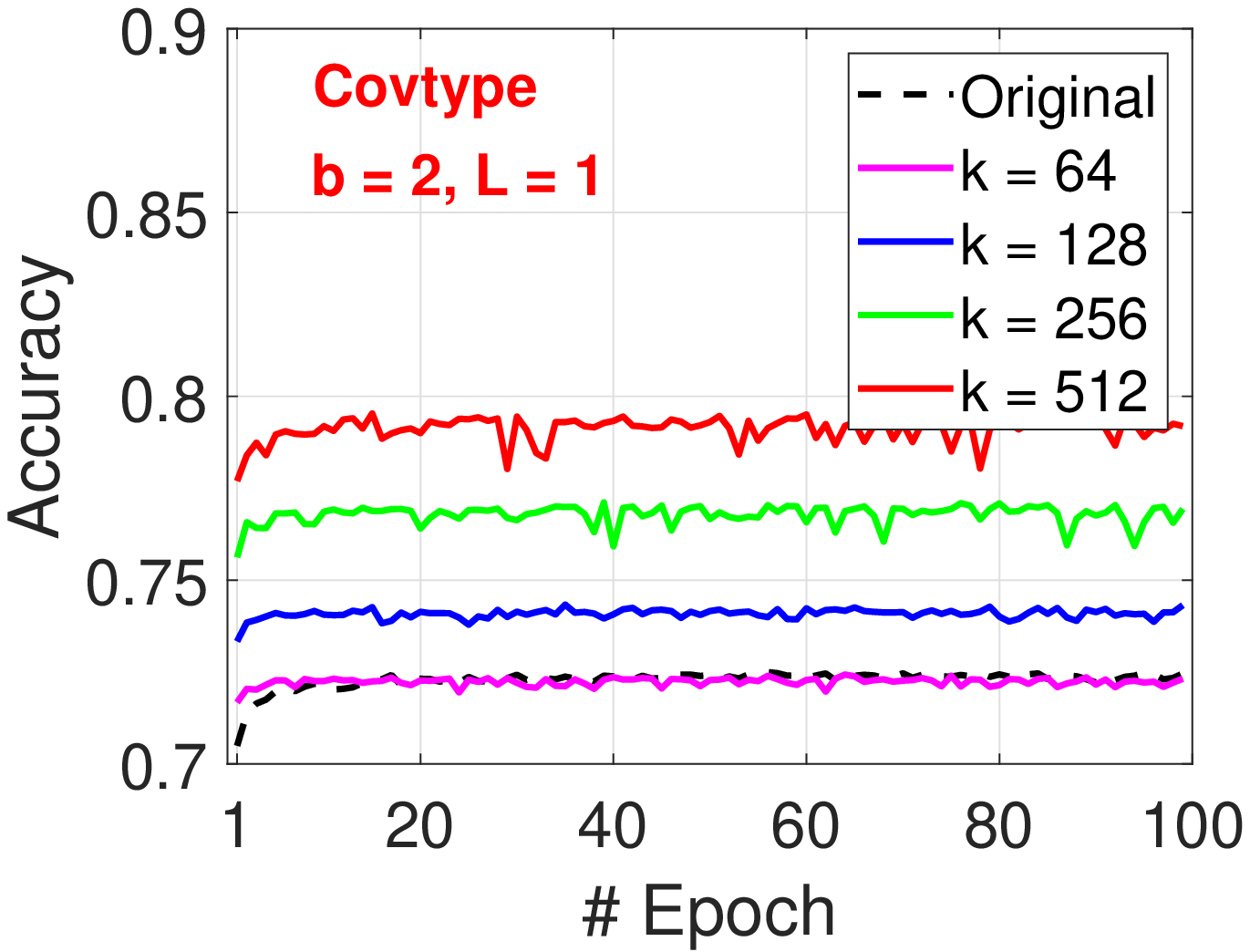}
\includegraphics[width=2.2in]{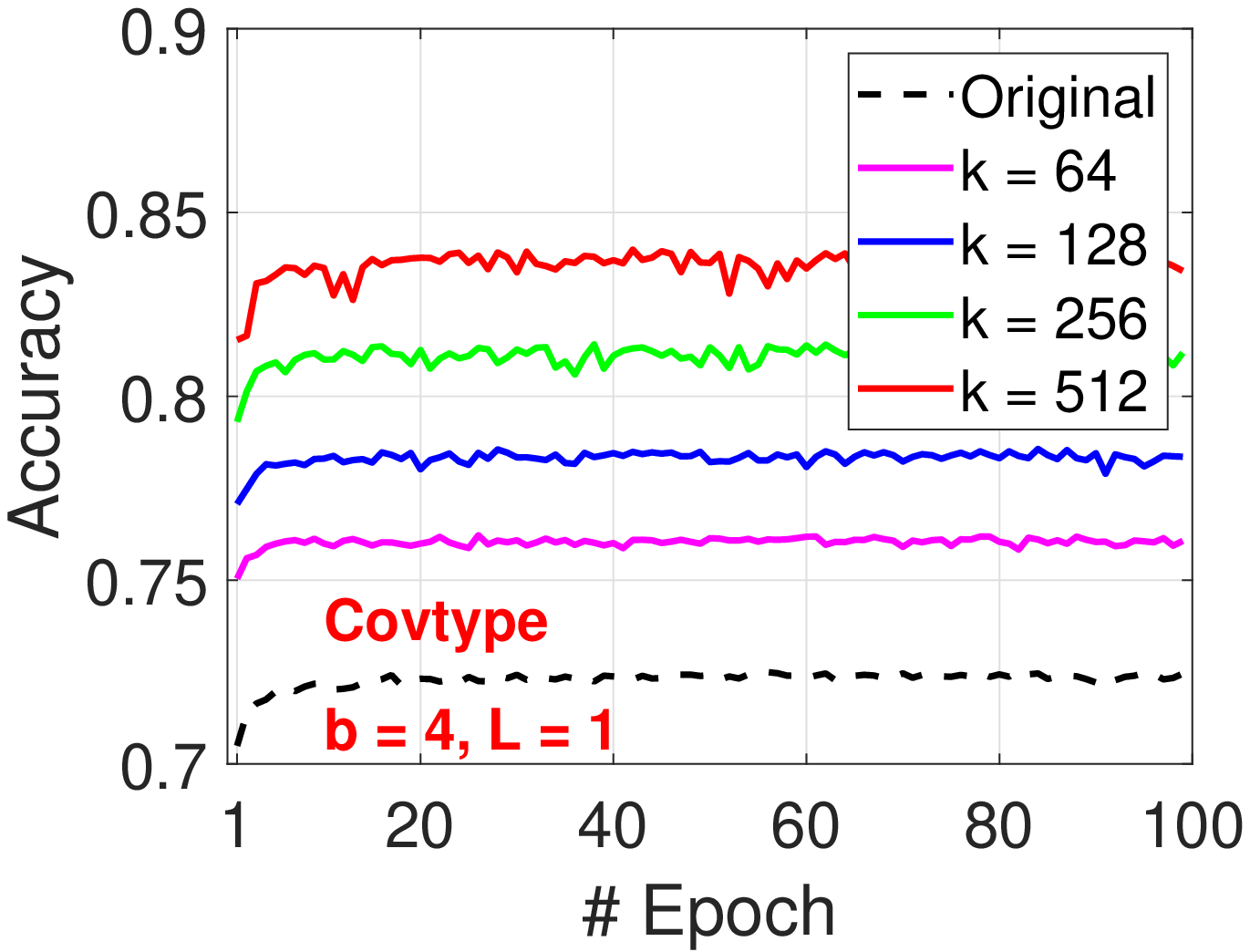}
\includegraphics[width=2.2in]{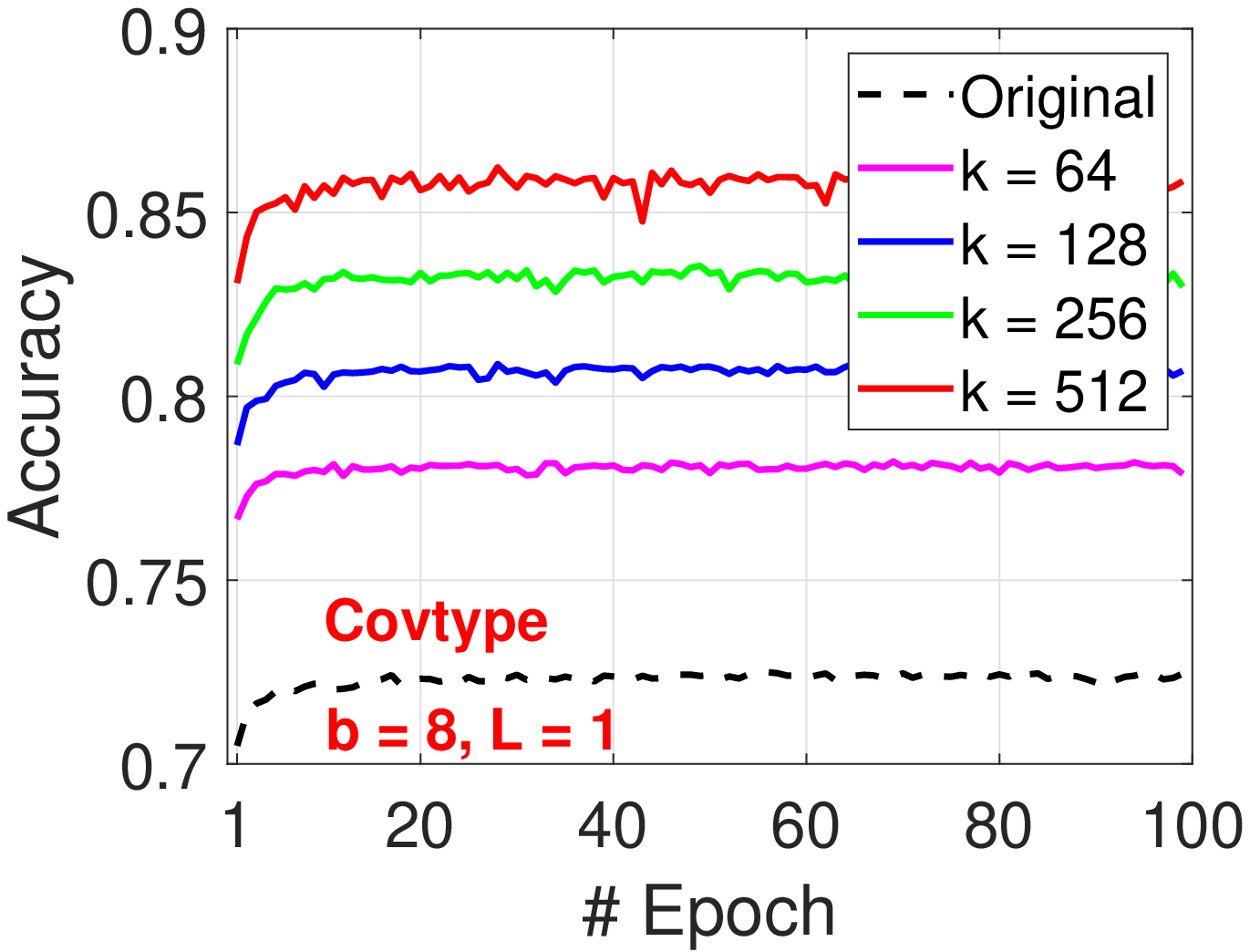}
}

\mbox{
\includegraphics[width=2.2in]{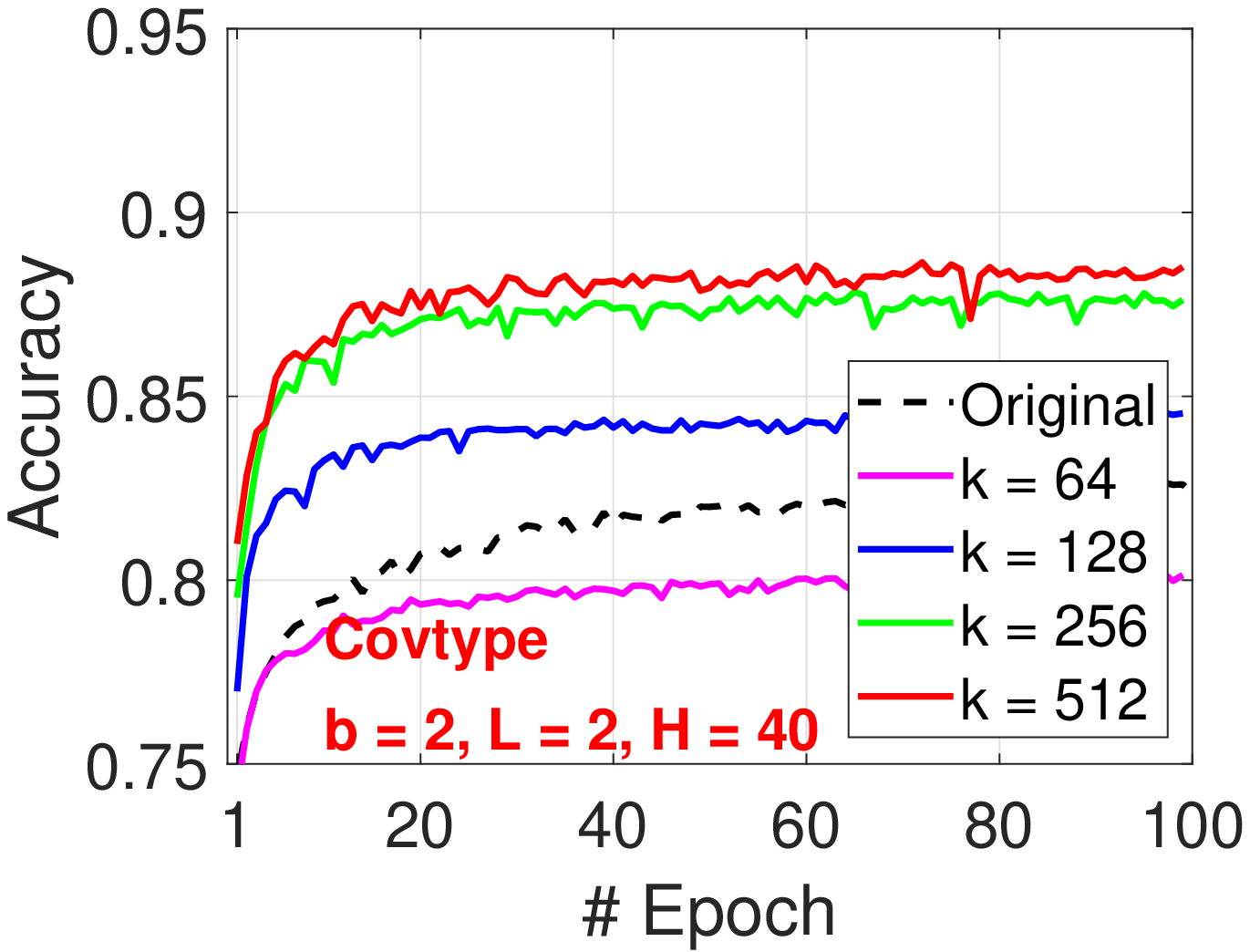}
\includegraphics[width=2.2in]{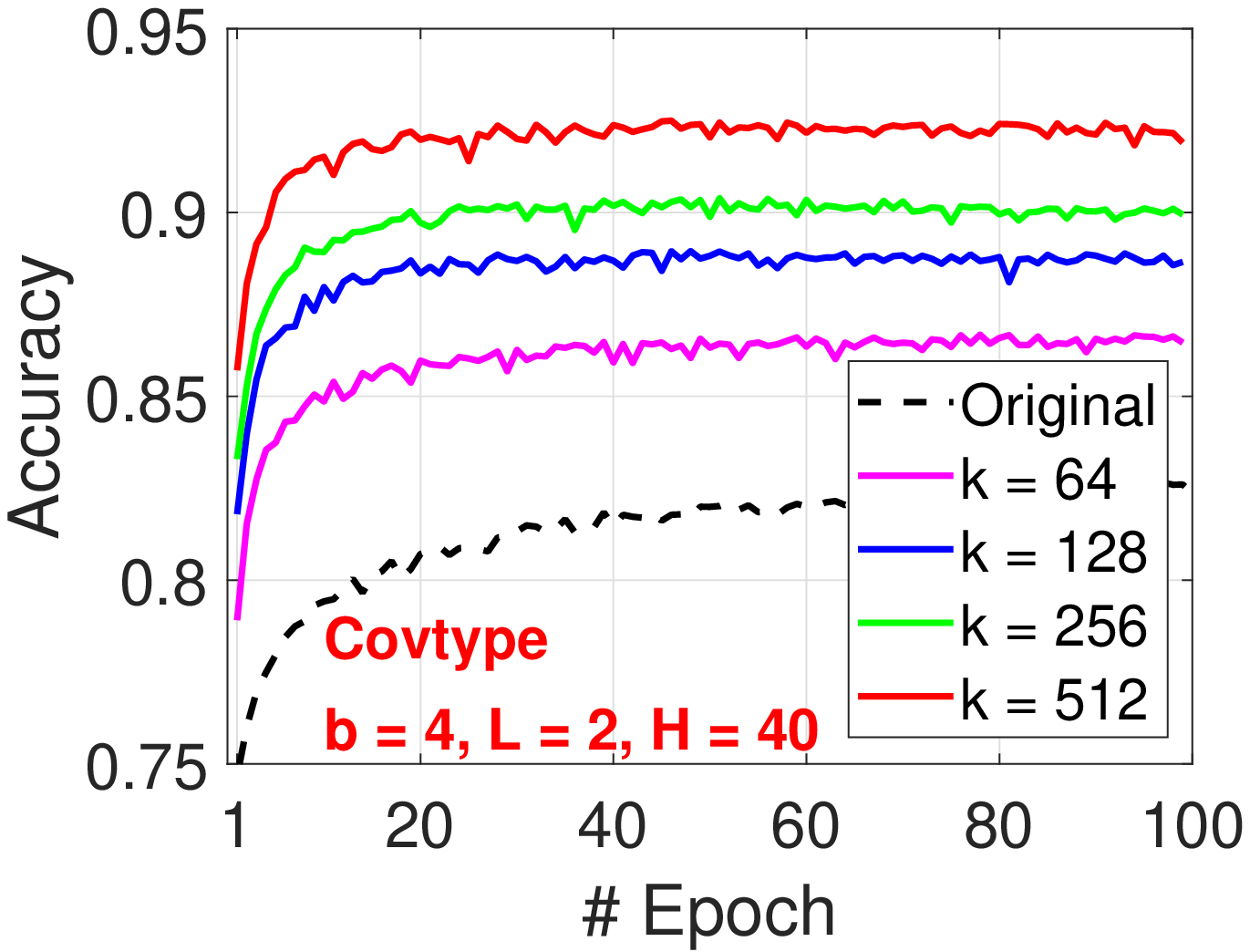}
\includegraphics[width=2.2in]{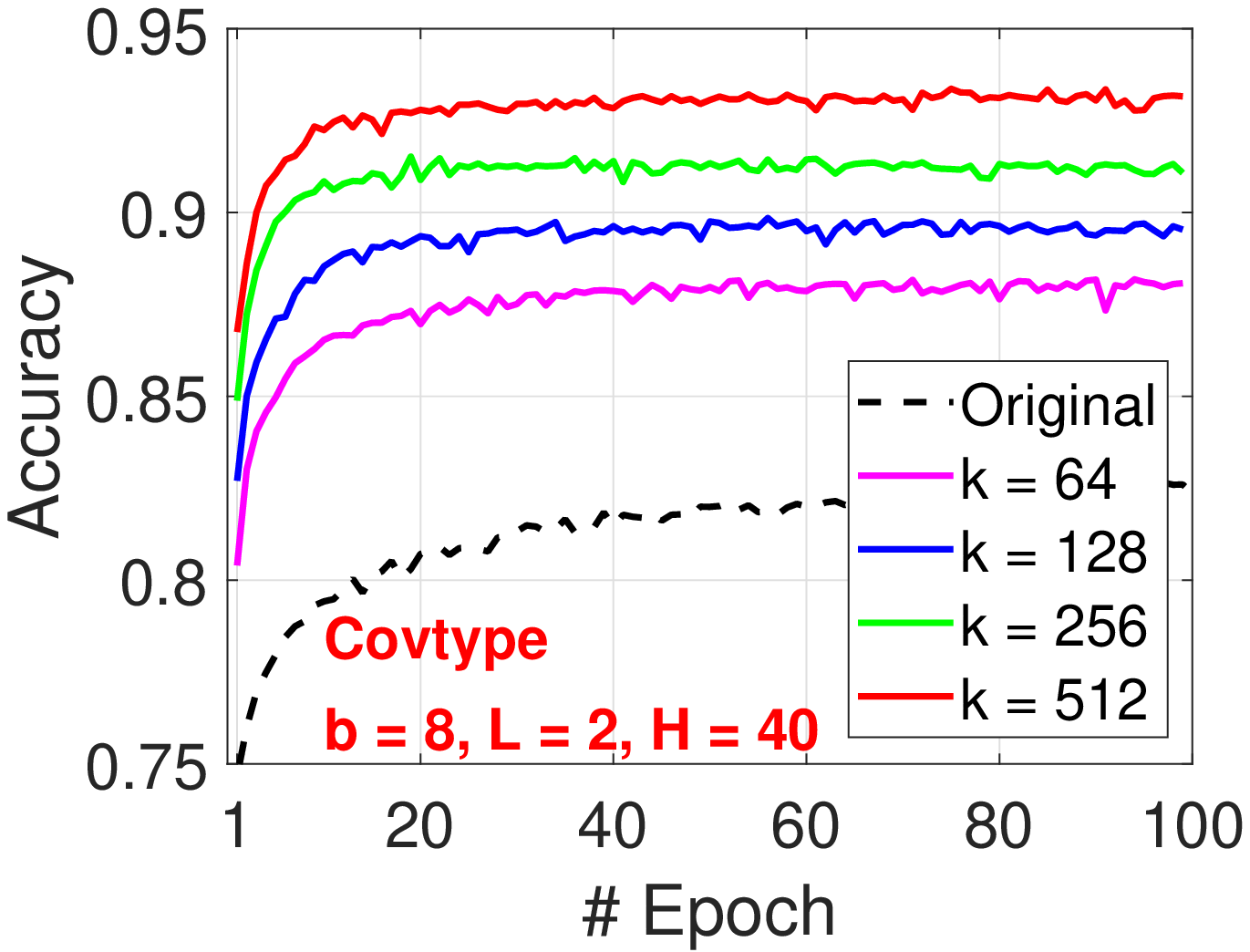}
}

\mbox{
\includegraphics[width=2.2in]{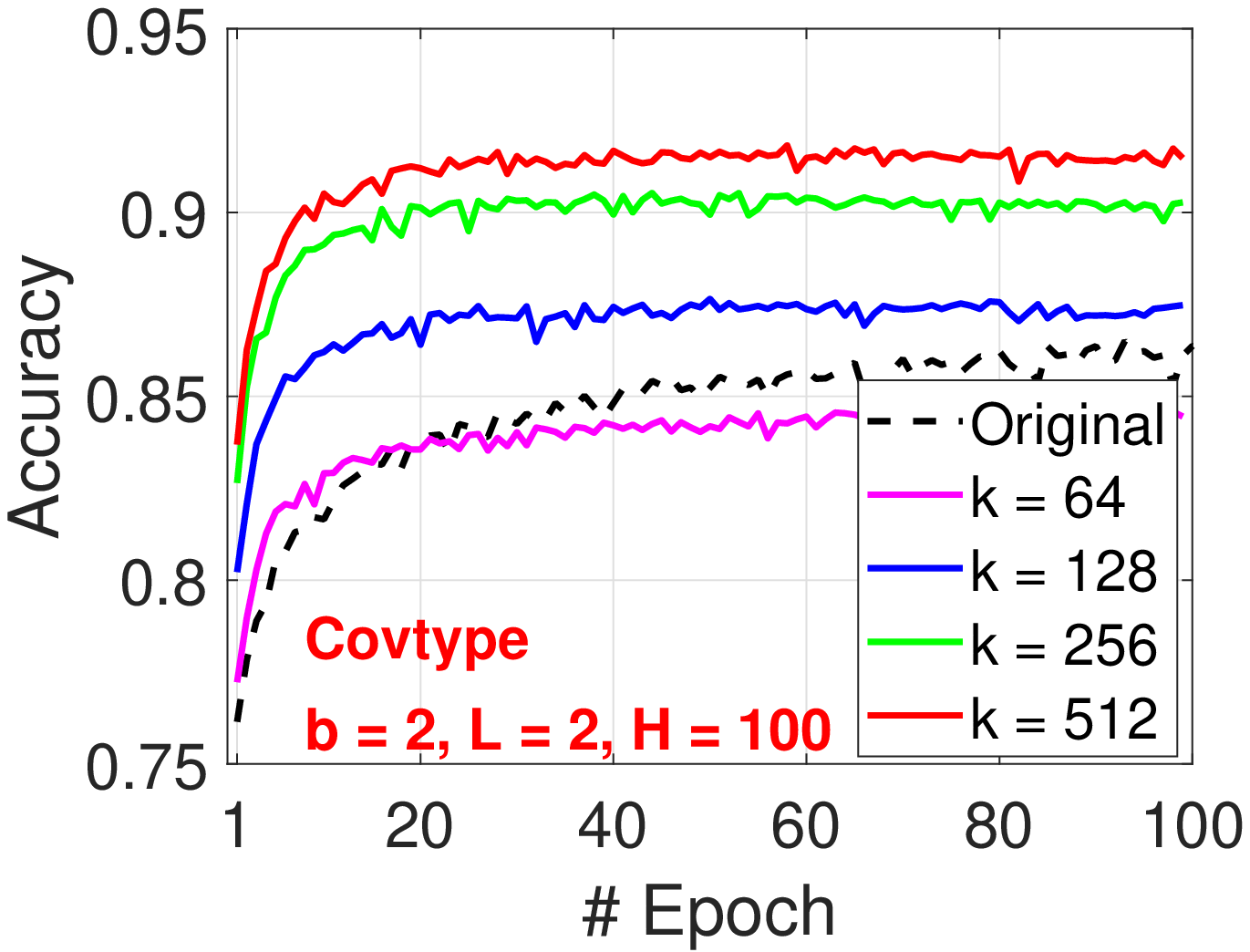}
\includegraphics[width=2.2in]{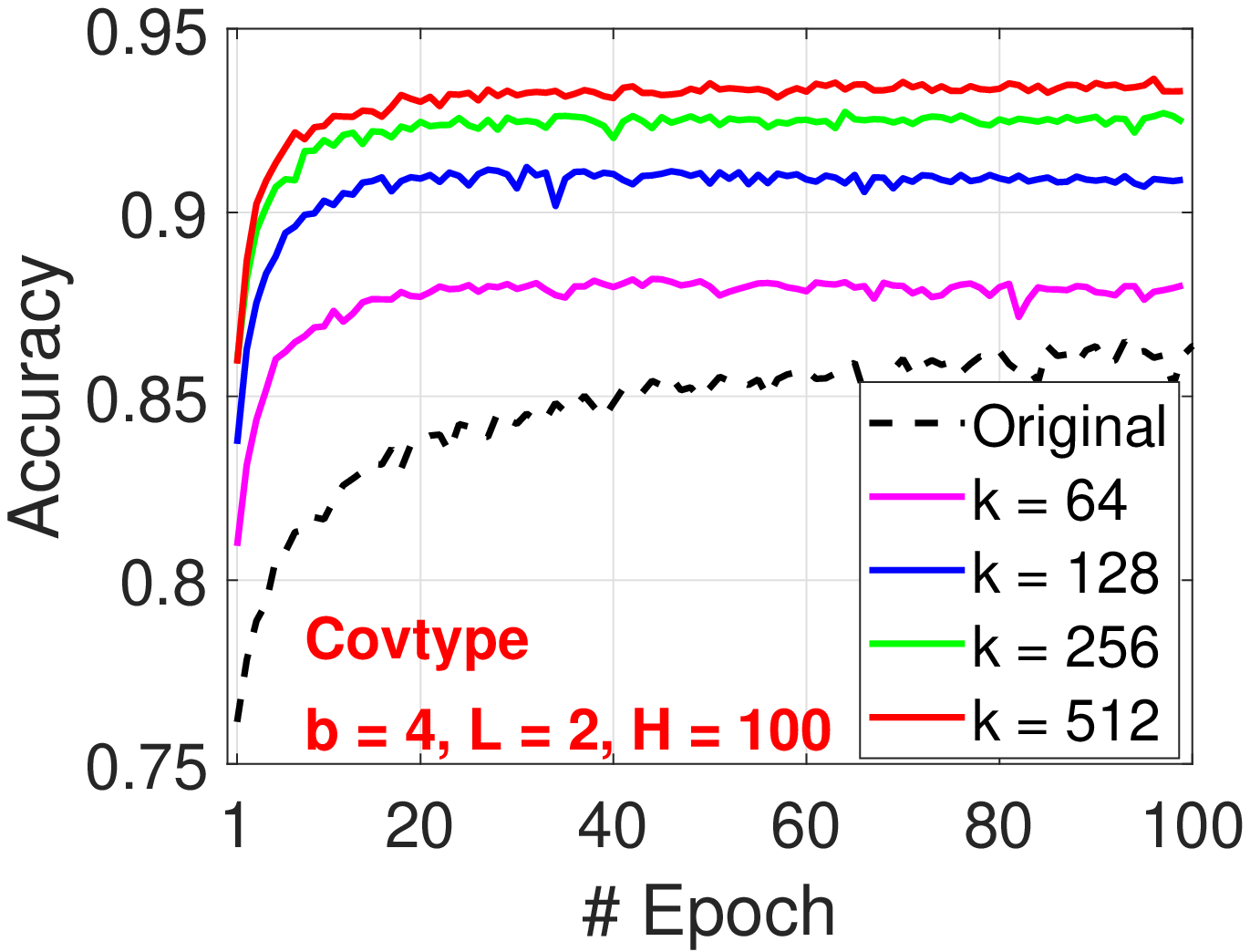}
\includegraphics[width=2.2in]{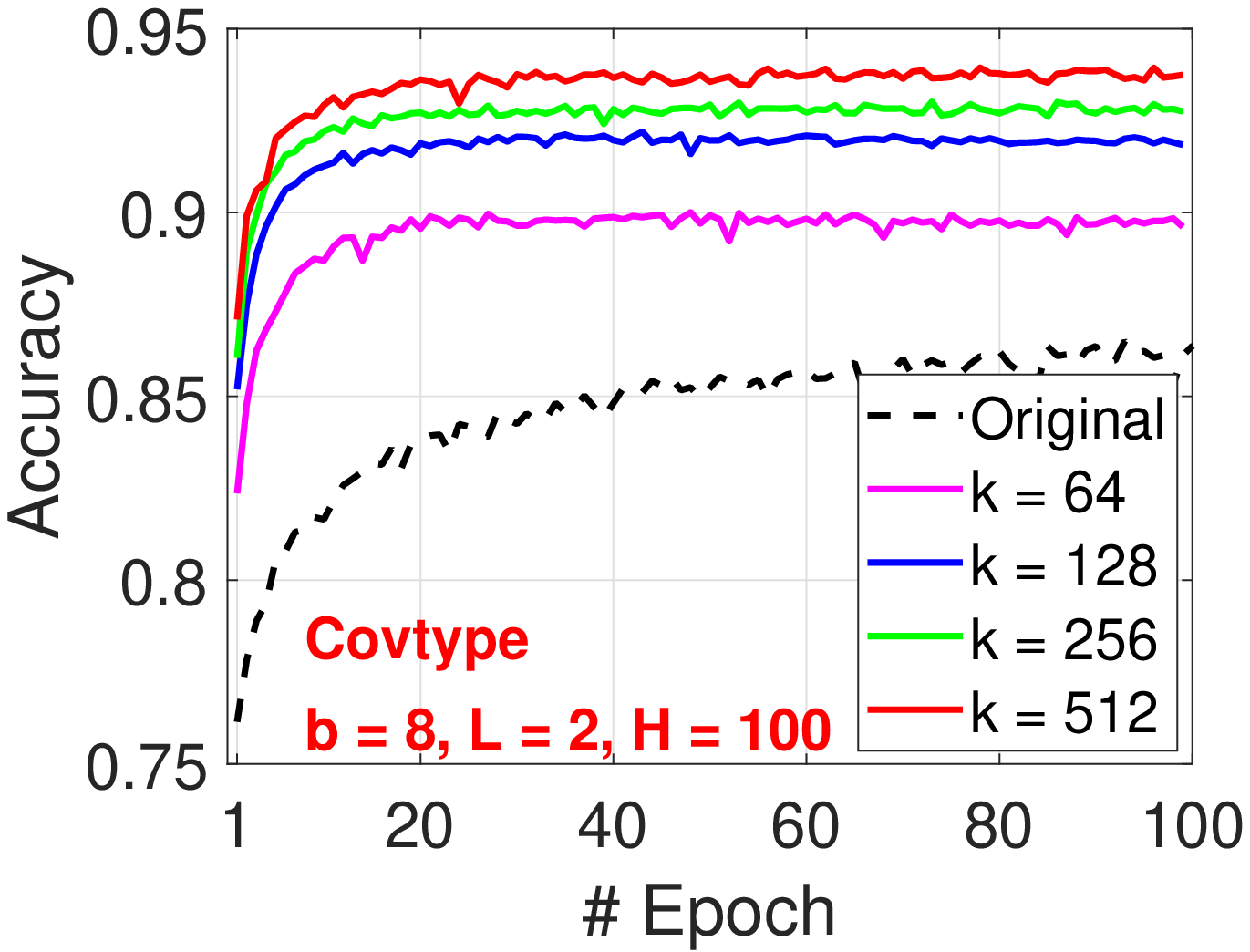}
}

\mbox{
\includegraphics[width=2.2in]{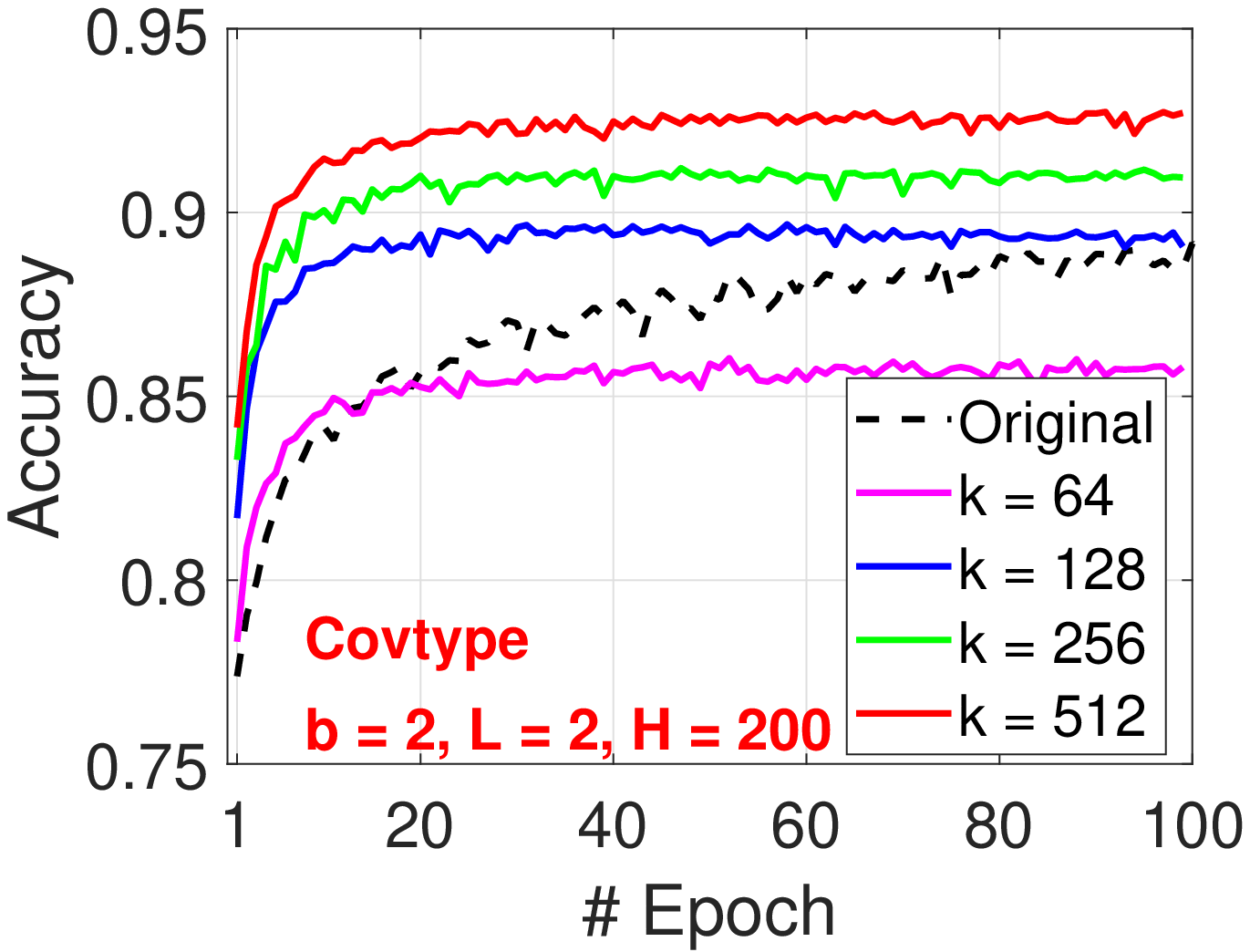}
\includegraphics[width=2.2in]{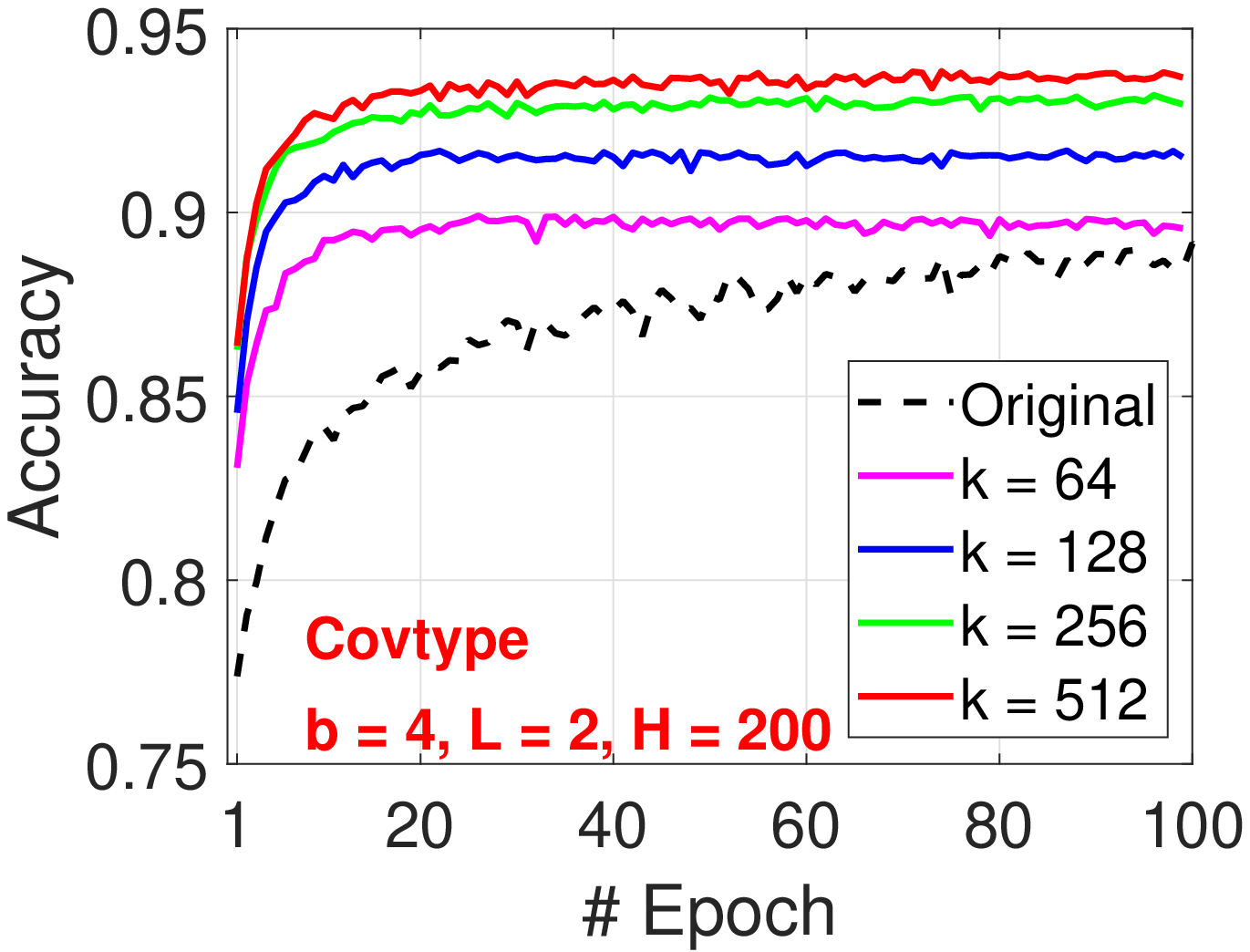}
\includegraphics[width=2.2in]{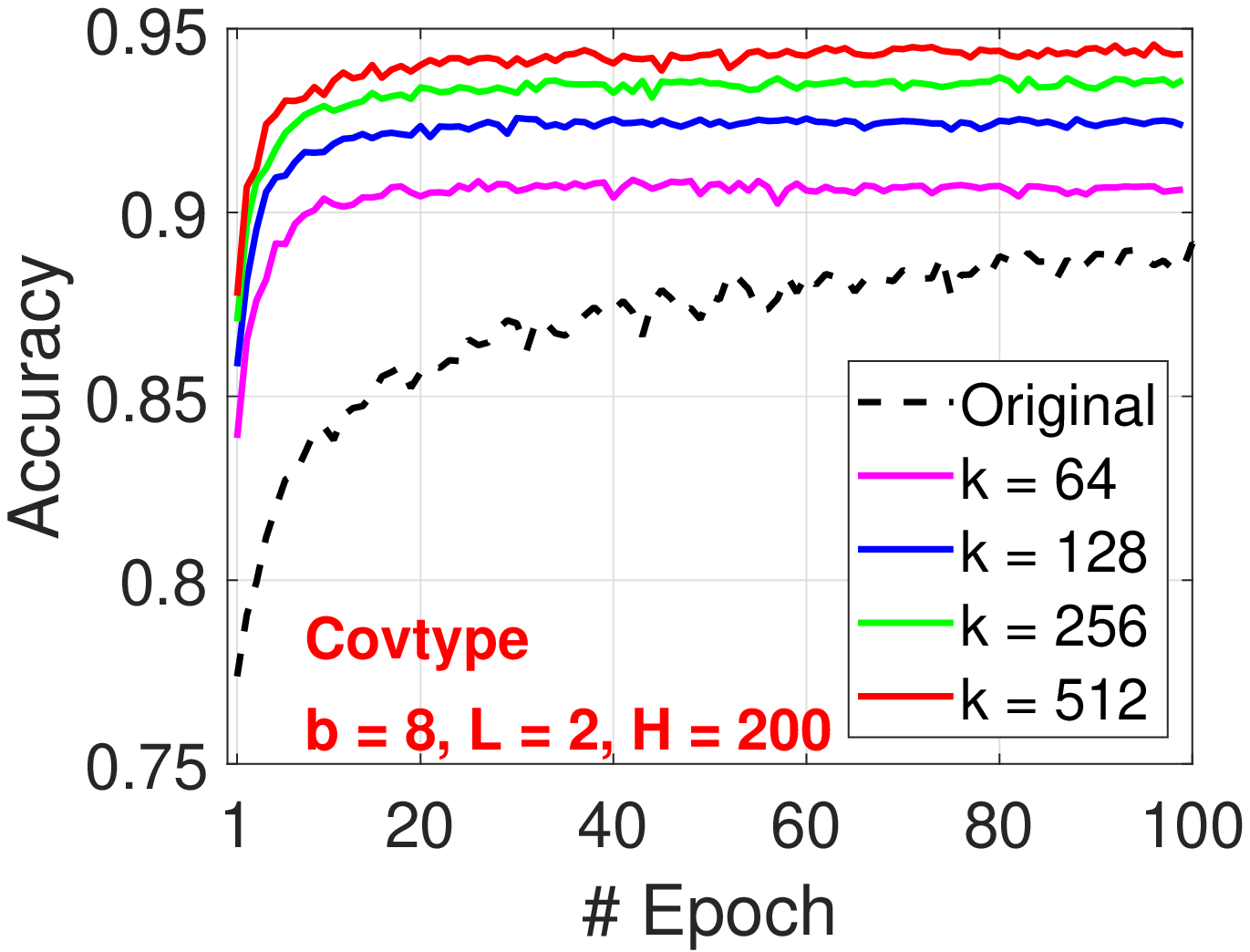}
}

\end{center}
\caption{GCWSNet with $p=1$, for $b\in\{2,4,8\}$ and $k\in\{64,128,256,512\}$, on the Covtype dataset, for 100 epochs. The results for the original data are reported as dashed (black) curves, which are substantially worse than the results of GCWSNet. Again, GCWSNet converges much faster than training on the original data.}\label{fig:Covtype_p1}
\end{figure}

\begin{figure}[h]

\begin{center}

\mbox{
\includegraphics[width=2.2in]{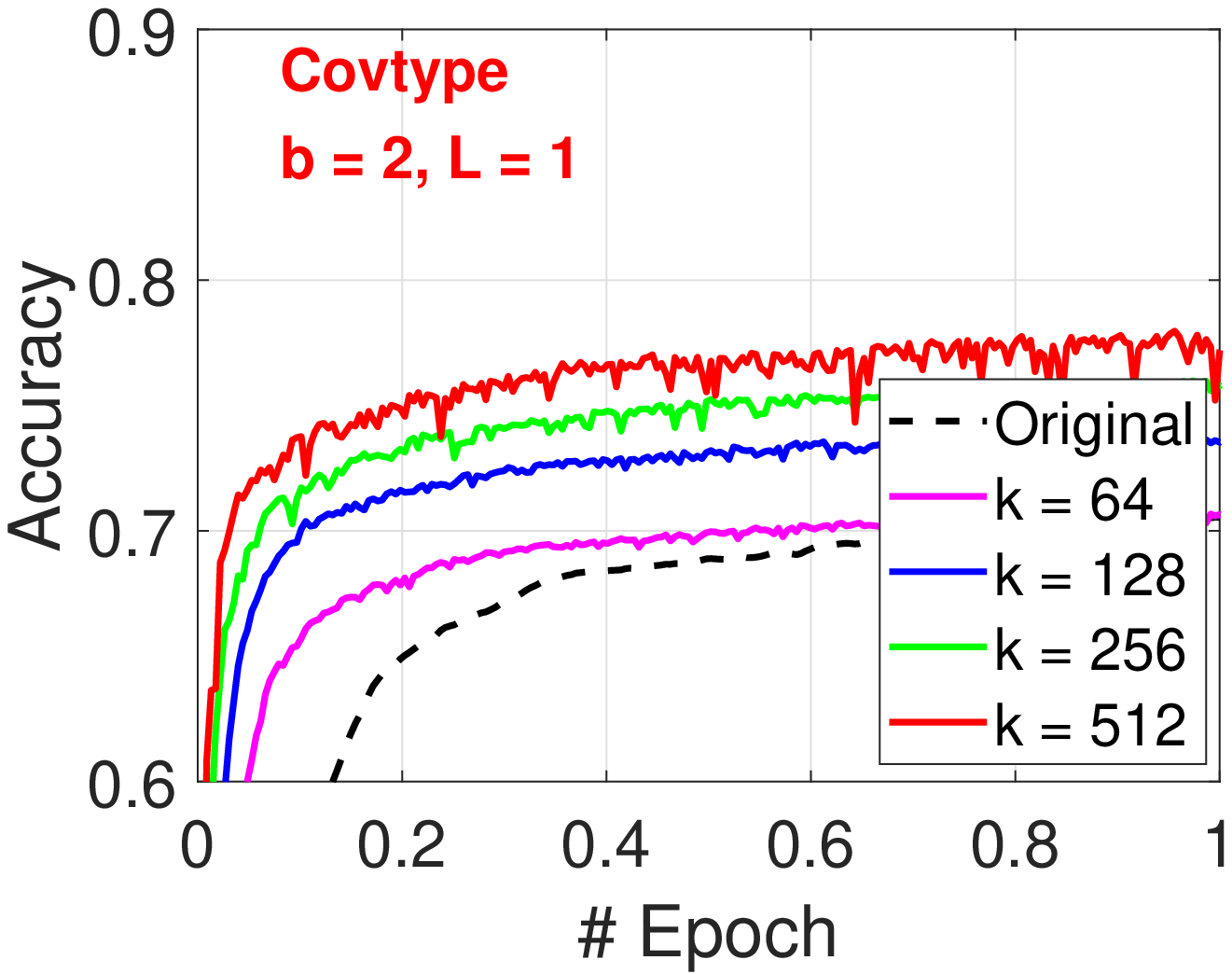}
\includegraphics[width=2.2in]{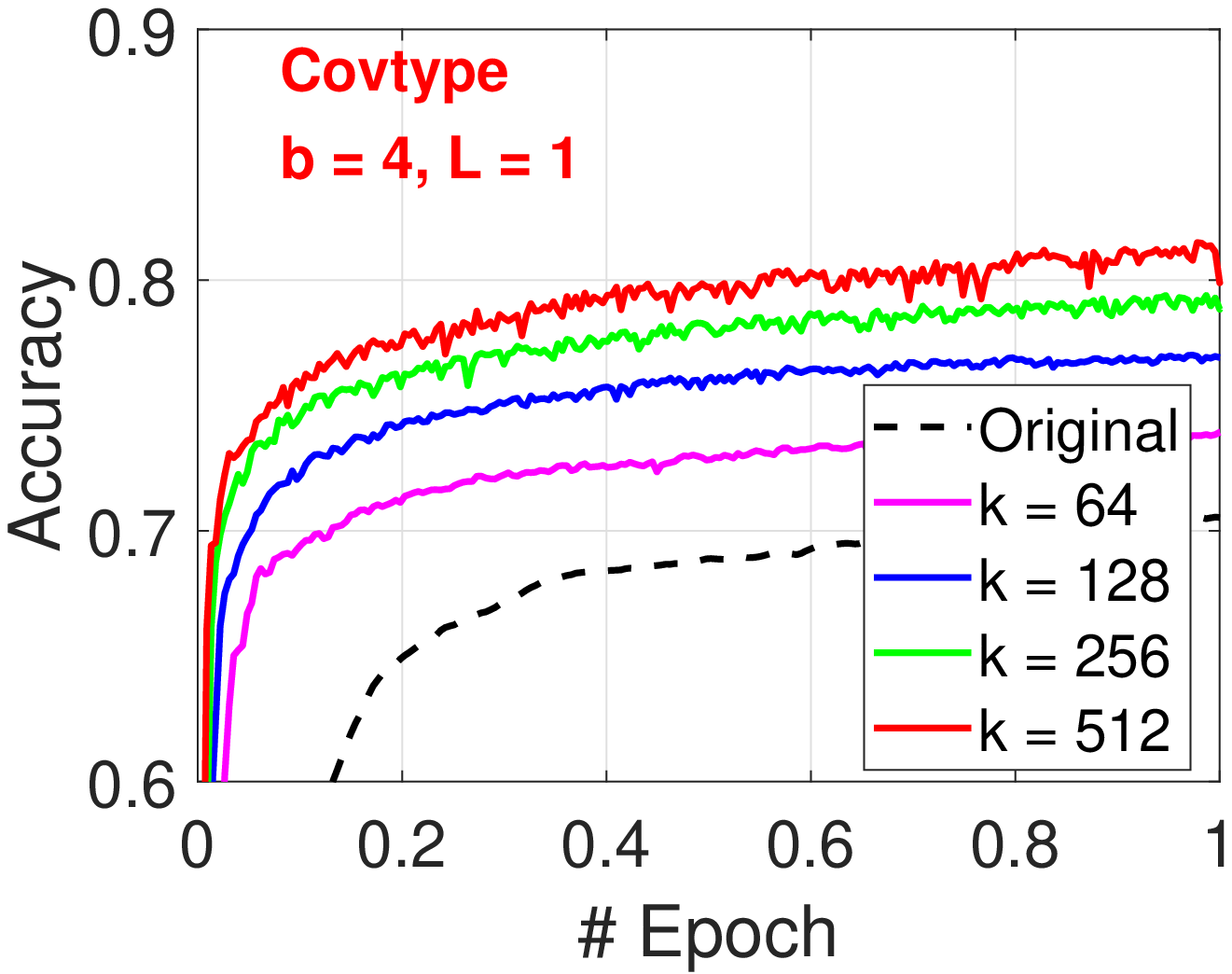}
\includegraphics[width=2.2in]{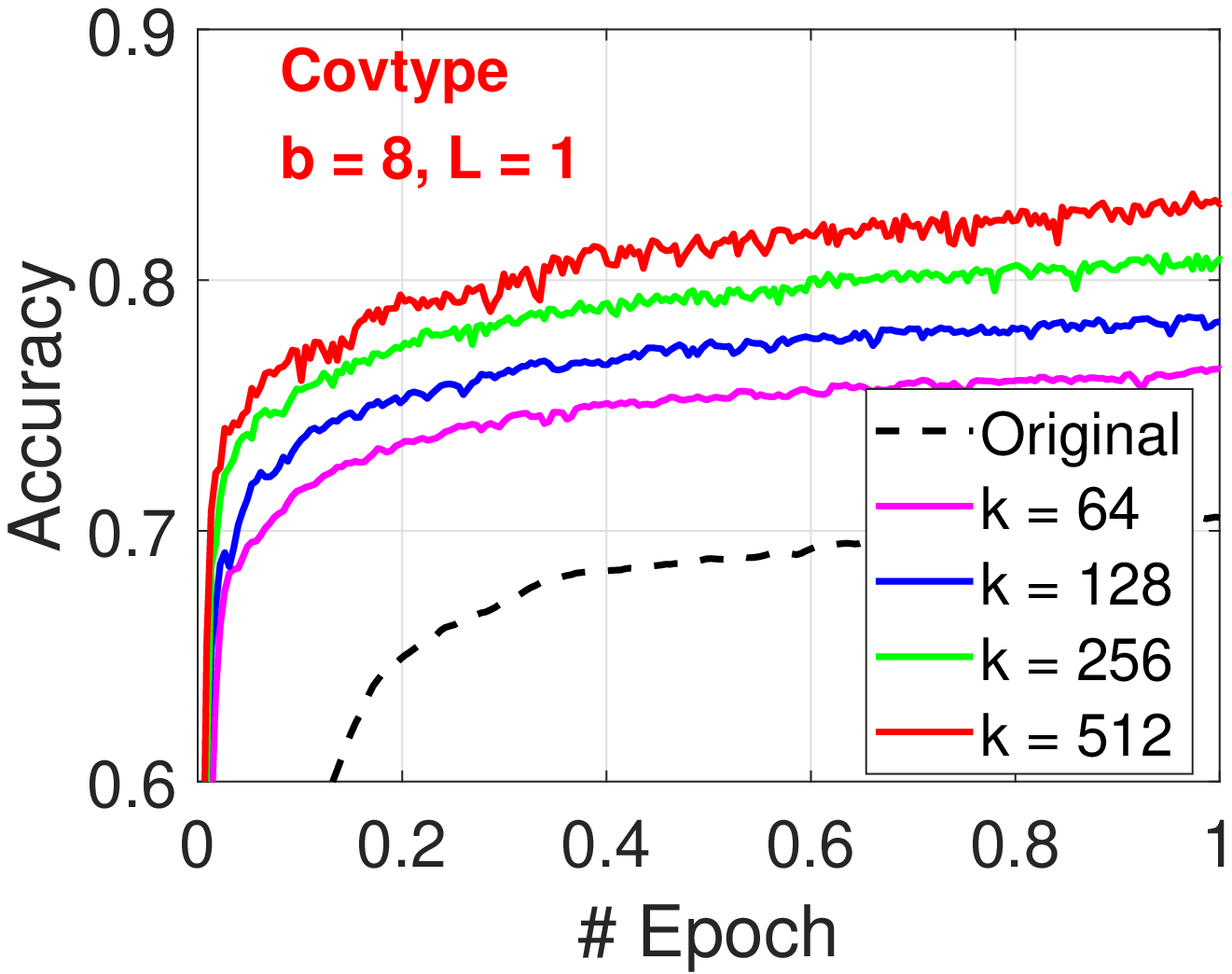}
}

\mbox{
\includegraphics[width=2.2in]{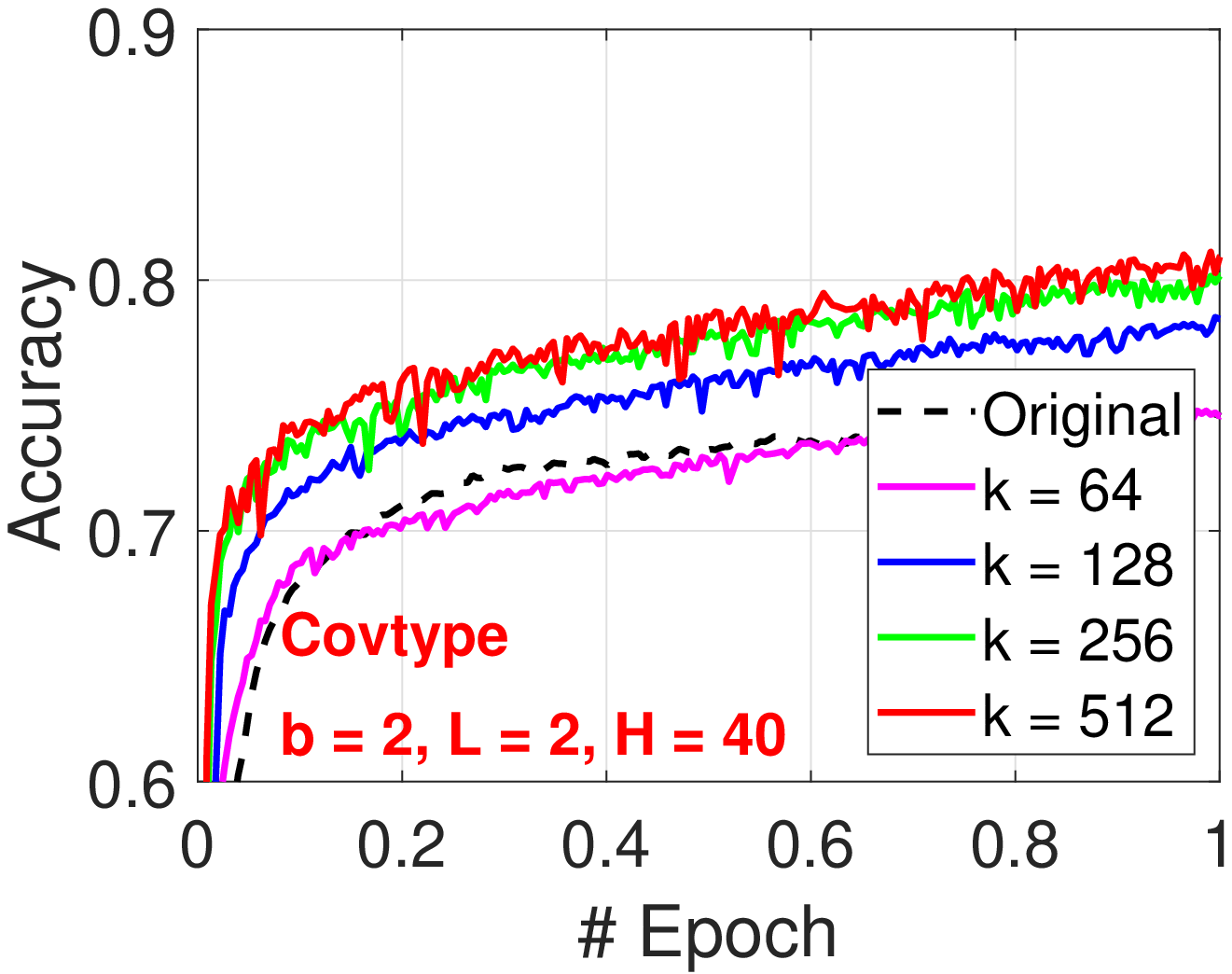}
\includegraphics[width=2.2in]{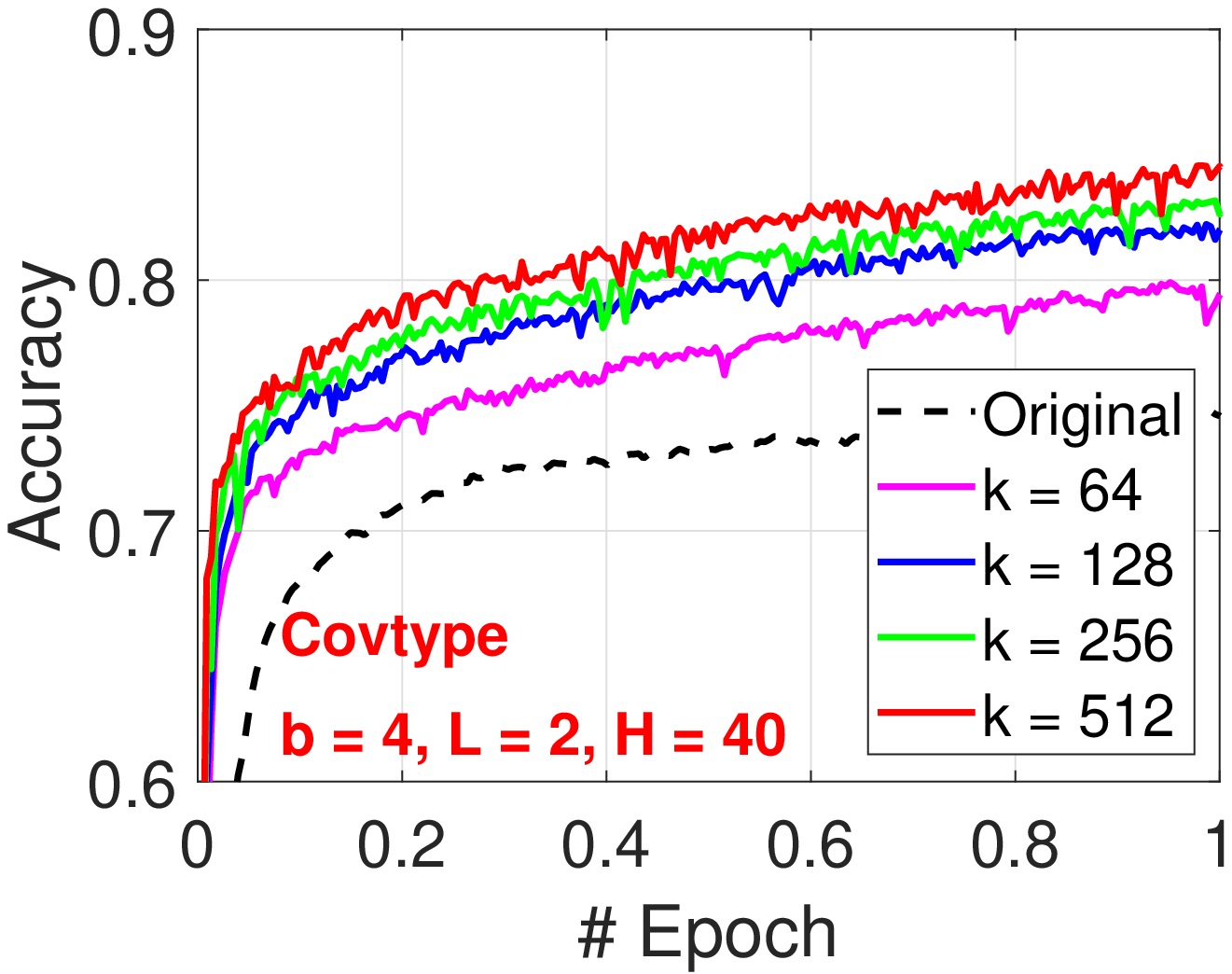}
\includegraphics[width=2.2in]{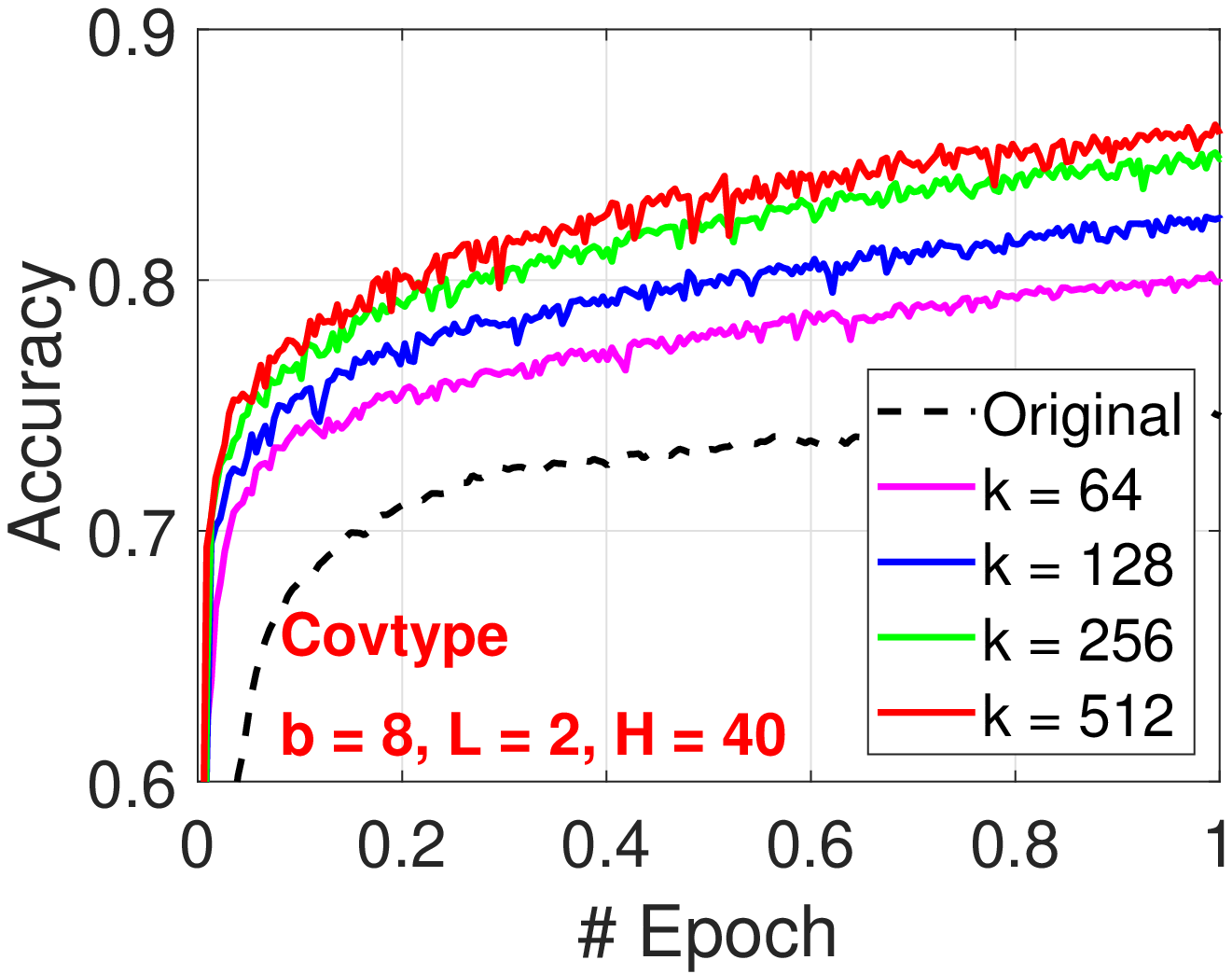}
}

\mbox{
\includegraphics[width=2.2in]{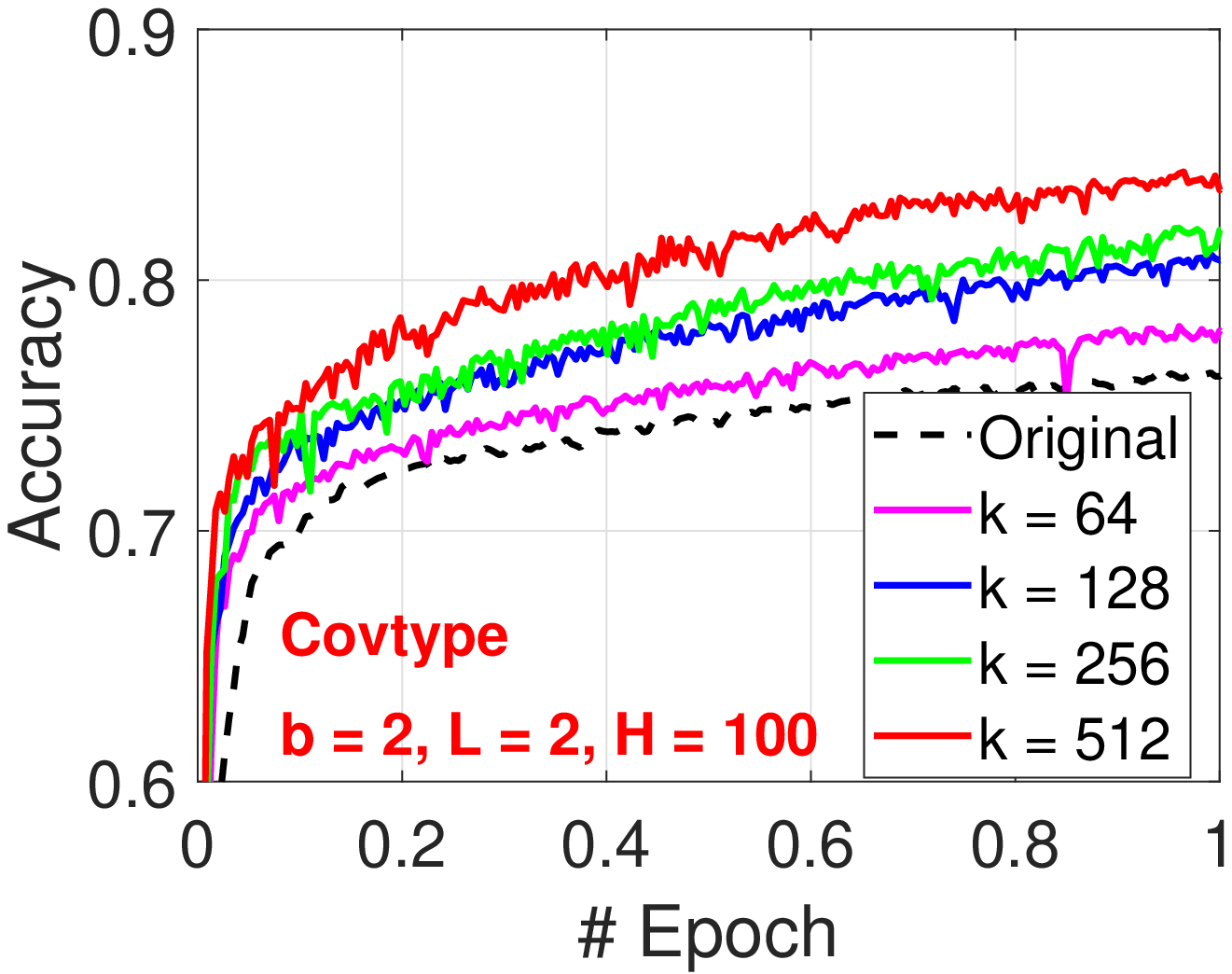}
\includegraphics[width=2.2in]{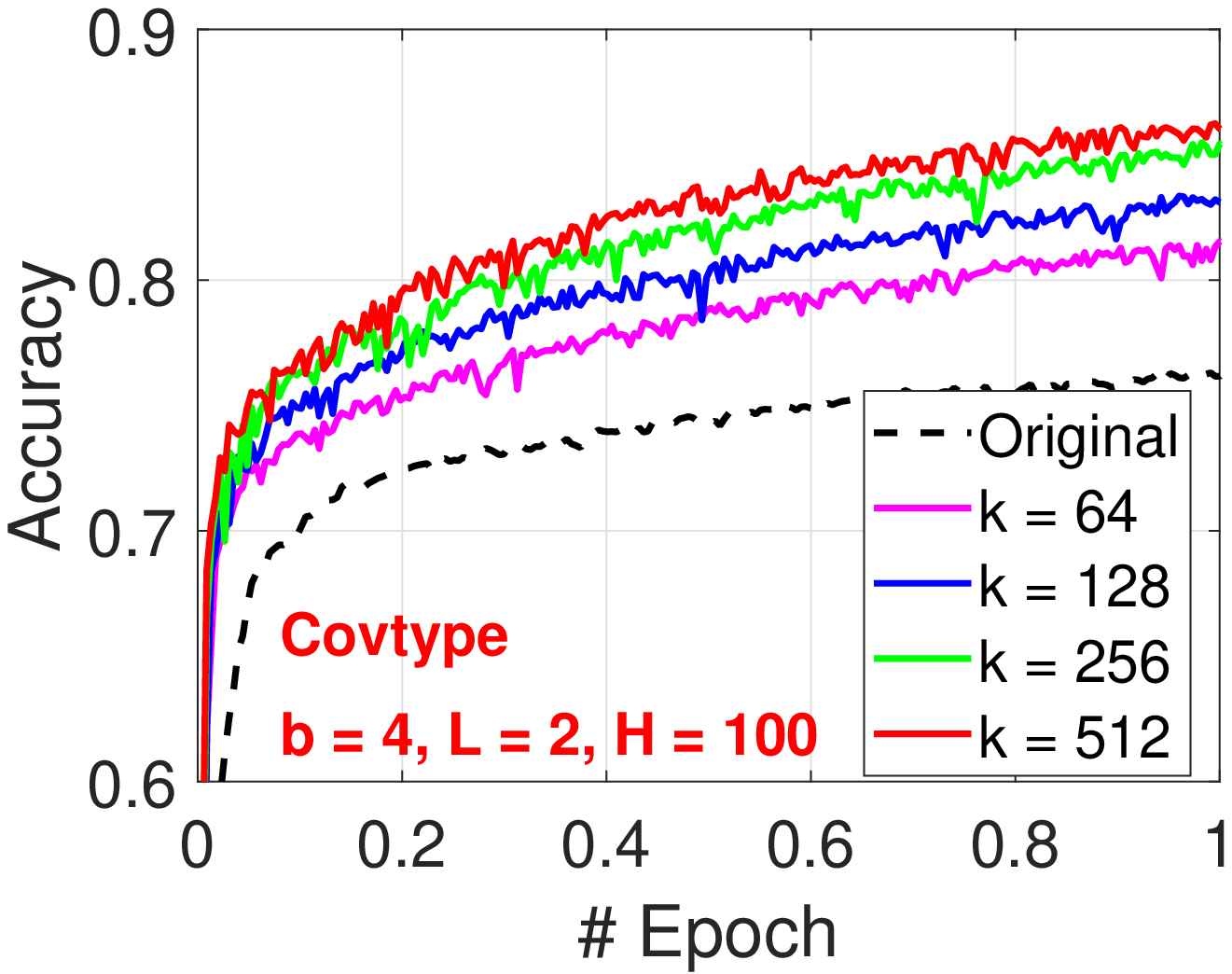}
\includegraphics[width=2.2in]{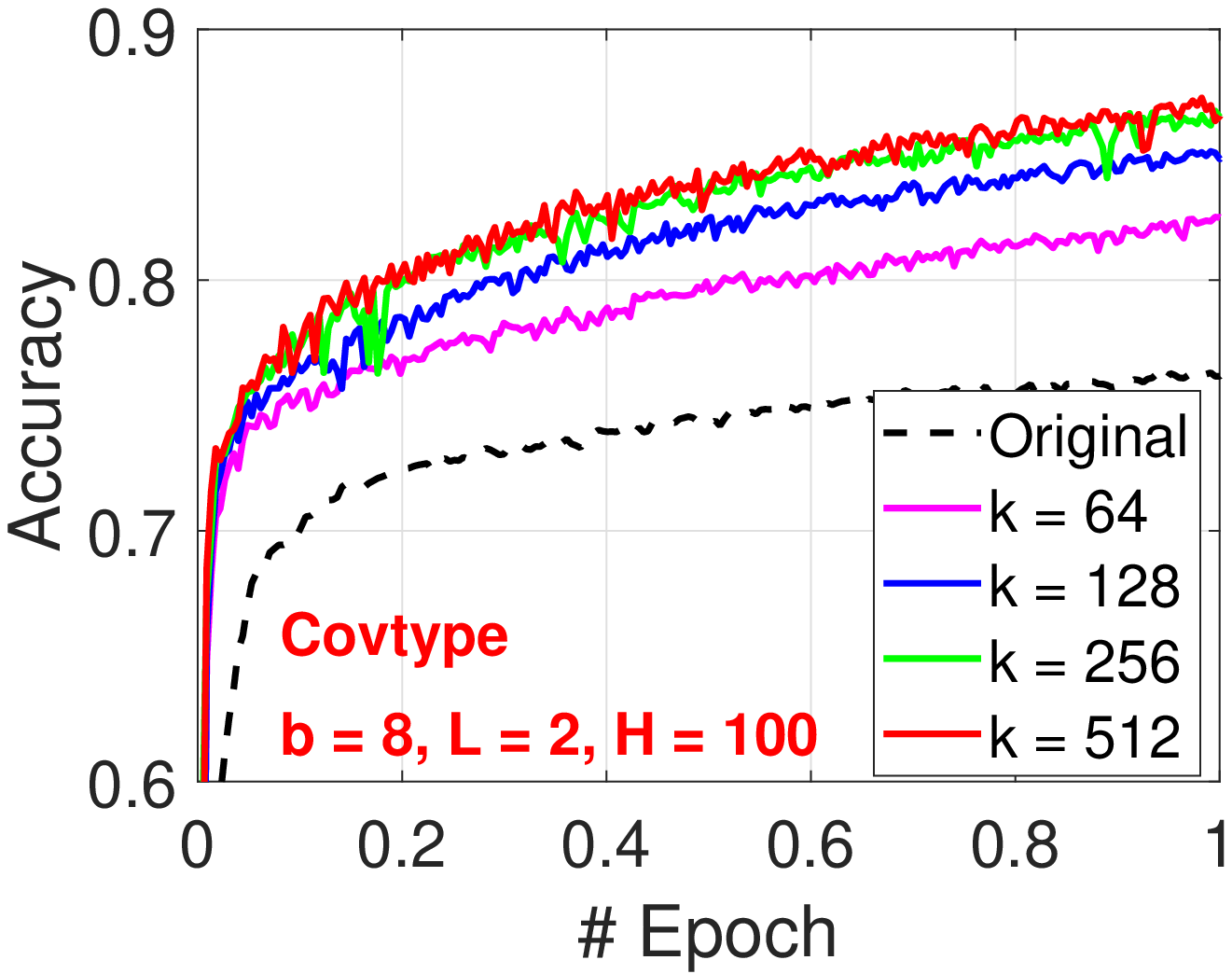}
}

\mbox{
\includegraphics[width=2.2in]{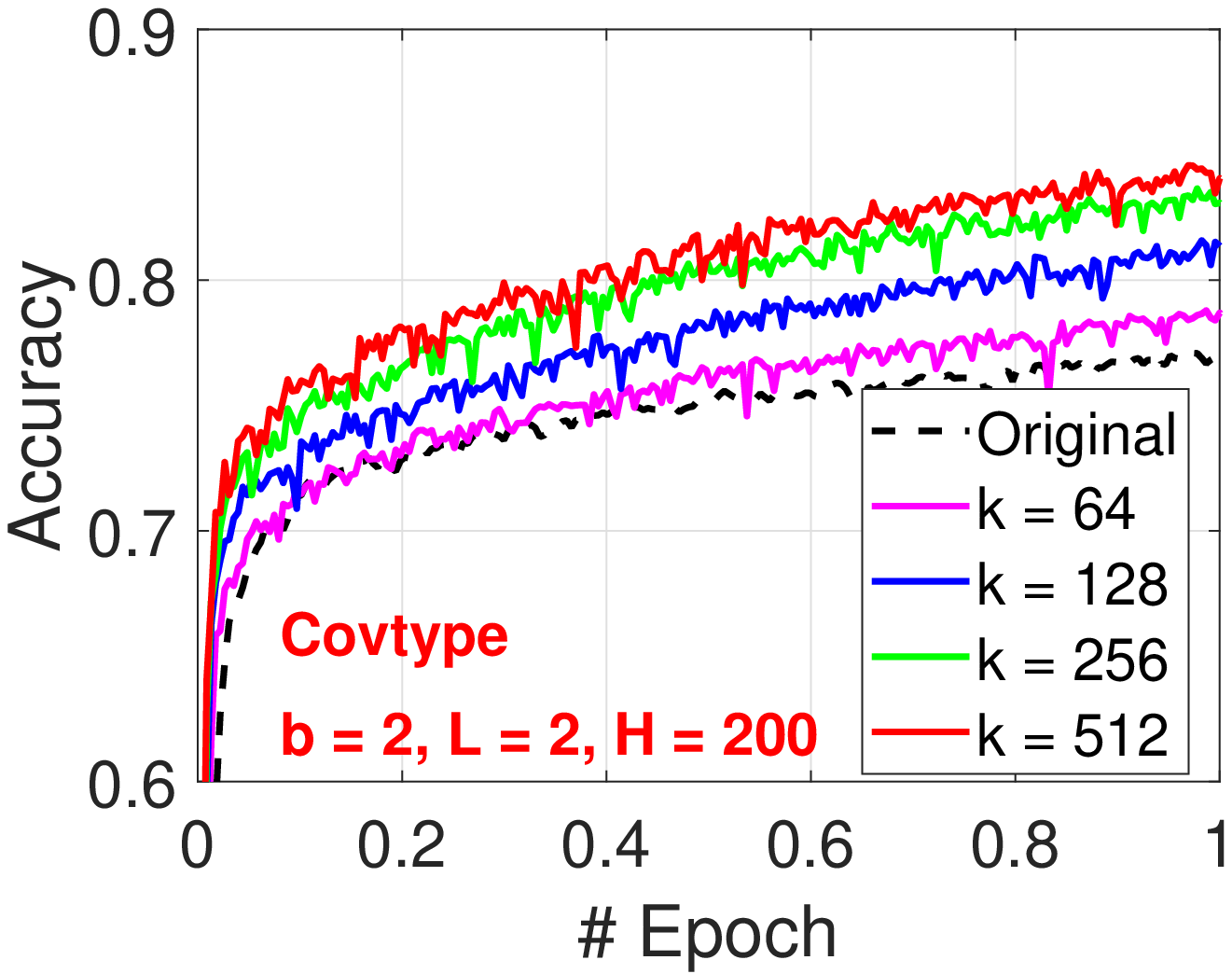}
\includegraphics[width=2.2in]{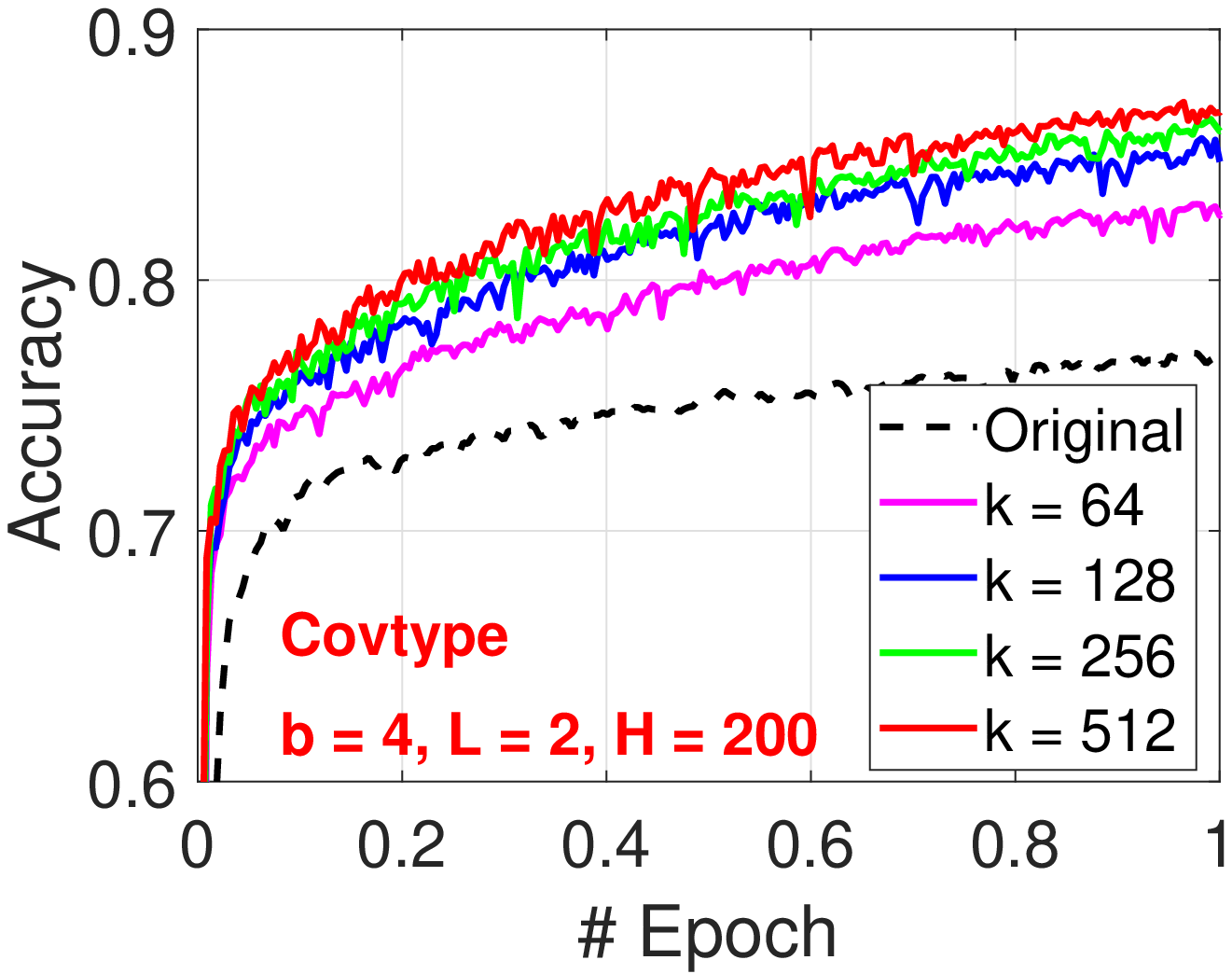}
\includegraphics[width=2.2in]{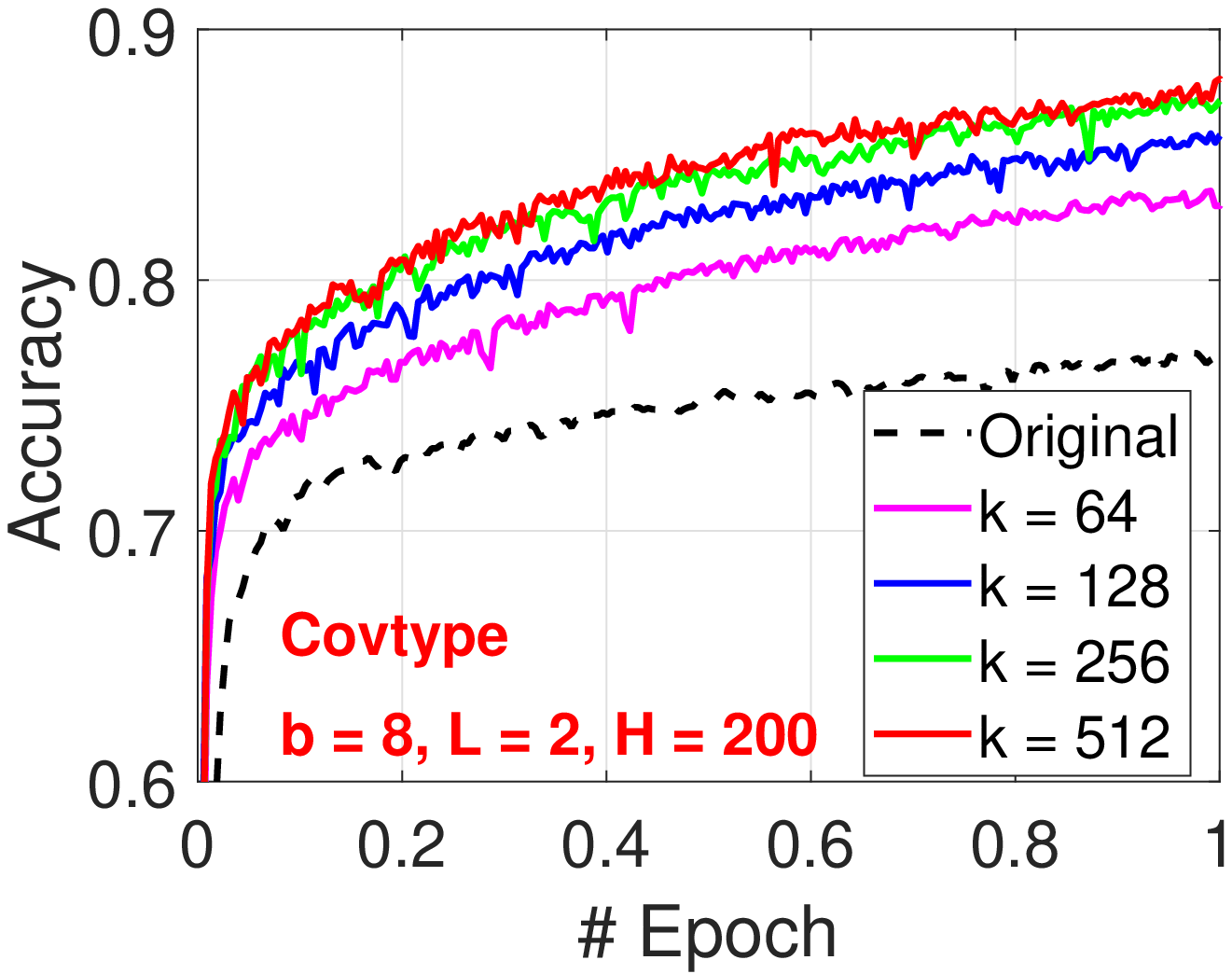}
}

\end{center}
\caption{GCWSNet with $p=1$, for $b\in\{2,4,8\}$ and $k\in\{64,128,256,512\}$, on the Covtype dataset, for just one epoch,  to better illustrate that GCWSNet converges much faster than training on the original data.}\label{fig:Covtype_p1_1ep}
\end{figure}

Next we examine the experimental results on the Covtype dataset, in Figure~\ref{fig:Covtype_p1} for 100 epochs. Again, we can see that GCWSNet converges substantially faster.  To view the convergence rate more clearly, we repeat the same plots in Figure~\ref{fig:Covtype_p1_1ep} but for just 1 epoch. As the Covtype dataset has 290,506 training samples, we cannot compute the pGMM kernel directly. LIBLINEAR reports an accuracy of $71.5\%$, which is more or less consistent with the results for $L=1$ (i.e., no hidden layer).

\vspace{0.1in}

The experiments on the Covtype dataset reveal a practically important issue, if applications only allow training for one epoch, even though training more epochs might lead to noticeably better accuracies. For example, consider the plots in the right-bottom corner in both  Figure~\ref{fig:Covtype_p1} and Figure~\ref{fig:Covtype_p1_1ep}, for $b=8$, $L=2$,  $H=200$. For the original data, if we stop after one epoch, the accuracy would drop from about $89\%$ to $77\%$, corresponding to a $13.5\%$ drop of accuracy. However, for GCWSNet with $k=64$, if the training stops after one epoch, the accuracy  would drop from $91\%$ to just $84\%$ (i.e., a $7.7\%$ drop). This again confirms the benefits of GCWSNet in the scenario of one-epoch training.

\newpage\clearpage

\begin{figure}[h]
\begin{center}
\mbox{
\includegraphics[width=2.2in]{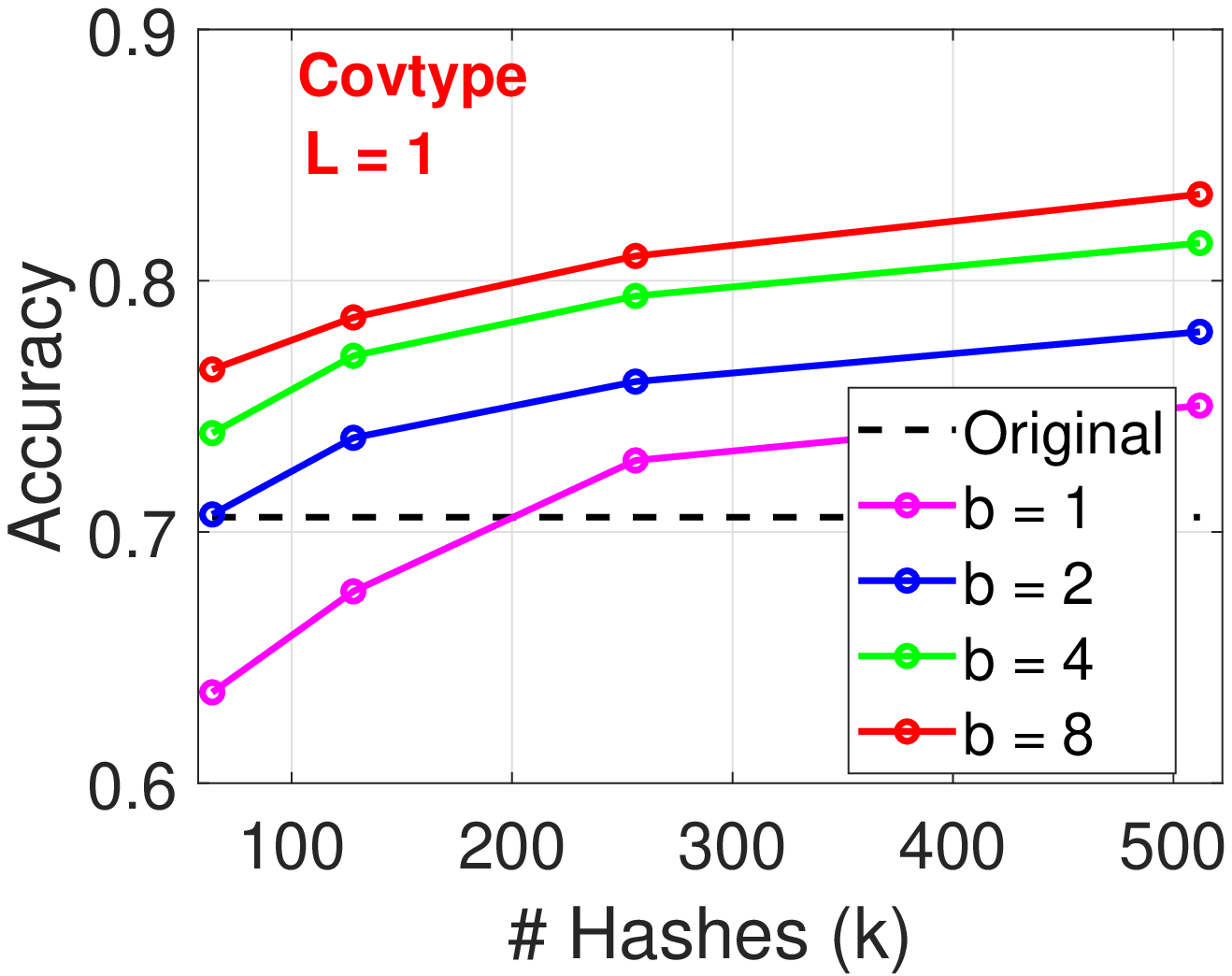}
\includegraphics[width=2.2in]{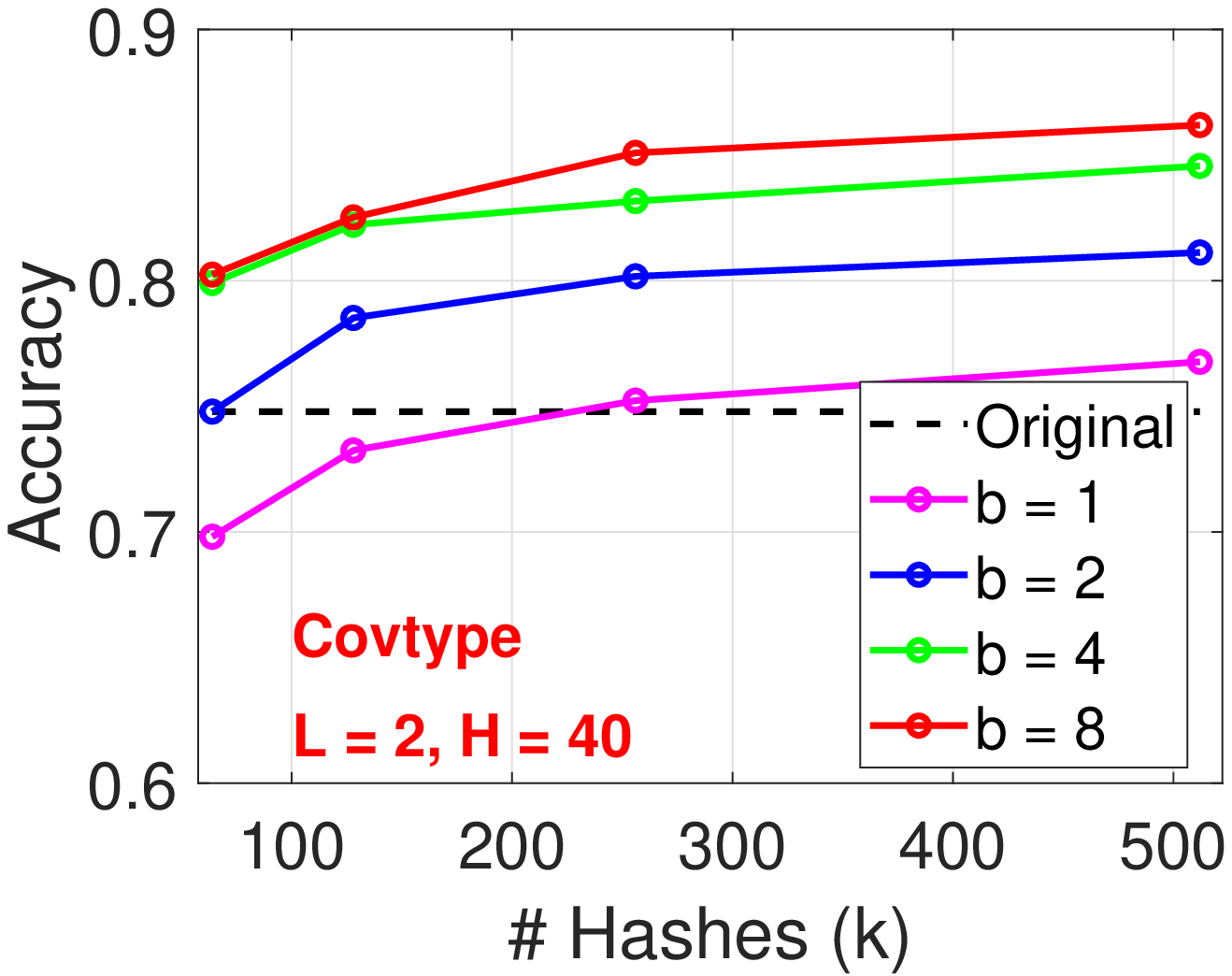}
\includegraphics[width=2.2in]{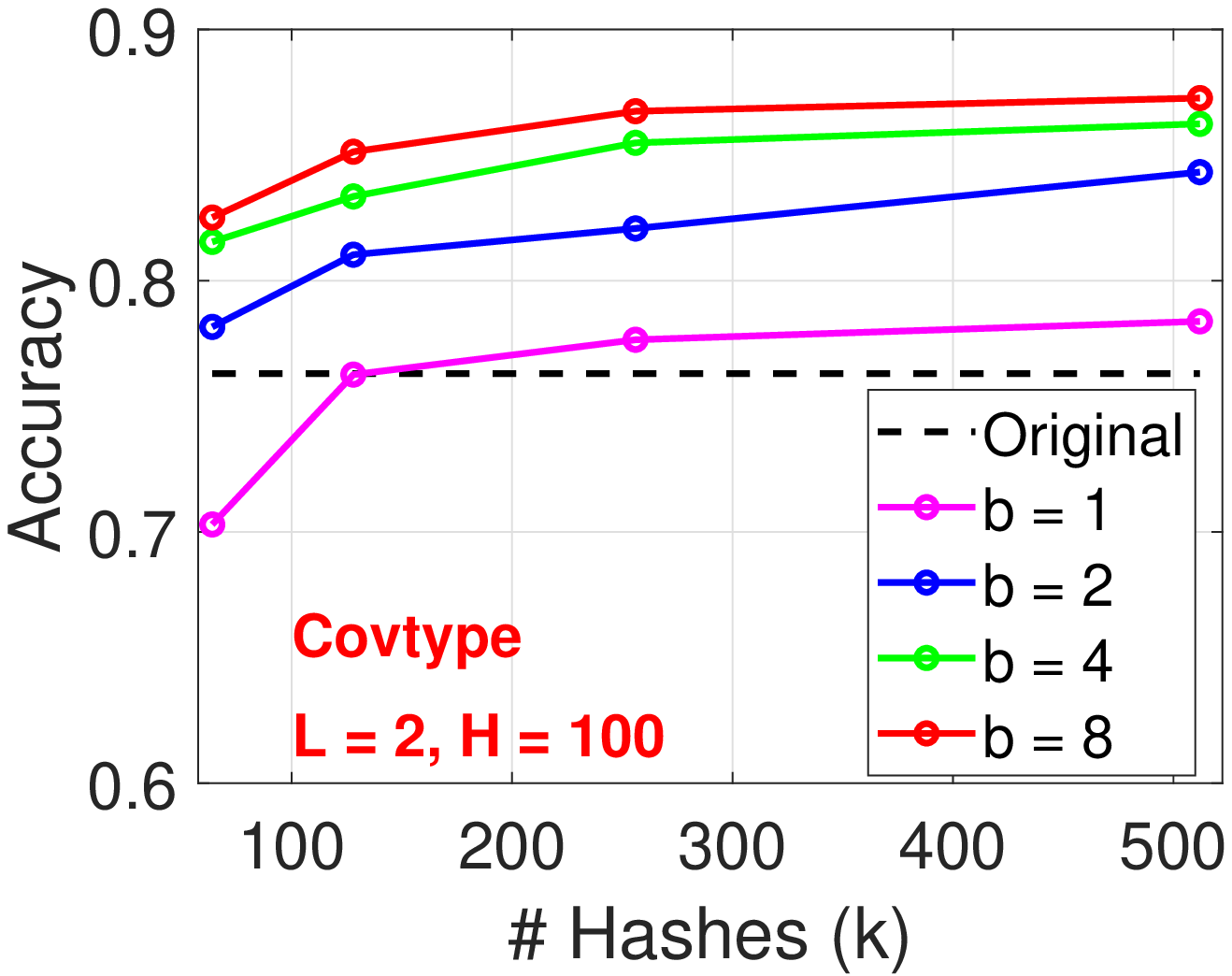}
}

\mbox{
\includegraphics[width=2.2in]{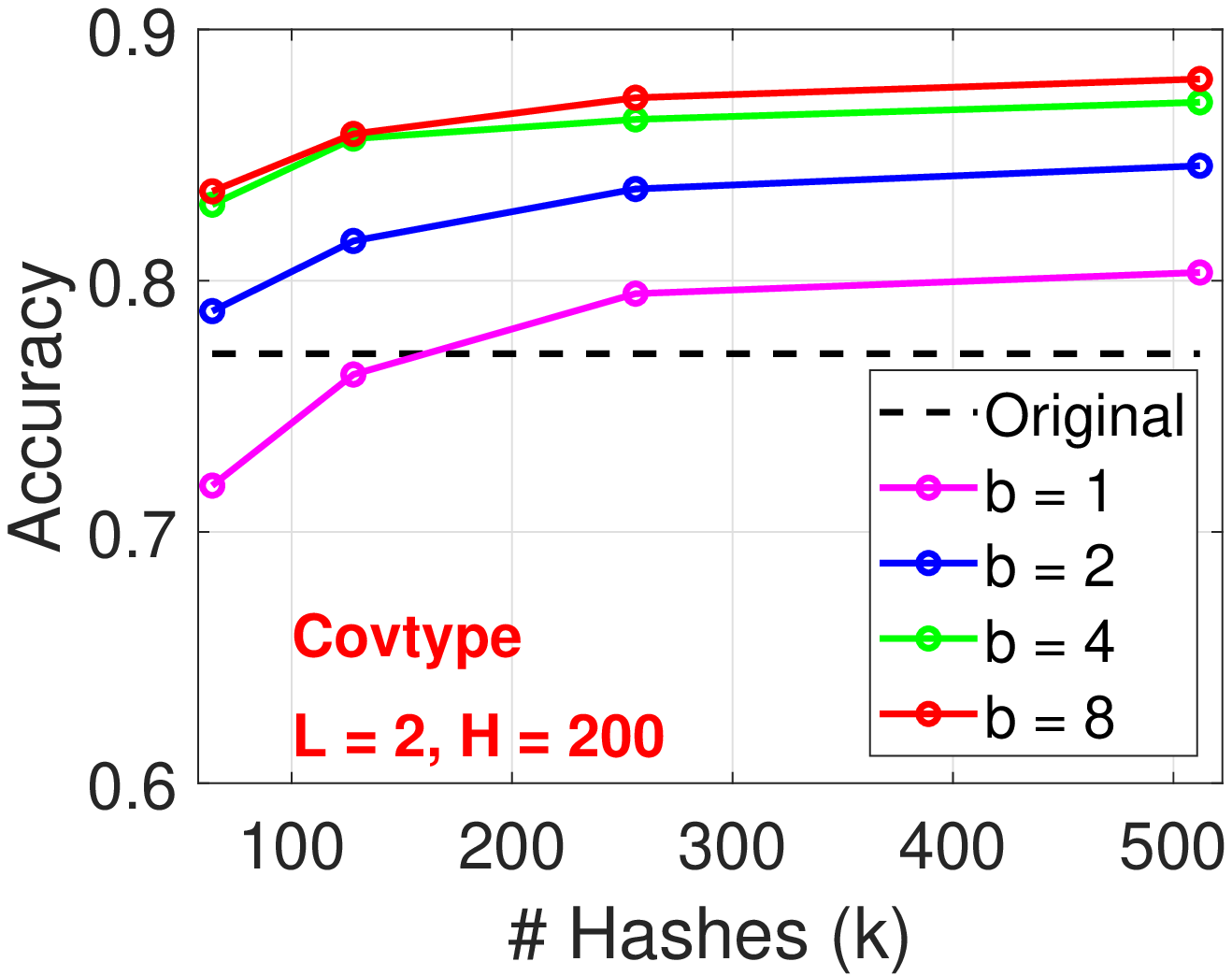}
\includegraphics[width=2.2in]{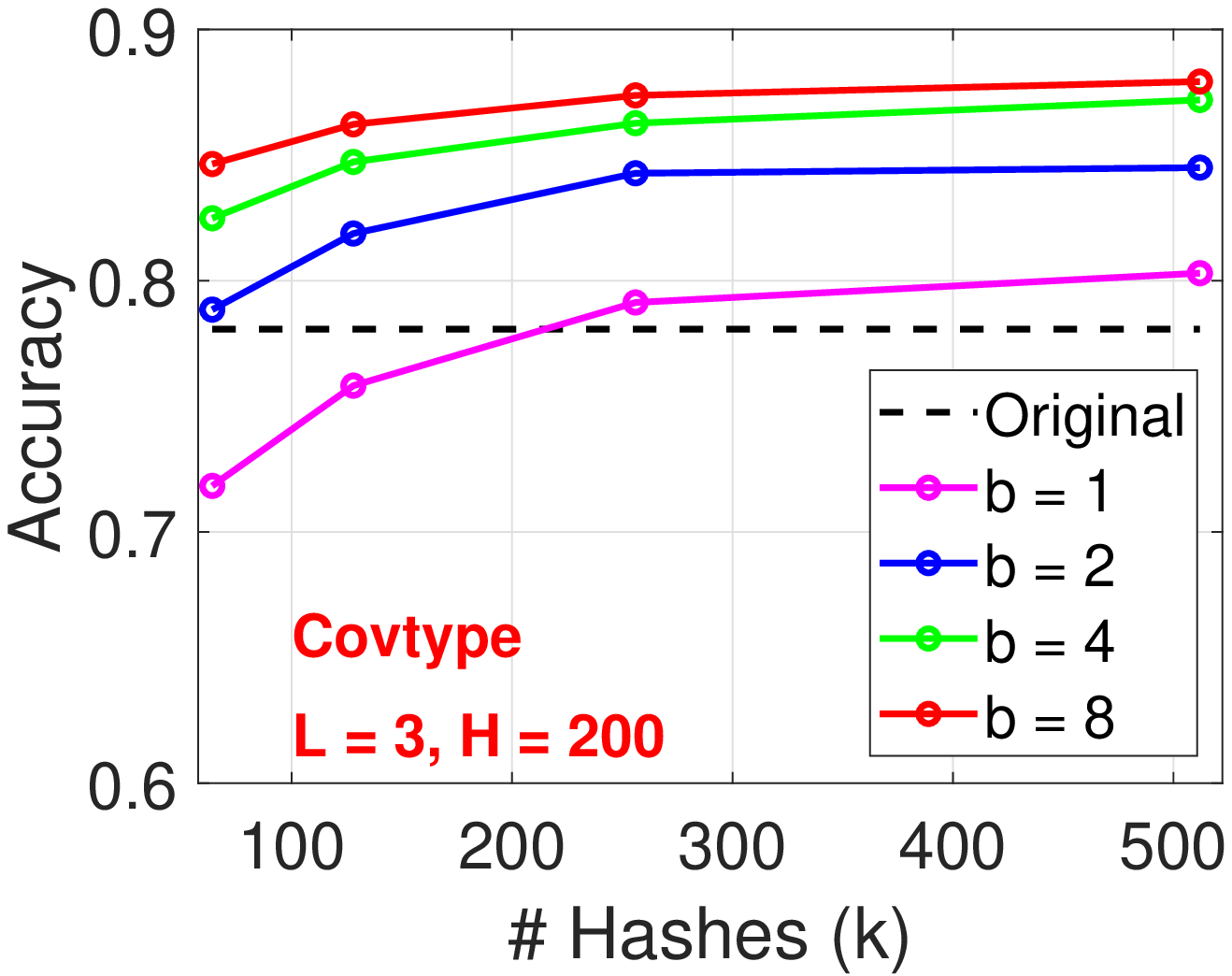}
\includegraphics[width=2.2in]{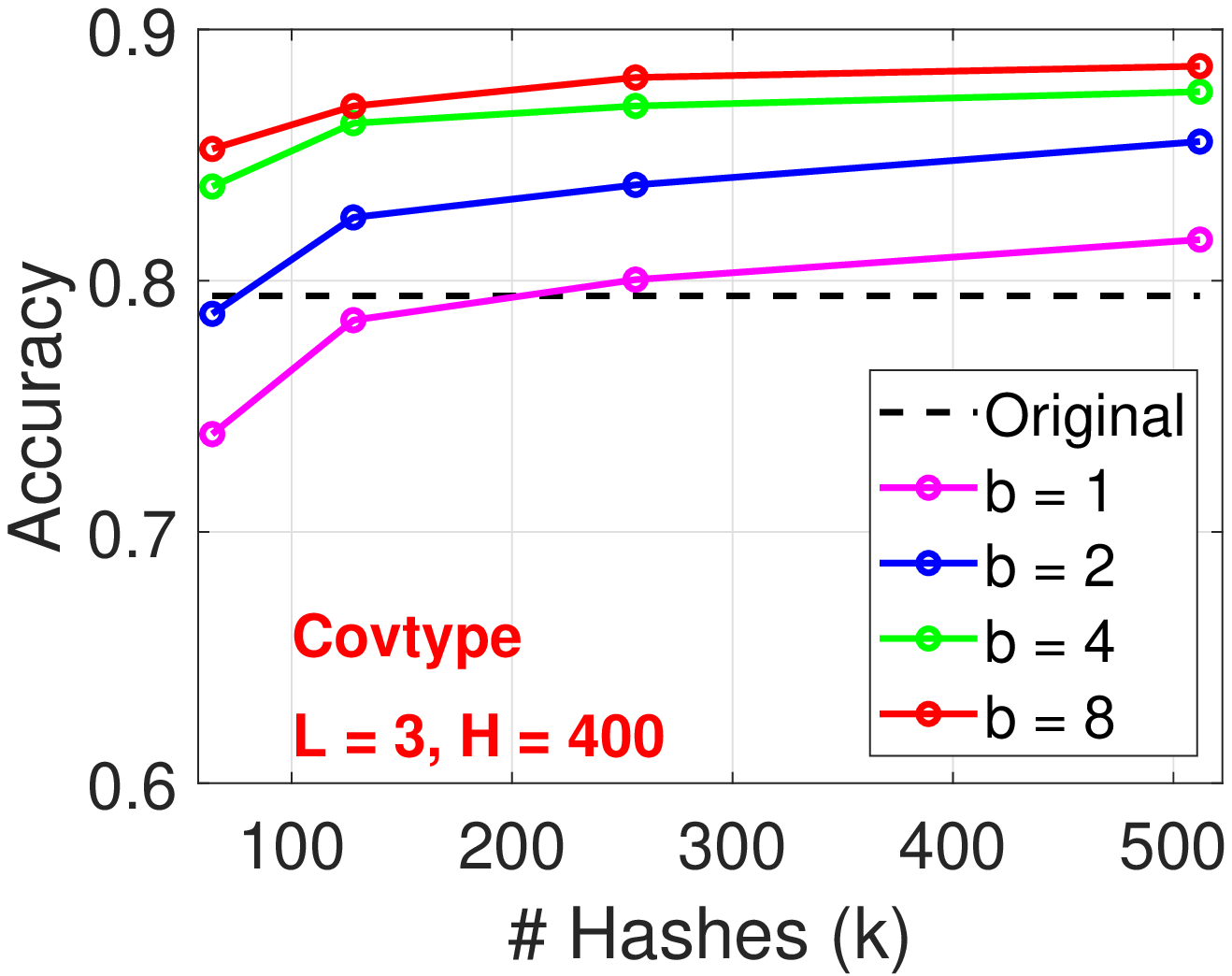}
}

\end{center}
\caption{Summary results of GCWSNet with $p=1$ on the Covtype dataset, at the end of the first epoch. }\label{fig:Covtype_p1_1ep_summary}\vspace{0.2in}
\end{figure}

Figure~\ref{fig:Covtype_p1_1ep_summary} reports the summary of the test accuracy for the Covtype dataset for just one epoch, to better illustrate the impact of $b$, $H$, $K$, as well as $L$. 

\vspace{0.1in}

\begin{figure}[h]

\begin{center}

\mbox{
\includegraphics[width=2.2in]{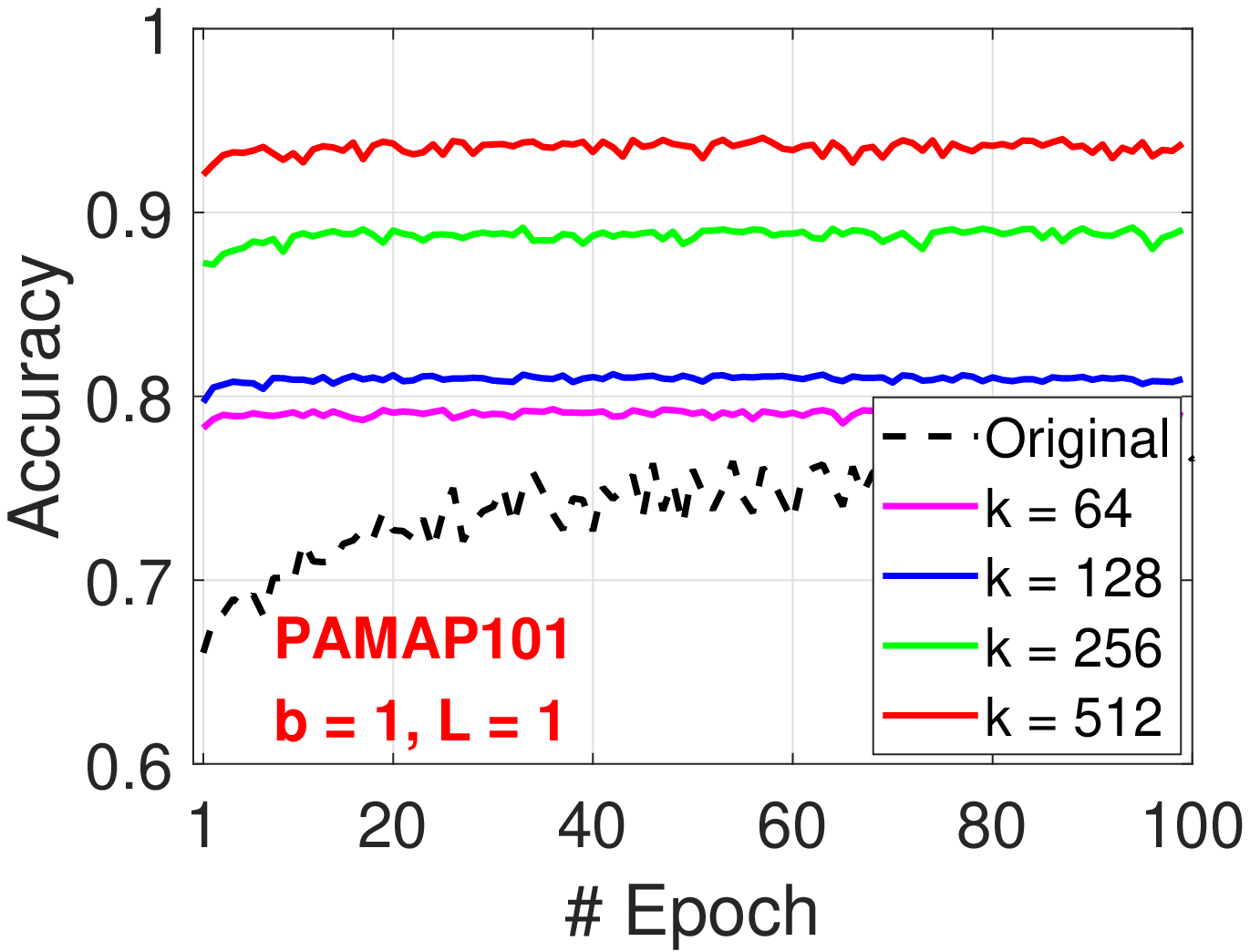}
\includegraphics[width=2.2in]{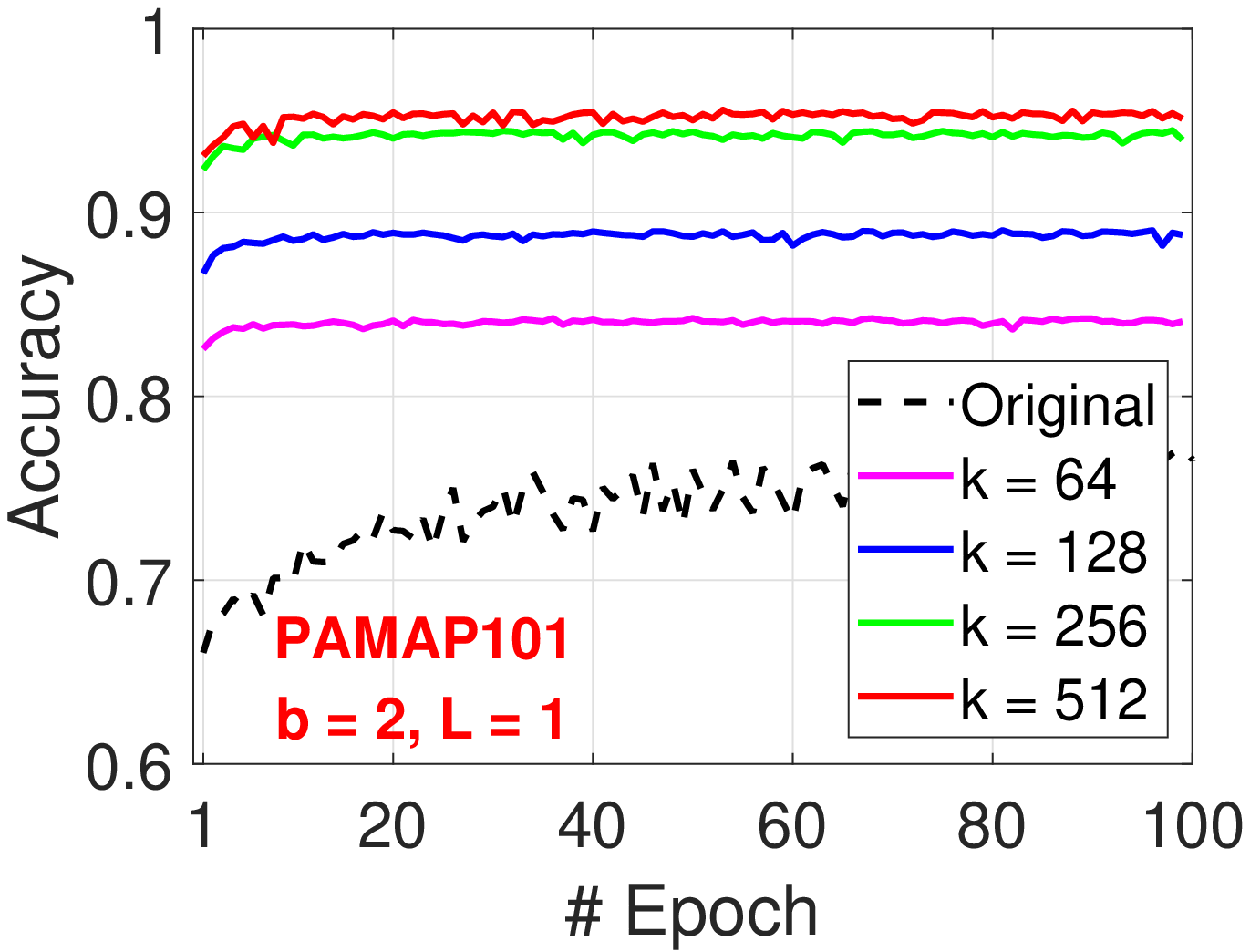}
\includegraphics[width=2.2in]{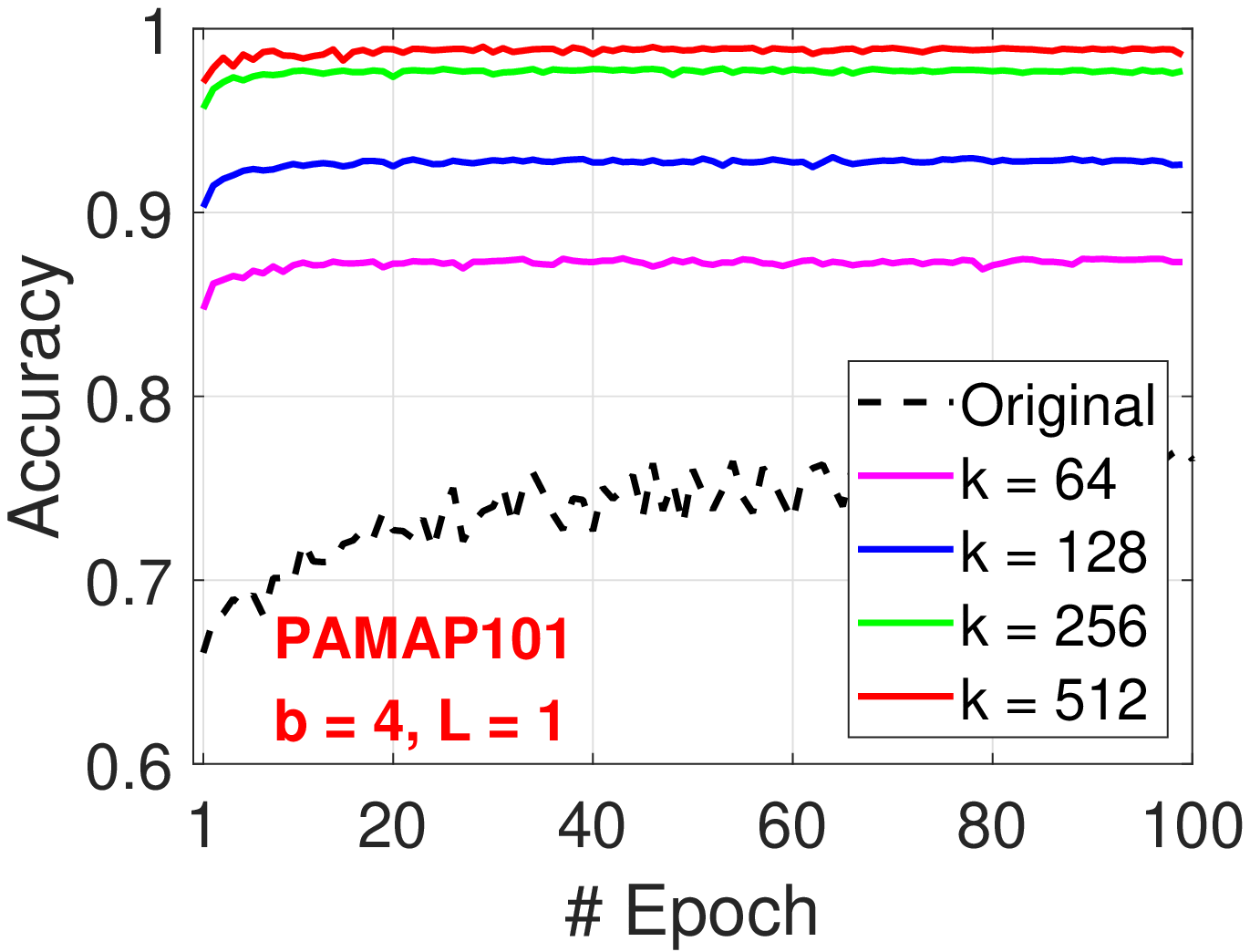}
}

\mbox{
\includegraphics[width=2.2in]{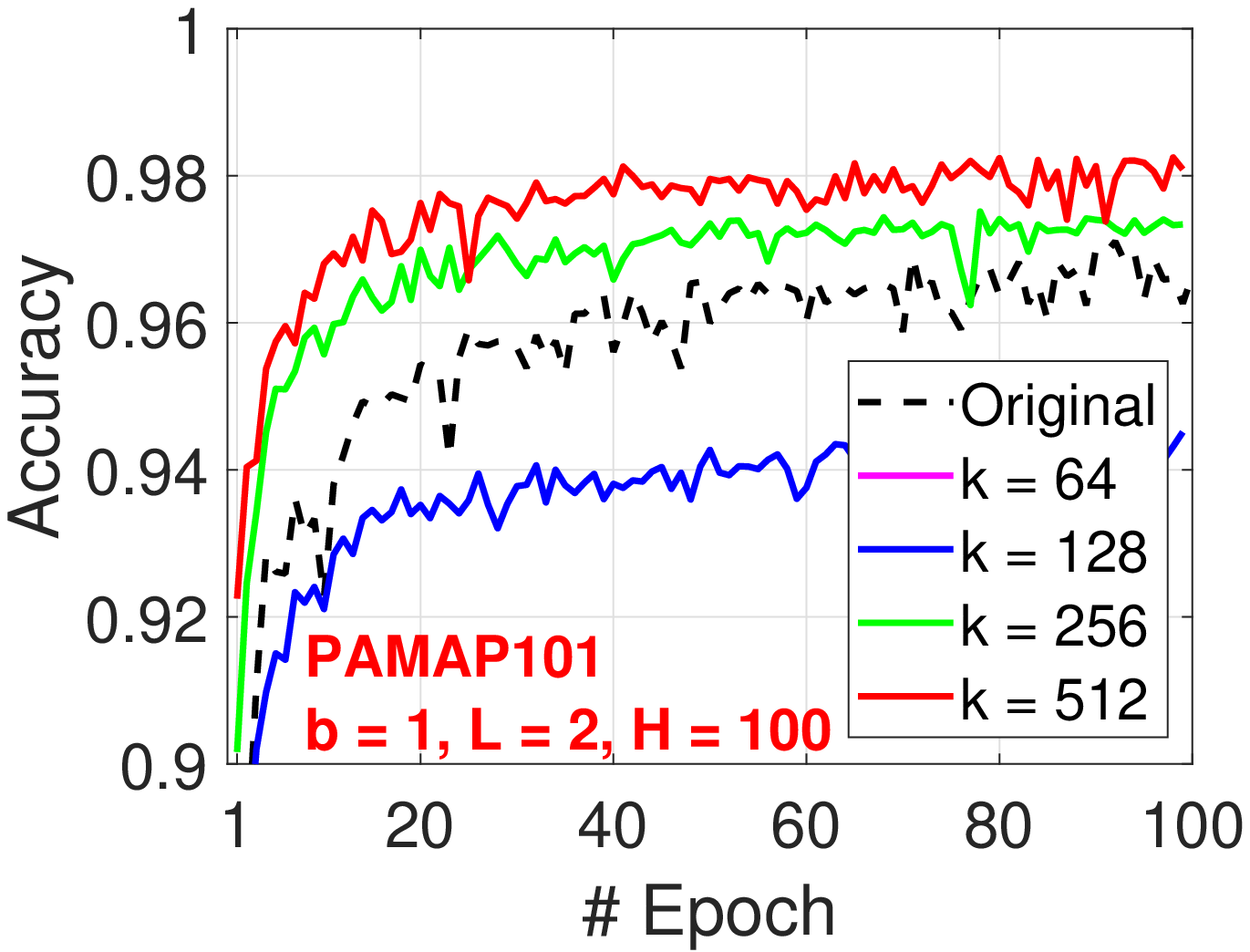}
\includegraphics[width=2.2in]{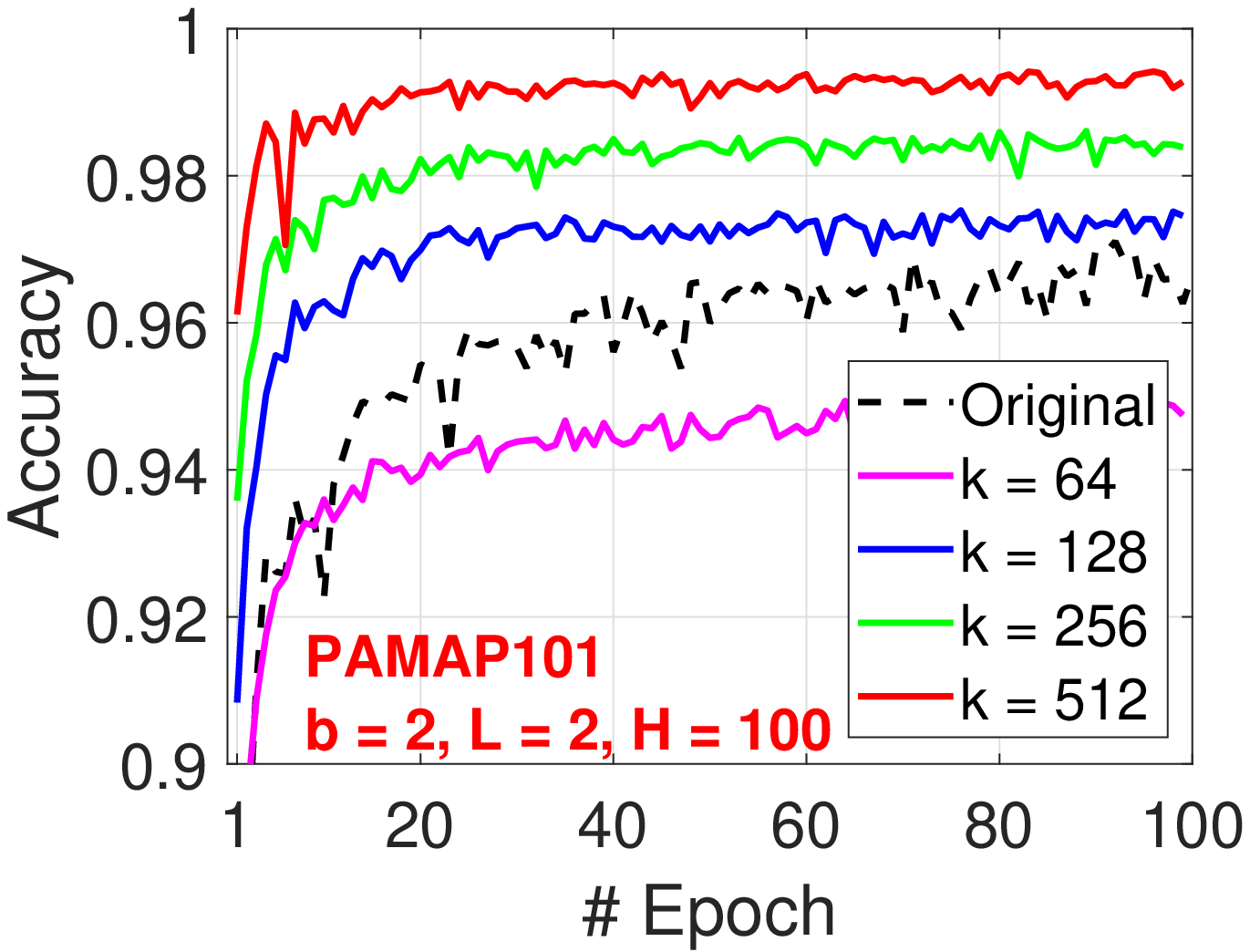}
\includegraphics[width=2.2in]{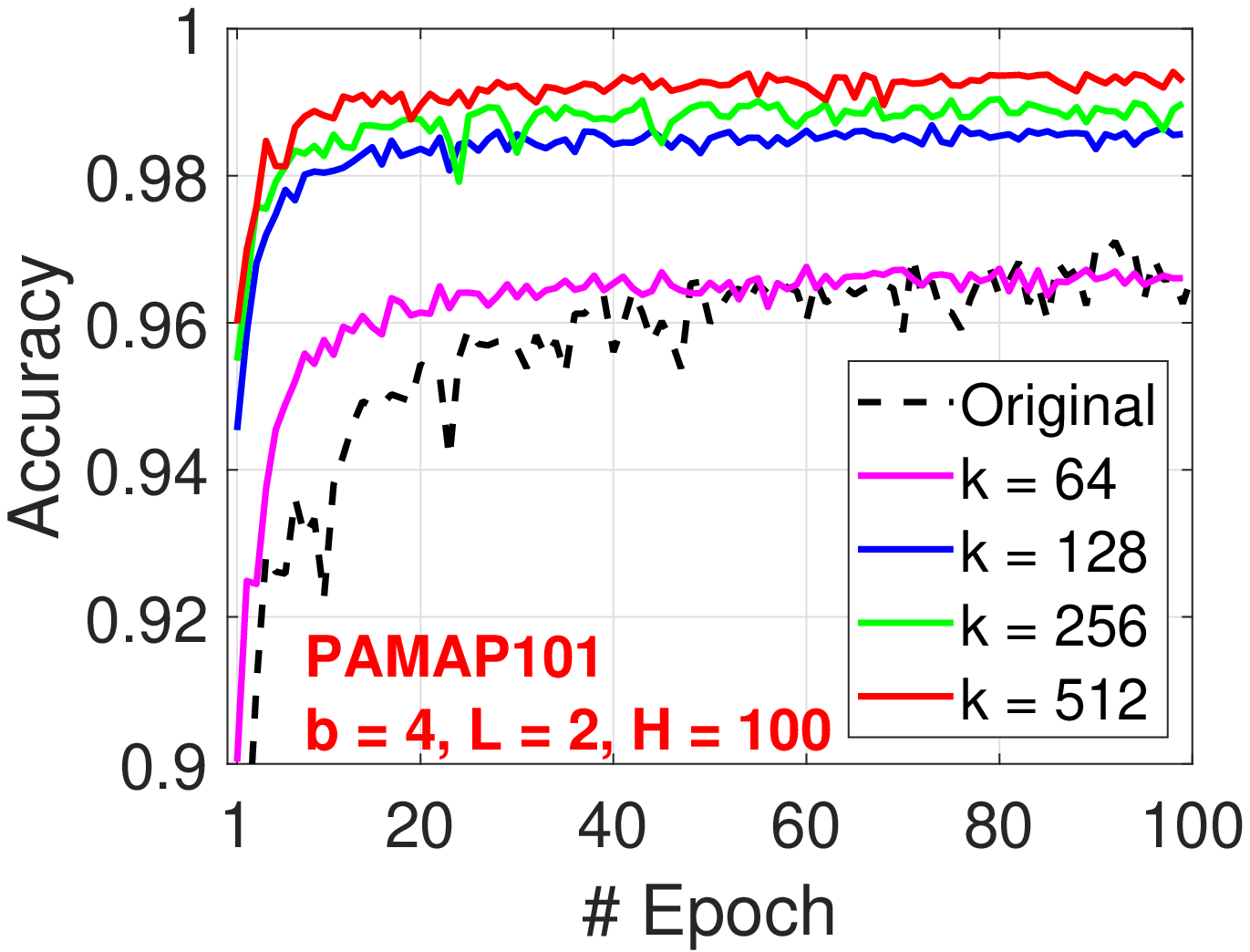}
}

\mbox{
\includegraphics[width=2.2in]{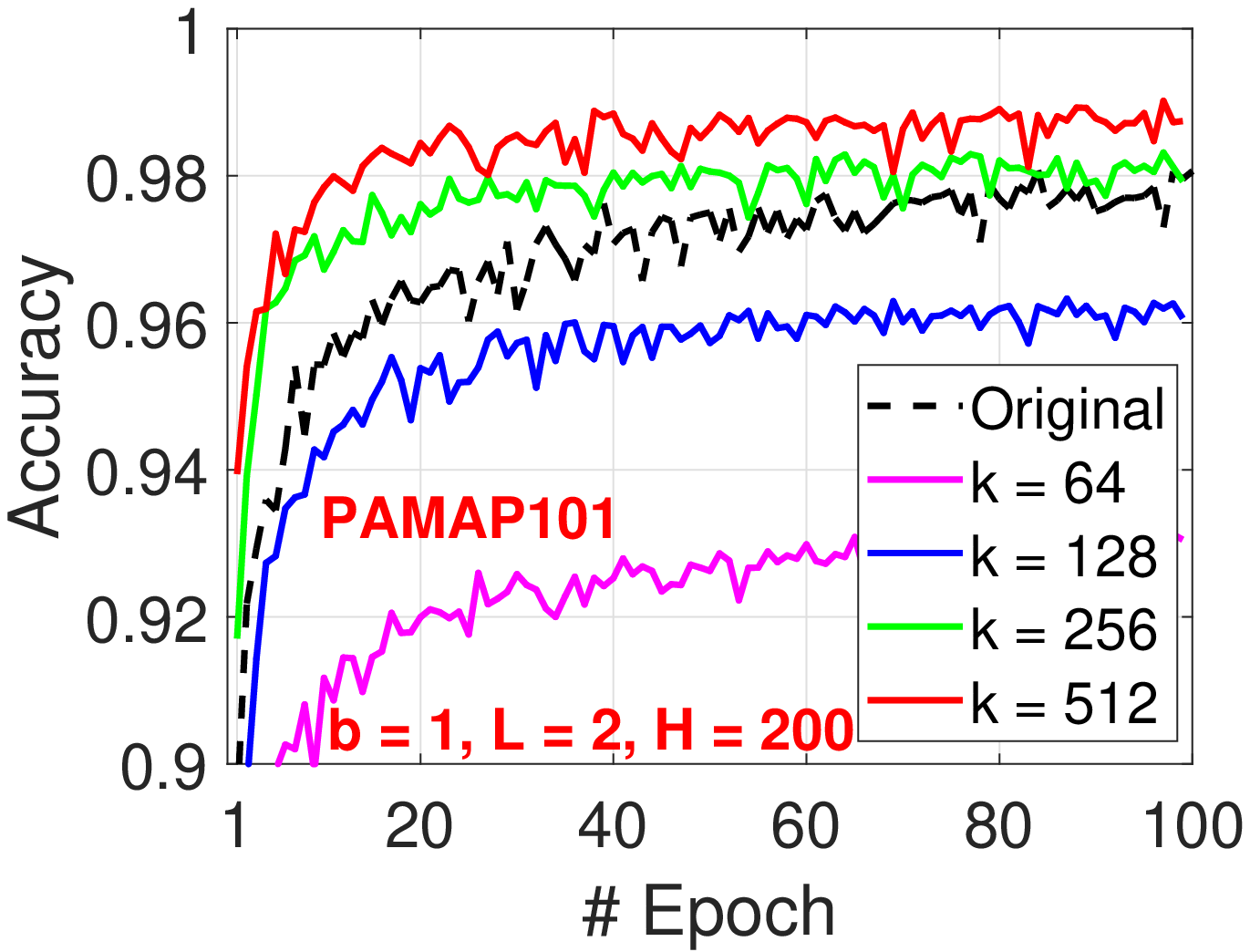}
\includegraphics[width=2.2in]{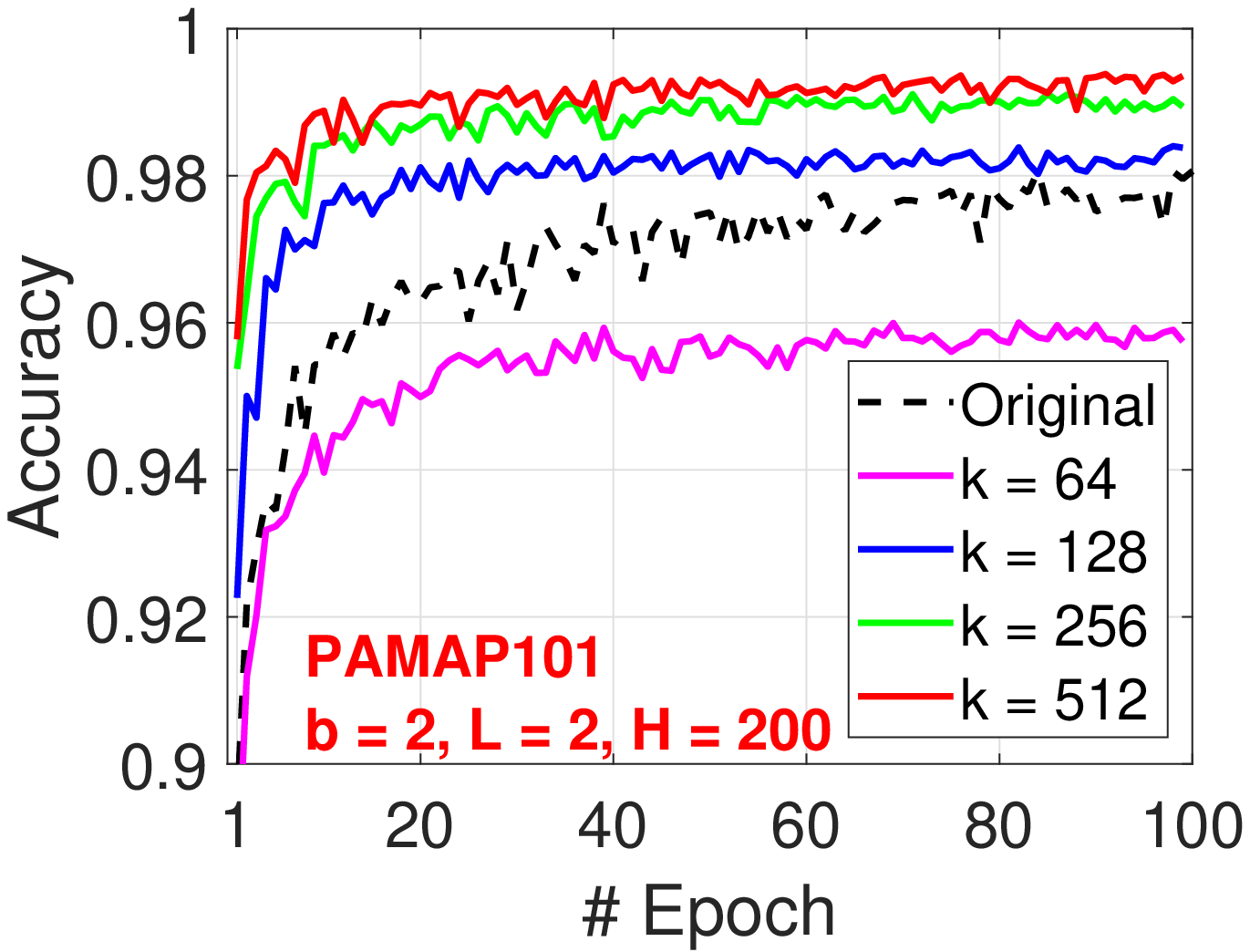}
\includegraphics[width=2.2in]{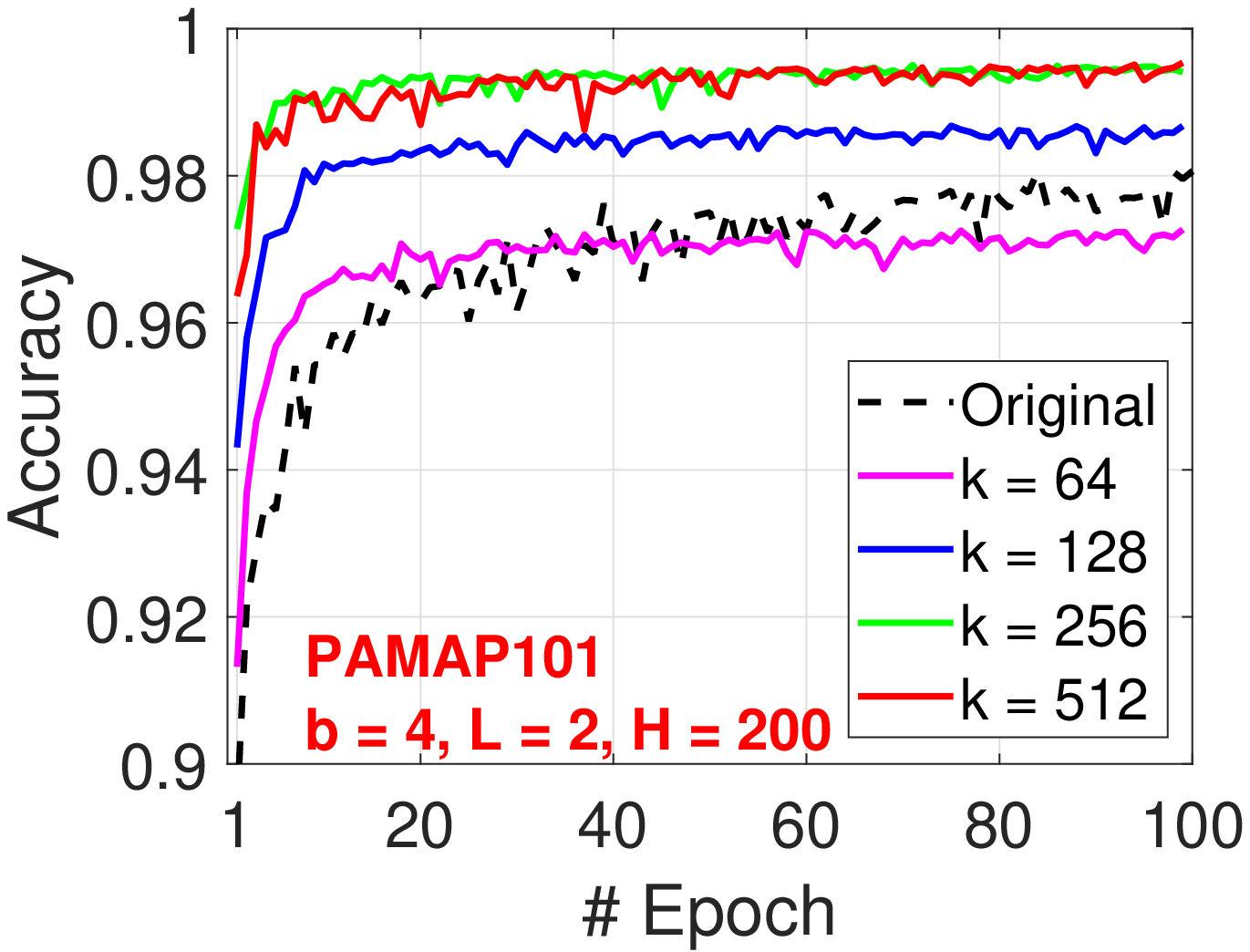}
}

\mbox{
\includegraphics[width=2.2in]{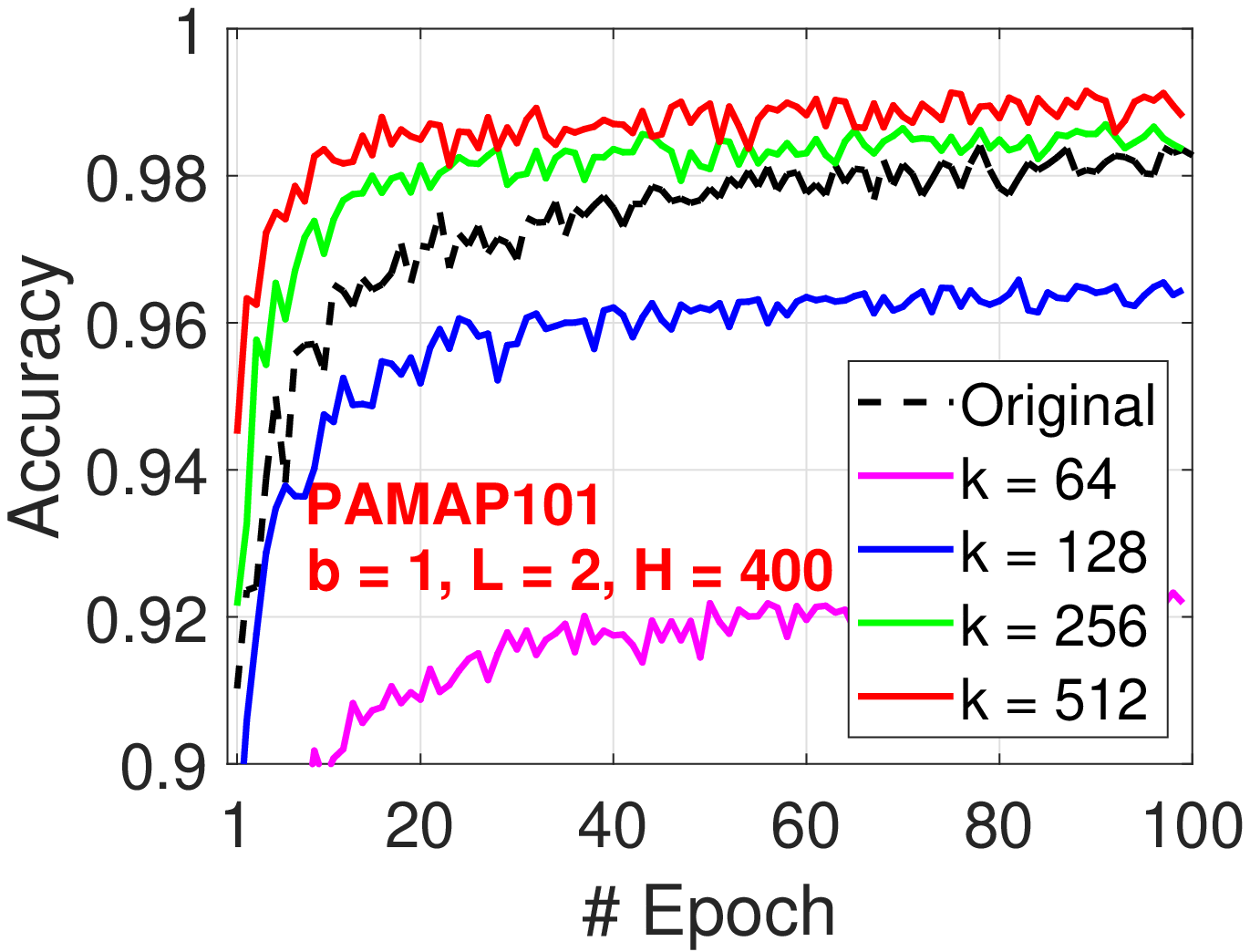}
\includegraphics[width=2.2in]{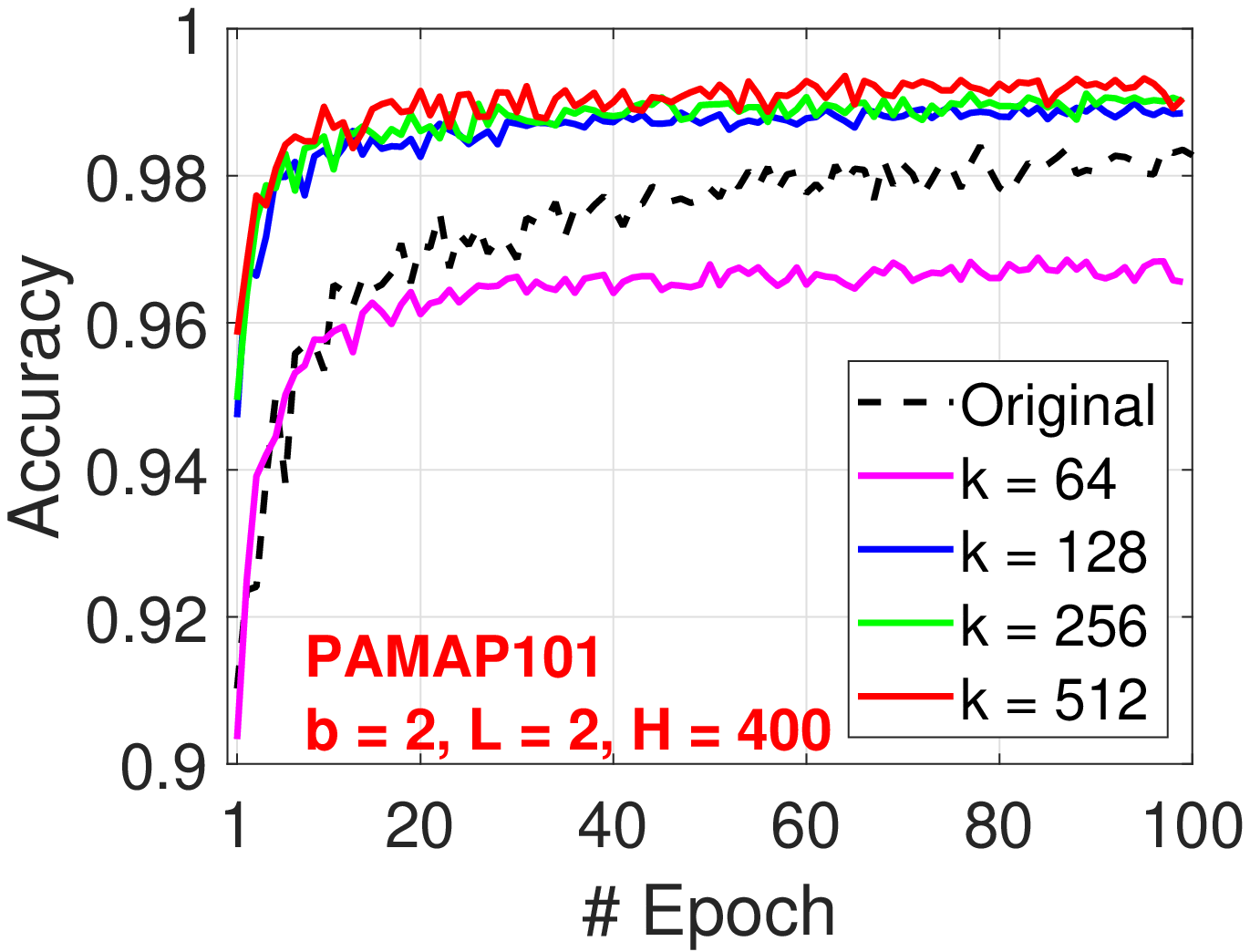}
\includegraphics[width=2.2in]{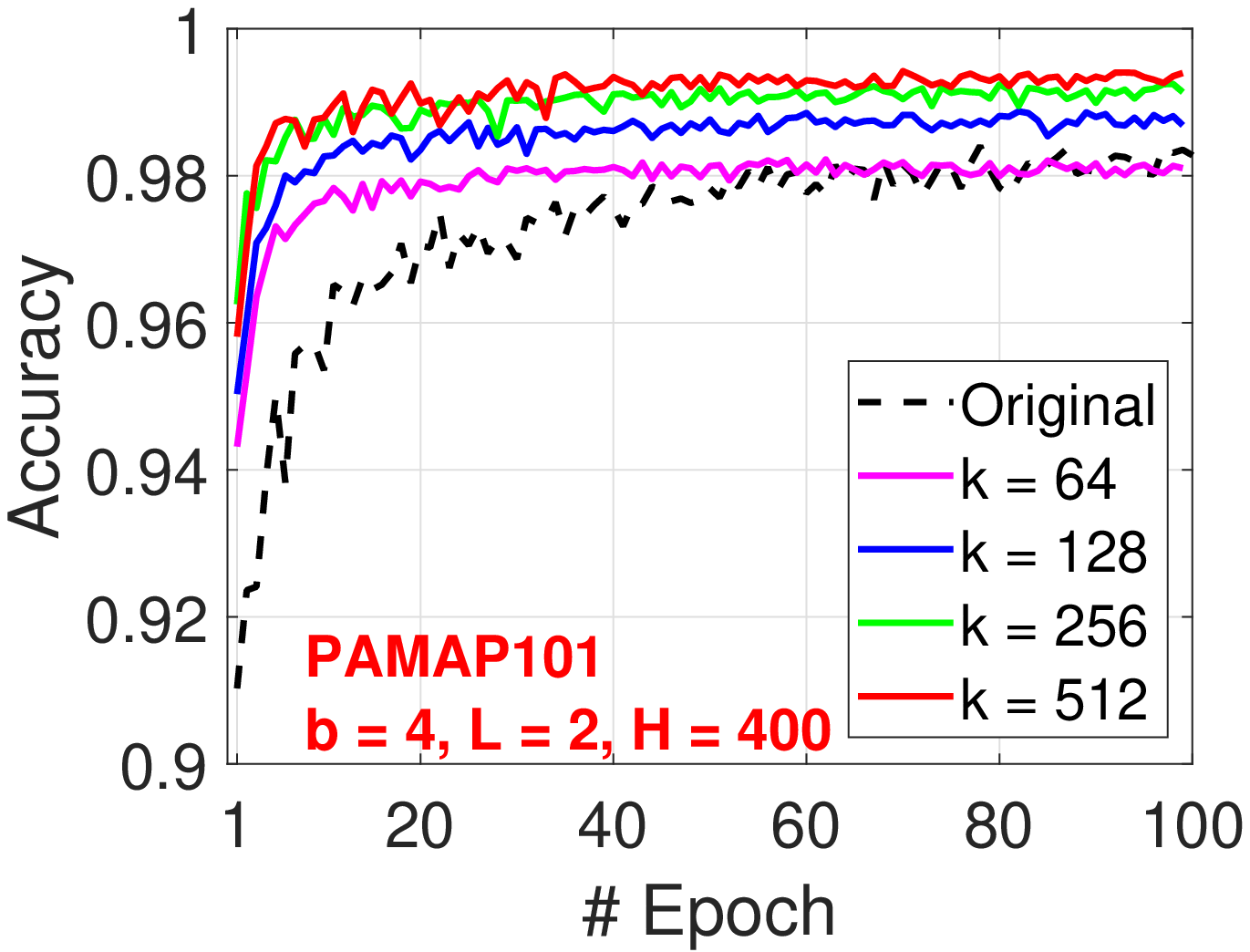}
}

\end{center}
\caption{GCWSNet with $p=1$, for $b\in\{1,2,4\}$ and $k\in\{64,128,256,512\}$, on the PAMAP101 dataset, for 100 epochs.  Again, we can see that GCWSNet converges much faster than training on the original data.}\label{fig:PAMAP101_p1}
\end{figure}

\begin{figure}[h]

\begin{center}

\mbox{
\includegraphics[width=2.2in]{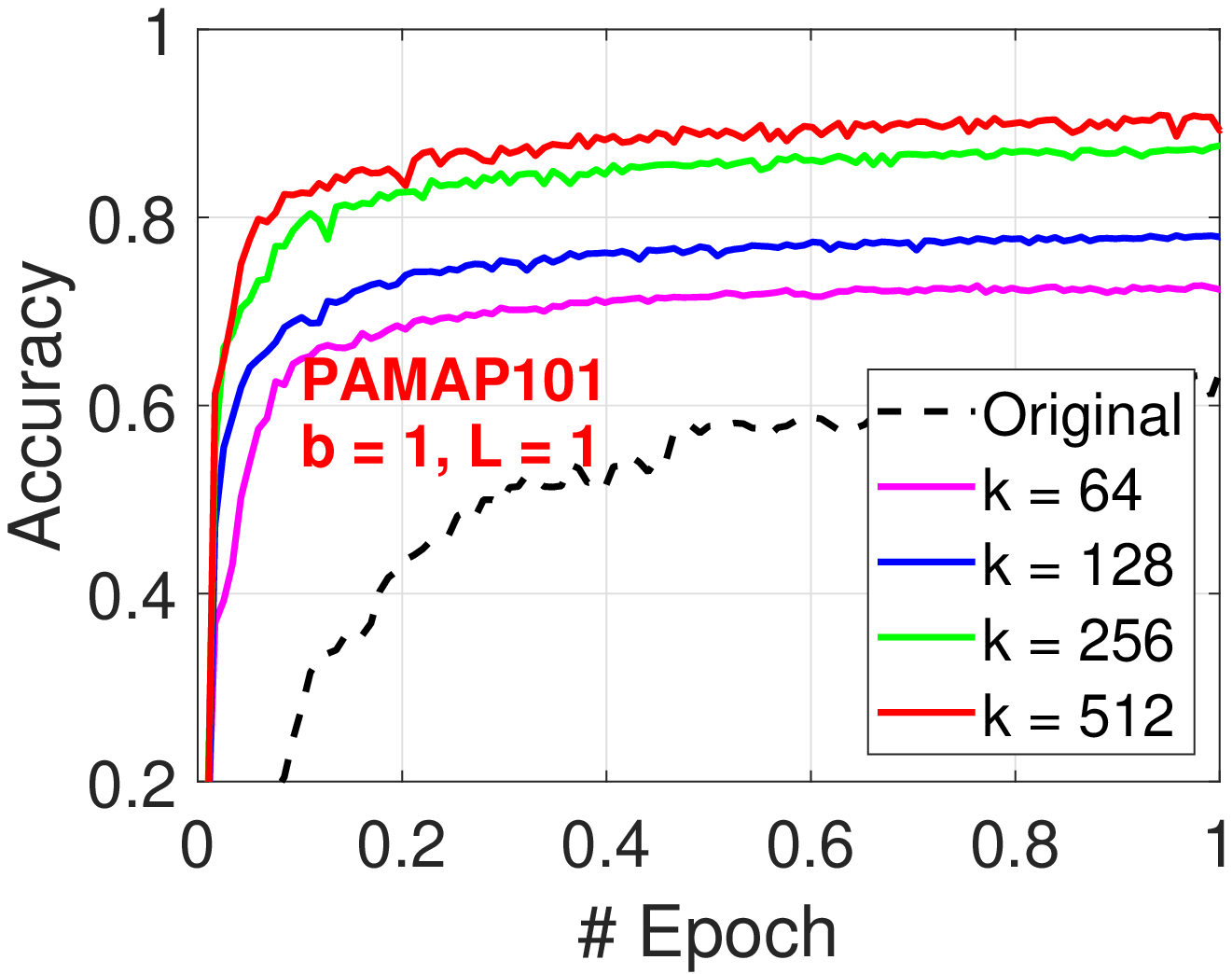}
\includegraphics[width=2.2in]{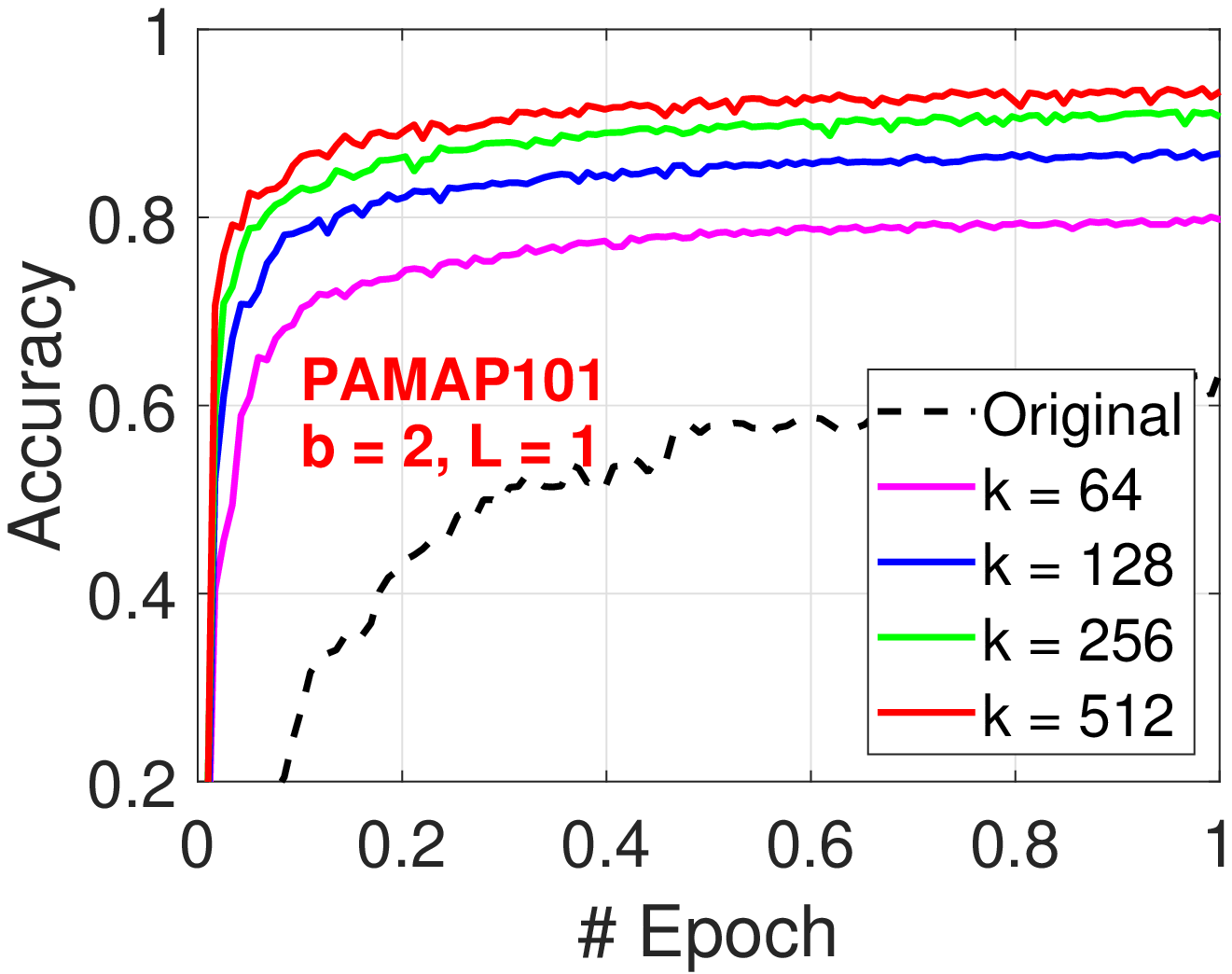}
\includegraphics[width=2.2in]{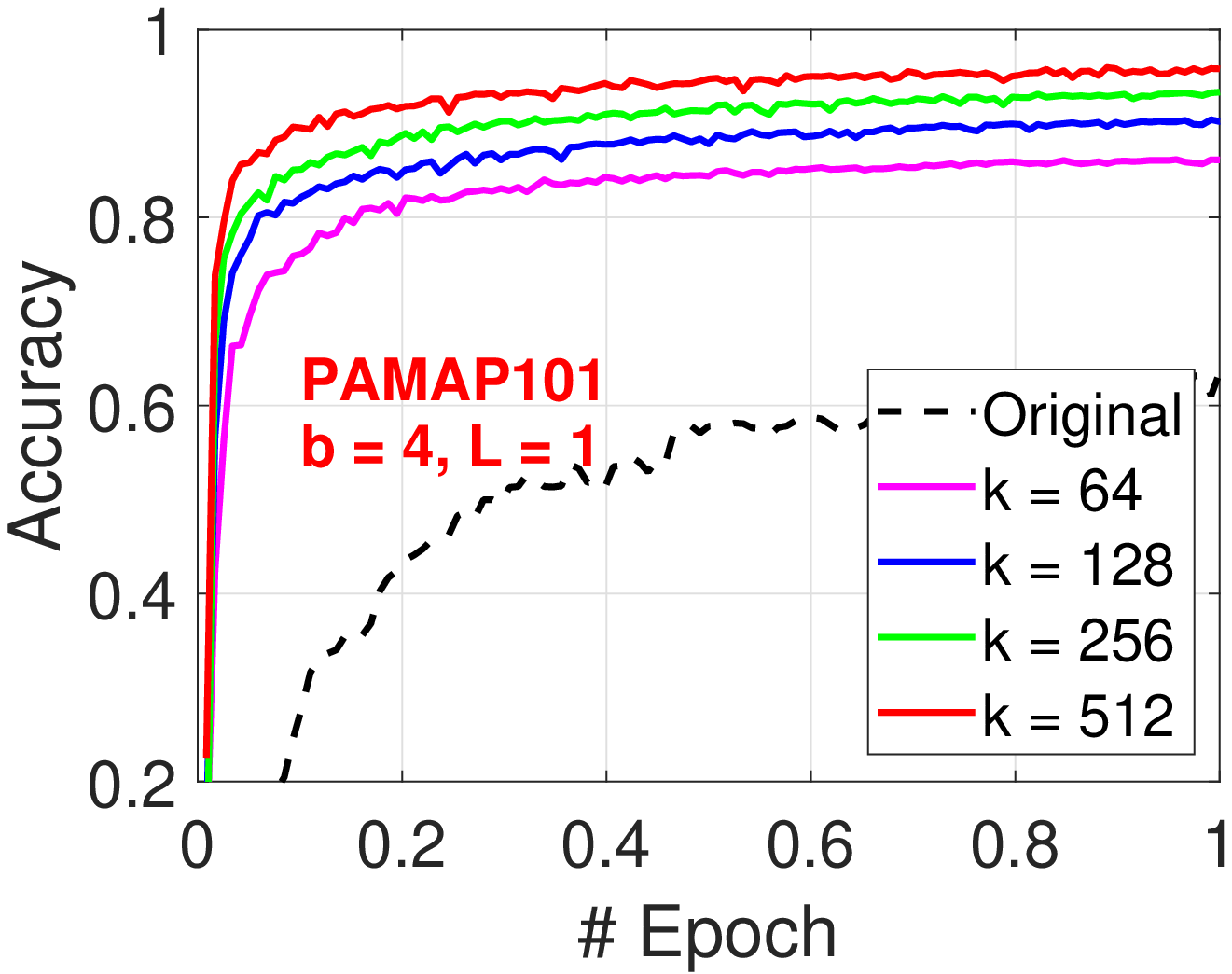}
}

\mbox{
\includegraphics[width=2.2in]{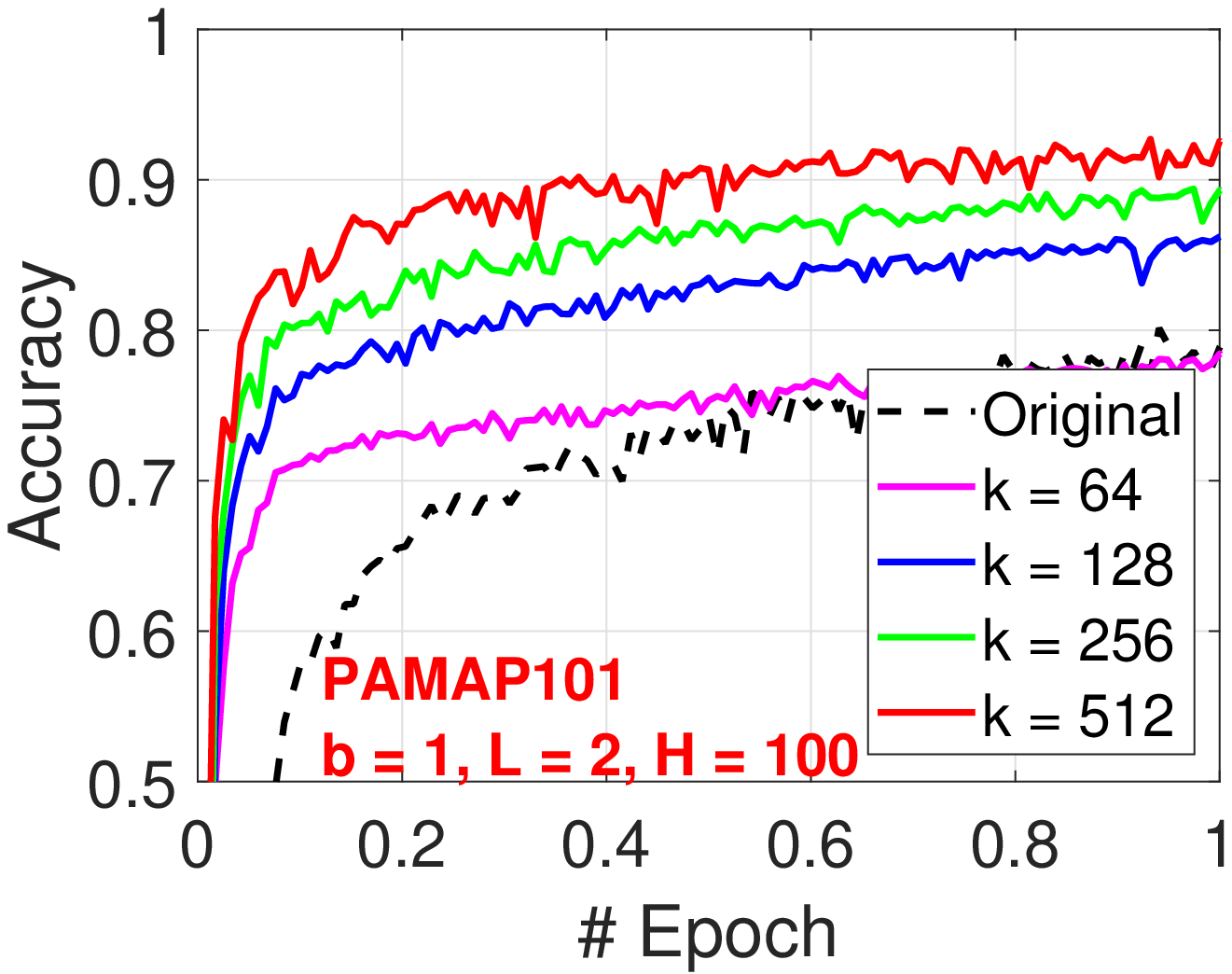}
\includegraphics[width=2.2in]{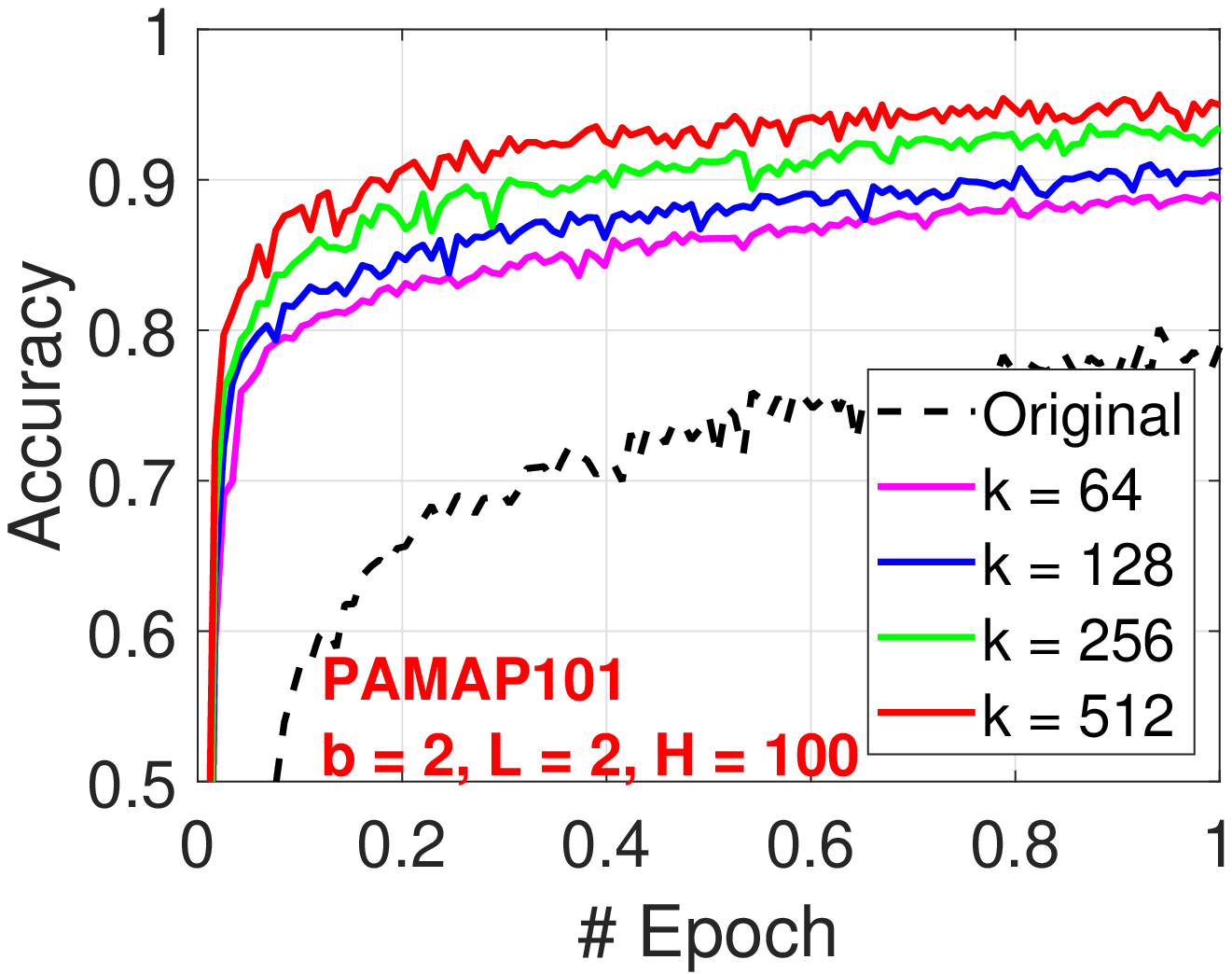}
\includegraphics[width=2.2in]{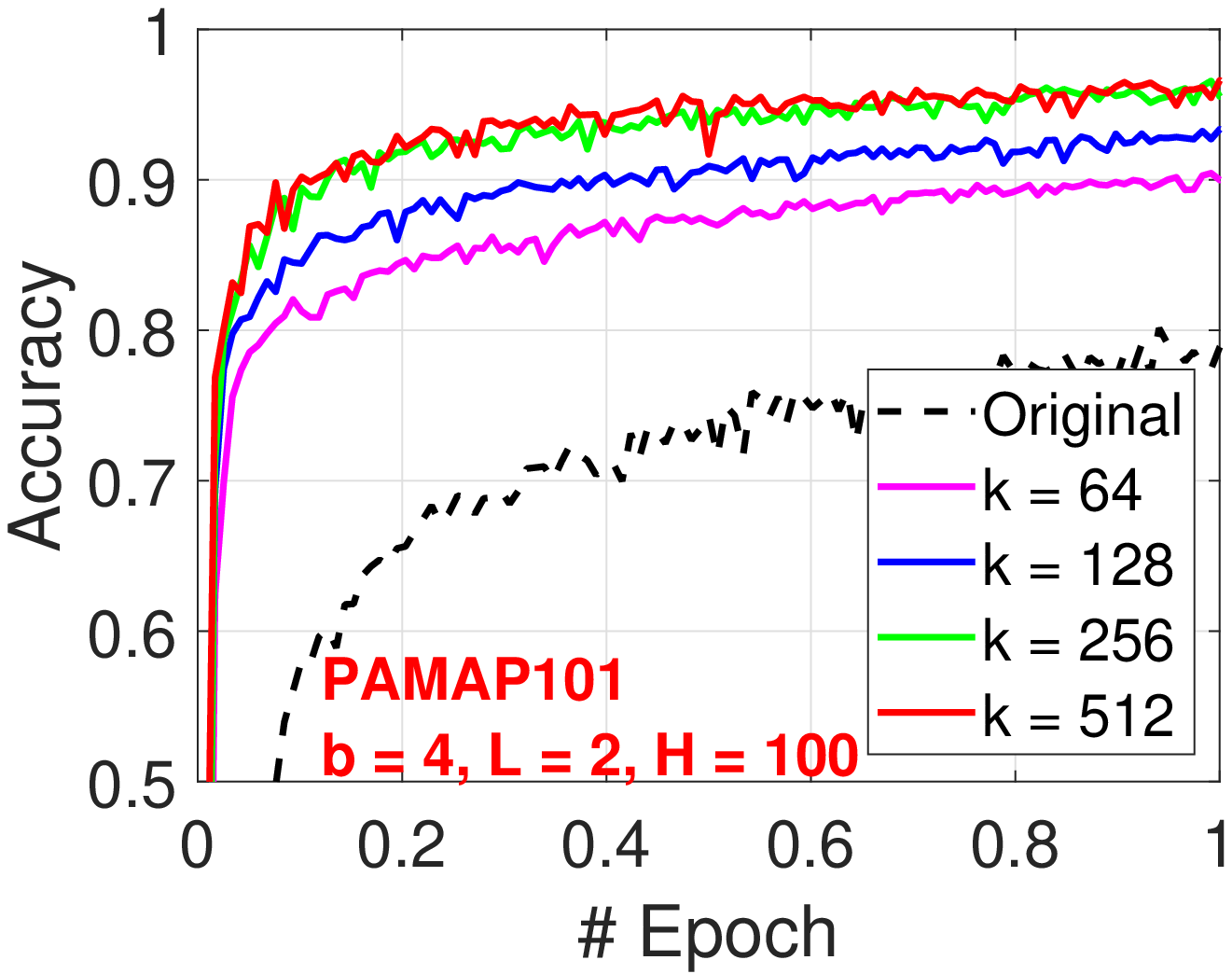}
}

\mbox{
\includegraphics[width=2.2in]{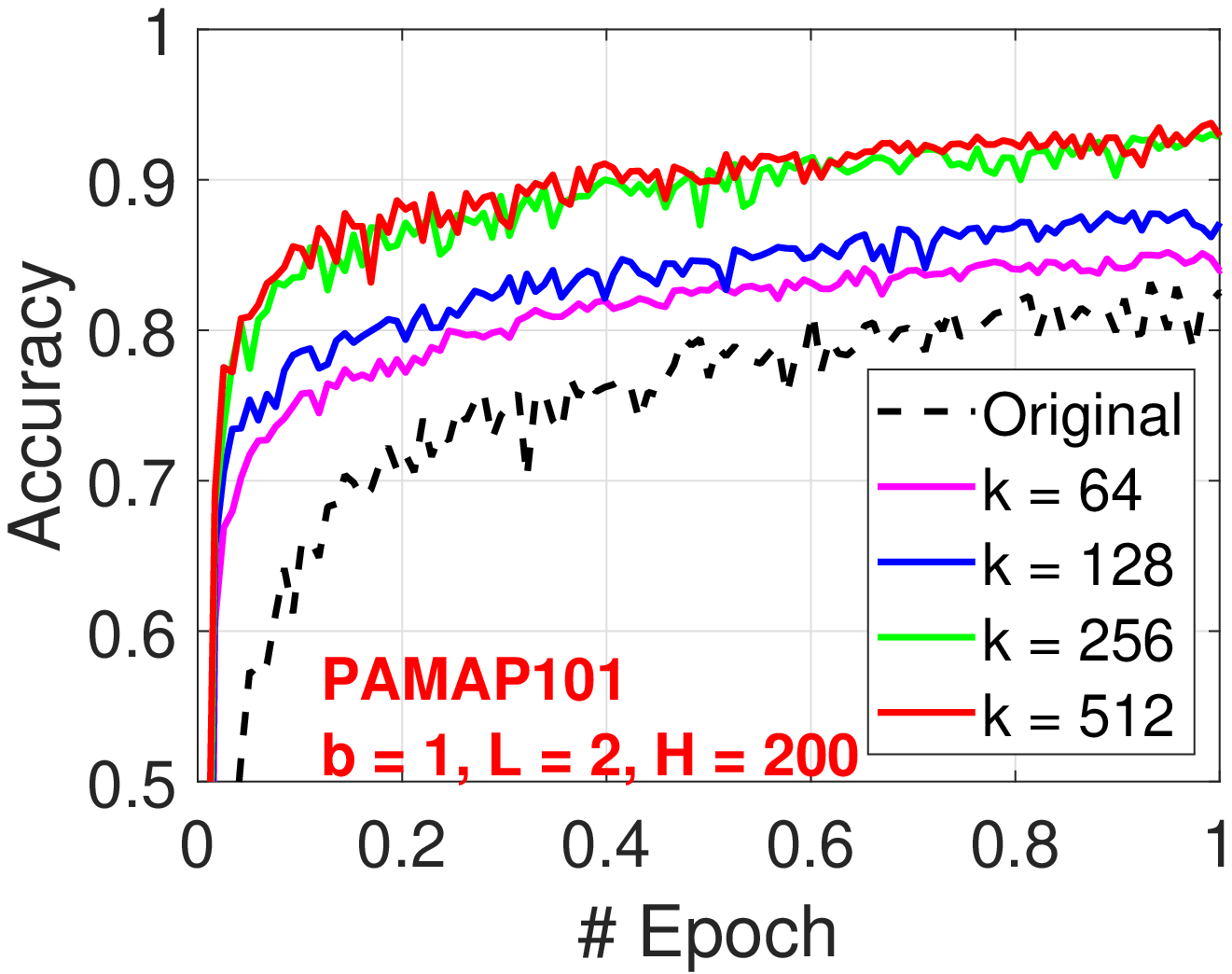}
\includegraphics[width=2.2in]{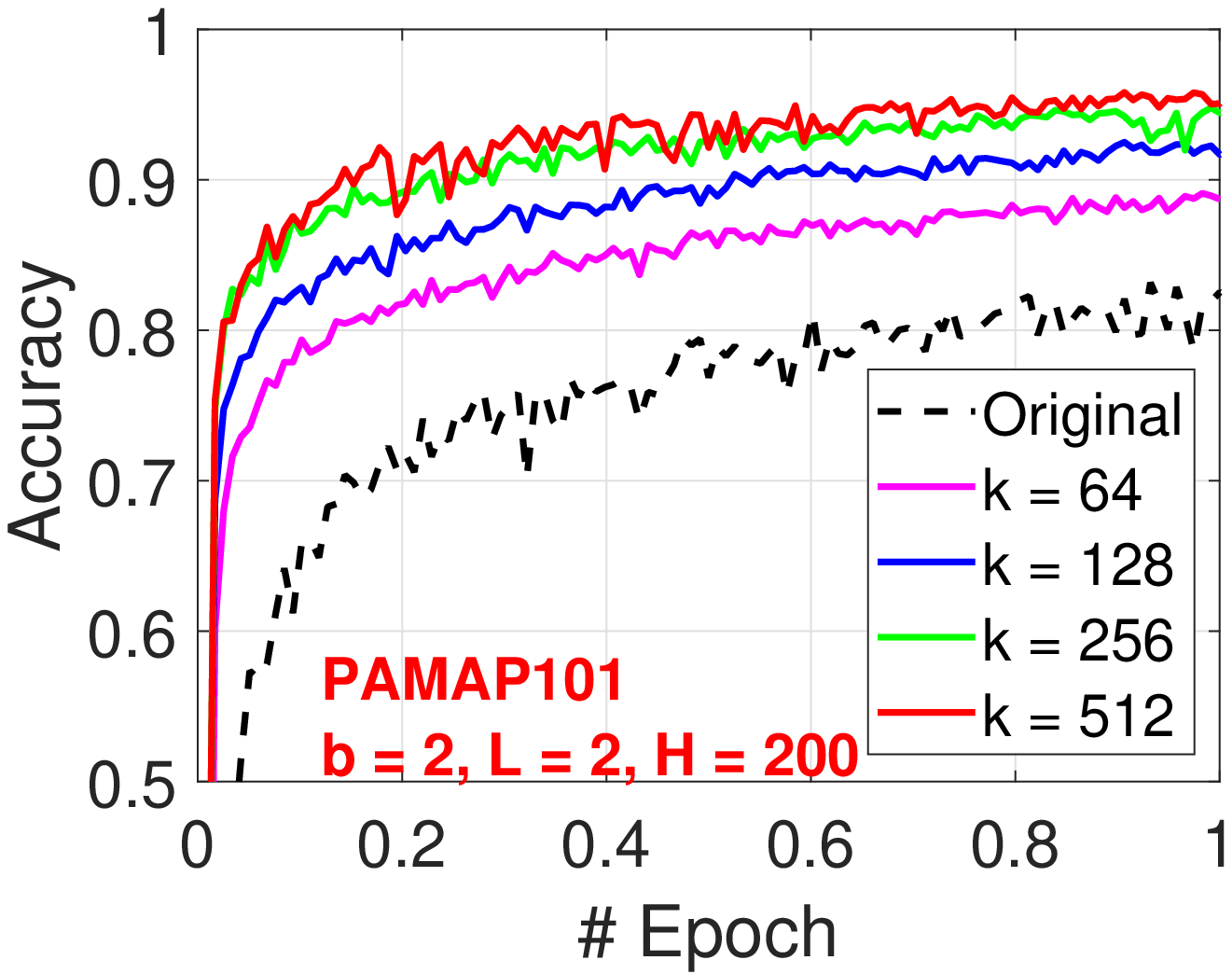}
\includegraphics[width=2.2in]{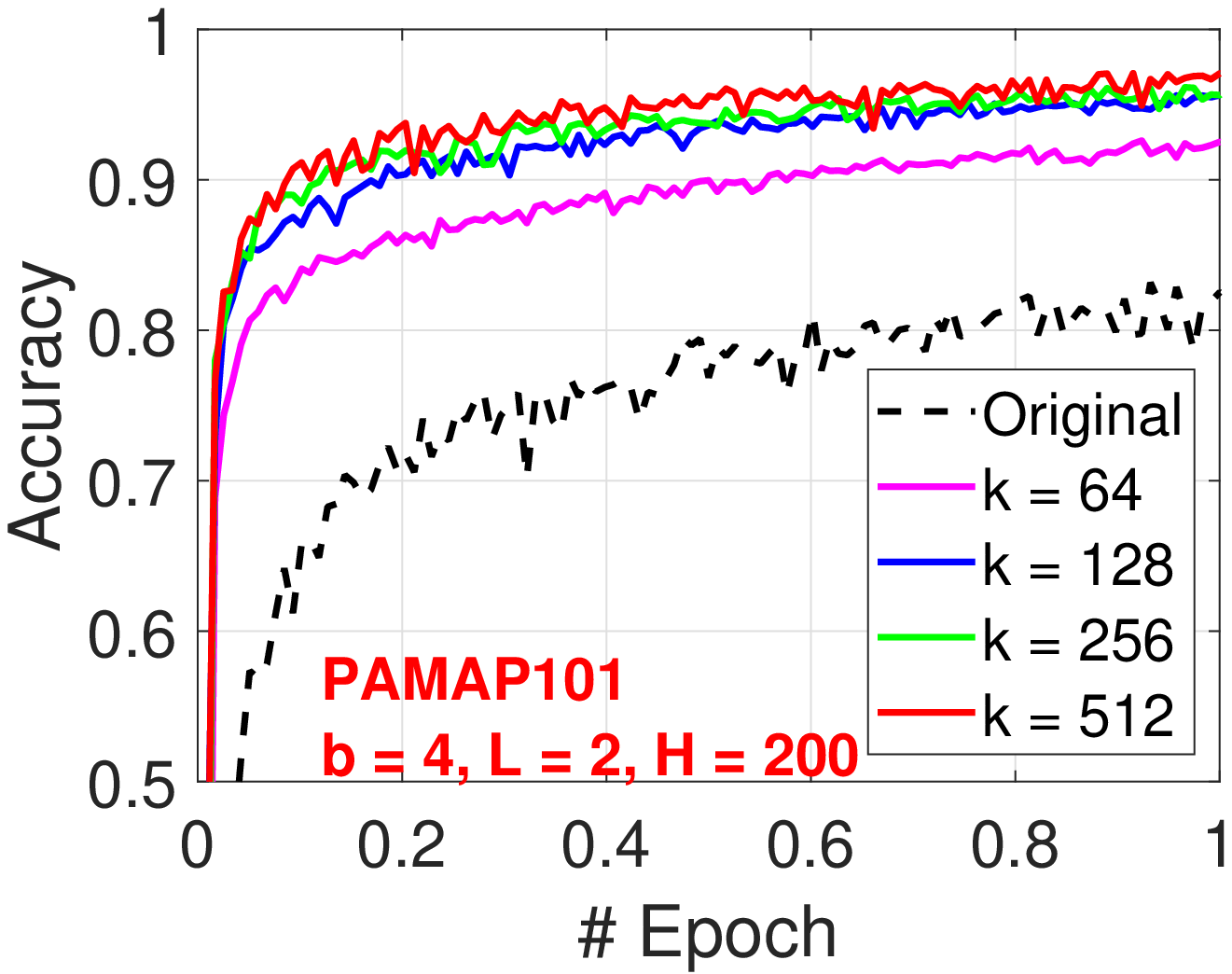}
}

\mbox{
\includegraphics[width=2.2in]{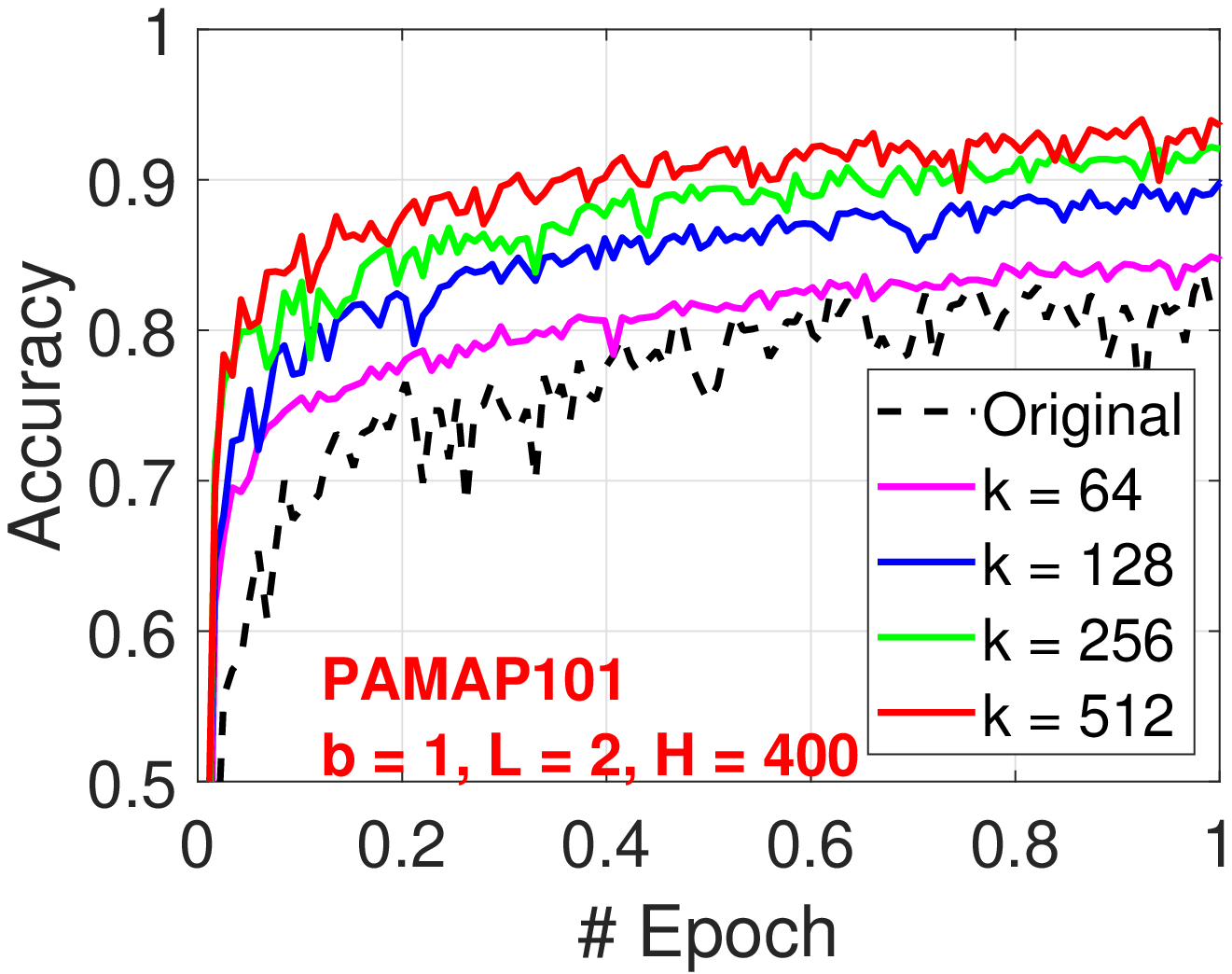}
\includegraphics[width=2.2in]{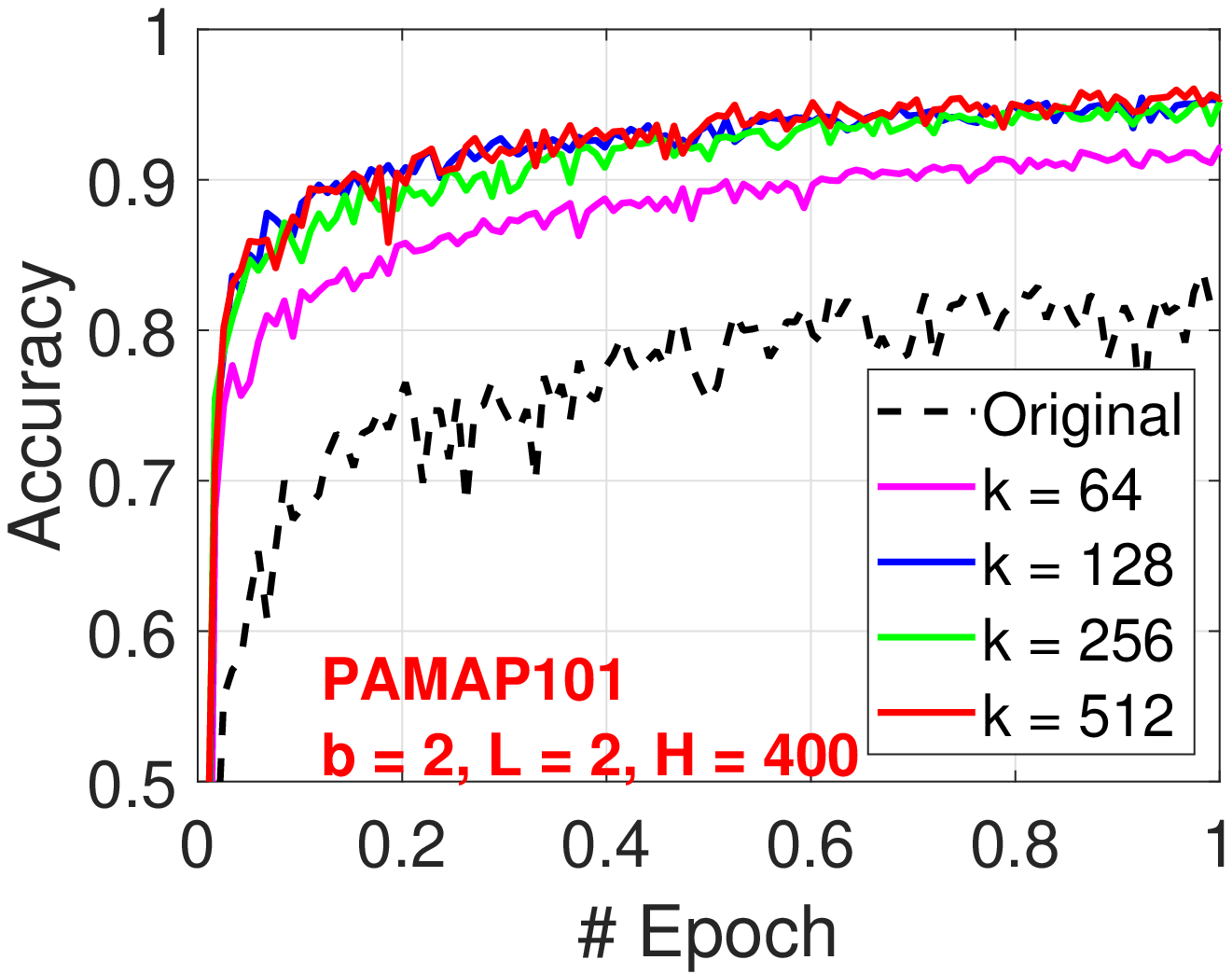}
\includegraphics[width=2.2in]{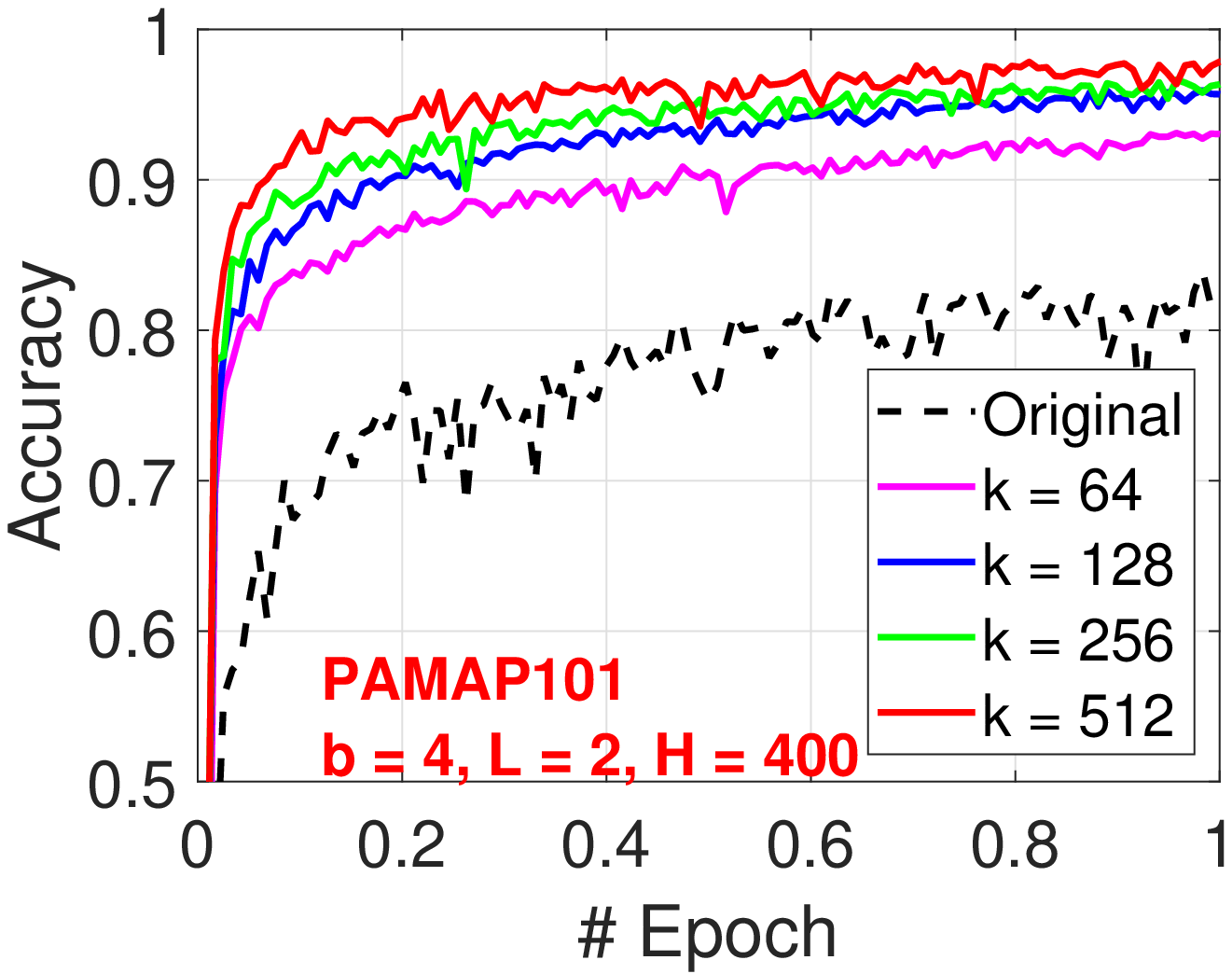}
}

\end{center}
\caption{GCWSNet with $p=1$, for $b\in\{1,2,4\}$ and $k\in\{64,128,256,512\}$, on the PAMAP101 dataset, just for 1 epoch. Again, we can see that GCWSNet converges much faster than training on the original data.}\label{fig:PAMAP101_p1_1ep}
\end{figure}

Next, Figure~\ref{fig:PAMAP101_p1} and Figure~\ref{fig:PAMAP101_p1_1ep} present the experimental results on the PAMAP101 dataset, for  We report the results for 100 epochs in Figure~\ref{fig:PAMAP101_p1} and the results for just 1 epoch in Figure~\ref{fig:PAMAP101_p1_1ep}.  Again, we can see that GCWSNet converges much faster and can reach a reasonable accuracy even with only one epoch of training. 

\newpage\clearpage

\begin{figure}[h!]
\begin{center}
\mbox{
\includegraphics[width=2.2in]{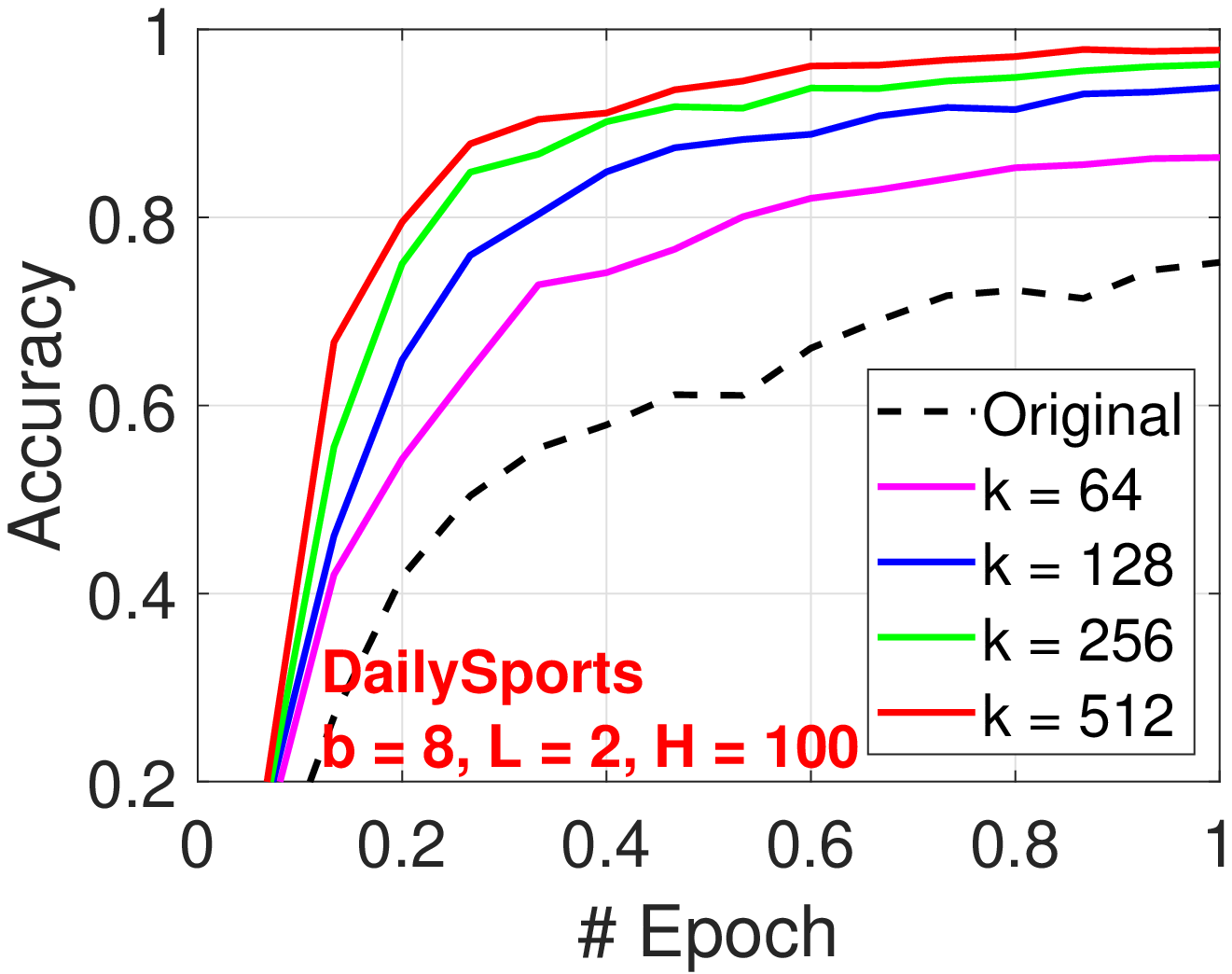}
\includegraphics[width=2.2in]{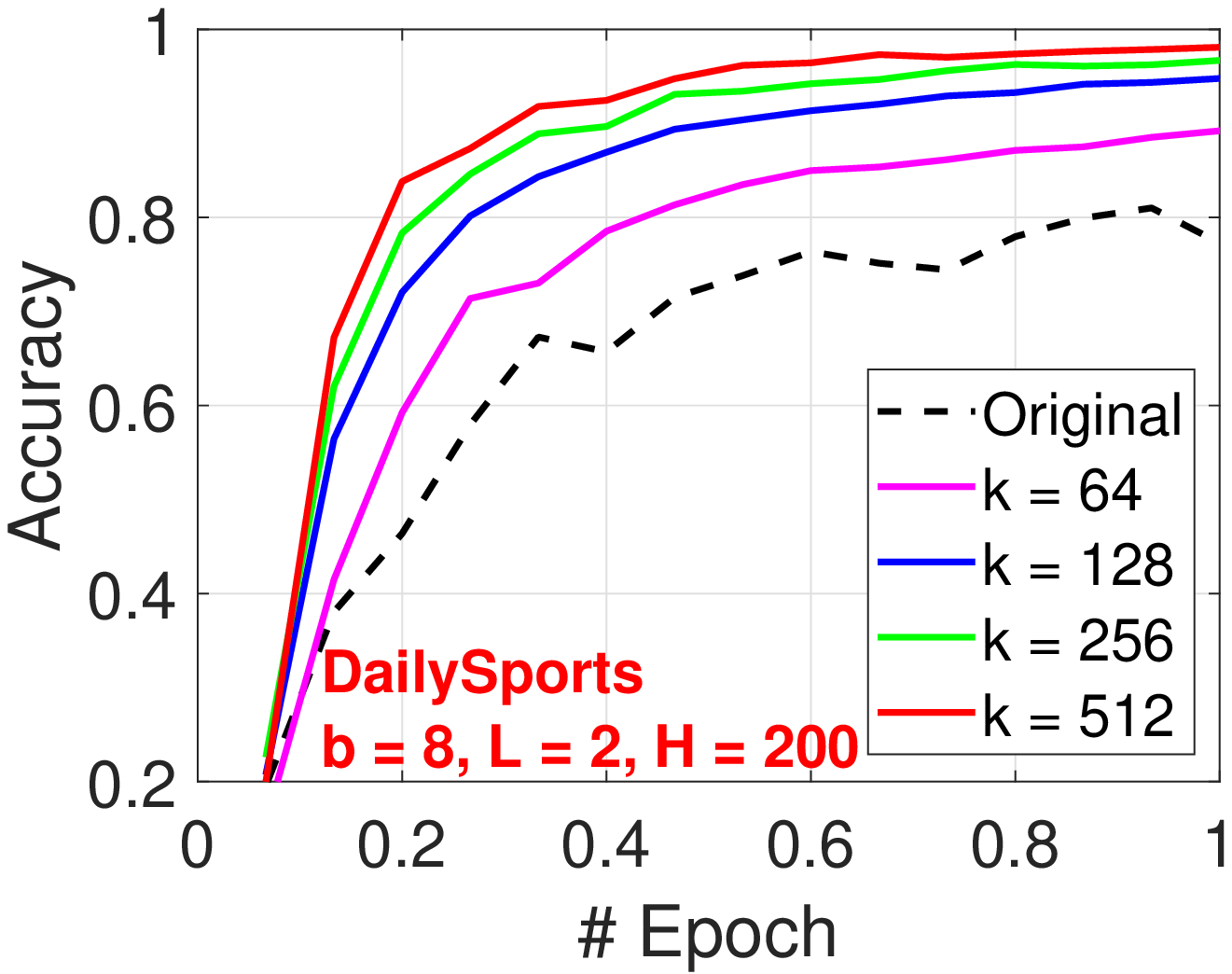}
\includegraphics[width=2.2in]{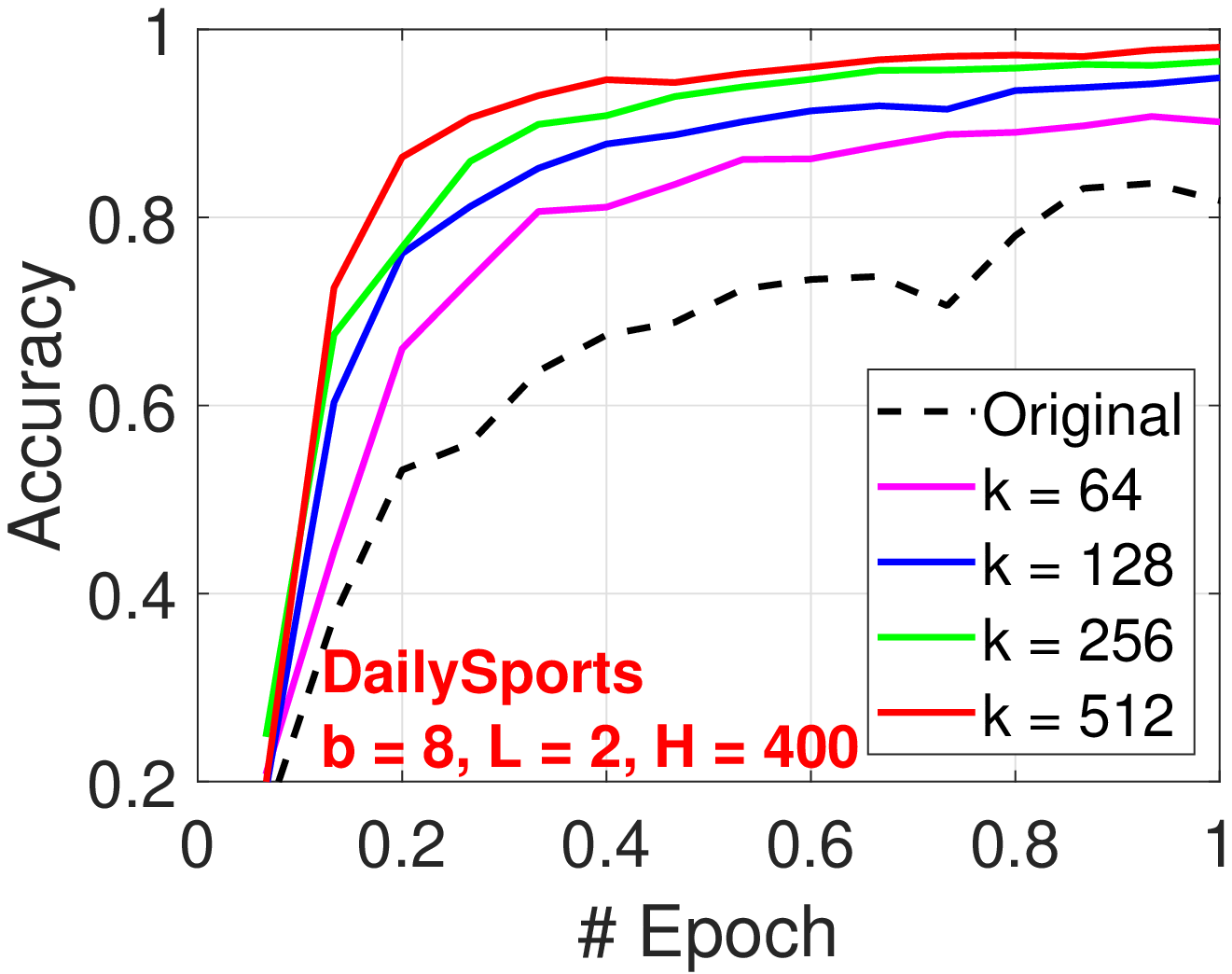}
}

\mbox{
\includegraphics[width=2.2in]{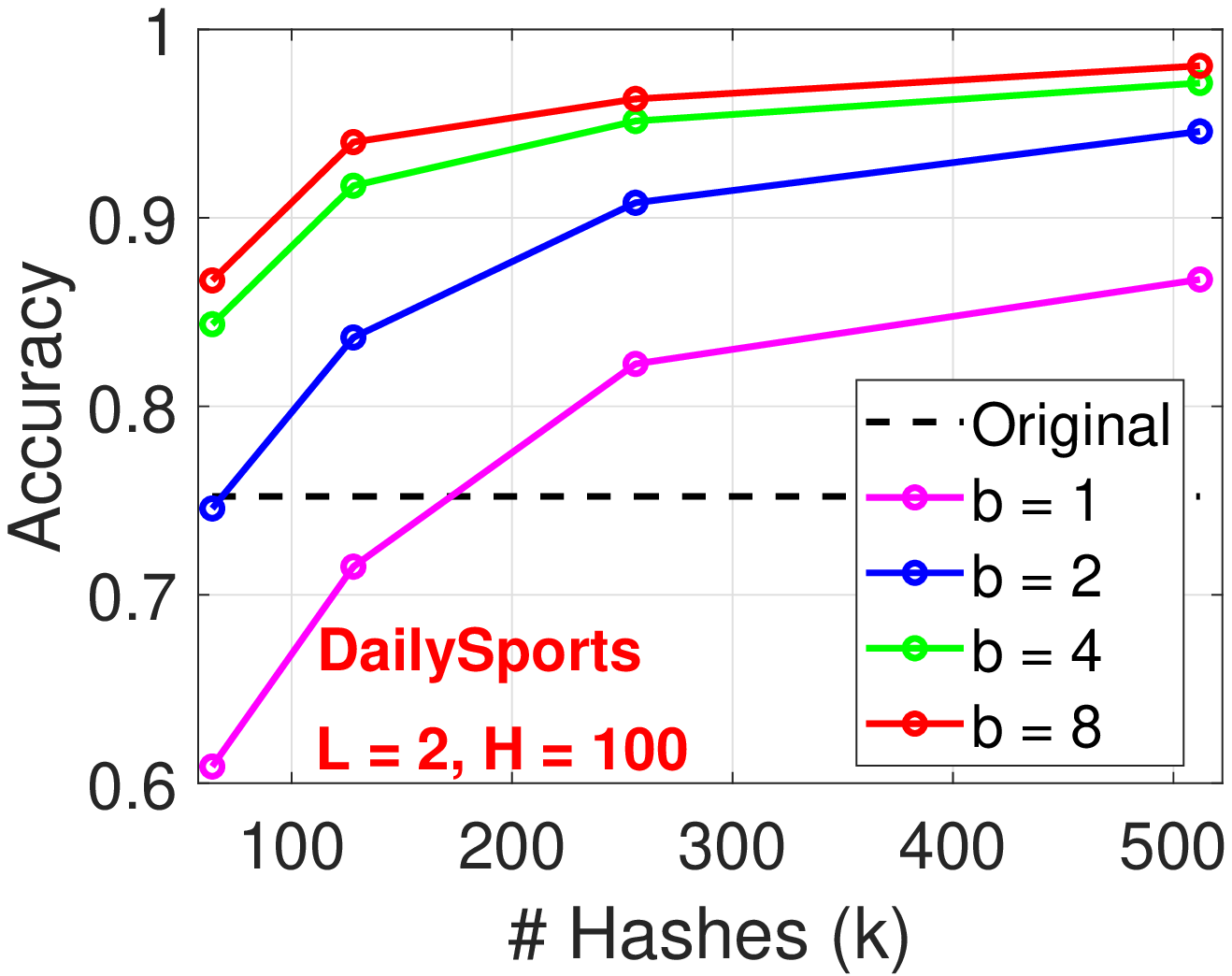}
\includegraphics[width=2.2in]{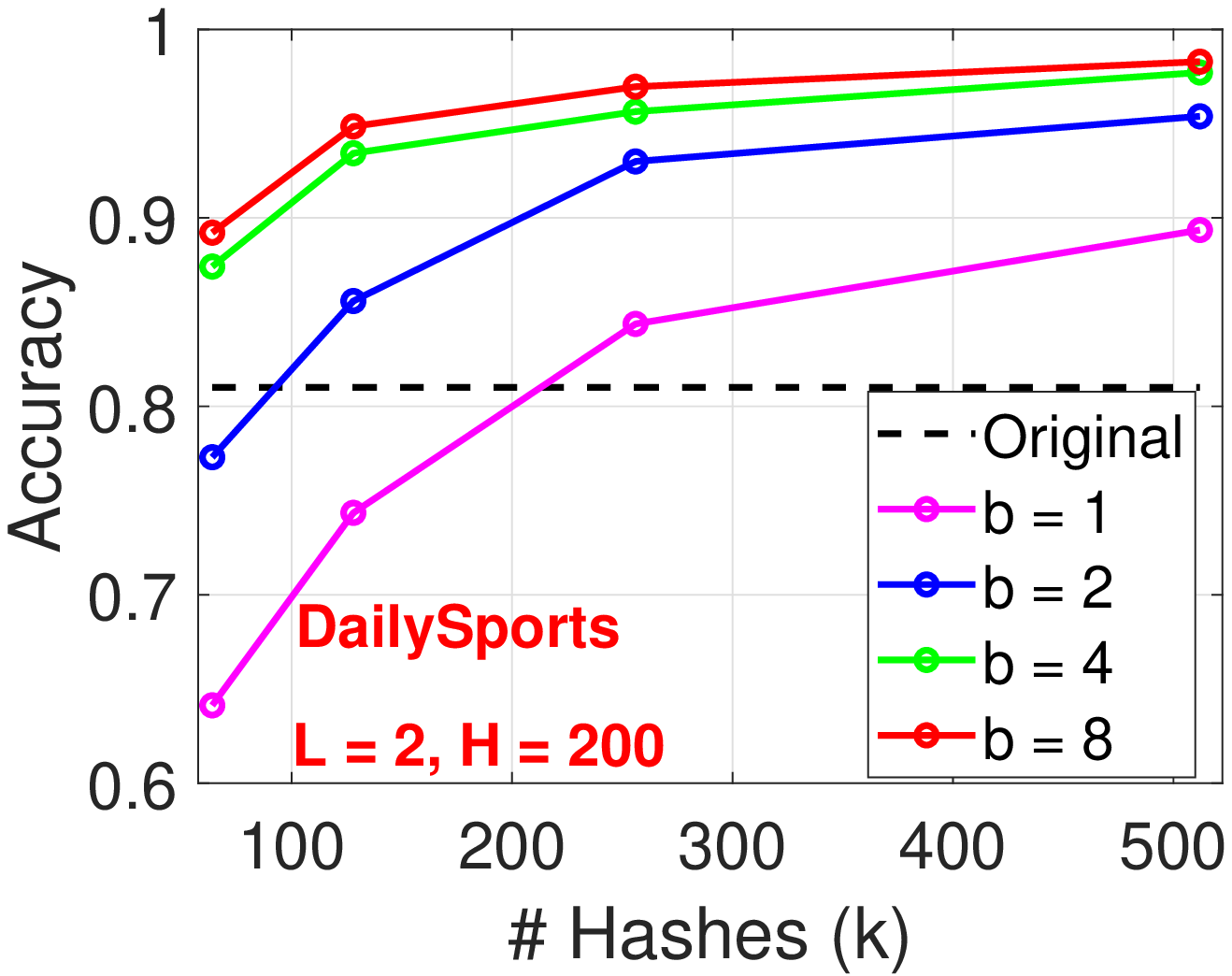}
\includegraphics[width=2.2in]{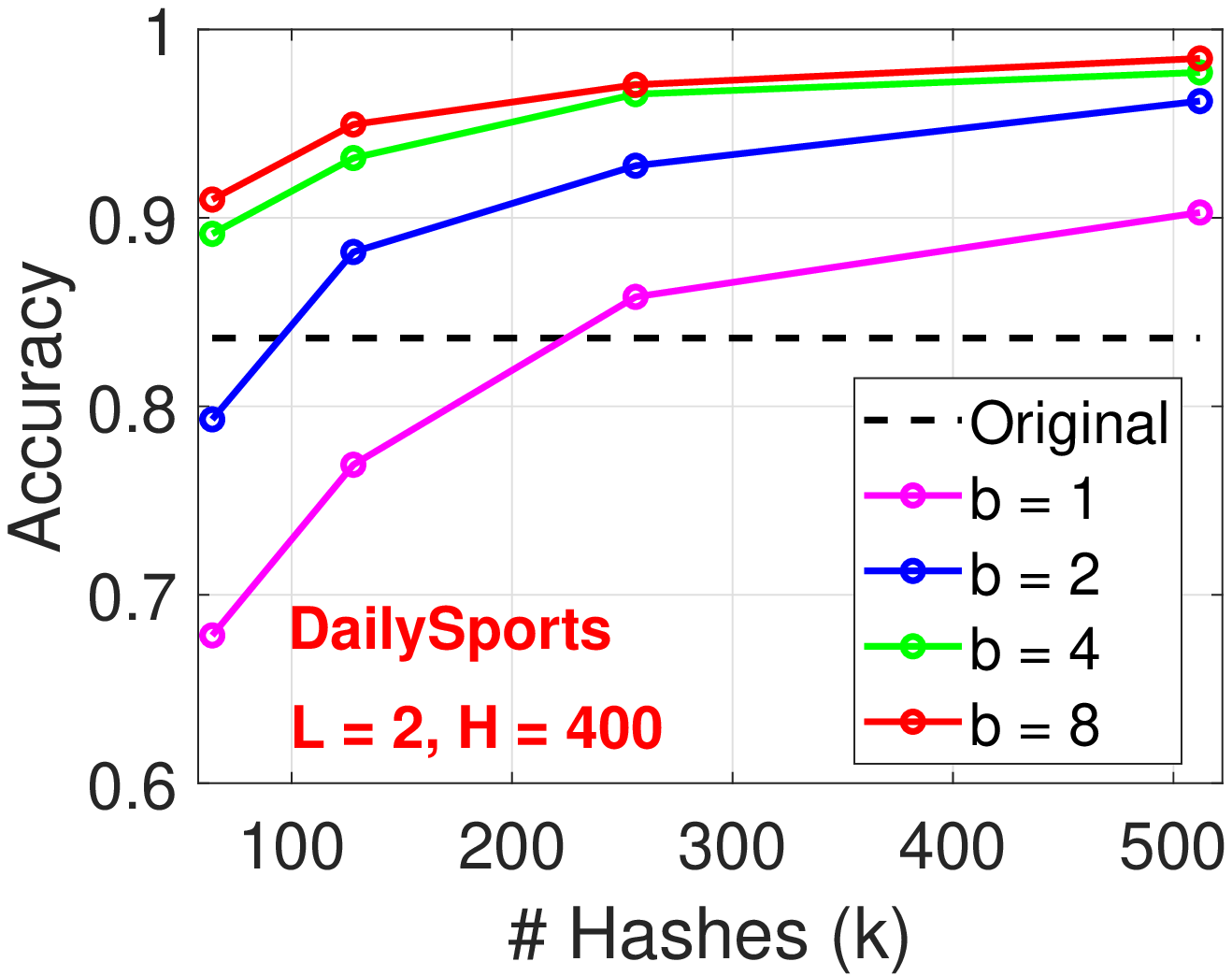}
}

\end{center}

\vspace{-0.2in}

\caption{GCWSNet with $p=1$ on the DailySports dataset, for just 1 epoch.  }\label{fig:DailySports_p1_1ep}\vspace{-0.2in}
\end{figure}

\begin{figure}[h!]
\begin{center}
\mbox{
\includegraphics[width=2.2in]{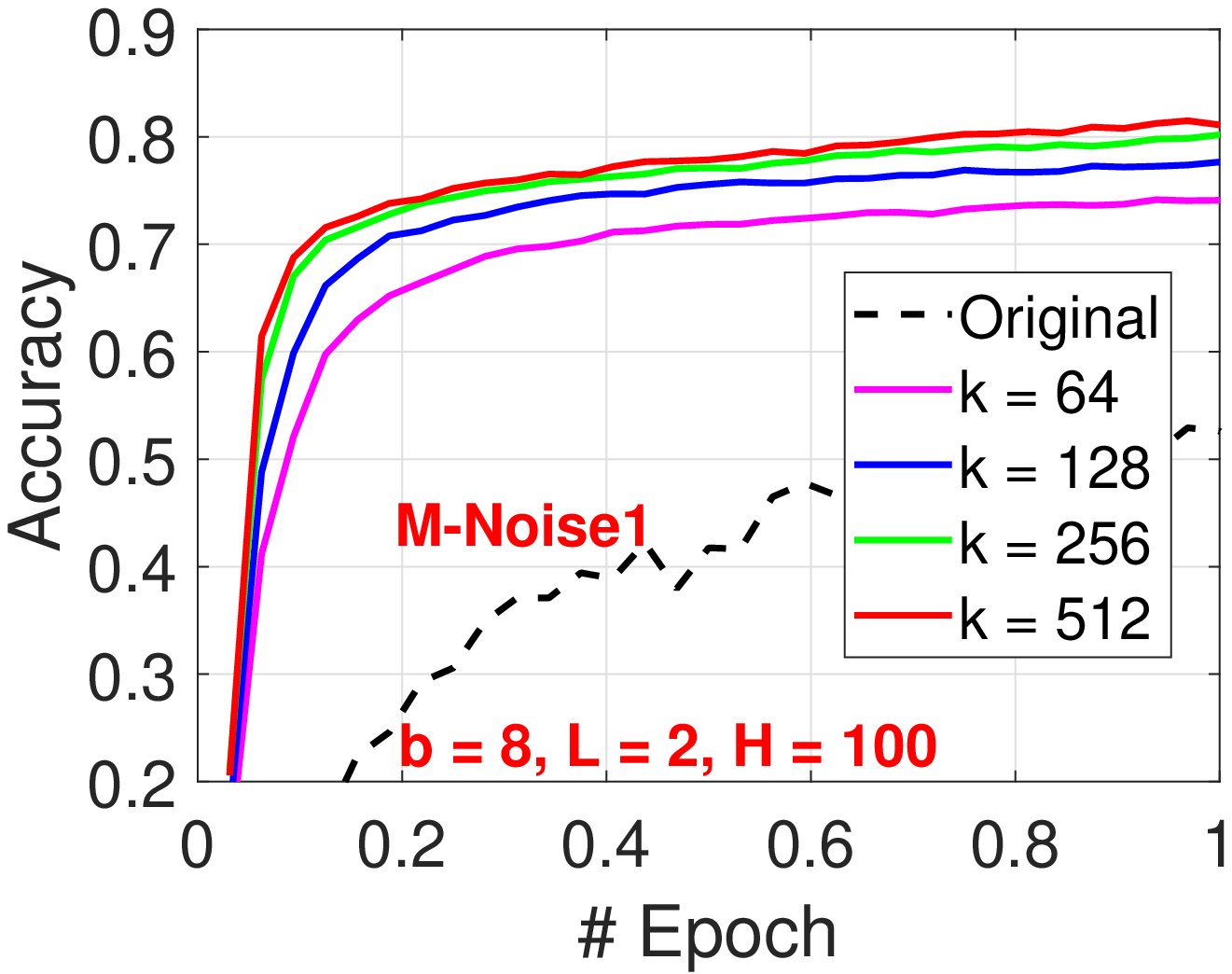}
\includegraphics[width=2.2in]{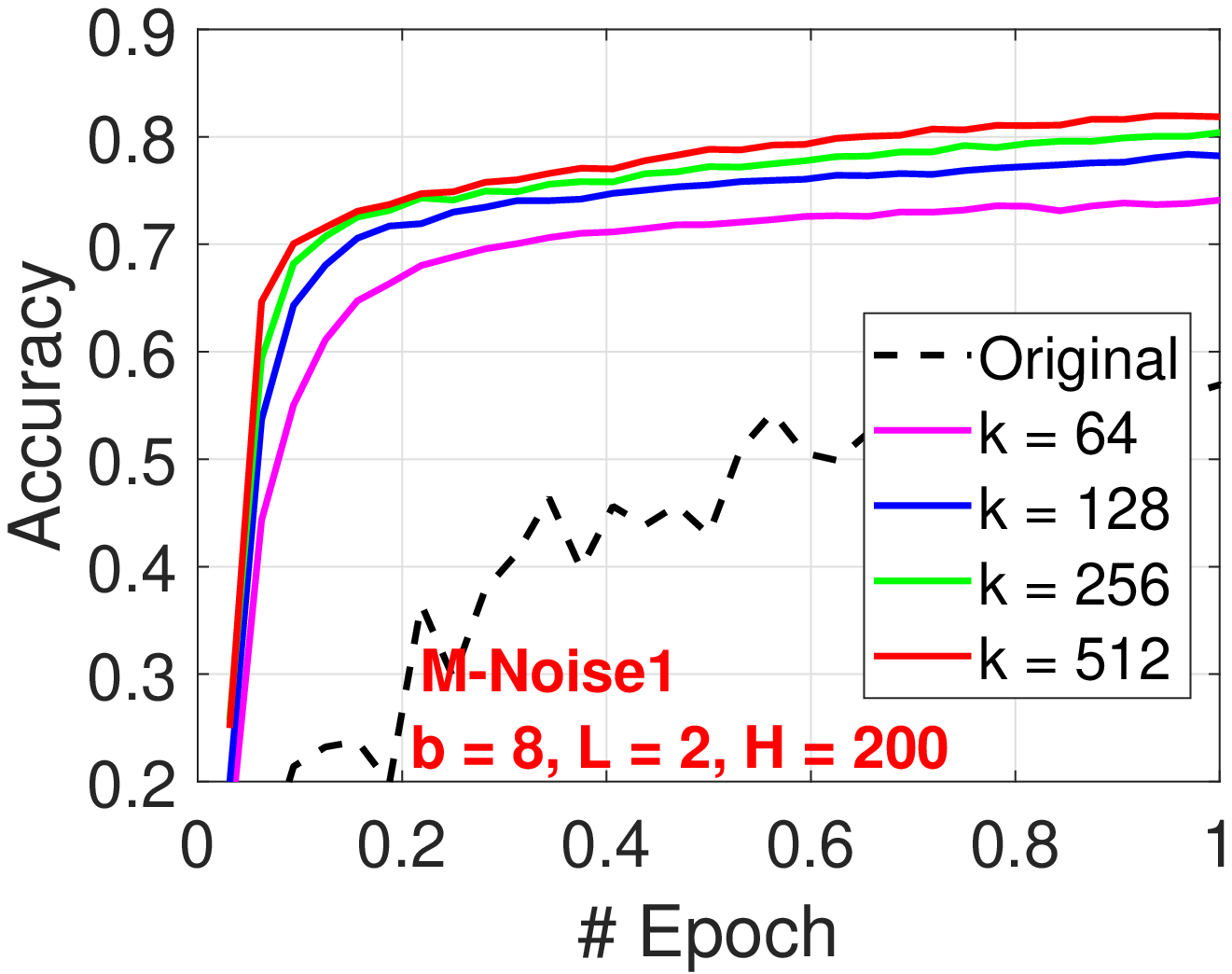}
\includegraphics[width=2.2in]{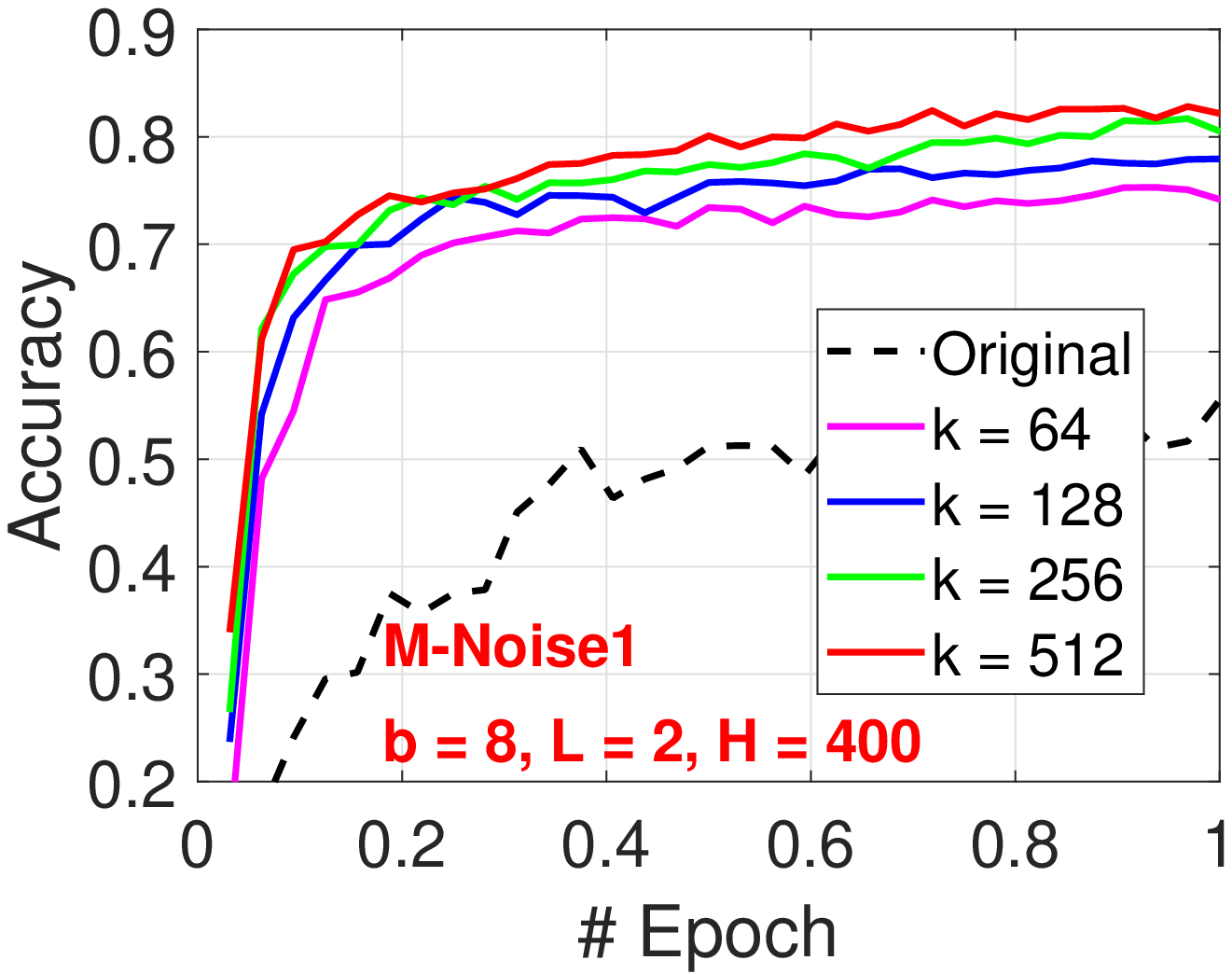}
}
\mbox{
\includegraphics[width=2.2in]{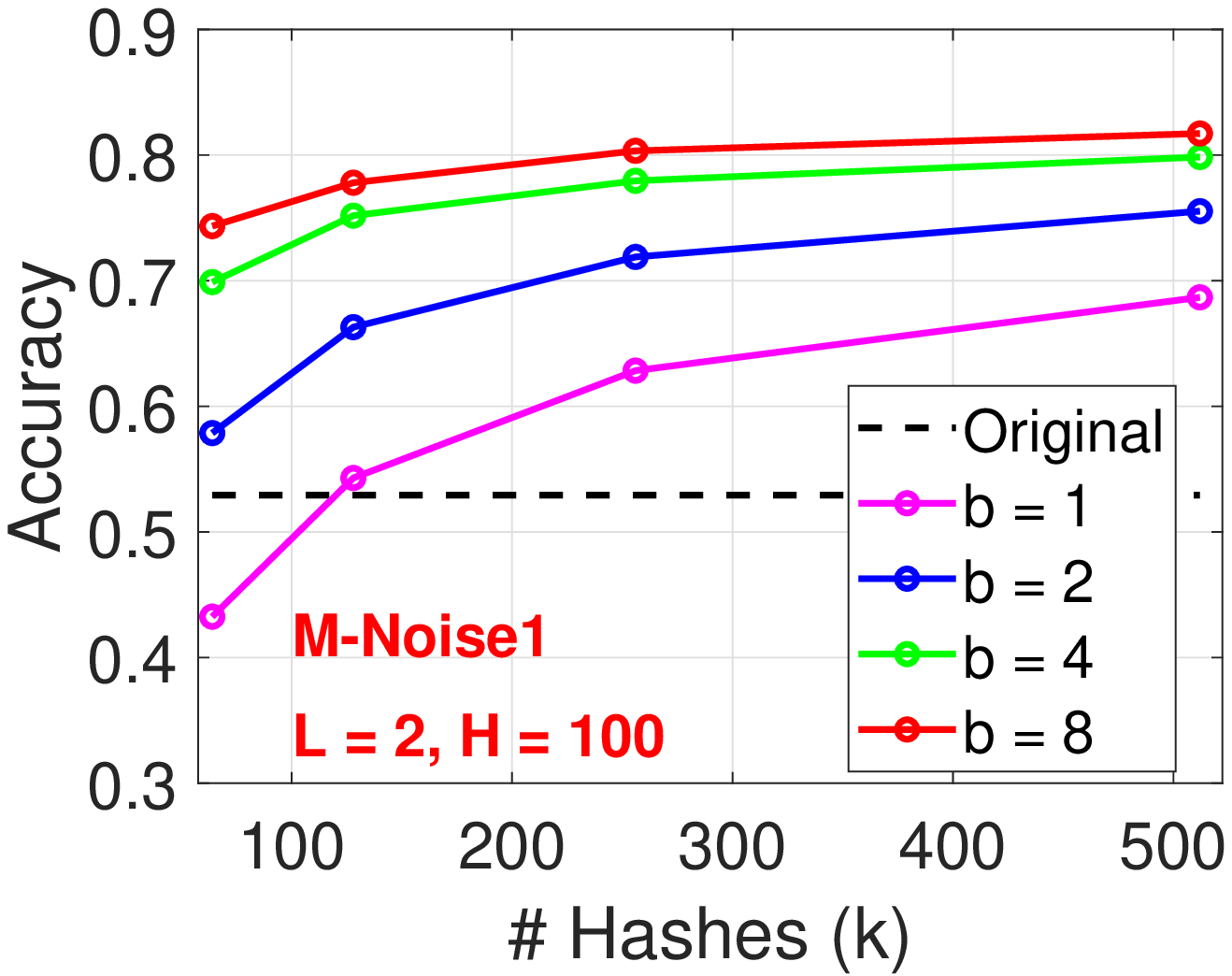}
\includegraphics[width=2.2in]{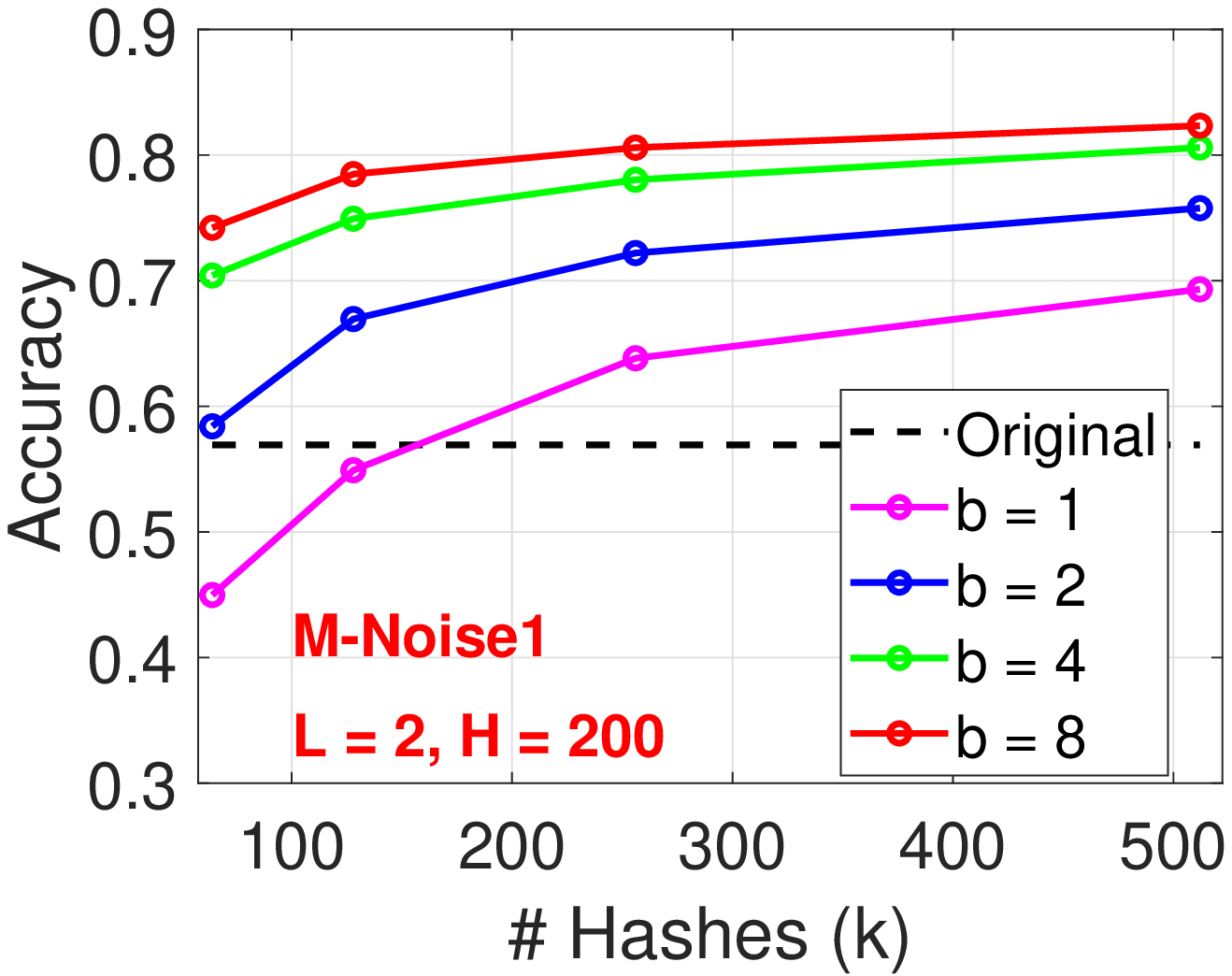}
\includegraphics[width=2.2in]{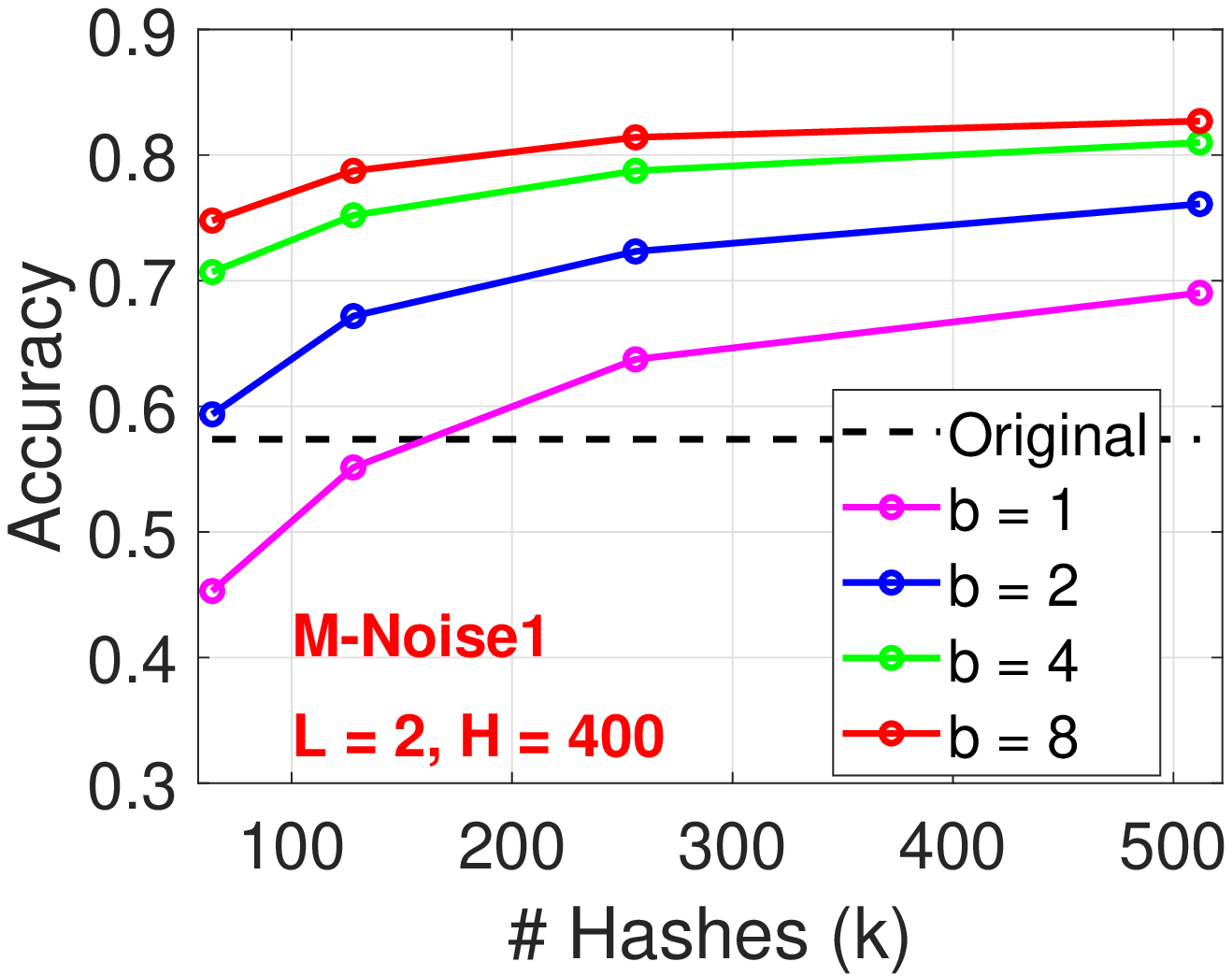}
}

\caption{GCWSNet with $p=80$ on the M-Noise1 dataset, for just 1 epoch.  }\label{fig:MNoise1_p1_1ep}
\end{center}\vspace{-0.1in}
\end{figure}

\begin{figure}[h]
\begin{center}
\mbox{
\includegraphics[width=2.2in]{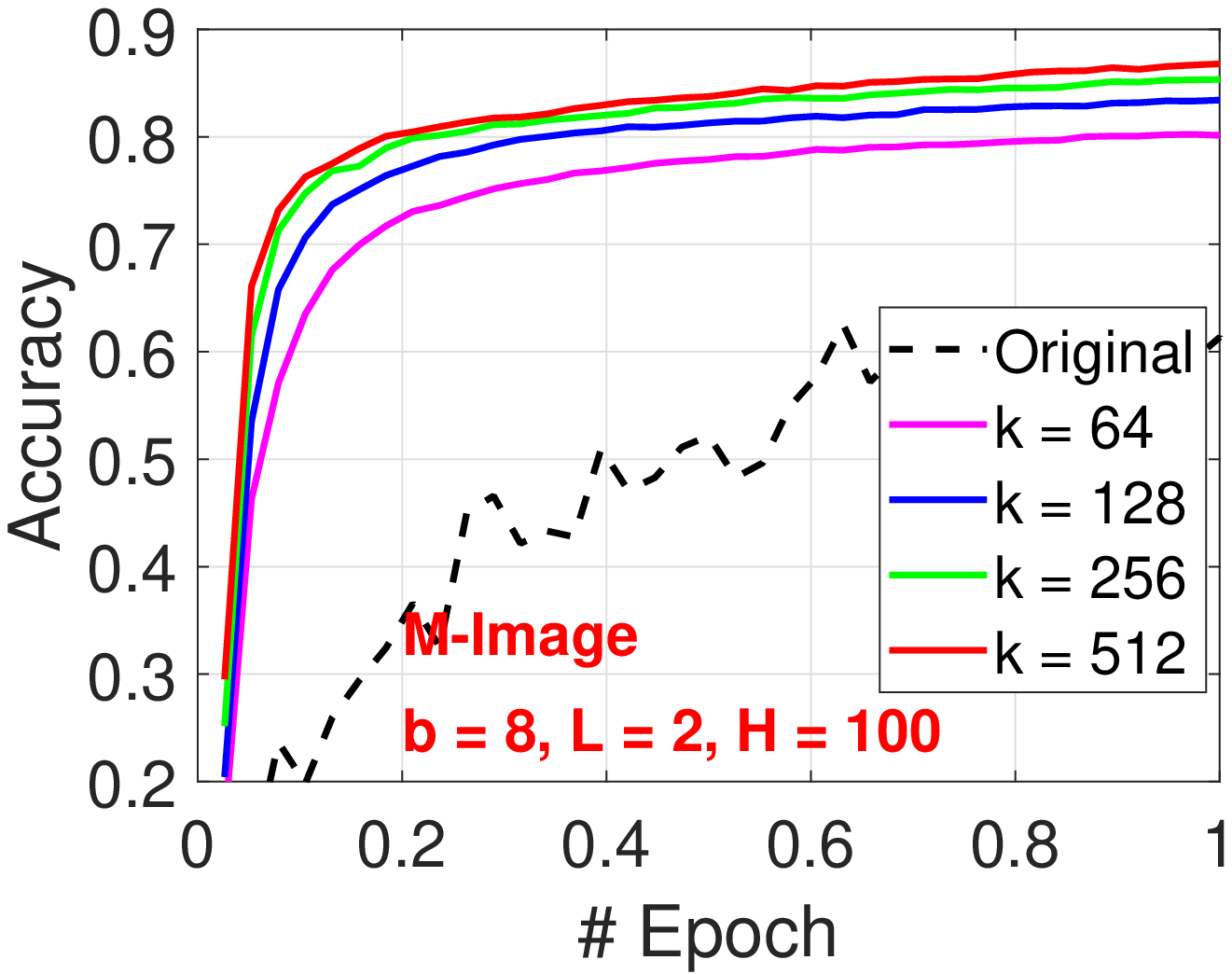}
\includegraphics[width=2.2in]{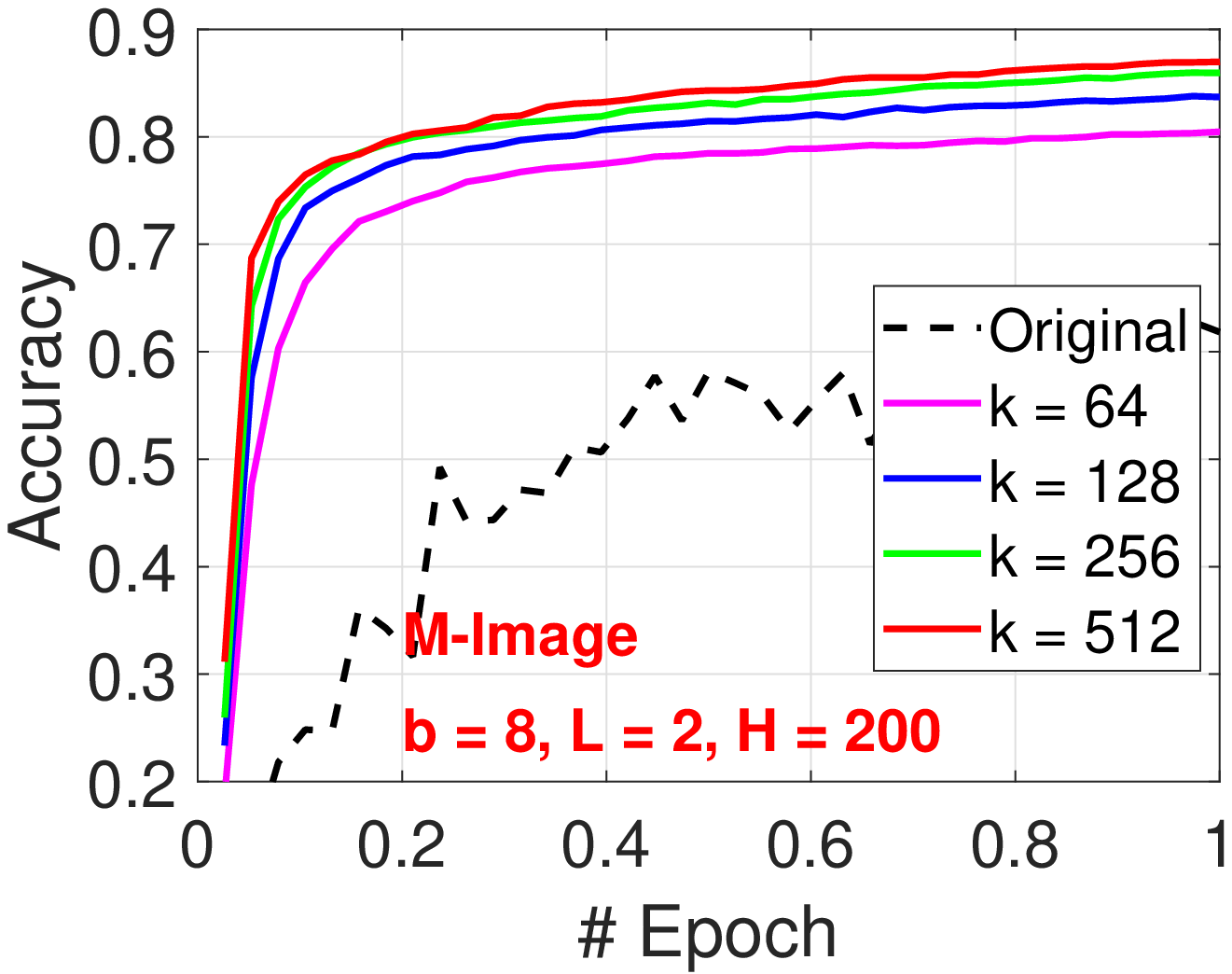}
\includegraphics[width=2.2in]{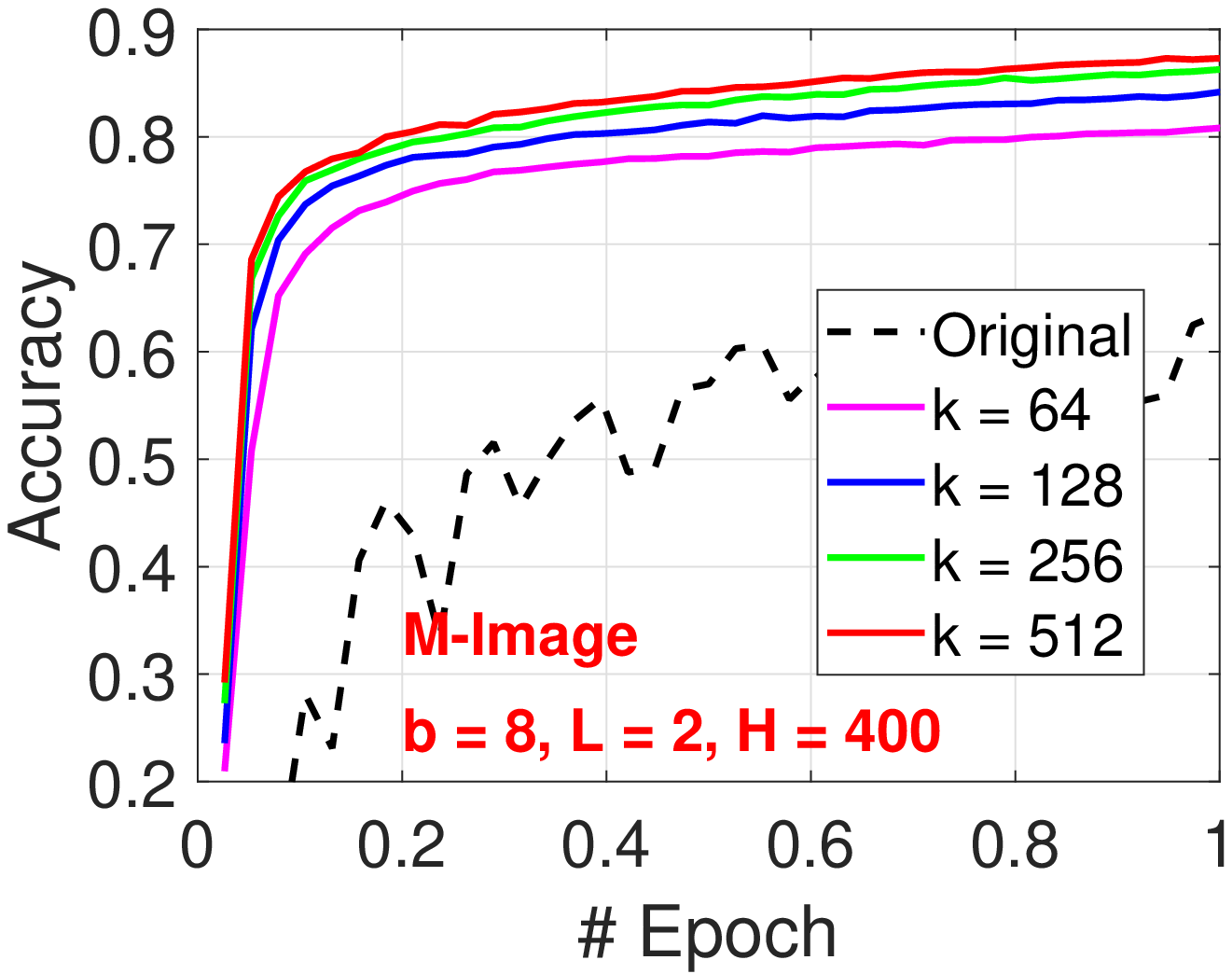}
}
\mbox{
\includegraphics[width=2.2in]{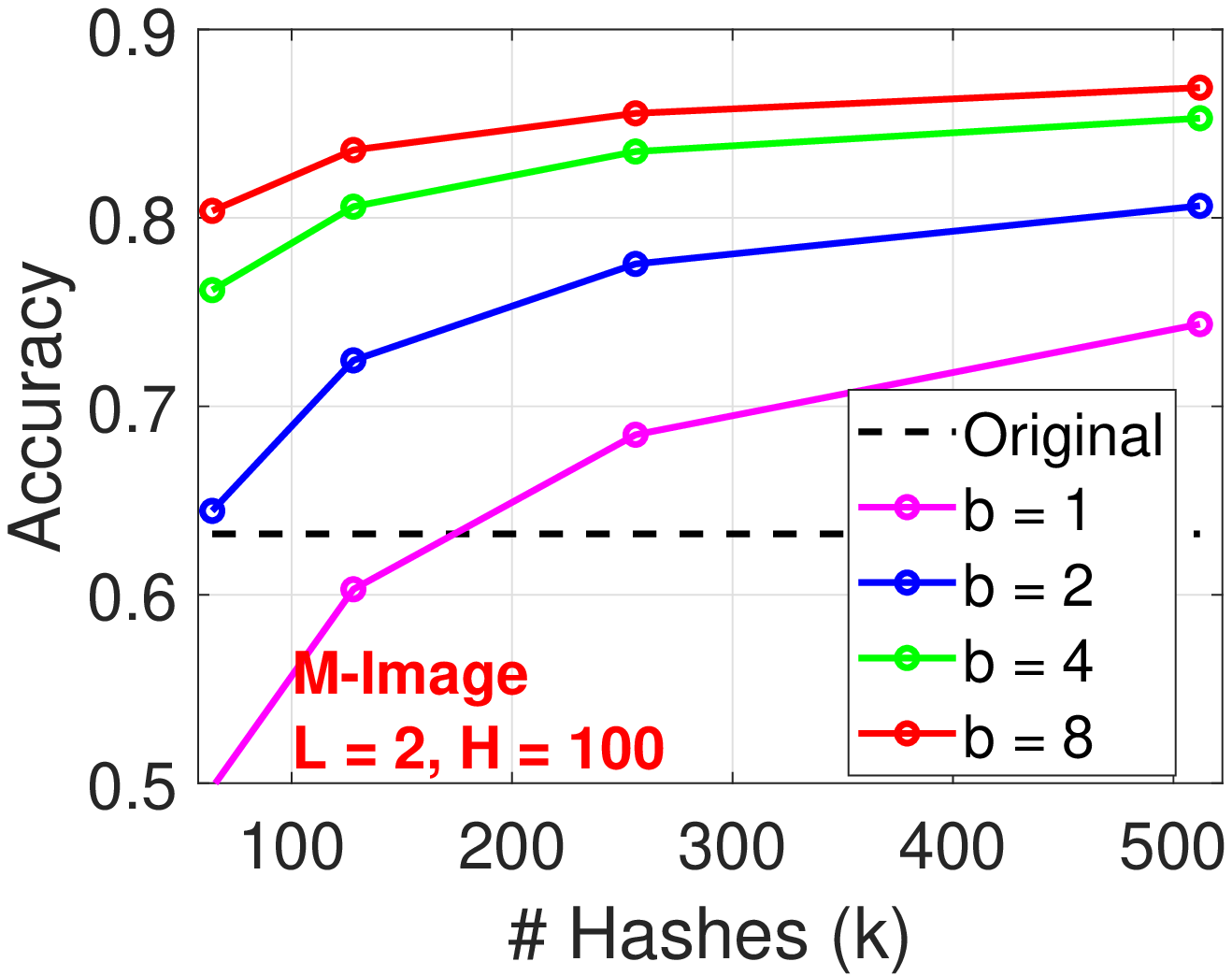}
\includegraphics[width=2.2in]{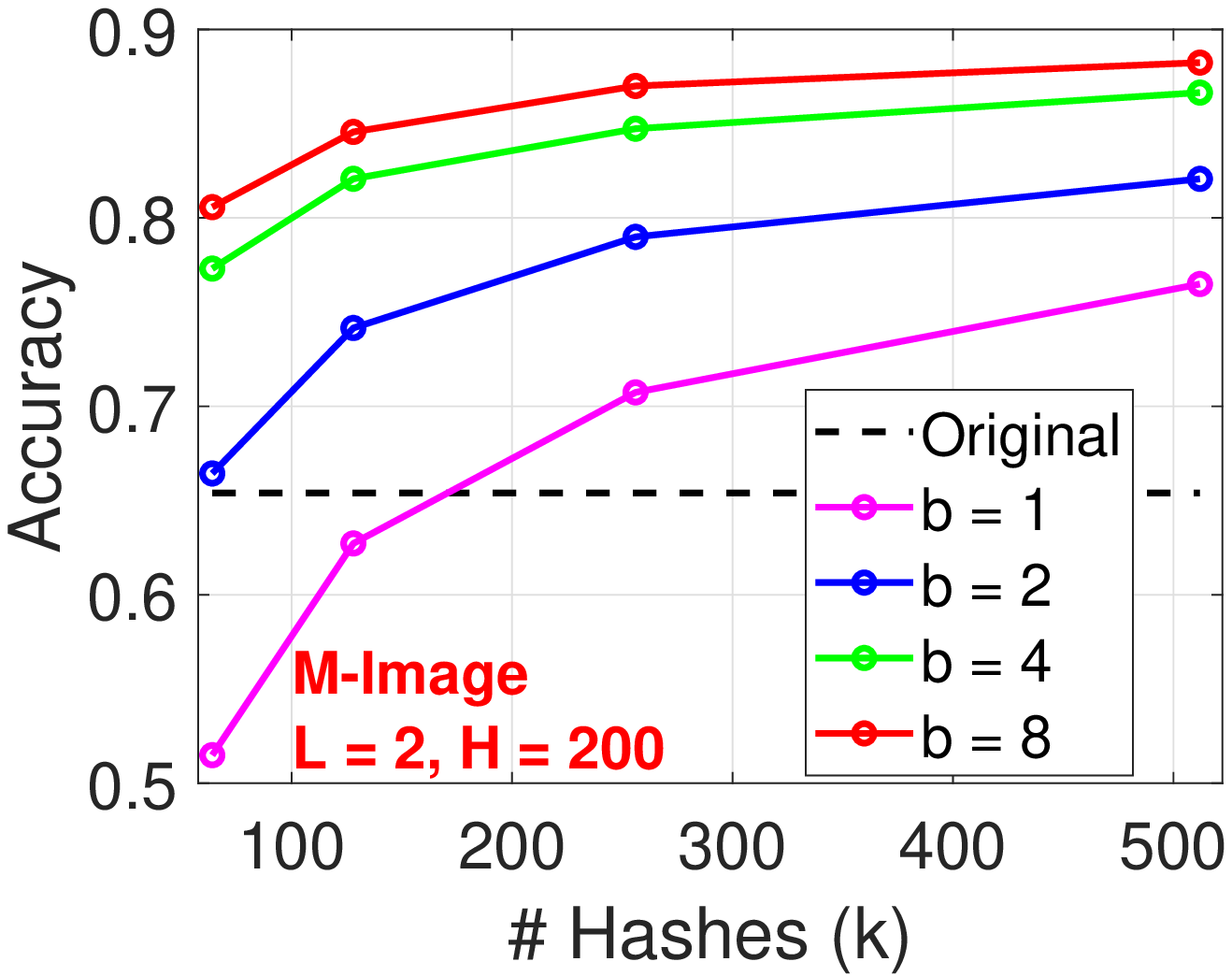}
\includegraphics[width=2.2in]{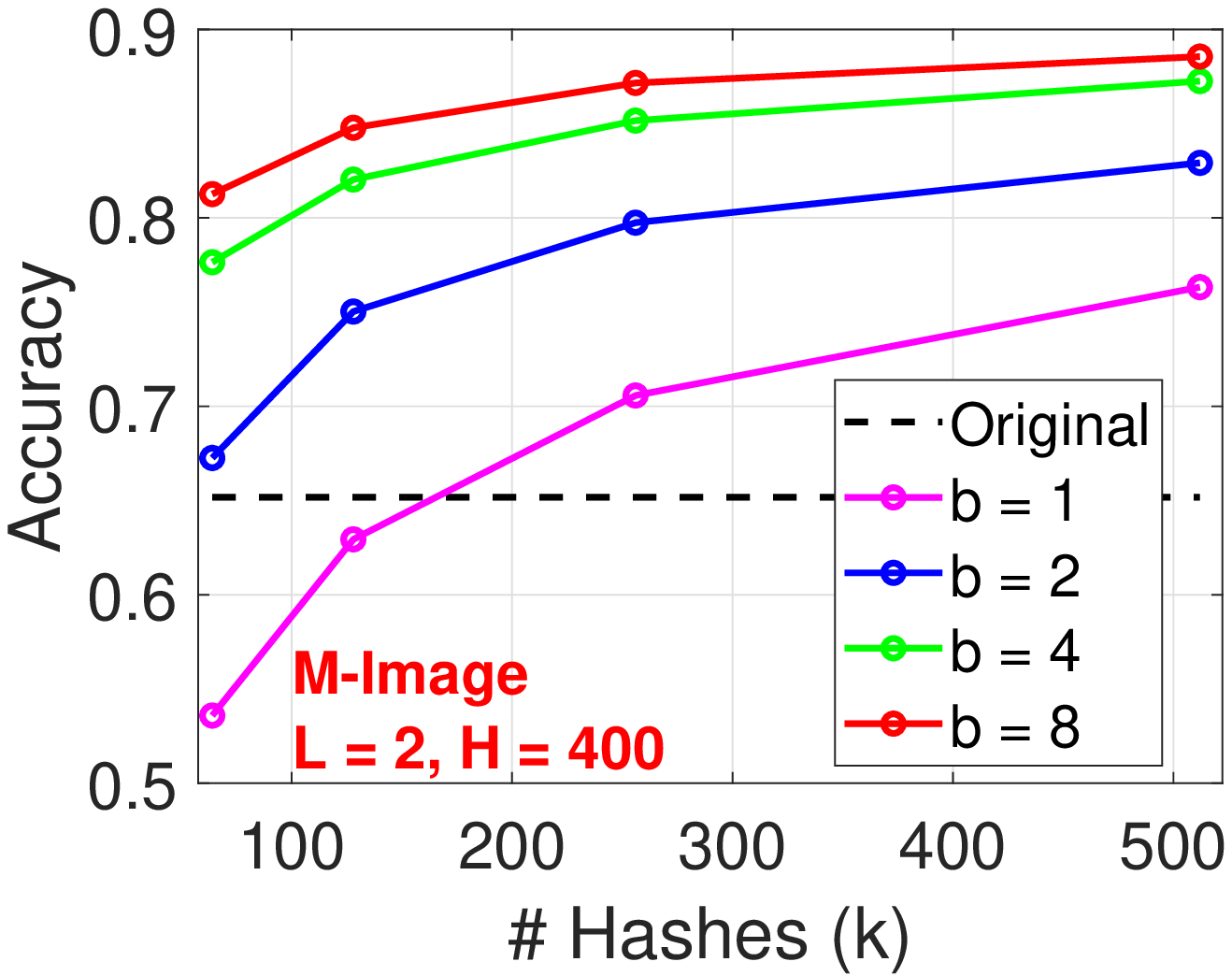}
}
\end{center}

\vspace{-0.2in}

\caption{GCWSNet with $p=50$ on the M-Image dataset, for just 1 epoch.  }\label{fig:MImage_p1_1ep}\vspace{-0.3in}
\end{figure}

Next, we summarize the results on the DailySports dataset, the M-Noise1 dataset, and the M-Image dataset, in Figure~\ref{fig:DailySports_p1_1ep}, Figure~\ref{fig:MNoise1_p1_1ep}, and Figure~\ref{fig:MImage_p1_1ep}, respectively, for just 1 epoch.  For DailySports, we let $p=1$. For M-Noise1 and M-Image, we use $p=80$ and $p=50$, respectively,  as suggested in Table~\ref{tab:SVM}.

\newpage\clearpage

\section{Using Bits from $t^*$ }

\begin{figure}[b!]

\vspace{-0.2in}

\begin{center}
\mbox{
\includegraphics[width=2.2in]{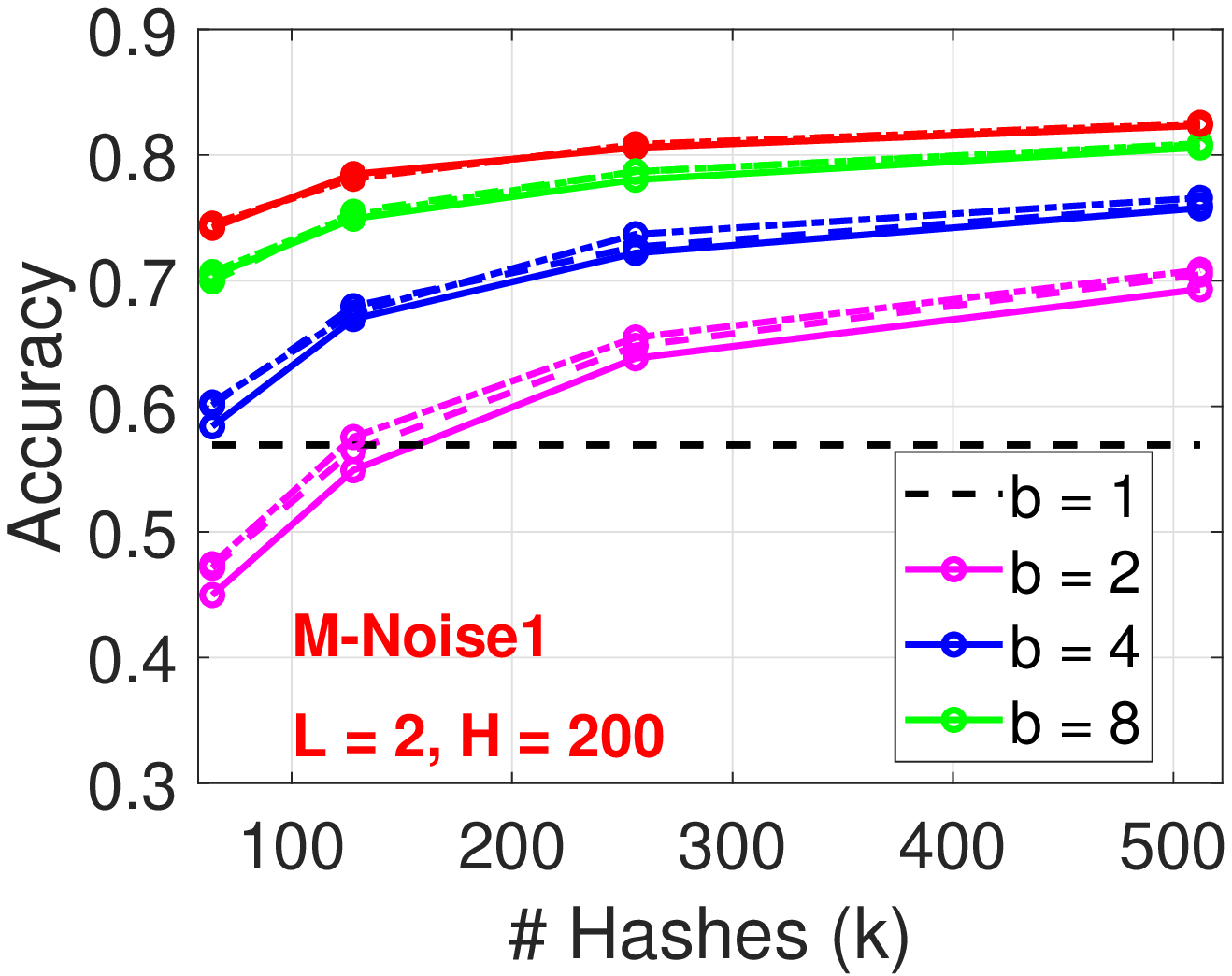}
\includegraphics[width=2.2in]{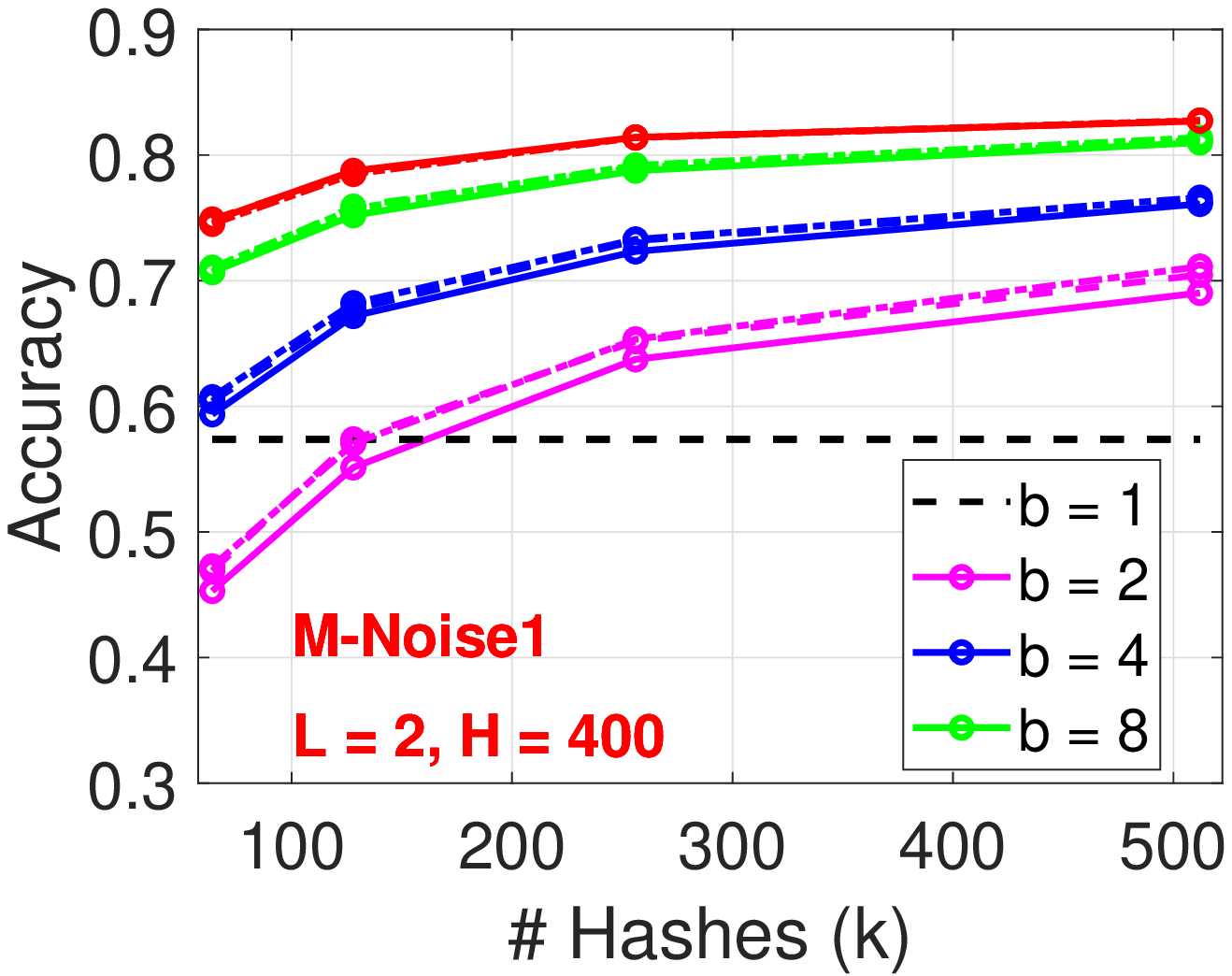}
\includegraphics[width=2.2in]{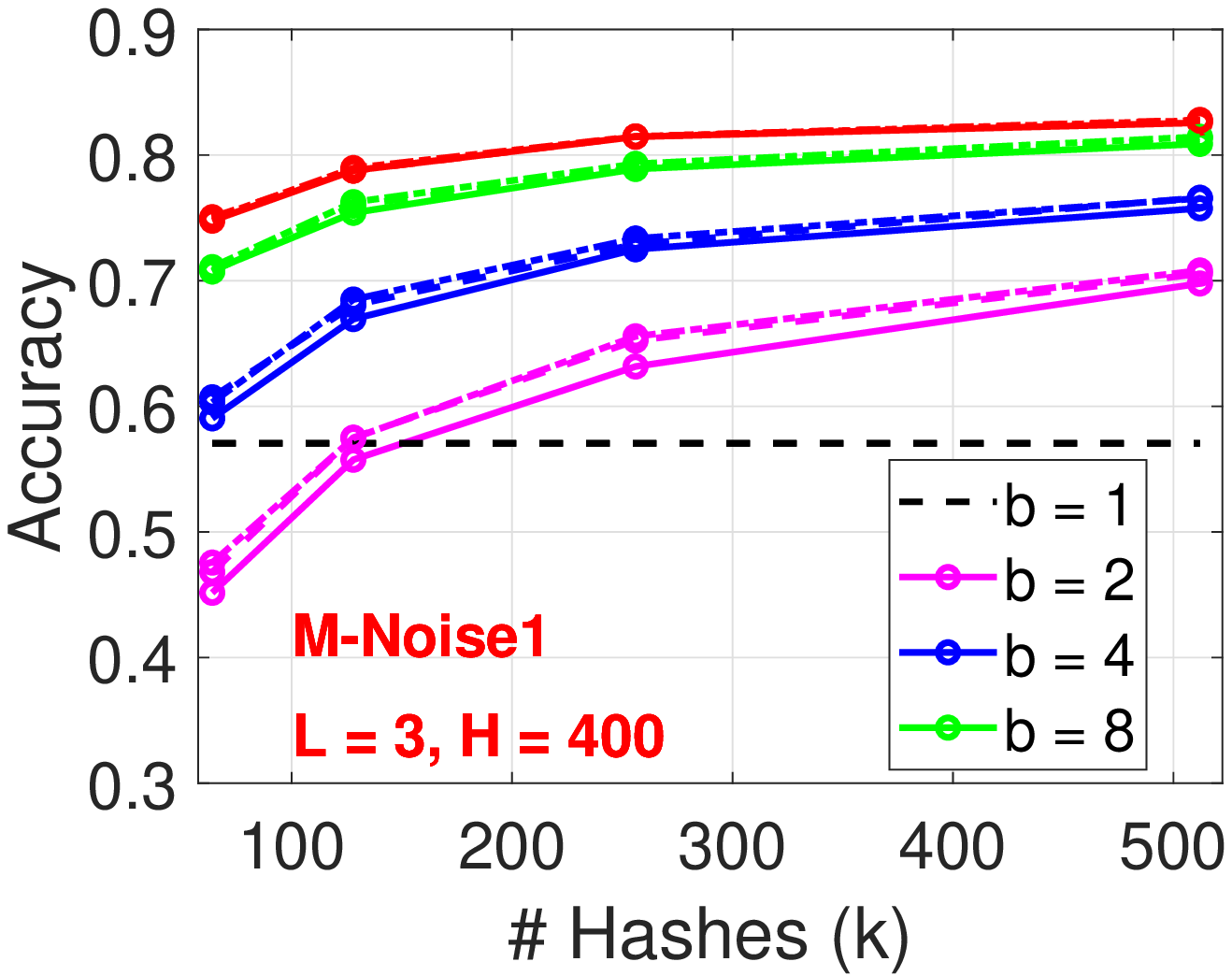}
}

\mbox{
\includegraphics[width=2.2in]{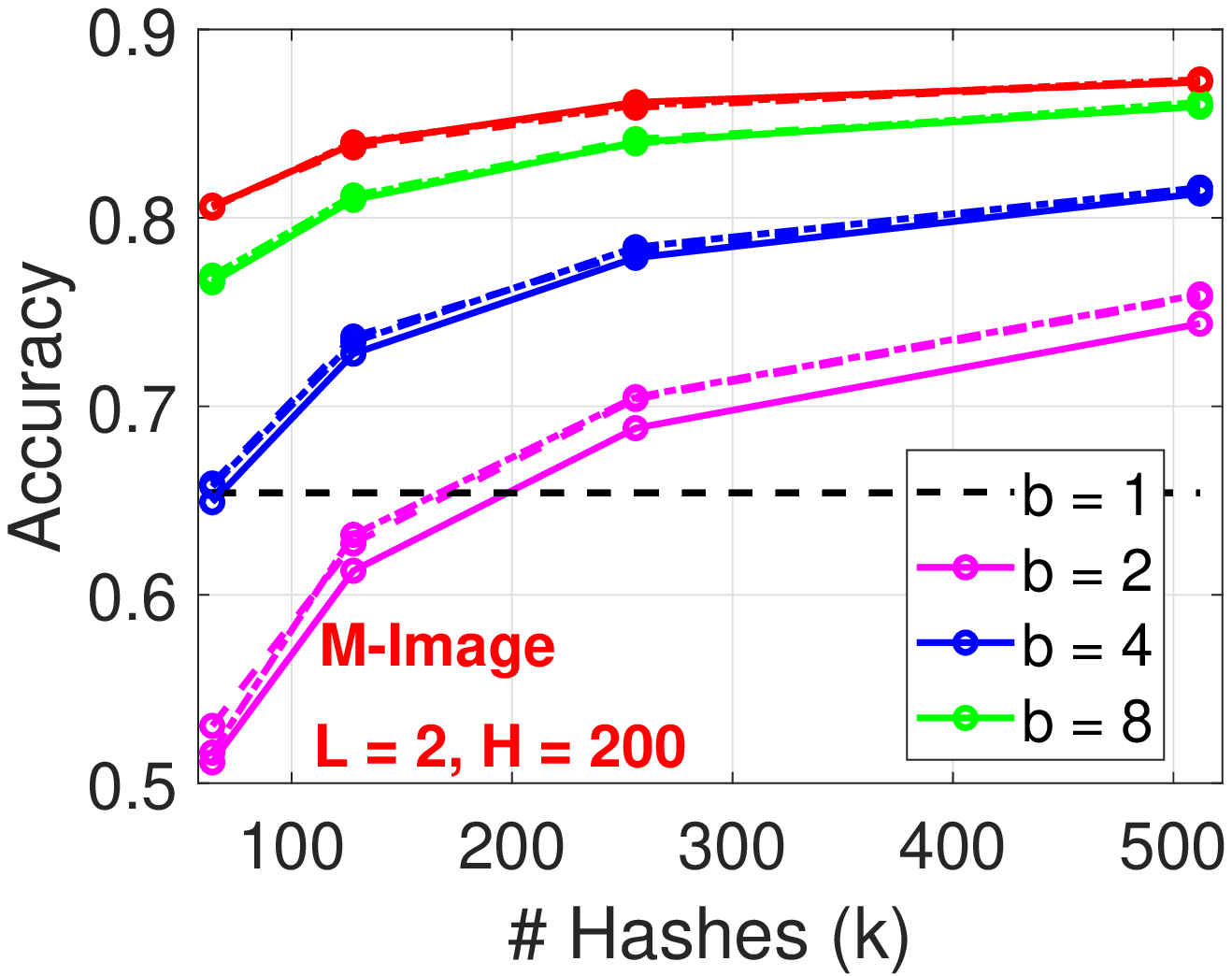}
\includegraphics[width=2.2in]{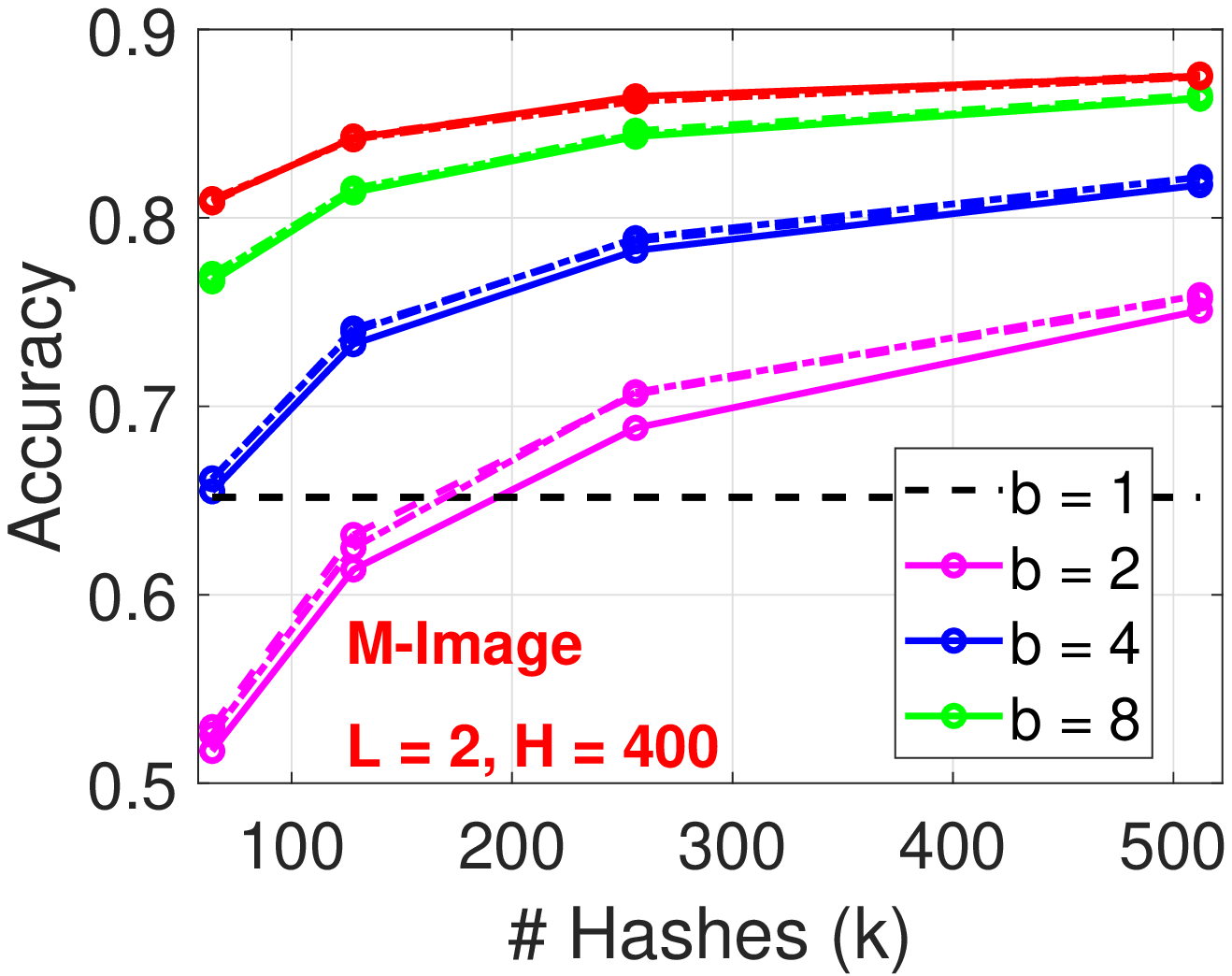}
\includegraphics[width=2.2in]{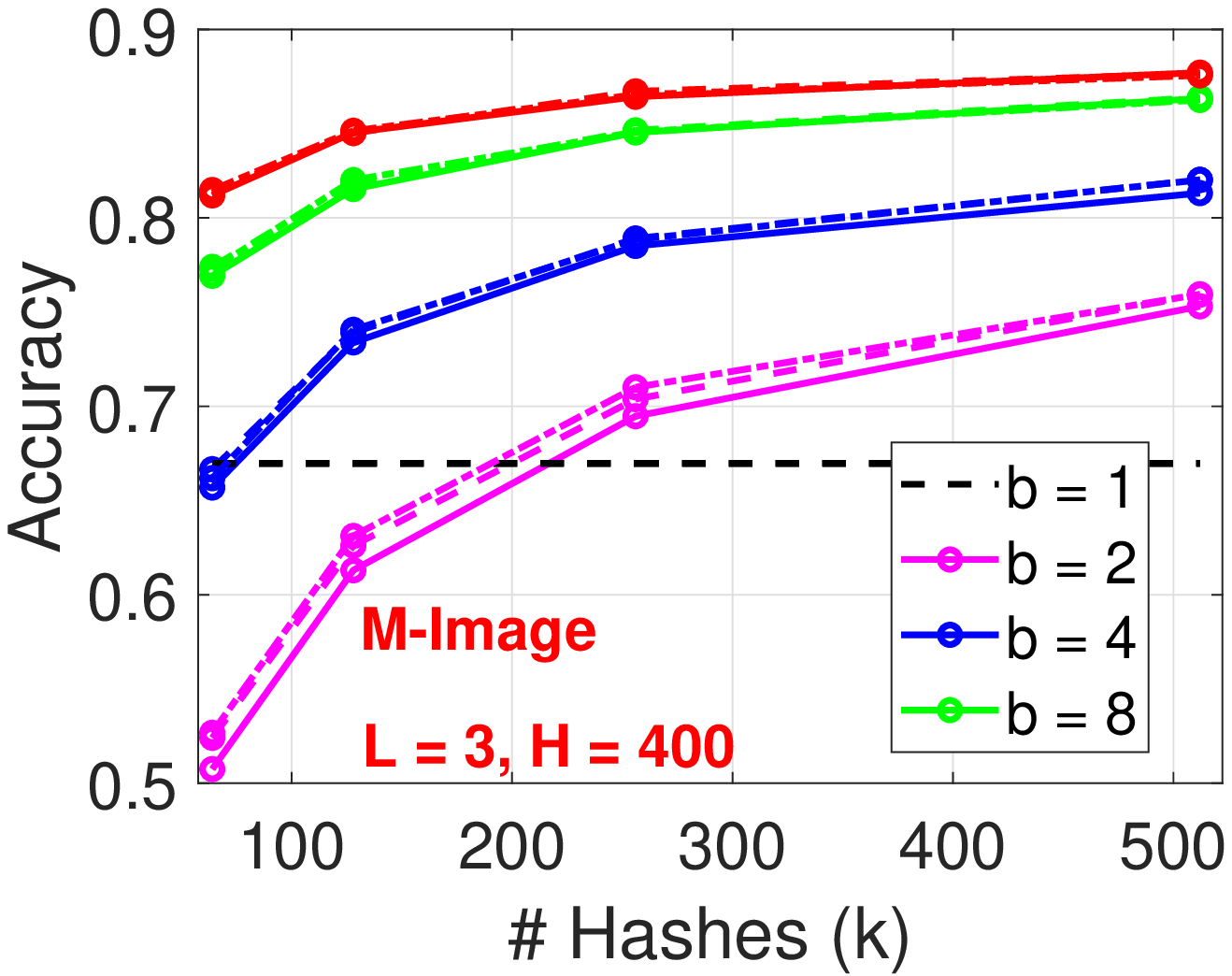}
}
\end{center}

\vspace{-0.2in}

\caption{GCWSNet  for M-Noise (with $p=80$) and M-Image (with $p=50$) by using 0 bit (solid), 1 bit (dashed), and 2 bits (dashed dot) for $t^*$ in the output of GCWS hashing $(i^*, t^*)$ }\label{fig:MoreBits}\vspace{-0.2in}
\end{figure}

In Figure~\ref{fig:MoreBits}, we report the experiments for using 1 or 2  bits from $t^*$, which  have not been used in all previously presented experiments. Recall the approximation in~\eqref{eqn:GCWS_Prob_approx} 
\begin{align}\notag
    P[i_u^*=i_v^*]\approx P[(i_u^*,t_u^*)=(i_v^*,t_v^*)] =\textit{pGMM}(u,v),
\end{align}
which, initially, was purely an  empirical observation~\citep{Proc:Li_KDD15}. The recent work~\citep{Proc:Li_WWW21} hoped to explain the above approximation by developing a new algorithm based on extremal processes which is closely related to CWS. Nevertheless, characterizing this approximation for CWS remains an open problem. Figure~\ref{fig:MoreBits} provides a validation study on M-Noise and M-Image, by adding the lowest 1 bit (dashed curve) or 2 bits (dashed dot curves) of $t^*$. We can see that adding 1 bit of $t^*$  indeed helps slightly when we only use $b=1$ or $b=2$ bits for $i^*$. Using 2 bits for $t^*$ does not lead further improvements. When we already use $b=4$ or $b=8$ bits to encode $i^*$, then adding the information  for 1 bit from $t^*$ does not help in a noticeable manner.

In practice, we expect that practitioners would anyway need to use a sufficient number of bits for $i^*$ such as 8 bits. We therefore do not expect that using bits from $t^*$  would help. Nevertheless, if it is affordable, we would suggest using the lowest 1 bit of $t^*$ in addition to using $b$ bits for $i^*$.

\section{Combining GCWS with Count-Sketch}

GCWS generates high-dimensional binary  sparse inputs. With $k$ hashes and $b$ bits for encoding each hashed value, we obtain a vector of size $2^b \times k$ with exactly $k$ 1's, for each original data vector. While the (online) training cost is mainly determined by $k$ the number of nonzero entries, the model size is still proportional to $2^b\times k$. This might be an issue for the GPU memory when both $k$ and $b$ are large. In this scenario, the well-known method of ``count-sketch'' might be helpful for reducing the model size. 

Count-sketch~\citep{Article:Charikar_2004} was originally developed for recovering sparse signals (e.g., ``elephants'' or ``heavy hitters'' in network/database terminology). \cite{Proc:Weinberger_ICML2009} applied count-sketch as a dimension reduction tool. The work of~\cite{Proc:Li_NIPS11_Learning}, in addition to developing hash learning algorithm based on minwise hashing, also provided the thorough theoretical analysis for count-sketch in the context of estimating inner products. The conclusion from~\cite{Proc:Li_NIPS11_Learning}  is that,  to estimate inner products, we should use count-sketch (or very sparse random projections~\citep{Proc:Li_KDD07})  instead of the original (dense) random projections, because count-sketch is not only computationally much more efficient but also (slightly) more accurate, as far as the task of similarity estimation is concerned.

\vspace{0.1in}

We use this opportunity to review count-sketch~\citep{Article:Charikar_2004} and the detailed analysis in~\cite{Proc:Li_NIPS11_Learning}. The key step  is to independently and uniformly hash elements of the data vectors to buckets $\in\{1, 2, 3, ..., B\}$ and the hashed value is the weighted sum of the elements in the bucket, where the weights are generated from a random distribution which must be $\{-1, 1\}$ with equal probability (the reason will soon be clear).   That is, we have $h(i) = j$ with probability $\frac{1}{B}$, where $j \in\{1, 2, ..., B\}$. For convenience, we introduce an indicator function:
\begin{align}\notag
I_{ij} =\left\{\begin{array}{ll}
1 &\text{if } h(i) = j\\
0 &\text{otherwise}
\end{array}
\right.\end{align}
Consider two vectors $x, y\in\mathbb{R}^d$ and assume $d$ is divisible by $B$, without loss of generality.  We also generate a random vector $r_i$, $i  = 1 $ to $d$, i.i.d., with the following property:
\begin{align}
    E(r_i) = 0, \hspace{0.2in} E(r_i^2) = 1, \hspace{0.2in} E(r_i^3) = 0, \hspace{0.2in} E(r_i^4) = s
\end{align} 
Then we can generate $k$ (count-sketch) hashed values for $x$ and $y$, as follows, 
\begin{align}\label{eqn:z,w}
z_j = \sum_{i=1}^d x_{i} r_i I_{ij}, \hspace{0.5in} w_{j} = \sum_{i=1}^d y_{i} r_i I_{ij}, \ \ \ j = 1, 2, ..., B
\end{align}
The following Theorem~\ref{thm:cs} says that the inner product $<z,w> = \sum_{j=1}^B z_j w_j$ is an unbiased estimator of $<x,y>$. 
 
\begin{theorem}\citep{Proc:Li_NIPS11_Learning}\label{thm:cs}
\begin{align}
& E\{<z,w>\} = <x,y> = \sum_{i=1}^d x_{i}y_{i},\\
&\textit{Var}\{<z,w>\} = (s-1)\sum_{i=1}^dx_{i}^2y_{i}^2 + \frac{1}{B}\left[\sum_{i=1}^d x_{i}^2\sum_{i=1}^d y_{i}^2 + \left(\sum_{i=1}^d x_{i}y_{i}\right)^2-2\sum_{i=1}^d x_{i}^2y_{i}^2\right].
\end{align}
\end{theorem}

From the above theorem, we can see that we must have $E(r_i^4) = s = 1$. Otherwise, the variance $\textit{Var}\{<z,w>\}$ will end up with a positive term which does not vanish with increasing  $B$ the number of bins. There is only one such distribution, i.e., $\{-1, +1\}$ with equal probability. 

\vspace{0.1in}

Now we are ready to present our idea of combining GCWS with count-sketch. Recall that in the ideal scenario, we map the output of GCWS $(i^*, t^*)$ uniformly to a $b$-bit space of size $2^b$, denoted by $(i^*,t^*)_b$. For the original data vector $u$, we generate $k$ such (integer) hash values, denoted by $(i^*_{u,j},t^*_{u,j})_b$, $j = 1$ to $k$.  Then we concatenate the $k$ one-hot representations of $(i^*_{u,j},t^*_{u,j})_b$, we obtain a binary vector of size $d = 2^b\times k$ with exactly $k$ 1's.  Similarly, we have $(i^*_{v,j},t^*_{v,j})_b$, $j = 1$ to $k$ and the corresponding binary vector of length $2^b \times k$.  Once we have the sparse binary vectors, we can apply count-sketch with $B$ bins. For convenience, we denote 
\begin{align}
B = (2^b\times k)/m.
\end{align}
That is, $m$ represents the dimension reduction factor, and $m=1$ means no reduction. We can then generate the ($B$-bin) count-sketch samples, $z$ and $w$, as in~\eqref{eqn:z,w}.  We present the theoretical results on the inner product $<z,w>$ as the next theorem.

\begin{theorem}\label{thm:Pb_cs}
Consider original data vector $u$ and $v$. Denote $P_b = P\left\{(i_u^*, t^*_u) _b = (i_v^*, t_v^*)_b\right\}$. Vector $z$ of size $B$ is the ($B$-bin) count-sketch samples  generated from the concatenated binary vector  from the one-hot representation of $(i^*_{u,j},t^*_{u,j})_b$. Similarly, $w$ of size $B$ is the count-sketch samples for $v$. Also, denote 
\begin{align}
a = \sum_{j=1}^k 1\left\{(i^*_{u,j},t^*_{u,j})_b = (i^*_{v,j},t^*_{v,j})_b\right\}.
\end{align}
Conditioning on the GCWS output $(i^*_{u,j},t^*_{u,j})_b, (i^*_{v,j},t^*_{v,j})_b$ and using Theorem~\ref{thm:cs} with $s=1$, we  have
\begin{align}
& E\left\{<z,w>|\textit{GCWS}\right\} =a,\\
&Var\{<z,w>|\textit{GCWS}\} = \frac{1}{B}\left[k^2 + a^2-2a\right].
\end{align}
Note that $a$ is binomial $\textit{Bin}(k,P_b)$, i.e., $E(a) = kP_b, \textit{Var}(a) = kP_b(1-P_b)$.  Unconditionally, we have 
\begin{align}
& E\left\{<z,w>\right\} = kP_b,\\
&\textit{Var}\{<z,w>\} = \frac{1}{B}\left[k^2+k^2P_b^2-kP^2_b-kP_b\right] + kP_b(1-P_b).
\end{align}
Define the estimator $\hat{P}_b$ as
\begin{align}
\hat{P}_b = \frac{<z,w>}{k}.
\end{align}
Then
\begin{align}
&E\left(\hat{P}_b\right) = P_b,\\\label{eqn:var_Pb_cs}
&\textit{Var}\left(\hat{P}_b\right) = \frac{P_b(1-P_b)}{k} +  \frac{1}{B}\left[1+P^2_b - P_b^2/k - P_b/k\right]. 
\end{align}
\end{theorem}

\vspace{0.2in}

The variance  $\textit{Var}\left(\hat{P}_b\right)$ \eqref{eqn:var_Pb_cs} in Theorem~\ref{thm:Pb_cs} has two parts. The first term $\frac{P_b(1-P_b)}{k}$ is the usual variance without using count-sketch. The second term $\frac{1}{B}\left[1+P^2_b - P_b^2/k - P_b/k\right]$ represents the additional variance due to count-sketch. The hope is  that the additional variance would not be too large, compared with  $\frac{P_b(1-P_b)}{k}$. Let $B = (2^b\times k)/m$, then the second term can be written as $\frac{1}{k}\frac{m}{2^b}\left[1+P^2_b - P_b^2/k - P_b/k\right]$.

\newpage

Thus, to assess the impact of count-sketch, we just need to compare $P_b(1-P_b)$ with $\frac{m}{2^b}\left[1+P^2_b - P_b^2/k - P_b/k\right]$, which is approximately $\frac{m}{2^b}\left[1+P^2_b\right]$, as $k$ has to be sufficiently large. Recall from Theorem~\ref{thm:pGMM_b} $P_b = J + \frac{1}{2^b}(1-J)$, where $J = \textit{pGMM}(u,v;p) =  \frac{\sum_{i=1}^{2D}\left(\min\{\tilde{u}_i,\tilde{v}_i\}\right)^p}{\sum_{i=1}^{2D}\left( \max\{\tilde{u}_i,\tilde{v}_i\}\right)^p}$.  Therefore, we can resort to the following ``ratio'' to assess the impact of count-sketch on the variance: 
\begin{align}\label{eqn:cs_ratio}
R = R(b,J,m)= \frac{m}{2^b}\frac{1+P_b^2}{P_b(1-P_b)}, \hspace{0.2in} \text{where } P_b = J + \frac{1}{2^b}(1-J). 
\end{align}

\begin{figure}[h]


\mbox{\hspace{-0.25in}
\includegraphics[width=2.35in]{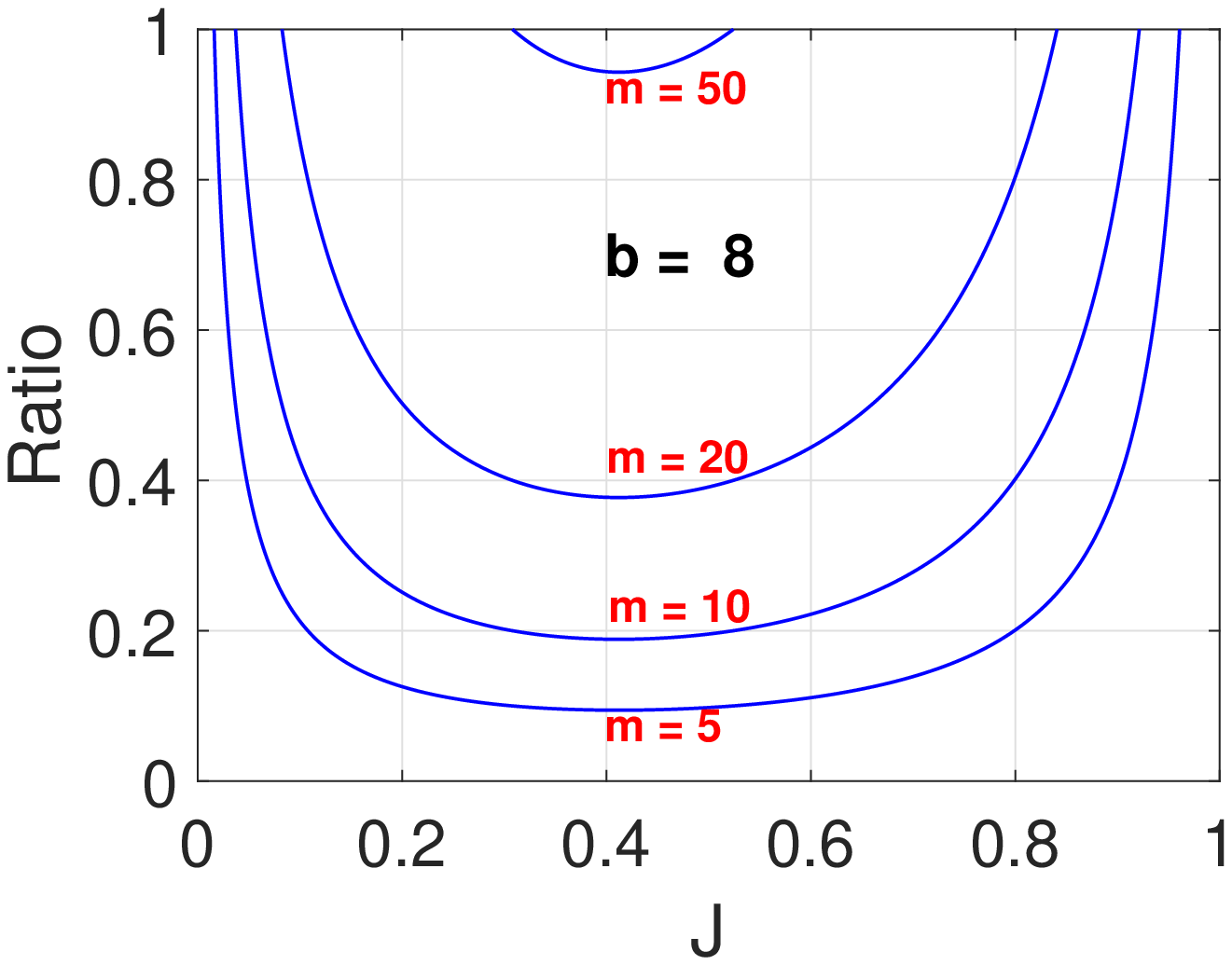}\hspace{-0.12in}
\includegraphics[width=2.35in]{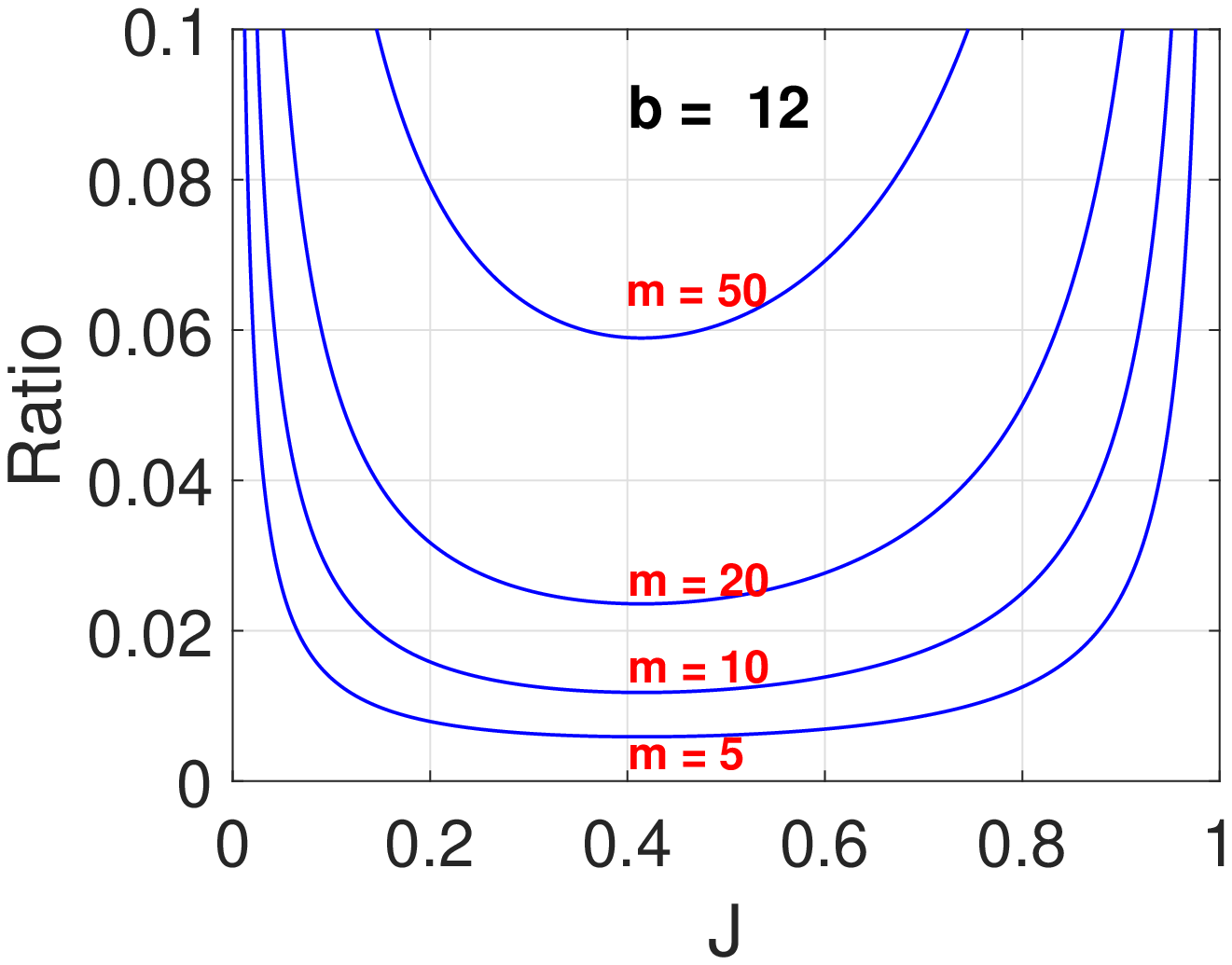}\hspace{-0.12in}
\includegraphics[width=2.35in]{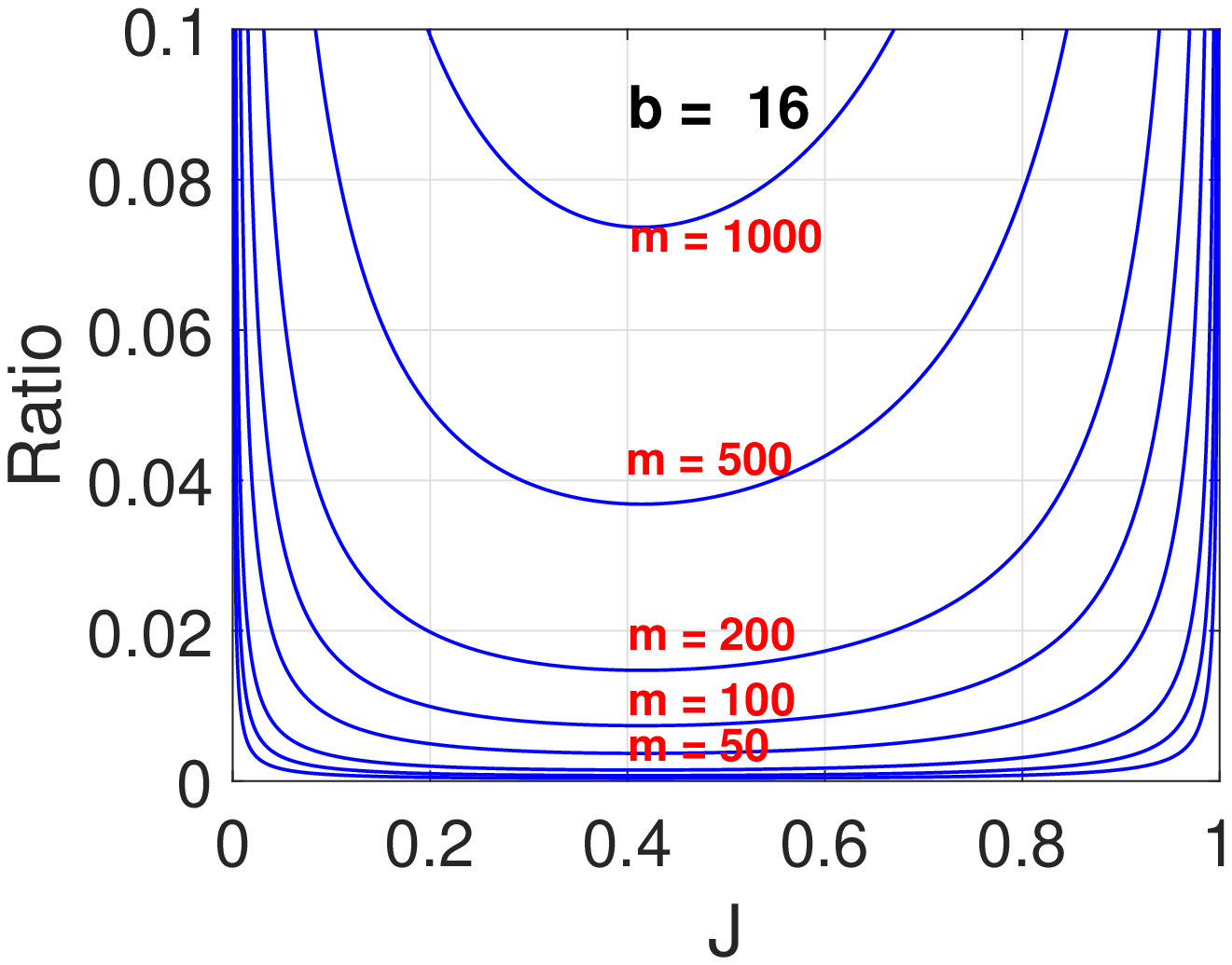}
}

\vspace{-0.15in}

\caption{The ratio~\eqref{eqn:cs_ratio} $R = \frac{m}{2^b}\frac{1+P_b^2}{P_b(1-P_b)}$, where $P_b = J + \frac{1}{2^b}(1-J)$ and $J$ the pGMM similarity of interest, for $b \in\{8, 12, 16\}$ and a series of $m$ values. Ideally, we would like to see large $m$ and small ratio values.}\label{fig:cs_ratio}\vspace{0.2in}
\end{figure}


As shown in Figure~\ref{fig:cs_ratio}, when $b=16$, the ratio is small even for $m=1000$. However, when $b=8$, the ratio is not too small after $m>10$; and hence  we  should only expect a saving by an order magnitude if $b=8$ is used. The choice of $b$ has, to a good extent, to do with  $D$, the original data dimension. When the data are extremely high-dimensional, say $D=2^{30}$, we expect an excellent storage savings can be achieved by using a large $b$  and a large $m$.

\begin{figure}[h]
\begin{center}

\mbox{
\includegraphics[width=2.4in]{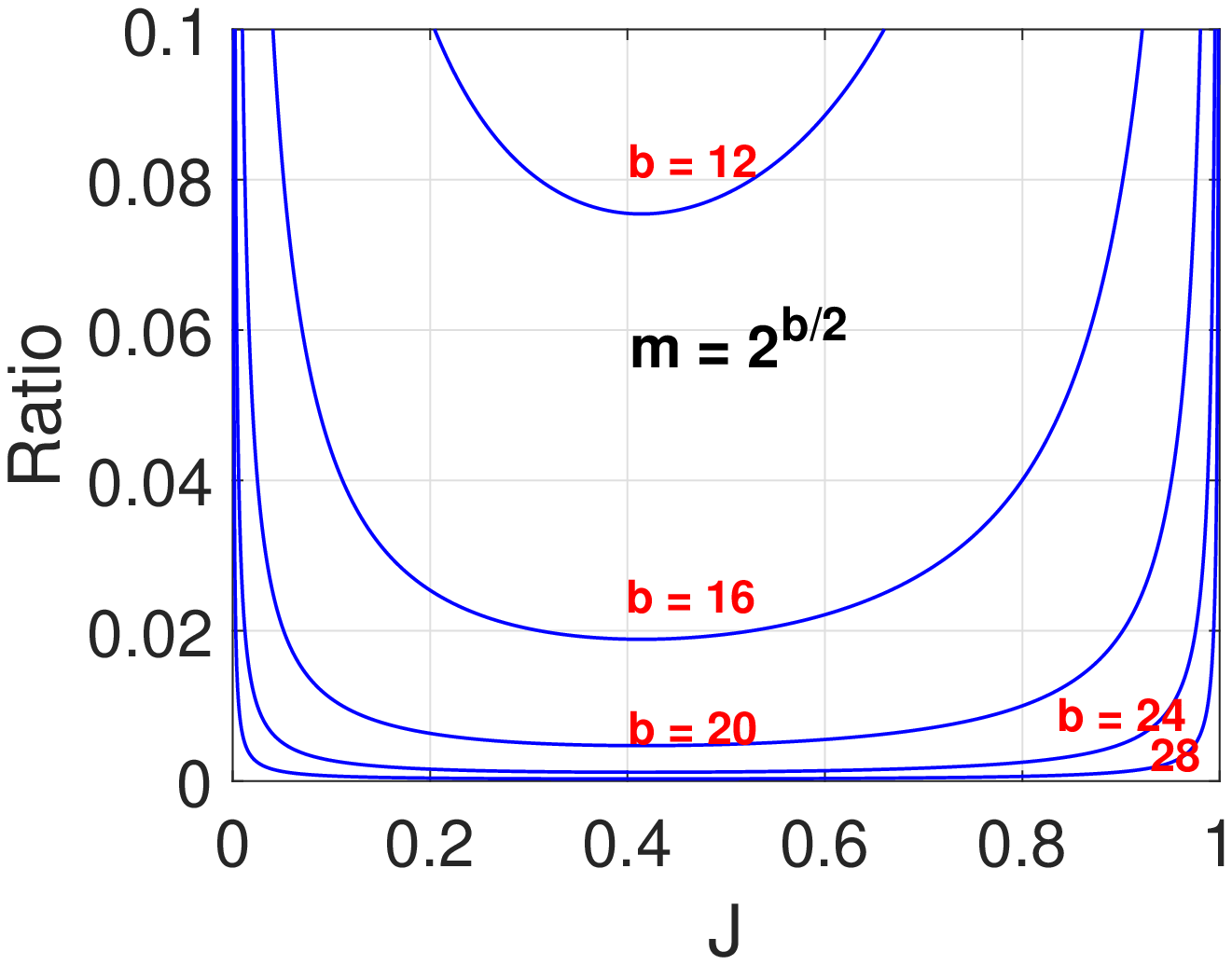}
\includegraphics[width=2.4in]{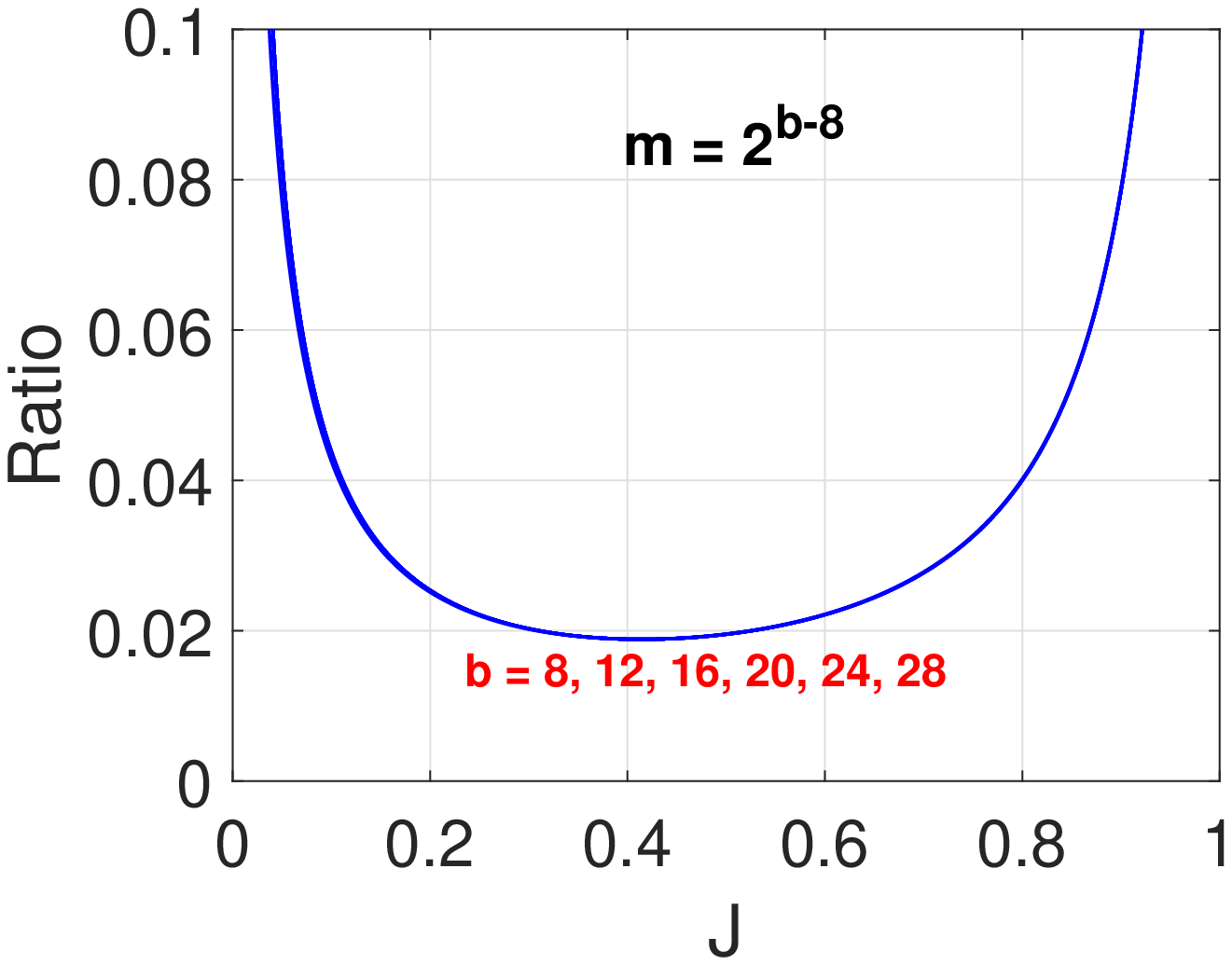}
}

\end{center}

\vspace{-0.2in}

\caption{The ratio~\eqref{eqn:cs_ratio} for two strategy of choosing $m$. In the left panel, we let $m = 2^{b/2}$ (i.e., always using half of the bits). In the right panel, for $b\geq 8$ we let $m = 2^{b-8}$ (i.e., always using  8 bits).  }\label{fig:cs_ratio2}
\end{figure}

For practical consideration, we consider two strategies for choosing $m$, as illustrated in  Figure~\ref{fig:cs_ratio2}. The first strategy (left panel) is ``always using half of the bits'', that is, we let $m = 2^{b/2}$. This method is probably a bit conservative for $b\geq 16$. The second strategy (right panel) is ``always using 8 bits'' which corresponds to almost like a single curve, because $R = \frac{1}{2^8}\frac{1+P_b^2}{P_b(1-P_b)}$ in this case and $P_b = J + (1-J)\frac{1}{2^b}\approx J$ when $b\geq 8$.

\vspace{0.2in}

We believe the ``always using 8 bits'' strategy might be a good practice. In implementation, it is convenient to use 8 bits (i.e., one byte). Even when we really just need 5 or 6 bits in some cases, we might as well simply use one byte to avoid the trouble of performing bits-packing.

\vspace{0.1in}

\begin{figure}[t]
\begin{center}
\mbox{
\includegraphics[width=2.2in]{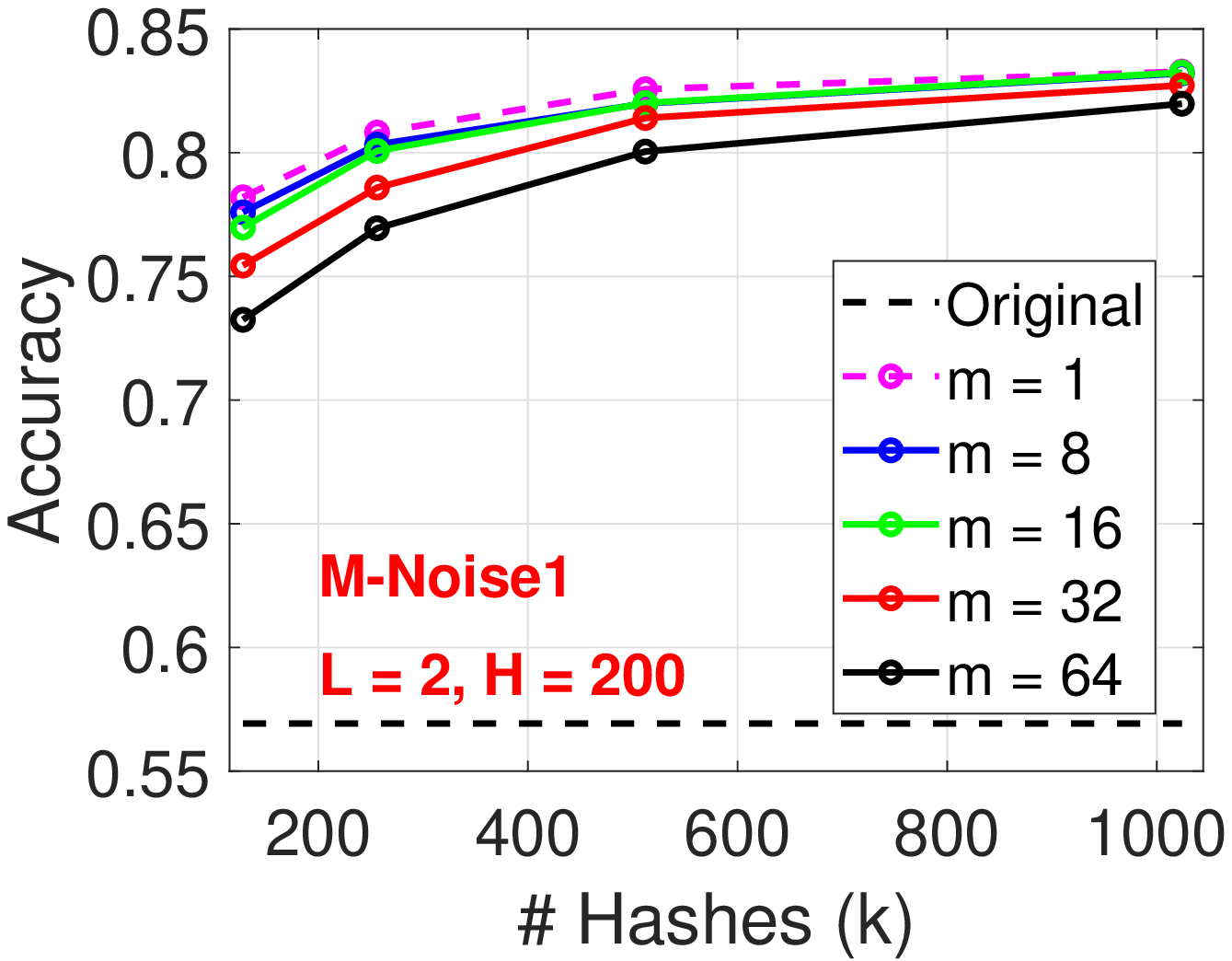}
\includegraphics[width=2.2in]{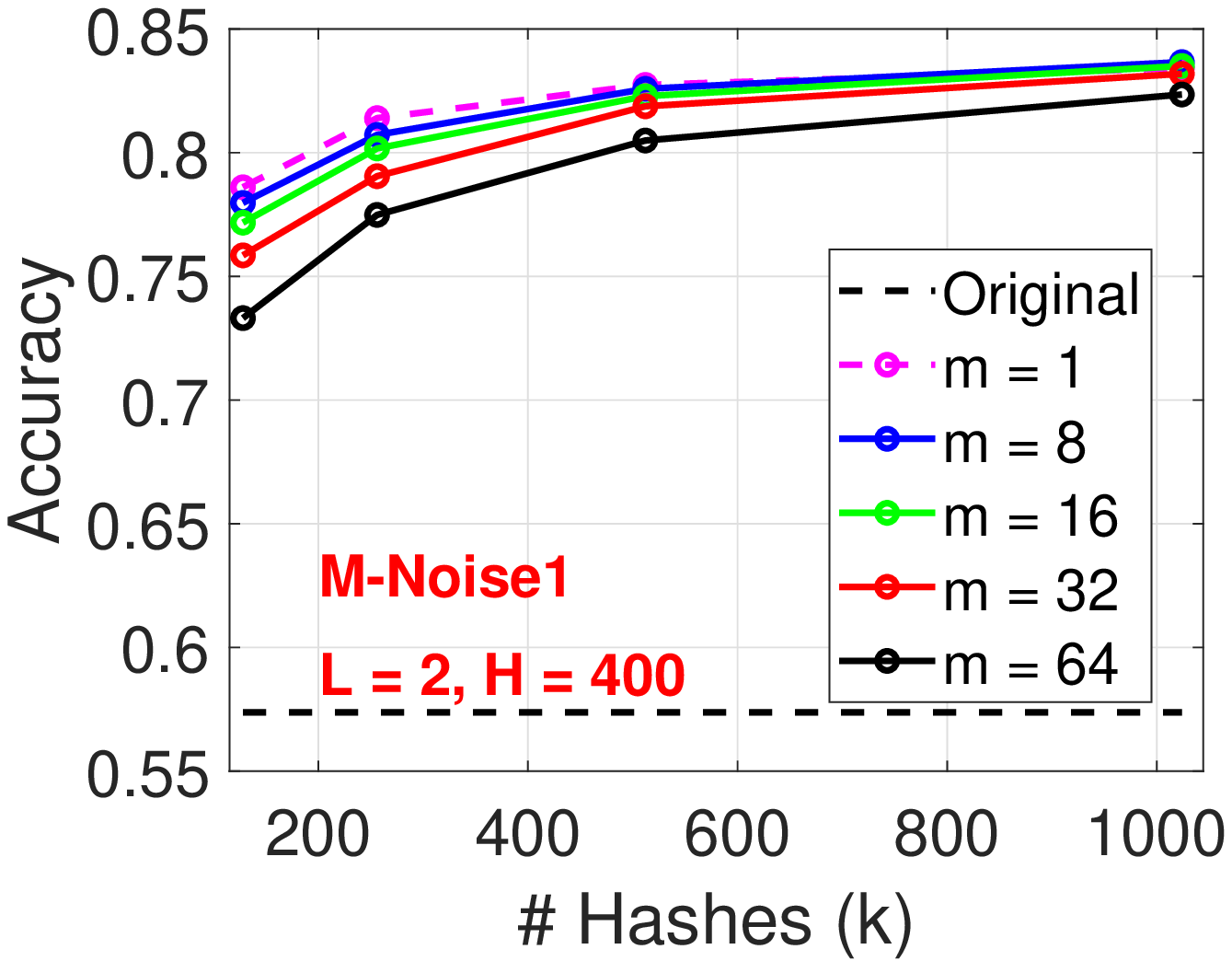}
\includegraphics[width=2.2in]{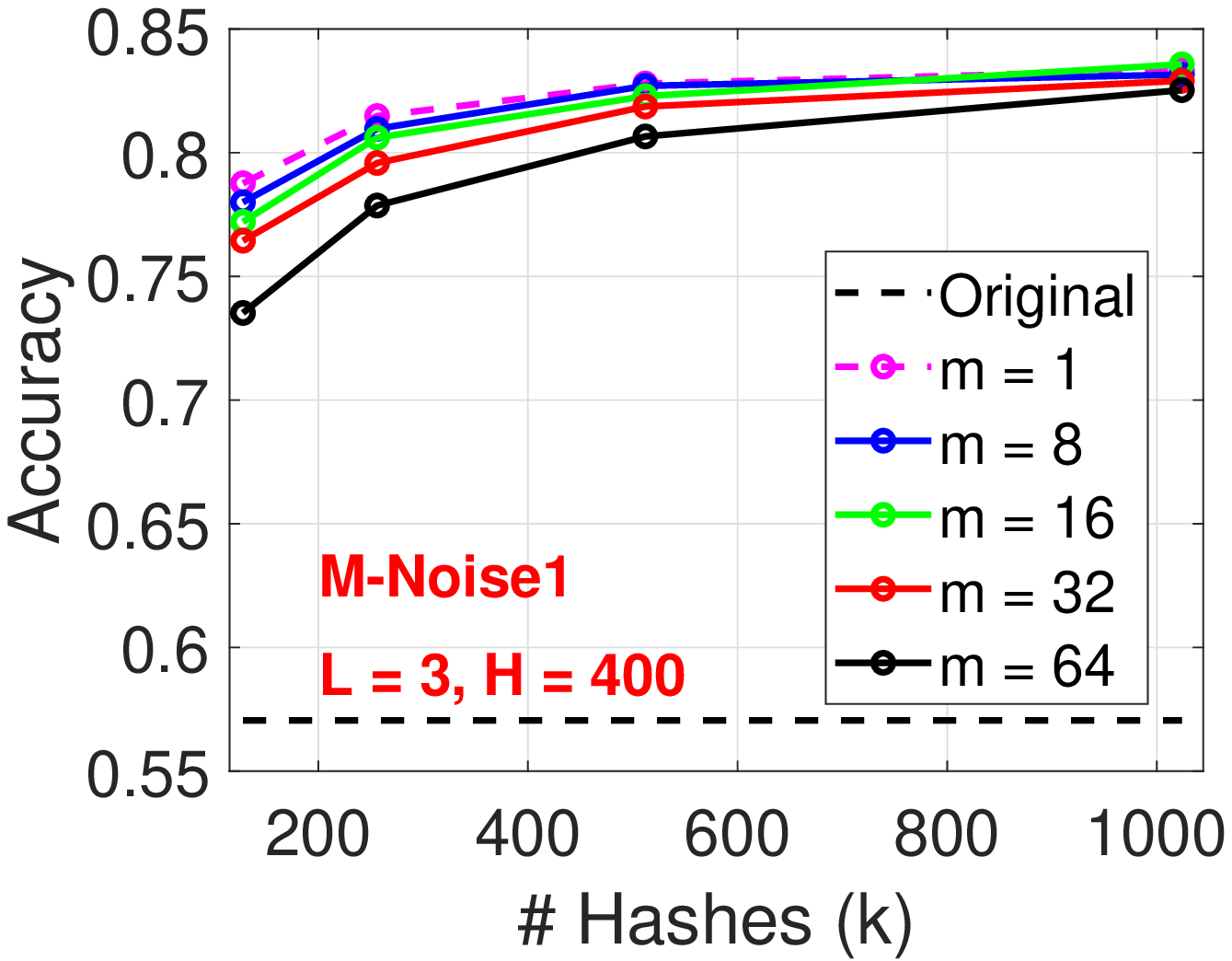}
}

\mbox{
\includegraphics[width=2.2in]{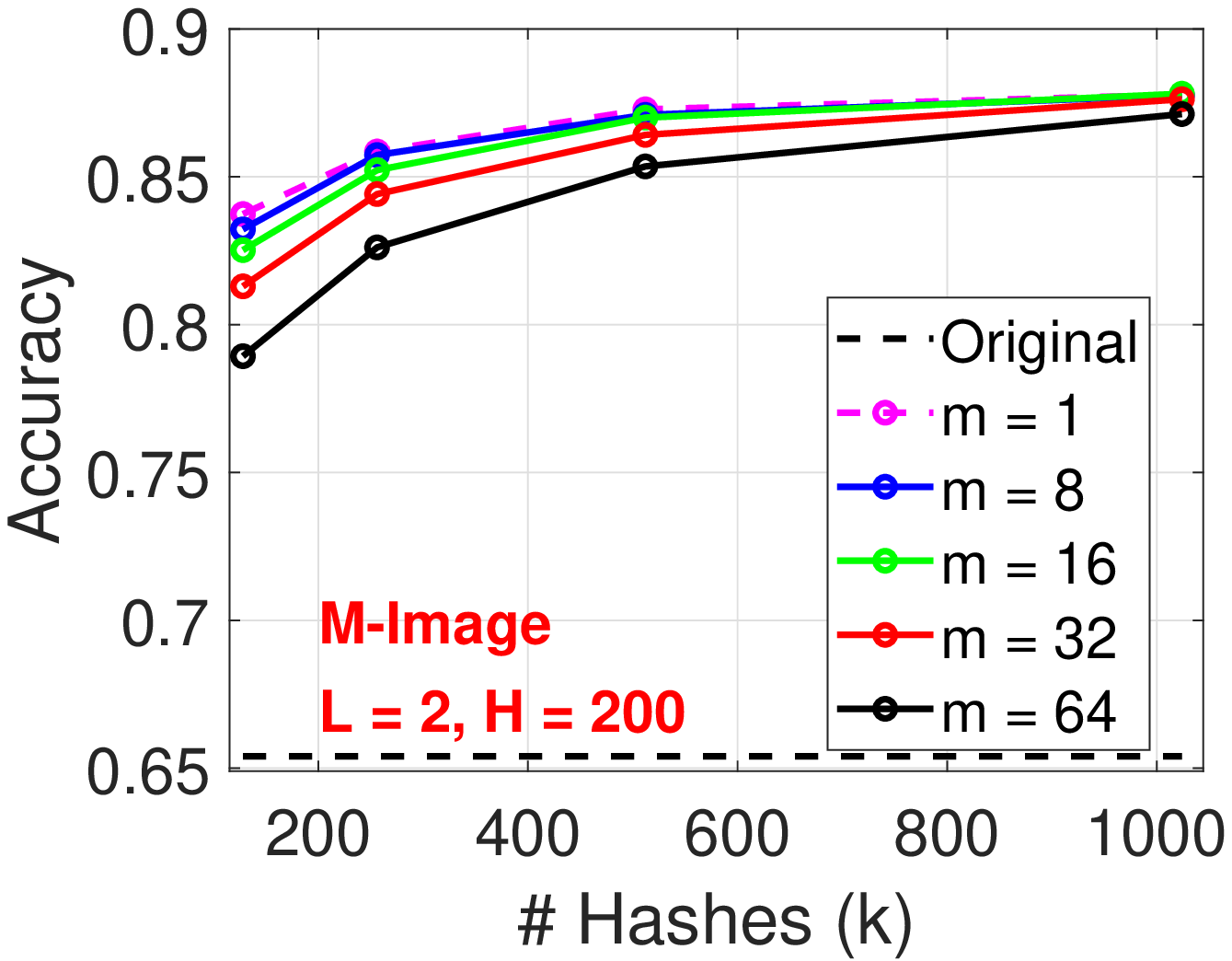}
\includegraphics[width=2.2in]{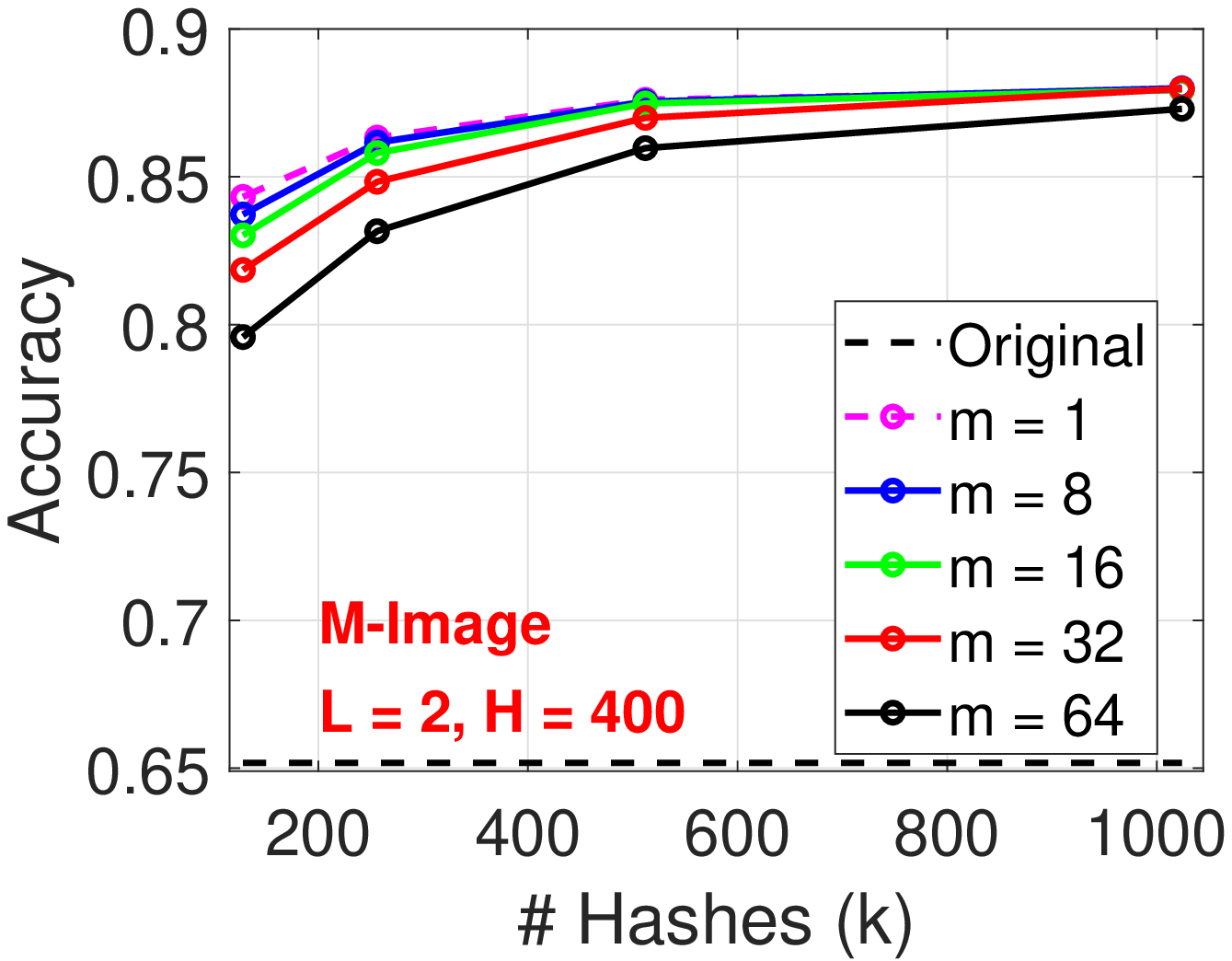}
\includegraphics[width=2.2in]{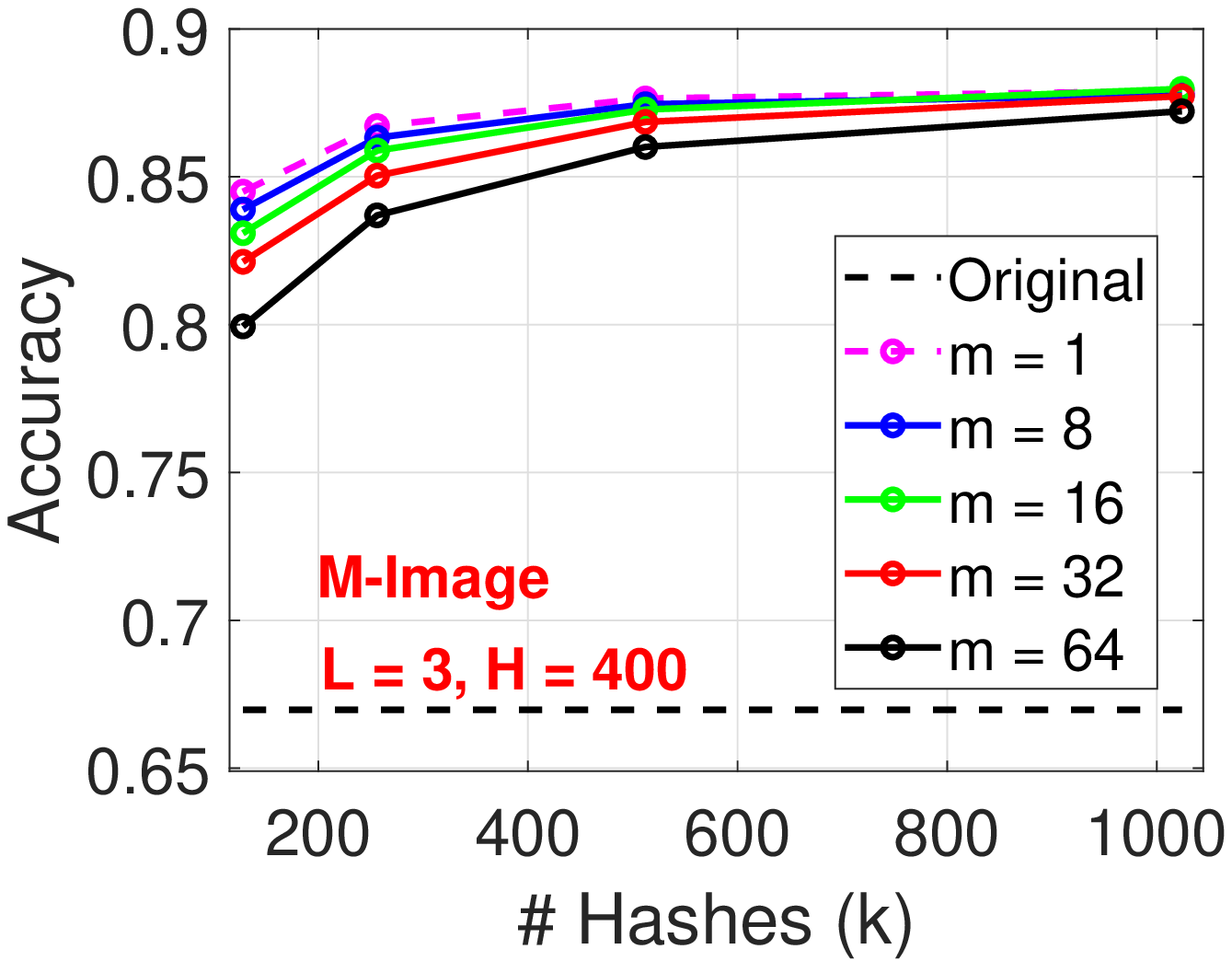}
}
\end{center}

\vspace{-0.15in}

\caption{GCWS combined with count-sketch for reducing the model size by a factor of $m$, where $m = (2^b\times k)/B$ and $B$ is the number of bins in the count-sketch step.  $m=1$ means there is no reduction. We can see that at least with $m=8$ or $m=16$, there is  no obvious loss of accuracy.  }\label{fig:cs}
\end{figure}

We conclude this section by providing an empirical study on M-Noise1 and M-Image and summarize the results in Figure~\ref{fig:cs}. For both datasets, when $m<=16$, we do not see an obvious drop of accuracy, compared with $m=1$ (i.e., without using count-sketch). This confirms the effectiveness of our proposed procedure by combining GCWSNet with count-sketch. 

\section{GCWS as a Robust Power Transformation}

In Algorithm~\ref{alg:GCWS}, this step (on nonzero entries) 
\begin{align}
t_i\leftarrow \left\lfloor p\frac{\log \tilde{u}_i }{r_i}+\beta_i\right\rfloor
\end{align}
suggests that GCWS can be viewed as a robust power transformation on the data. It should be clear that GCWS is not simply taking the log-transformation on the original data.  We will compare  three different strategies for data preprocessing. (i) GCWS; (ii) directly feeding the power transformed data (e.g., $u^p$) to neural nets; (iii) directly feeding the log-power transformed data (e.g., $p\log u$) to neural nets. 

Even though GCWSNet is obviously not a tree algorithm, there might be some interesting hidden connections. Basically, in GCWS, data entries with (relatively) higher values would be more likely to be picked (i.e., whose locations are chosen as $i^*$'s) in a probabilistic manner. Recall that in  trees~\citep{Book:Friedman_83,Book:Hastie_Tib_Friedman}, only the relative orders of the data values matter, i.e., trees are invariant to any monotone transformation. GCWS is somewhat in-between. It is affected by monotone transformations but not ``too much''.

Naively applying power transformation on the original data might encounter practical problems. For example, when data entries contain large values such as ``1533'' or ``396'', applying a powerful transformation such as $p=80$ might lead to overflow problems and other issues such as loss of accuracy during computations. Using the log-power transformation, i.e., $p\times \log u$, should alleviate many problems but new issues might arise. For example, when data entries ($u$) are small. How to handle zeros is another headache with log-power transformation.

\begin{figure}[h]
\begin{center}

\vspace{-0.15in}

\mbox{
\includegraphics[width=2.2in]{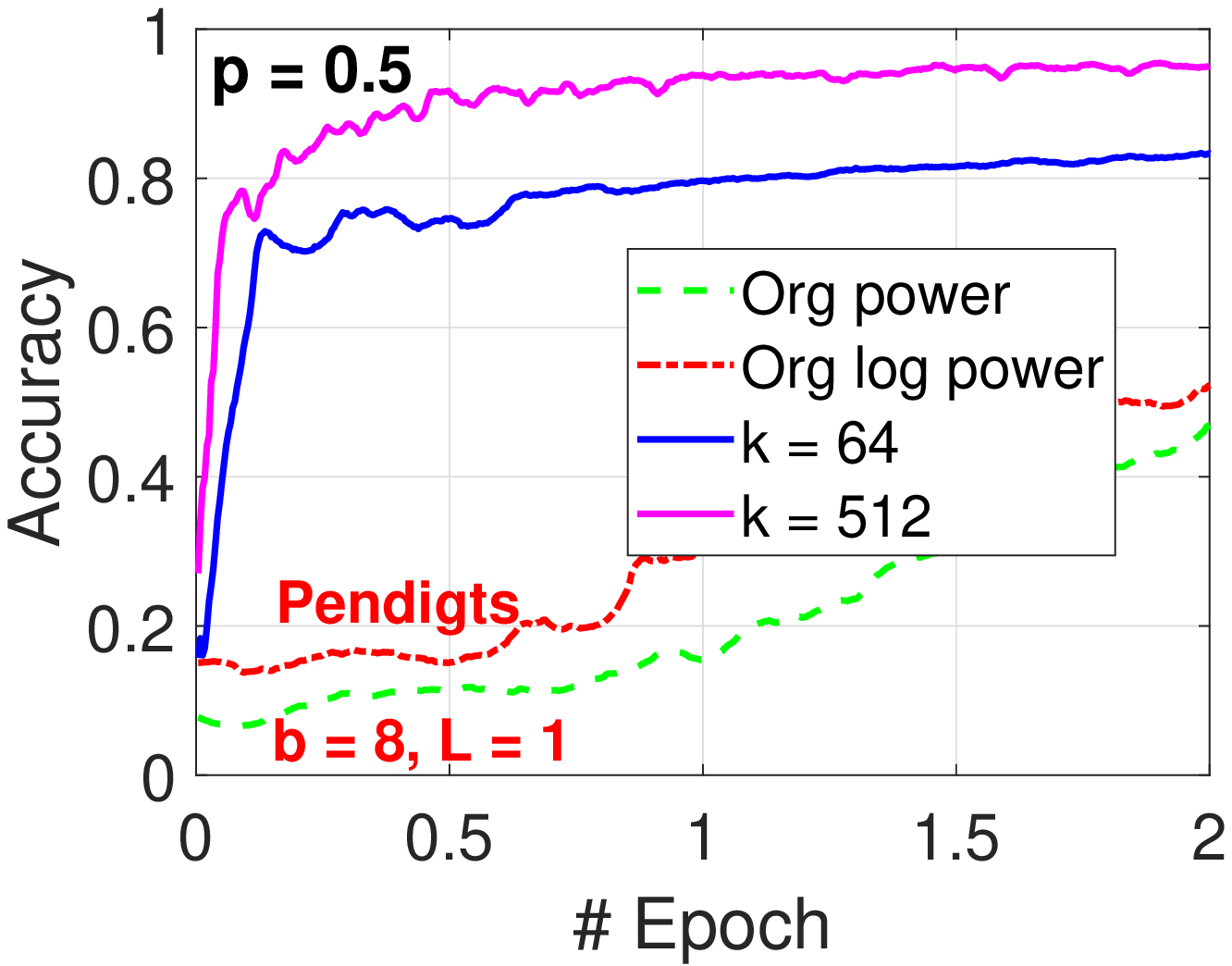}
\includegraphics[width=2.2in]{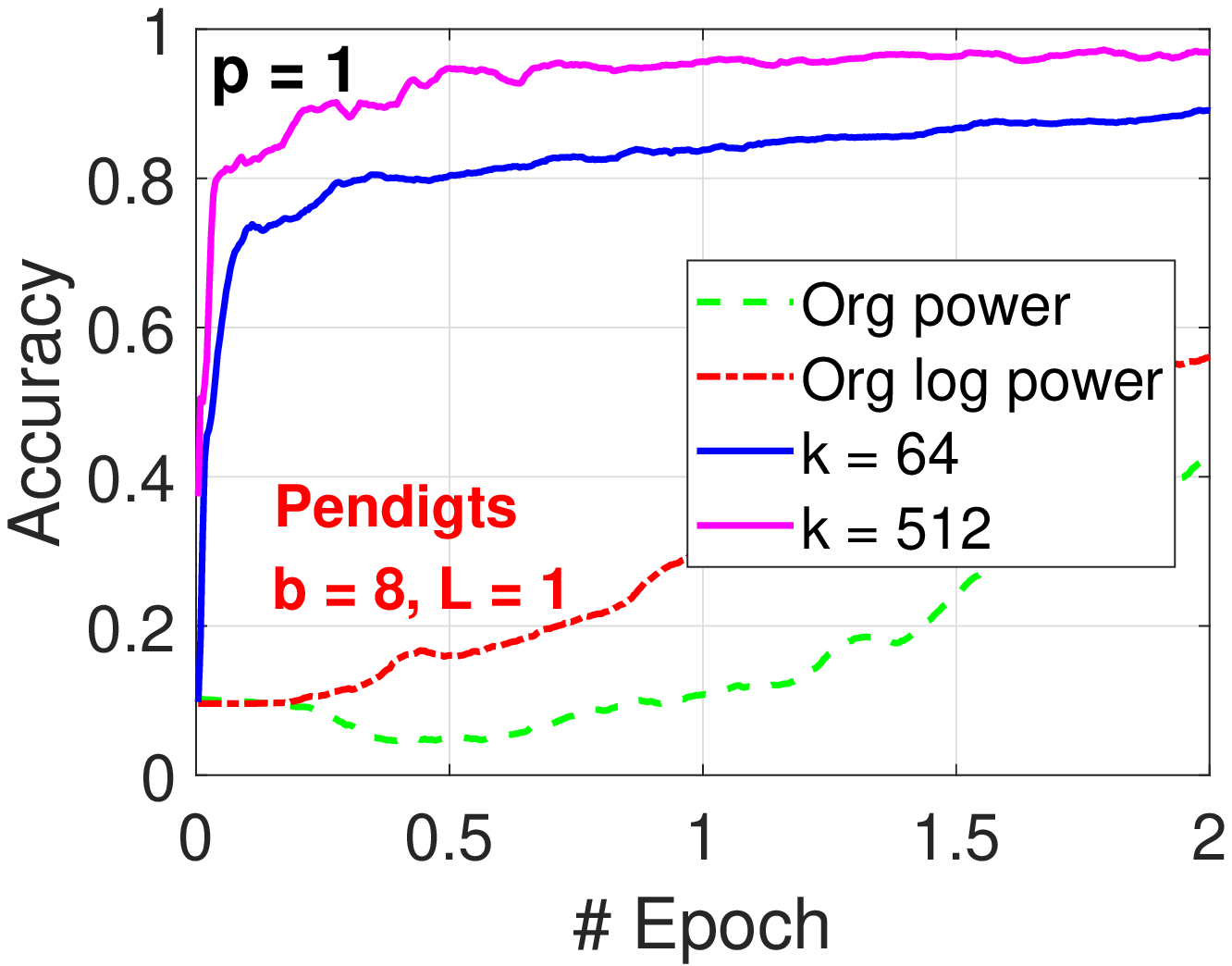}
\includegraphics[width=2.2in]{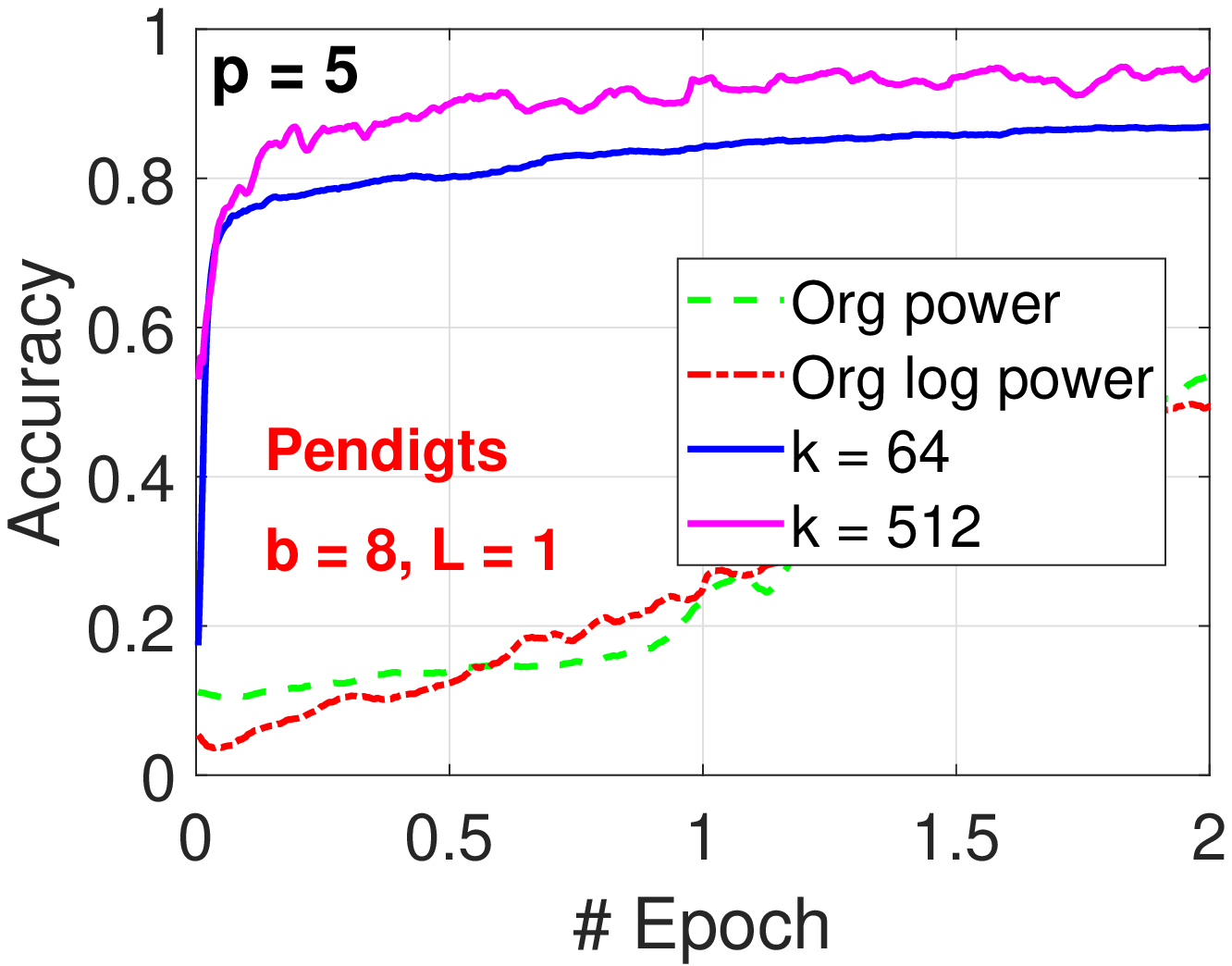}
}

\mbox{
\includegraphics[width=2.2in]{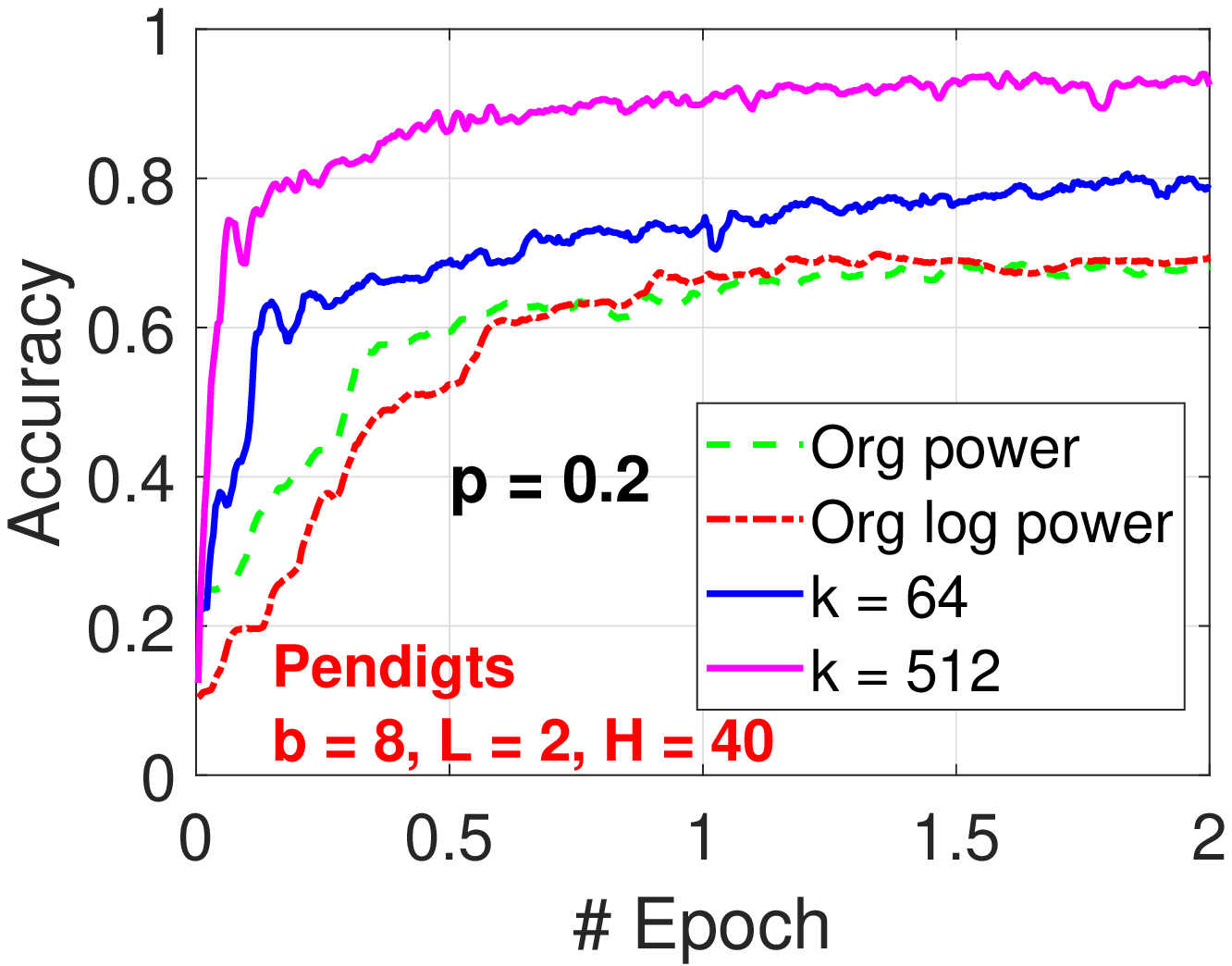}
\includegraphics[width=2.2in]{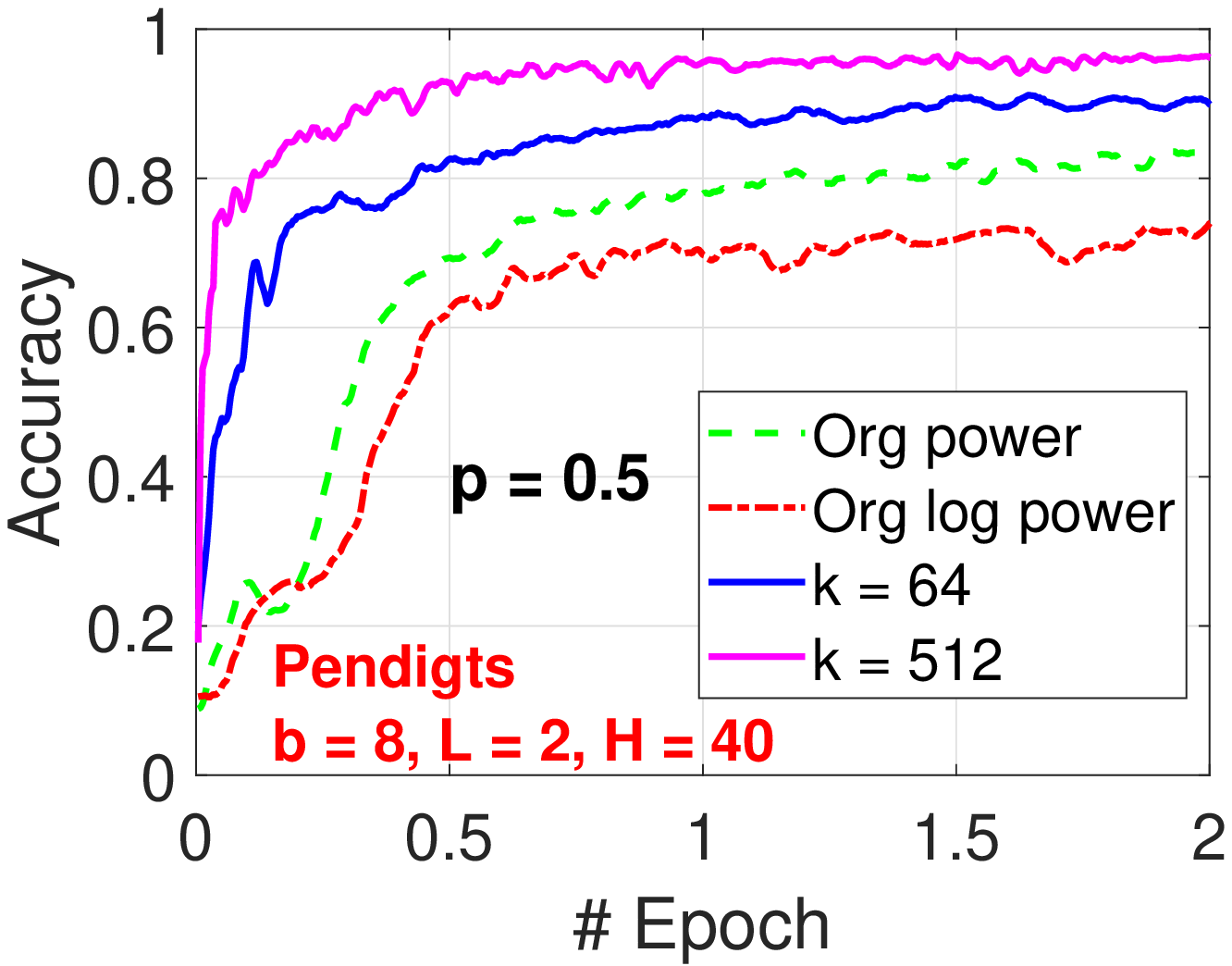}
\includegraphics[width=2.2in]{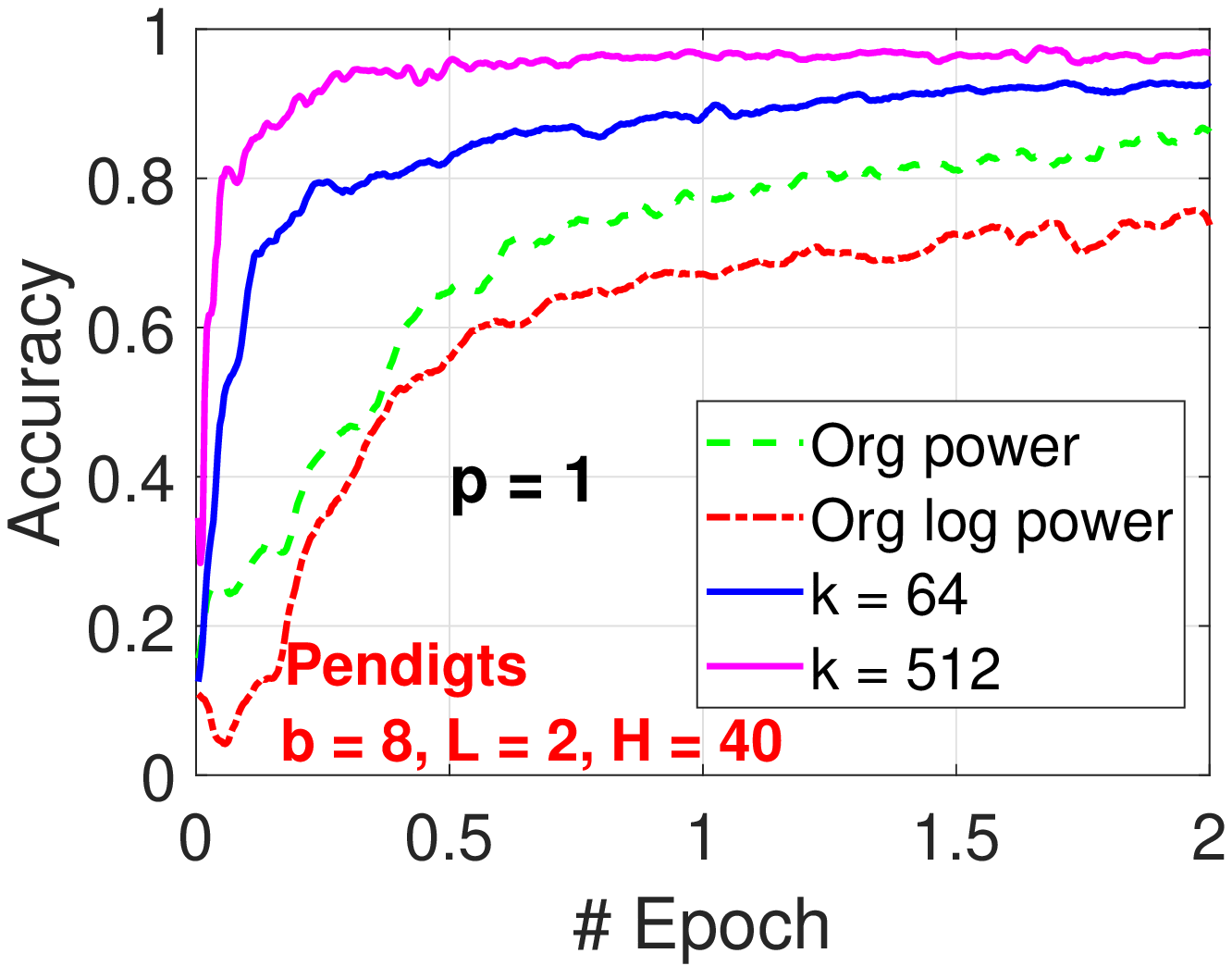}
}

\mbox{
\includegraphics[width=2.2in]{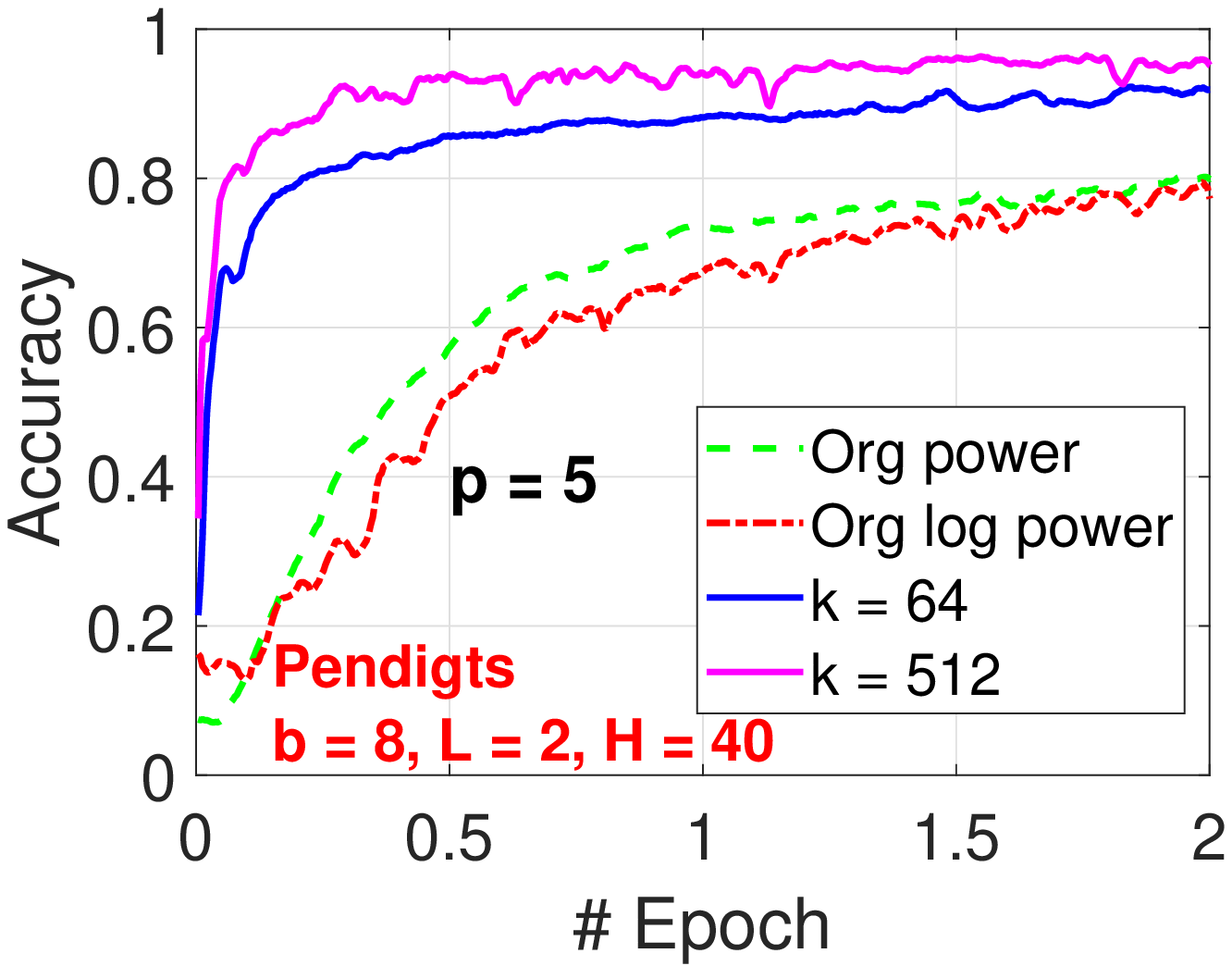}
\includegraphics[width=2.2in]{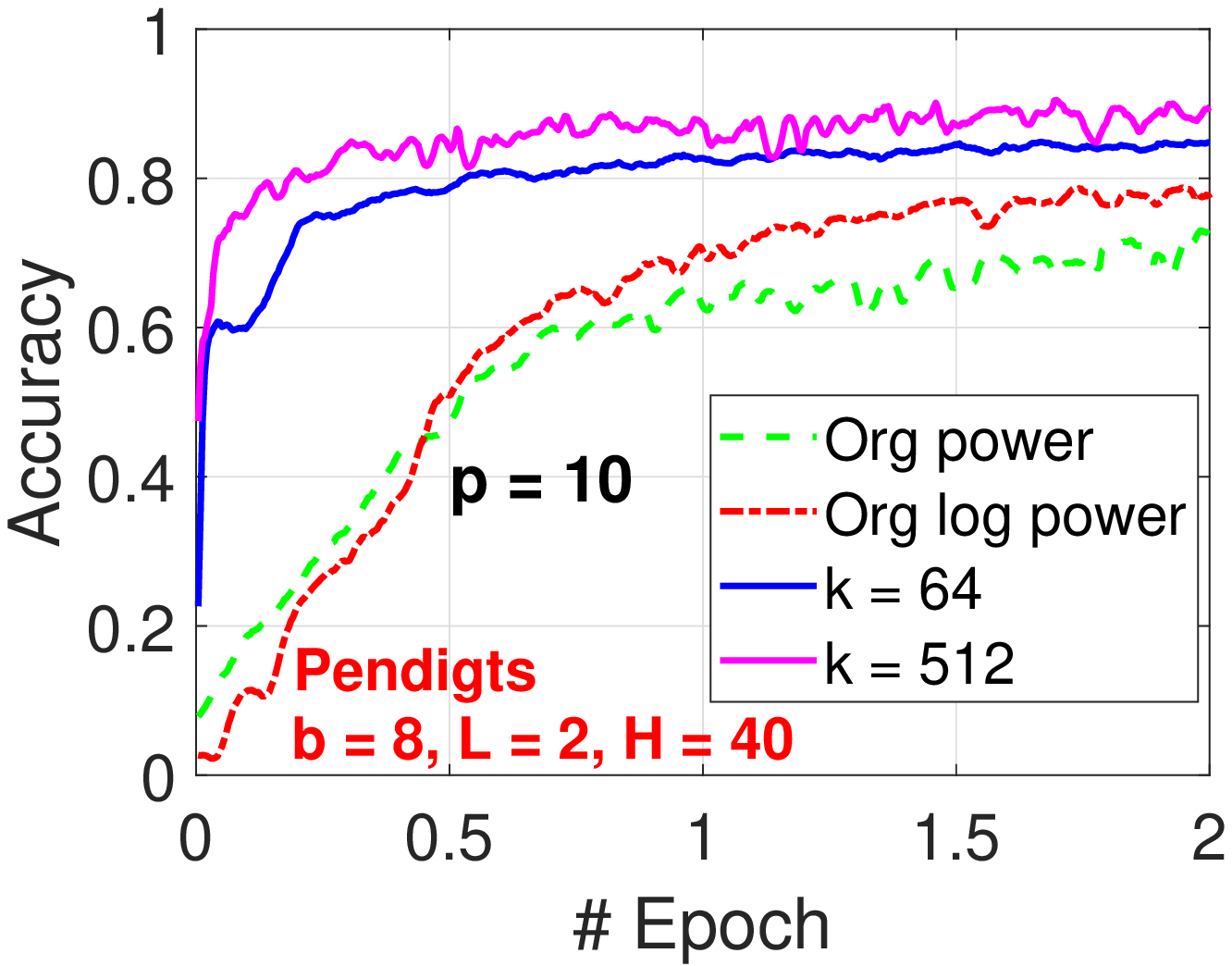}
\includegraphics[width=2.2in]{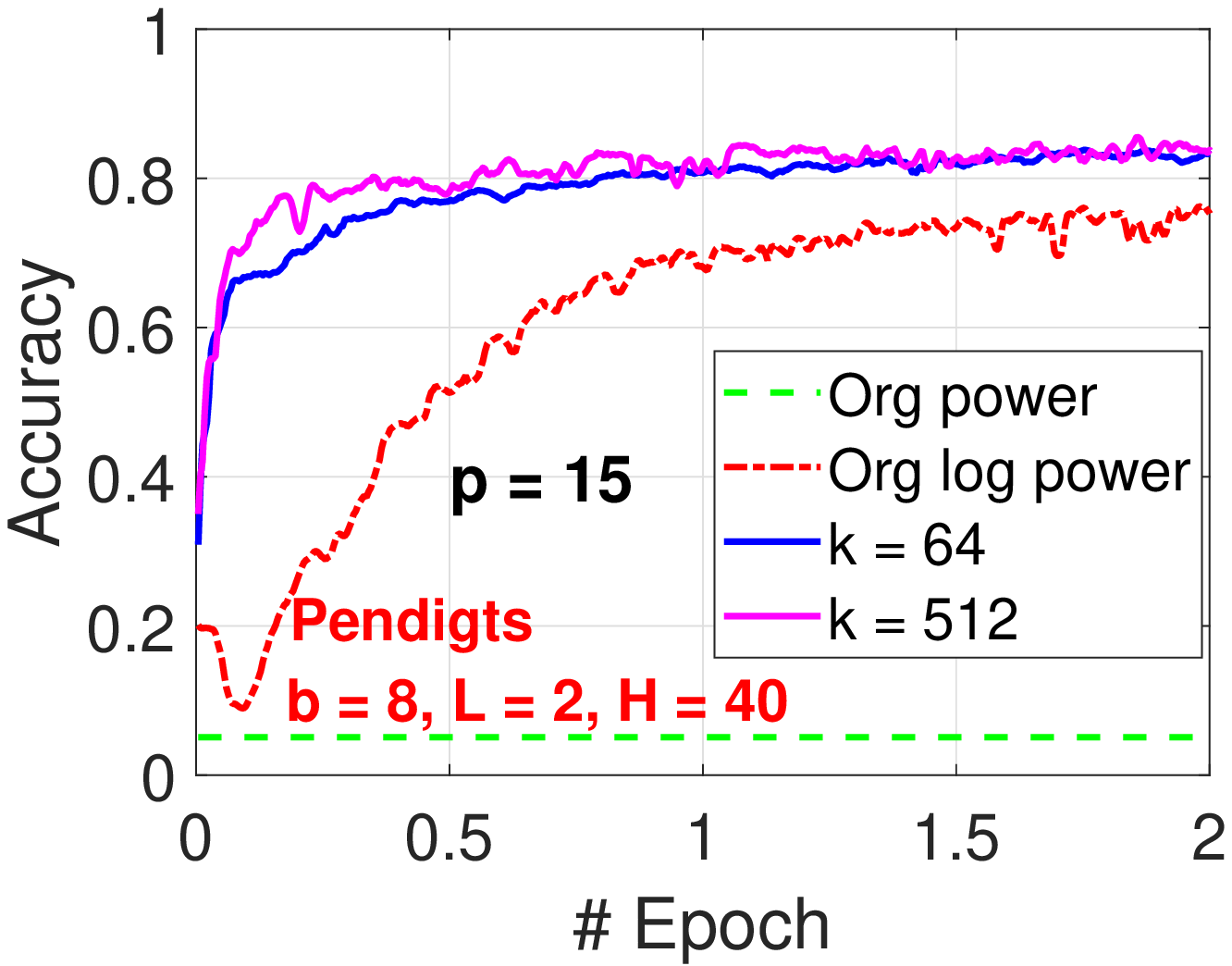}
}

\end{center}

\vspace{-0.2in}

\caption{GCWS with a wide range of $p$ values, on the UCI Pendigits dataset, for $b=8$ and $k\in\{64, 512\}$.  We can see that using $p\in[0.5,\ 5]$ produces the best results. Also, the results of GCWSNet are still reasonable even with very large or very small $p$ values. In comparison, we apply the power transformation (i.e., $u^p$) and log-power transformation (i.e., $p\log u$, treating $u\log 0 = 0$) on the original data and feed the transformed data directly to neural nets. Their results are not as accurate compared with GCWSNet. Also, the power transformation encountered numerical issues with~$p\geq 15$.  }\label{fig:Pendigits_power}\vspace{0.2in}
\end{figure}

\vspace{0.1in}
Here, we provide an experimental study on the UCI Pendigits dataset, which contains positive integers features (and many zeros). The first a few entries of the training dataset are ``8 1:47 2:100 3:27 4:81 5:57 6:37 7:26 10:23 11:56''  (in LIBSVM format). As shown in Figure~\ref{fig:Pendigits_power}, after $p\geq 15$, directly applying the power transformation (i.e., $u^p$) makes the training fail. The log-transformation (i.e., $p\log u$) seems to be quite robust in this dataset, because we use a trick by letting $p\log 0 = 0$, which seems to work well for this dataset. On the other hand, GCWS with any $p$ produces reasonable predictions, although it appears that $p\in[0.5,\  5]$ (which is quite a wide range) is the optimal range for GCWSNet. This set of experiments on the Pendigits dataset confirms the robustness of GCWSNet with any $p$. Additionally, the experiments once again verify that GCWSNet converges really fast and achieves a reasonable accuracy as early as in 0.1 epoch.

\newpage

\begin{figure}[t]
\begin{center}

\mbox{
\includegraphics[width=2.2in]{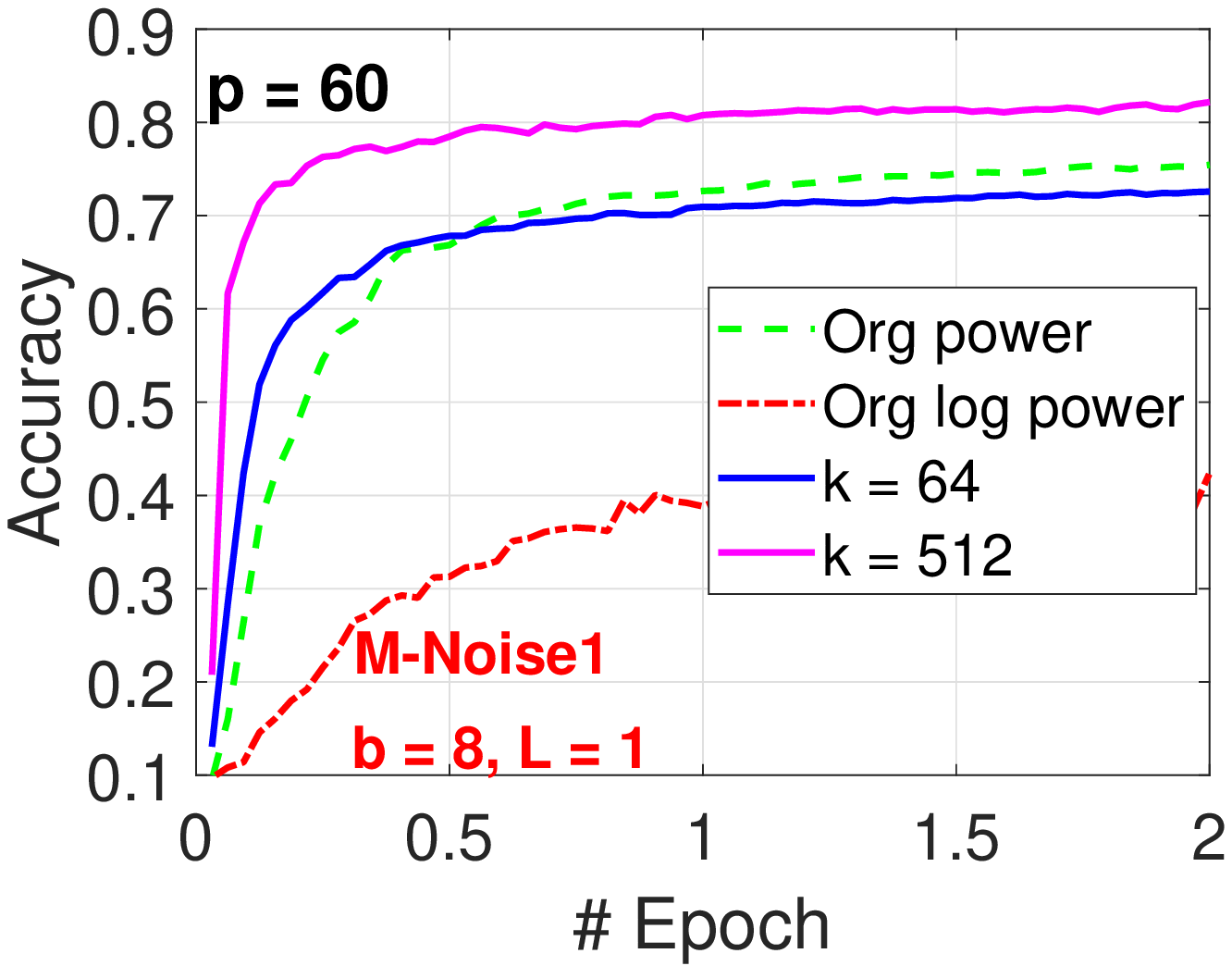}
\includegraphics[width=2.2in]{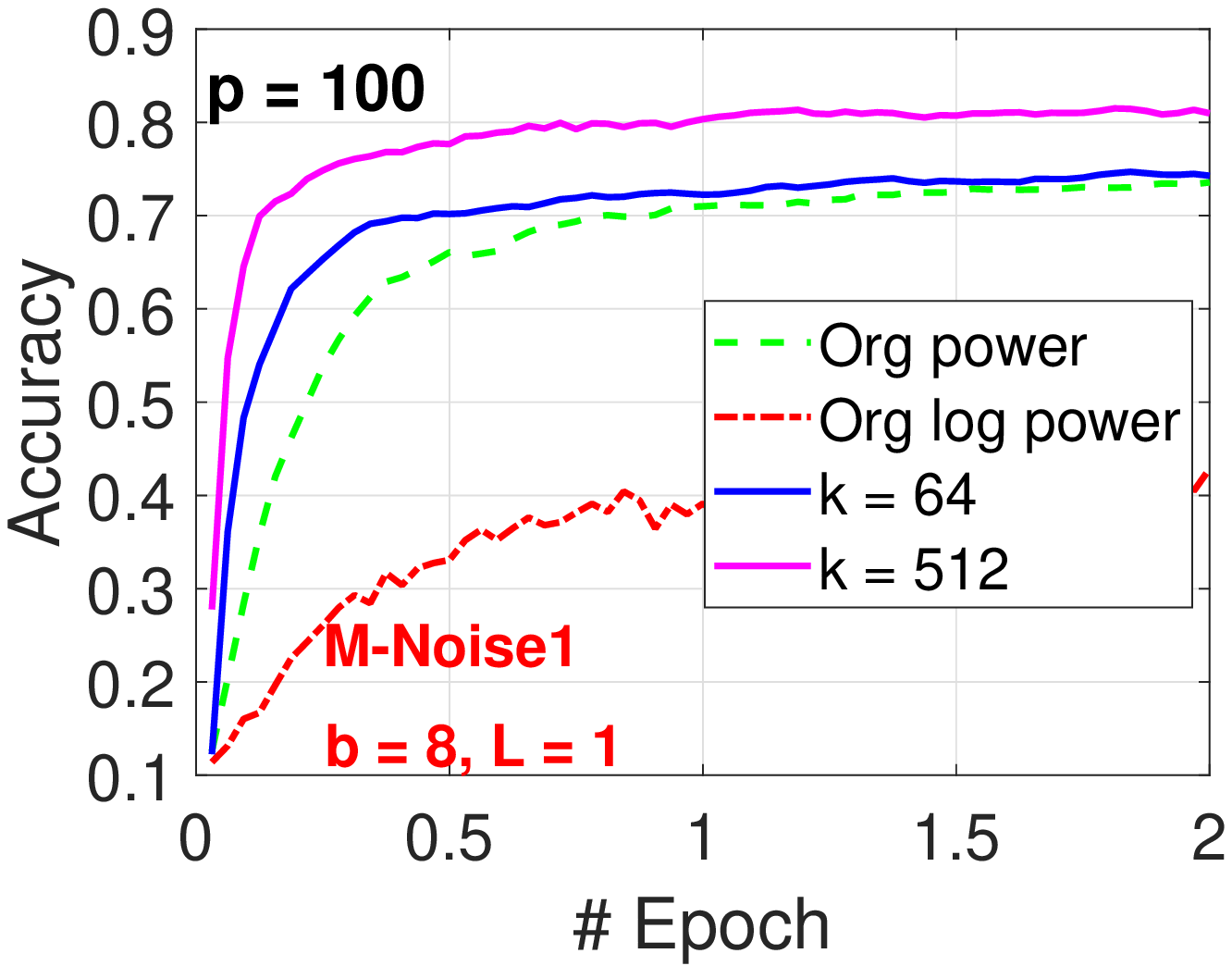}
\includegraphics[width=2.2in]{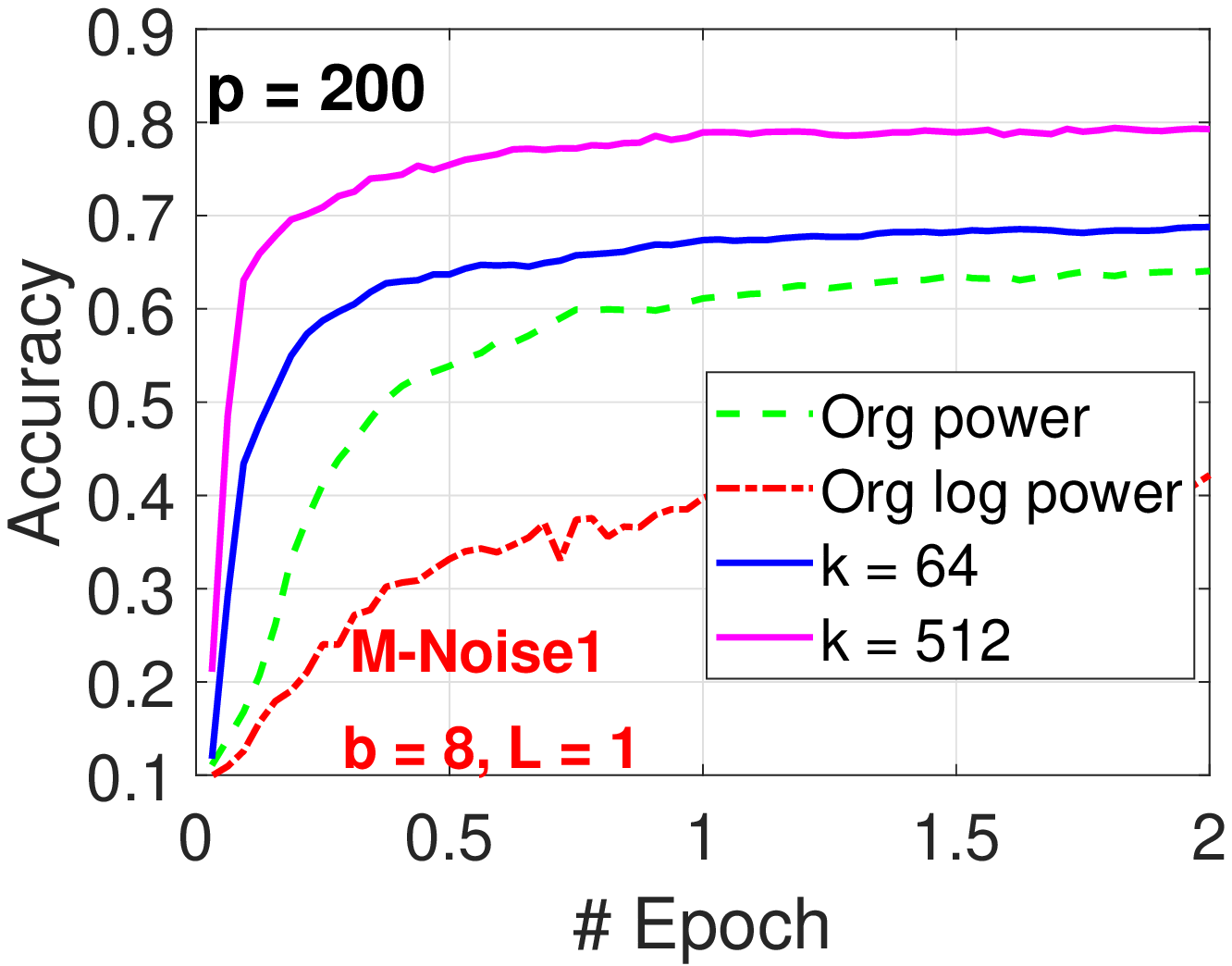}
}

\mbox{
\includegraphics[width=2.2in]{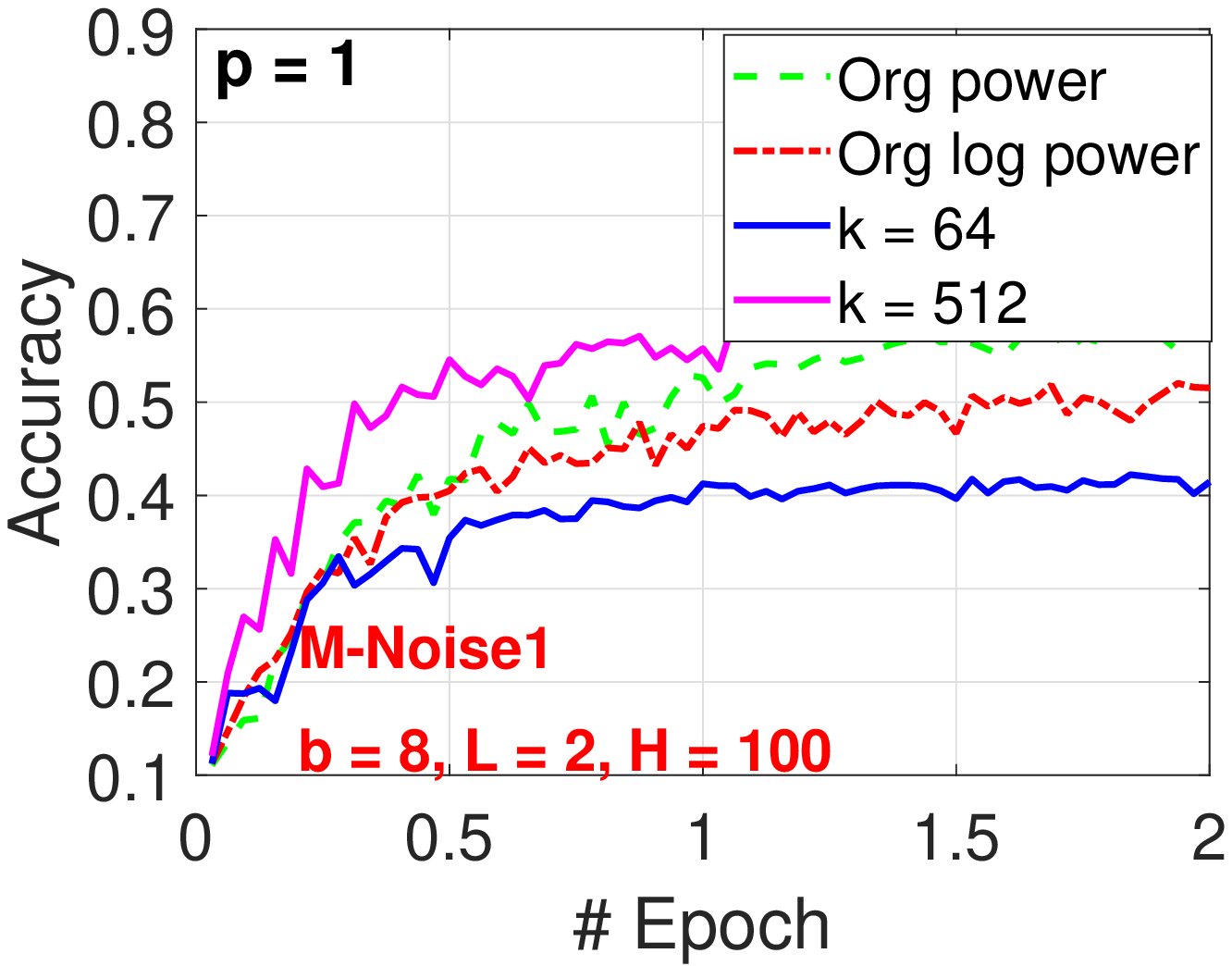}
\includegraphics[width=2.2in]{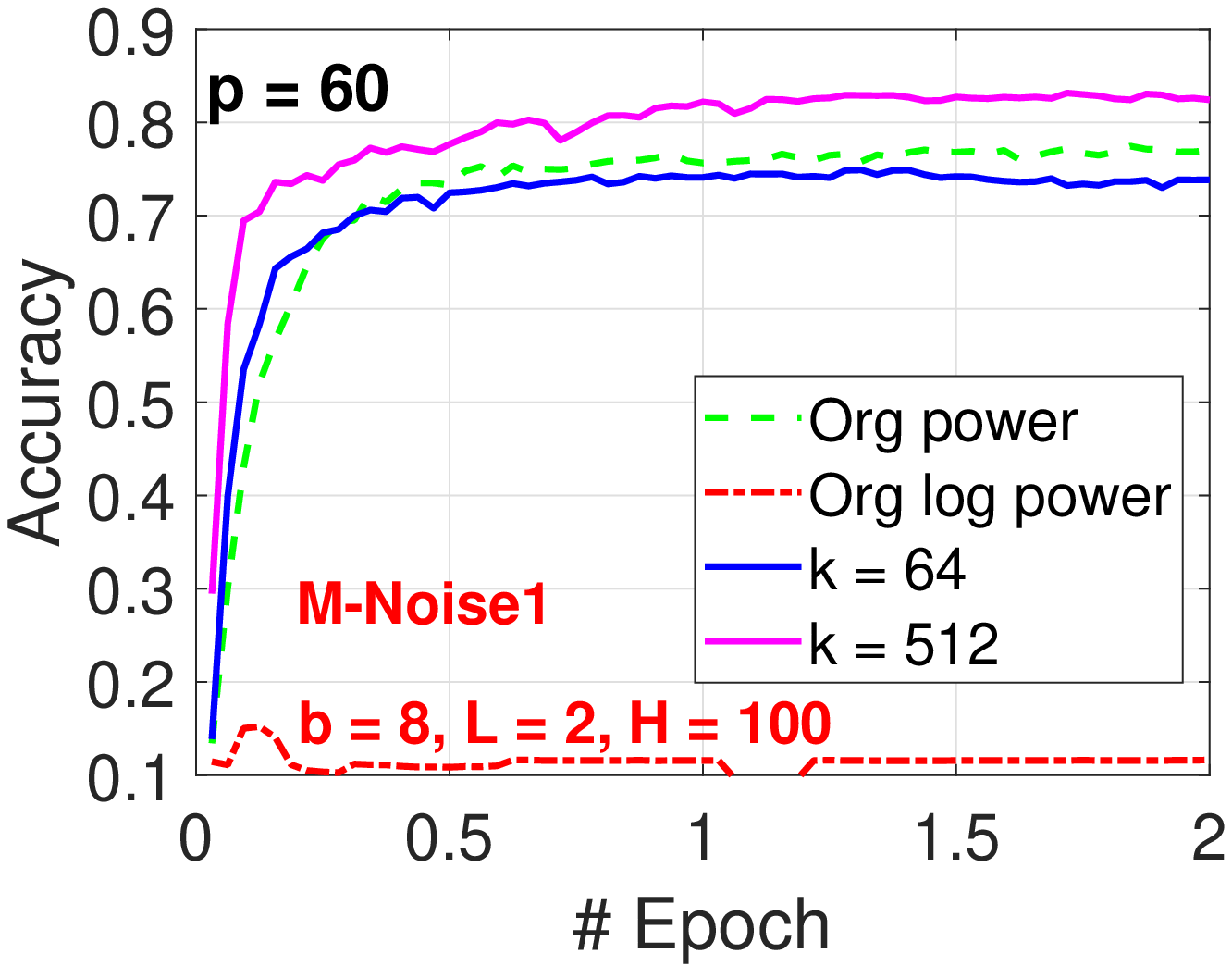}
\includegraphics[width=2.2in]{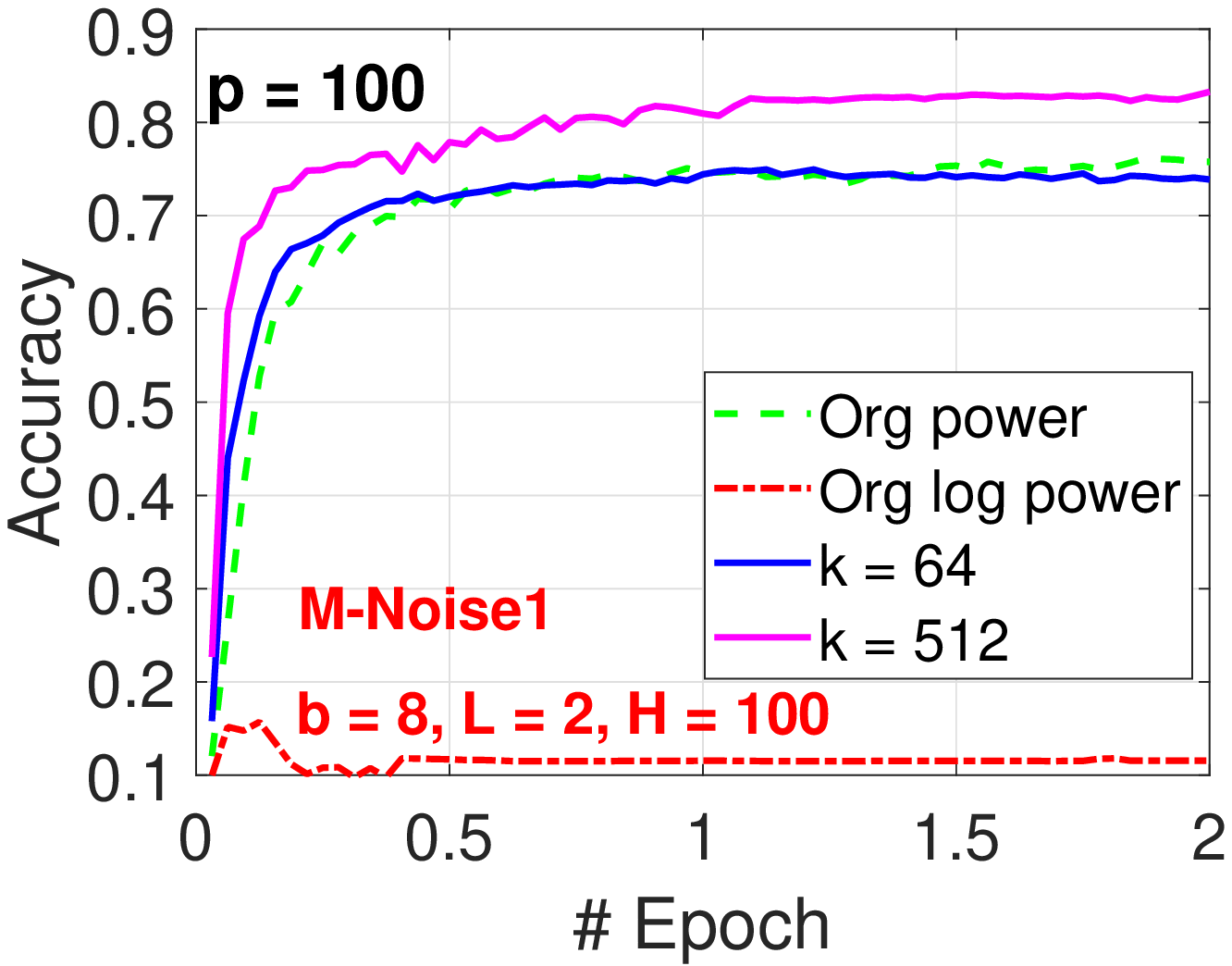}
}

\mbox{
\includegraphics[width=2.2in]{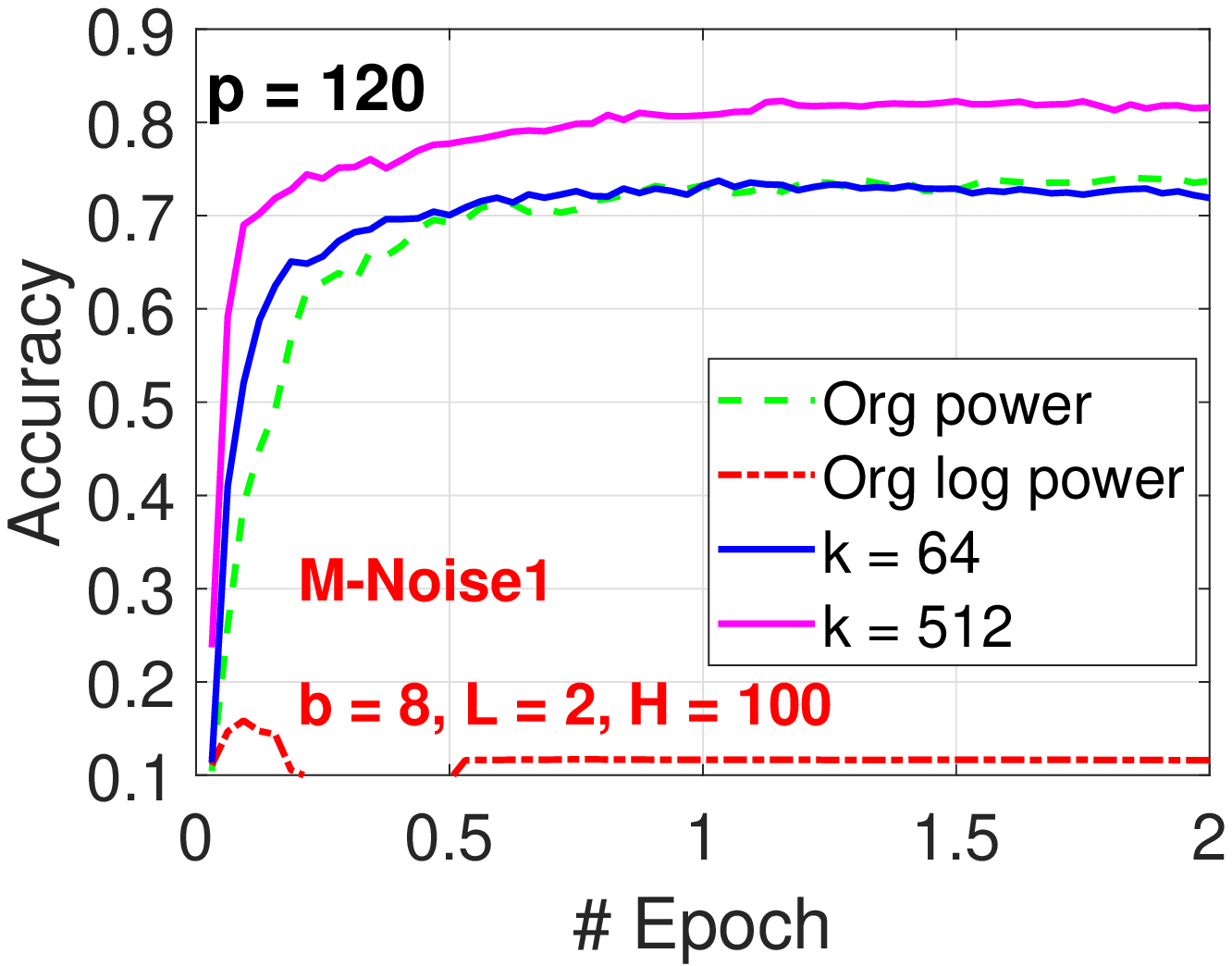}
\includegraphics[width=2.2in]{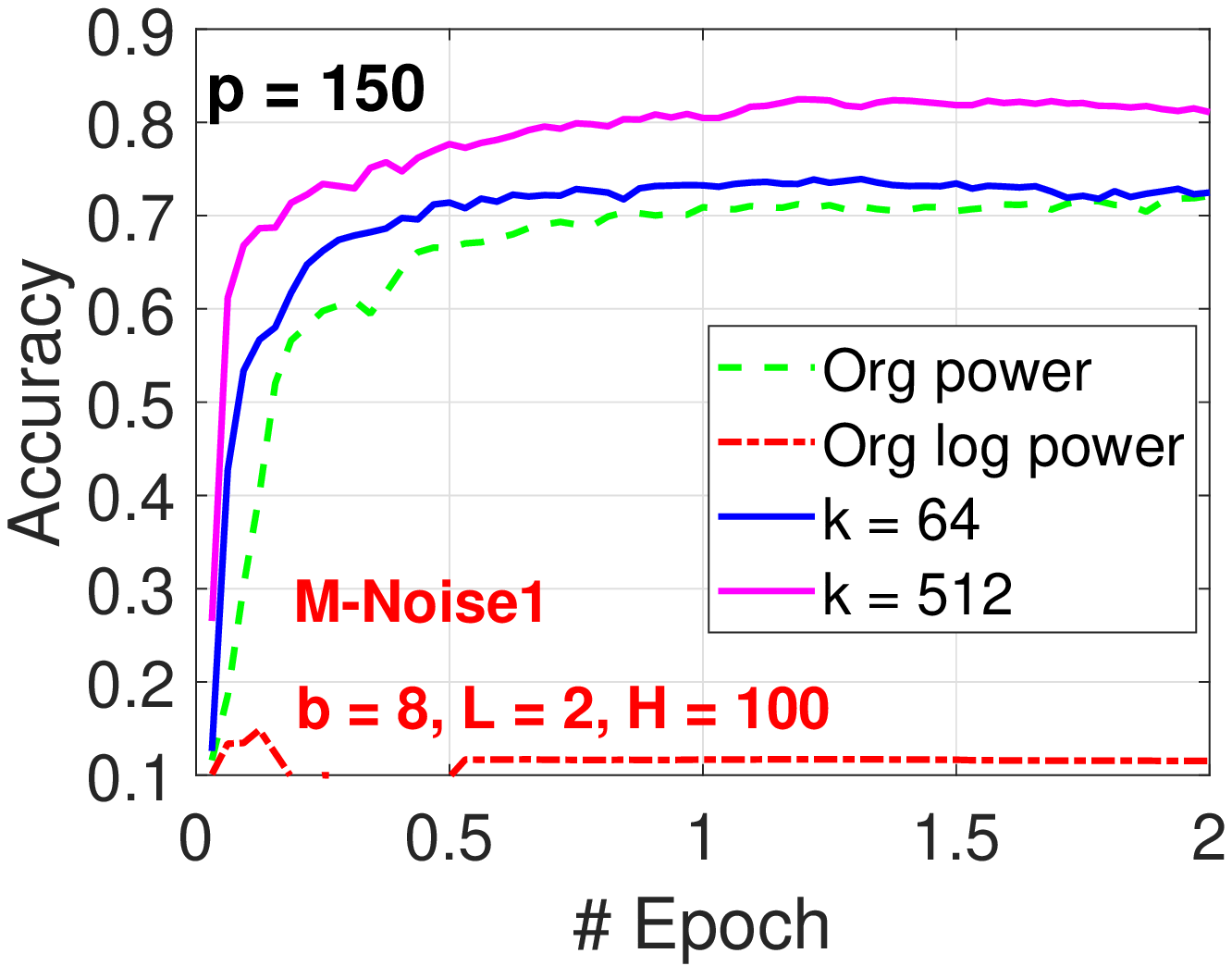}
\includegraphics[width=2.2in]{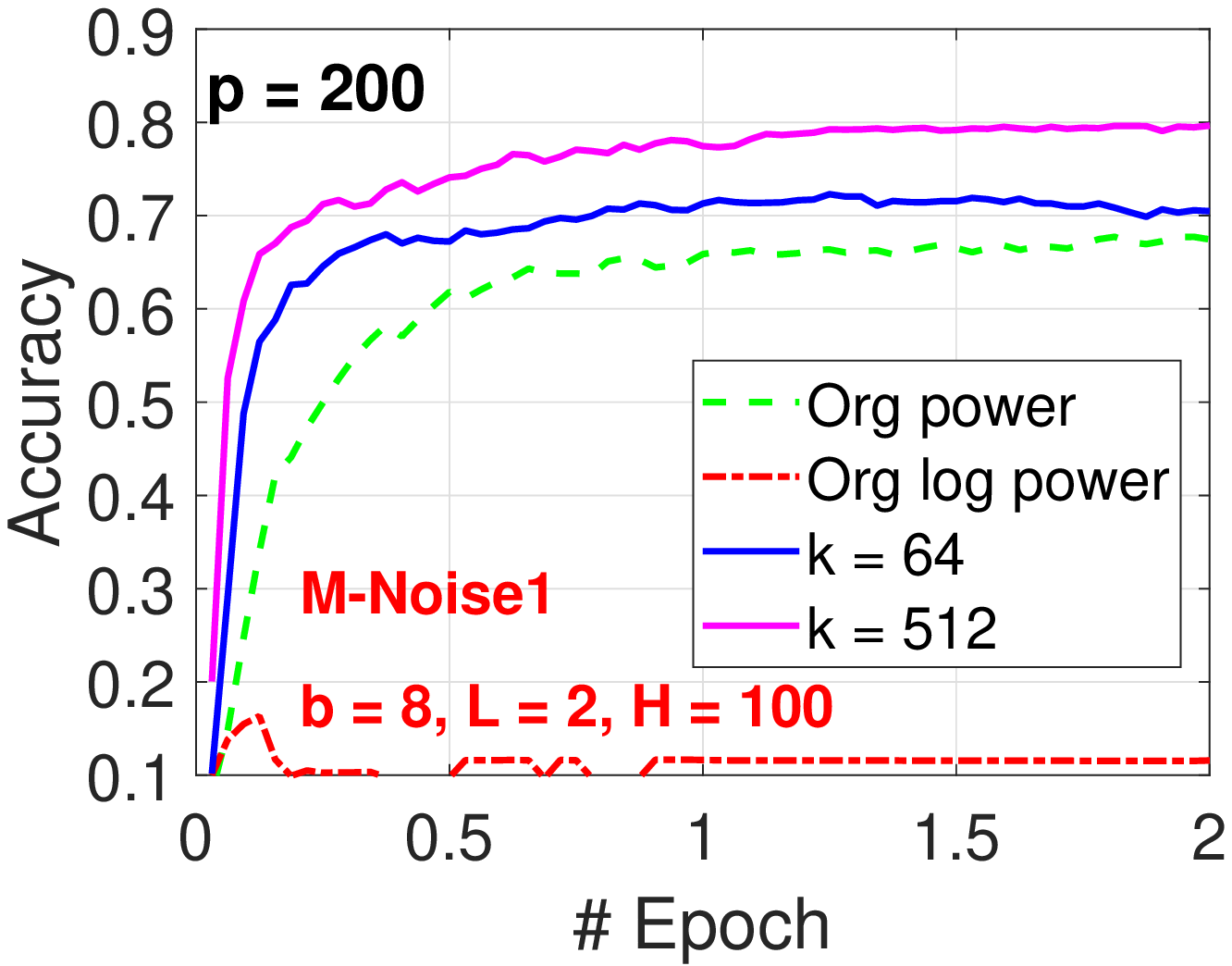}
}

\end{center}

\vspace{-0.2in}

\caption{GCWS with different $p$ values, on the M-Noise1 dataset, for $b=8$ and $k\in\{64, 512\}$.  We can see that using $p\in[60,\ 150]$ produces the best results. The results of GCWSNet are still reasonable even with $p=1$. Using the log-power transformation produces bad results except for $p$ close to 1.  Applying the power transformation directly on the original data produces pretty good results, which are still obviously worse than the results of GCWSNet. }\label{fig:MNoise1_power}\vspace{0.2in}
\end{figure}

Next, we present the experiments on the M-Noise1 dataset, which is quite different from the Pendigits dataset. The first a few entries are ``3 1:0.82962122 2:0.56410292 3:0.27904908 4:0.25310652 5:0.29387237'' (also in LIBSVM format). This dataset is actually dense, with (almost) no zero entries. As shown in Figure~\ref{fig:MNoise1_power}, using the log-power transformation produced very bad results, except for $p$ close 1. Directly applying the power transformation seems to work well on this dataset (unlike Pendigits), although we can see that GCWSNet still produces more accurate predictions.

\vspace{0.1in}

To conclude this section, we should emphasize that we do not claim that we have fully solved the data preprocessing problem, which is a highly crucial task for practical applications. We simply introduce GCWSNet with one single tuning parameter which happens to be quite robust, in comparison with obvious (and commonly used) alternatives for power transformations.

\newpage

\section{Applying GCWS on the Last Layer of Trained Networks}

\begin{figure}[h]

\vspace{-0.1in}

\begin{center}
\mbox{
\includegraphics[width=2.2in]{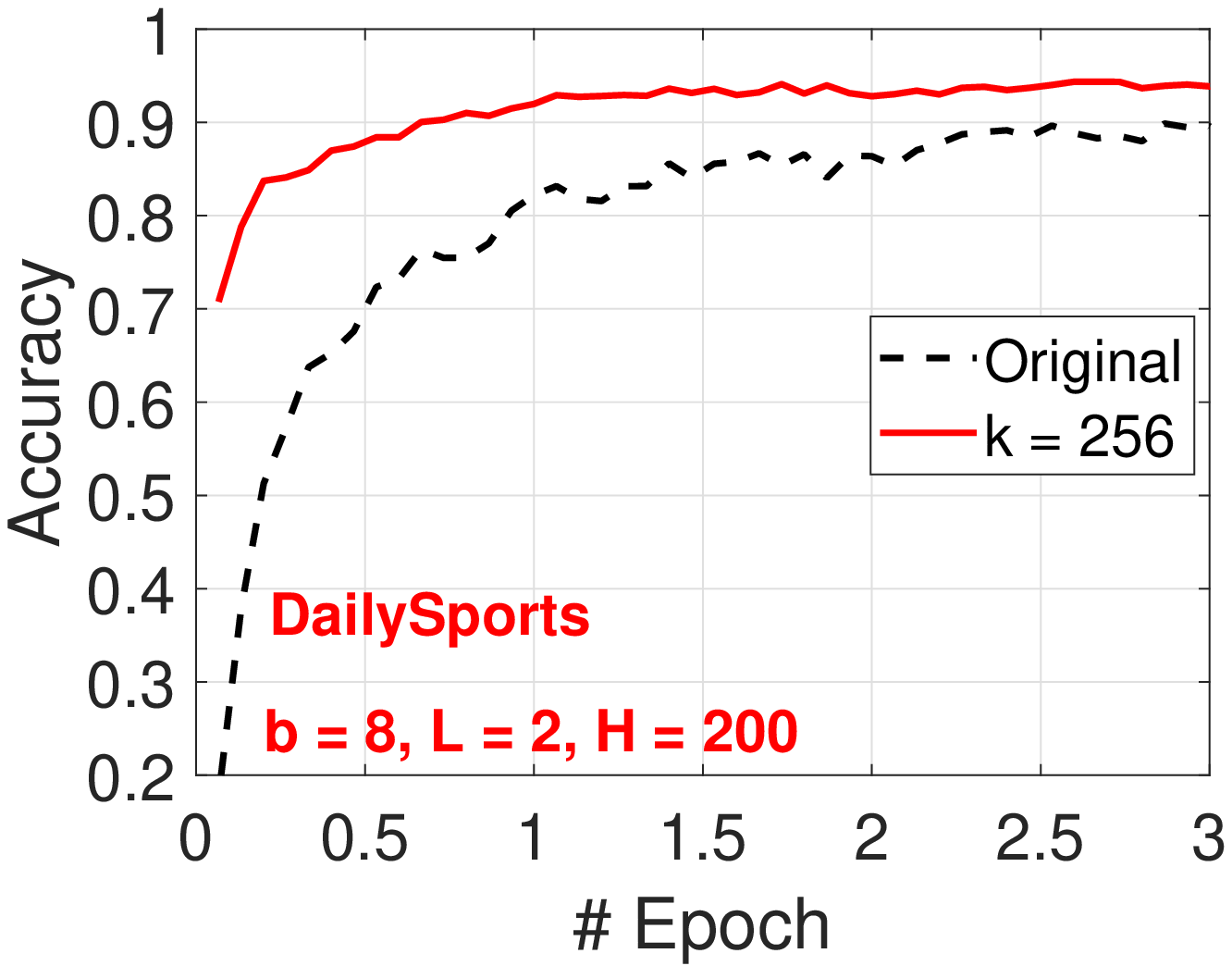}
\hspace{0.3in}
\includegraphics[width=2.2in]{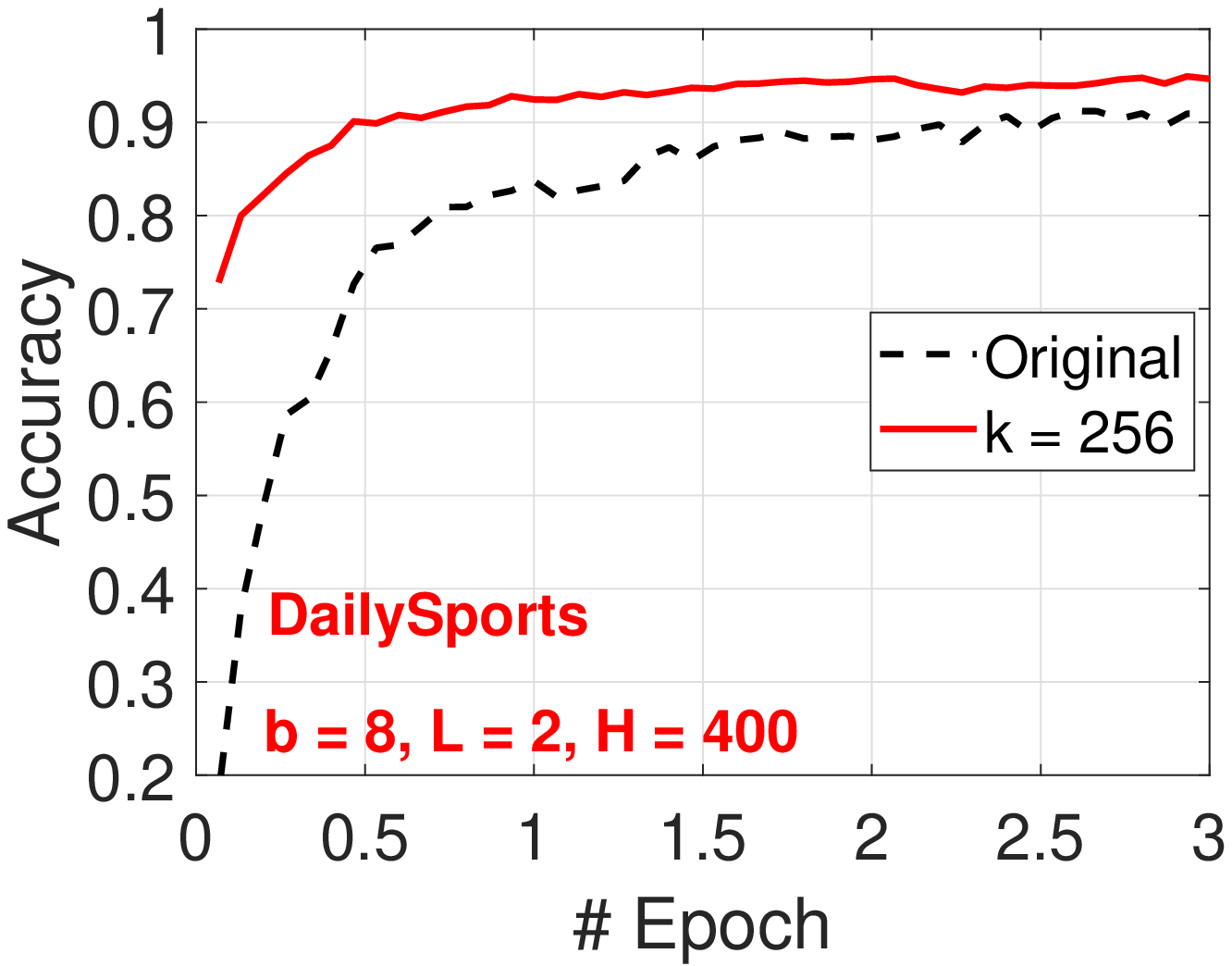}
}
\end{center}

\vspace{-0.2in}

\caption{We apply GCWS (with $k=256$ and $b=8$) on the output of the last layer of neural net. This case, the network has one hidden layer (i.e., $L =2$) and $H\in \{200,400\}$. We perform this operation at every batch, without affecting the accuracy of the original network. This way, we obtain a history of improvements.    }\label{fig:DailySports_nnthencws}\vspace{-0.1in}
\end{figure}

We can always apply GCWS on the trained embedding vectors. Figure~\ref{fig:DailySports_nnthencws} illustrates such an example. We train a neural net on the original data  with one hidden layer (i.e., $L=2$) of $H\in \{200,400\}$ hidden units. We directly apply GCWS on the output of the hidden layer and perform the classification task (i.e.,  a logistic regression)  using the output of GCWS. We do this for every iteration of the neural net training process so that we can compare the entire history. From the plots, we can see that GCWS (solid red curves) can drastically improve the original test accuracy (dashed black curves), especially at the beginning of the training process, for example, improving the test accuracy from $20\%$ to $70\%$ after the first batch. 

\section{Comparison with Normalized Random Fourier Features (NRFF)}

The method of random Fourier features (RFF)~\citep{Book:Rudin_90,Proc:Rahimi_NIPS07} is a popular randomized algorithm for approximating the RBF (Gaussian) kernel. See, for example, a recent work in~\cite{Proc:Li_AISTATS21} on quantizing RFFs for achieving storage savings and speeding up computations. Here, we first review the normalized RFF (NRFF) in~\cite{Proc:Li_KDD17} which recommended the following procedure when applying RFF.

\vspace{0.1in}

At first, the input data vectors are normalized to have unit $l_2$ norms, i.e., $\|u\| = \|v\| =1$. The RBF (Gaussian) kernel is then  $RBF(u,v;\gamma) = e^{-\gamma(1-\rho)}$,  where $\rho=\rho(u,v) = <u,v>$ is the cosine and $\gamma>0$ is a  tuning parameter. We sample $w\sim \textit{uniform}(0,2\pi)$, $r_{i}\sim N(0,1)$ i.i.d., and denote  $x = \sum_{i=1}^D u_i r_{ij}$, $y = \sum_{i=1}^D v_i r_{ij}$.  It is clear that $x \sim N(0,1)$ and $y \sim N(0,1)$. The procedure is repeated $k$ times to generate $k$ RFF samples for each data vector.

\vspace{0.1in}
\cite{Proc:Li_KDD17} proved the next Theorem for the so-called  ``normalized RFF'' (NRFF). 
\begin{theorem}\citep{Proc:Li_KDD17}\label{thm_NRFF}
Consider $k$ iid samples ($x_j, y_j, w_j$) where $x_j\sim N(0,1)$, $y_j\sim N(0,1)$,  $E(x_jy_j) = \rho$, $w_j\sim \textit{uniform}(0,2\pi)$, $j =1, 2, ..., k$. Let $X_j = \sqrt{2}\cos\left(\sqrt{\gamma}x_j + w_j\right)$ and $Y_j = \sqrt{2}\cos\left(\sqrt{\gamma}y_j + w_j\right)$. As $k\rightarrow\infty$, the following asymptotic normality holds:
\begin{align}\label{eqn_NRFF}
&\sqrt{k}\left(\frac{\sum_{j=1}^k X_j Y_j}{\sqrt{\sum_{j=1}^k X_j^2}\sqrt{\sum_{j=1}^k Y_j^2}} - e^{-\gamma(1-\rho)}\right)\overset{D}{\Longrightarrow}
N\left(0,V_{n,\rho,\gamma}\right)
\end{align}
\text{where}
\begin{align}\label{eqn_NRFF_Var}
&V_{n,\rho,\gamma} = V_{\rho,\gamma}- \frac{1}{4}e^{-2\gamma(1-\rho)} \left[3-e^{-4\gamma(1-\rho)}  \right]\\
&V_{\rho,\gamma} = \frac{1}{2}+\frac{1}{2}\left(1-e^{-2\gamma(1-\rho)}\right)^2
\end{align}
\end{theorem}

\vspace{0.1in}

In the above theorem, $V_{\rho,\gamma}$ is the corresponding variance term without normalizing the output RFFs. Obviously, $V_{n,\rho,\gamma} < V_{\rho,\gamma}$, meaning that it is always a good idea to normalize the output RFFs before feeding NRFF samples to the subsequent tasks. 

\vspace{0.1in}

\begin{figure}[t]
\begin{center}
\mbox{

\includegraphics[width=2.2in]{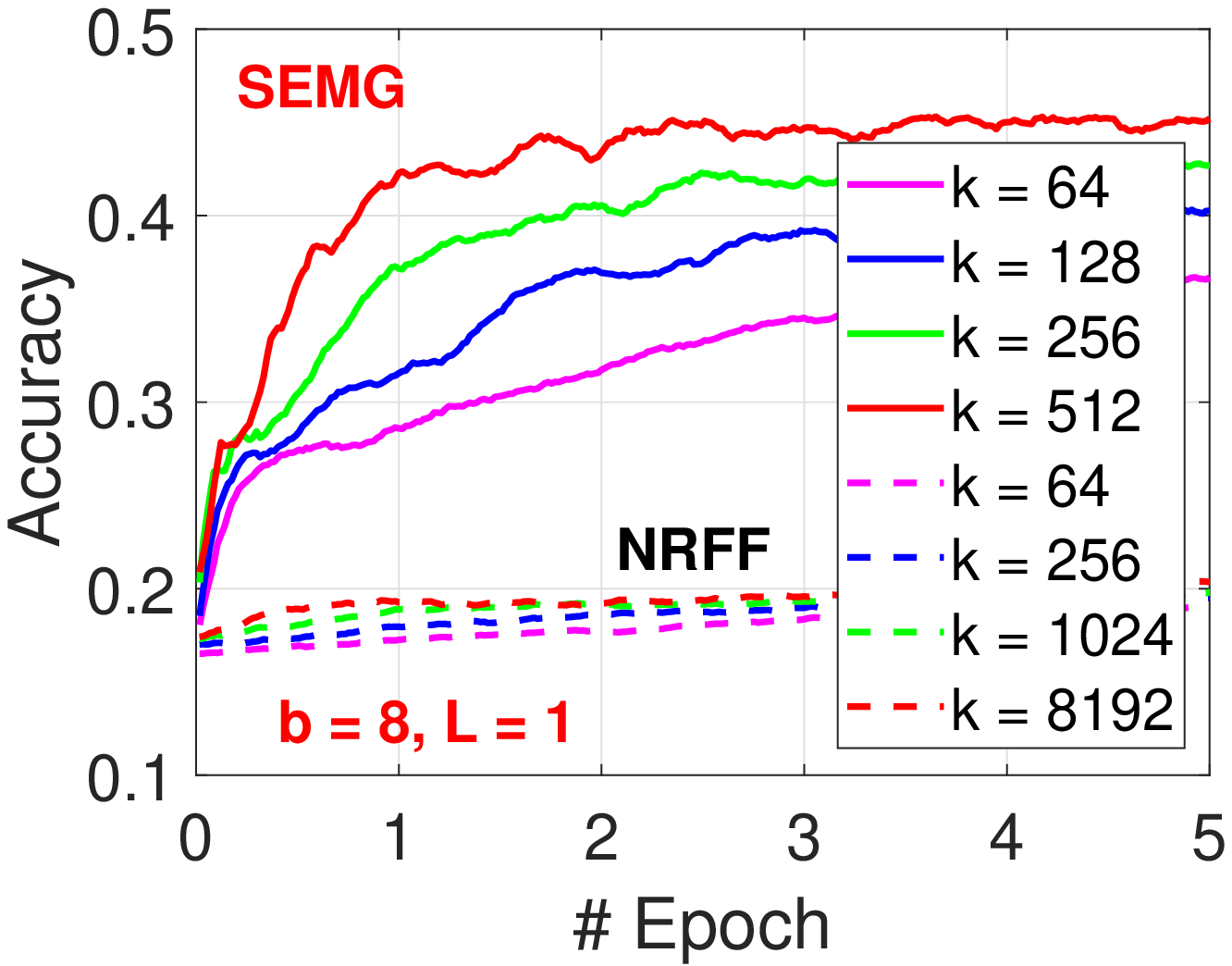}
\includegraphics[width=2.2in]{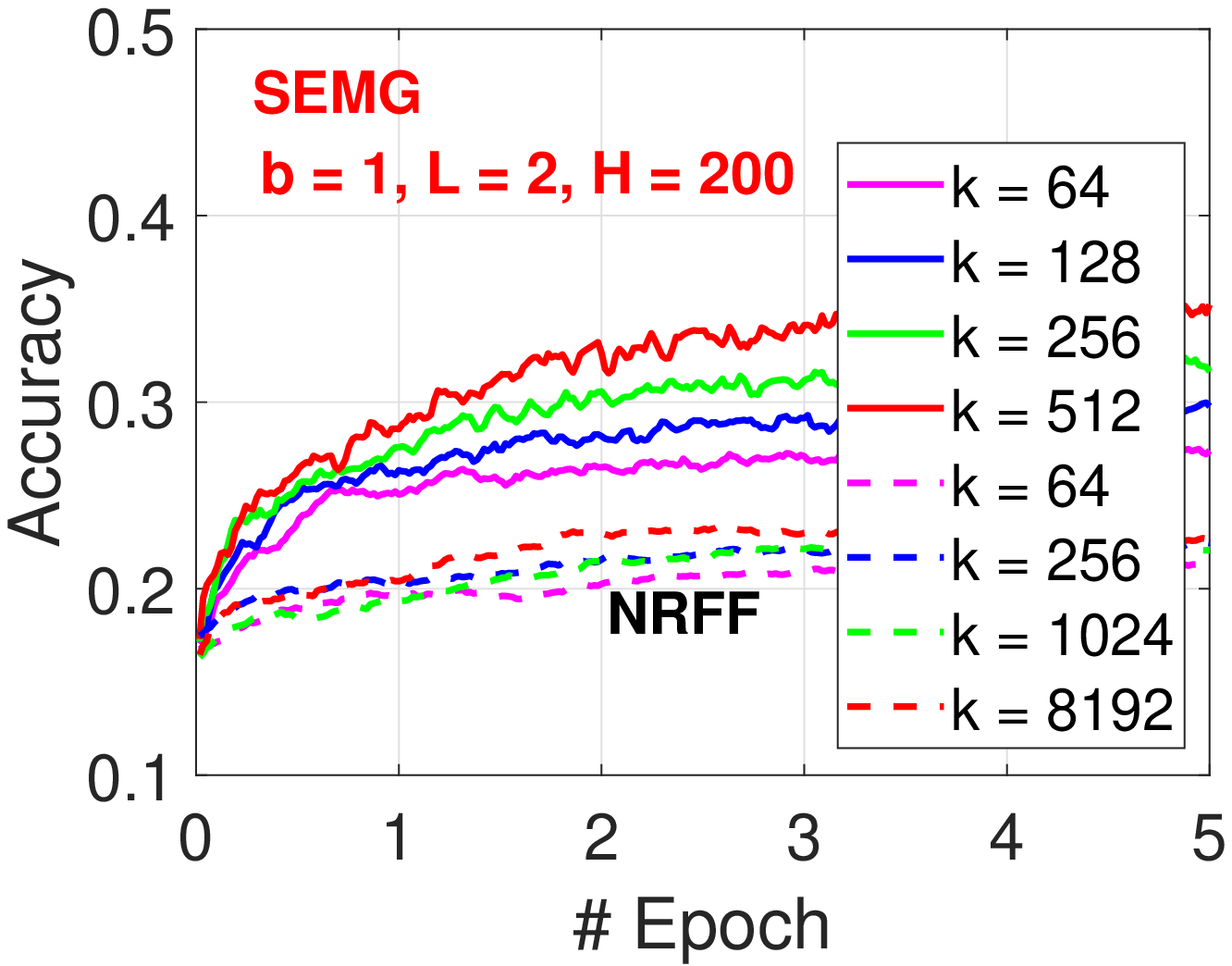}
\includegraphics[width=2.2in]{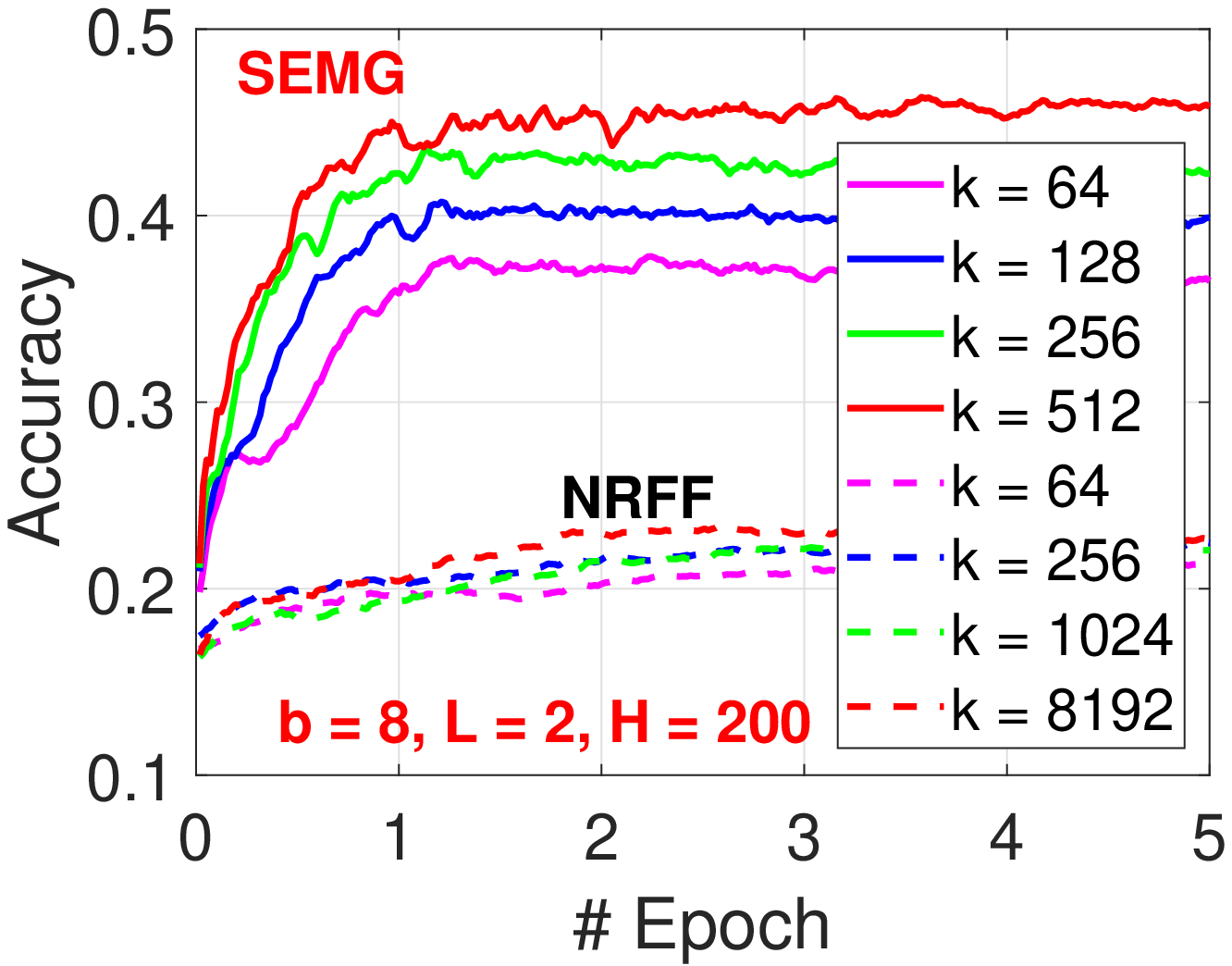}
}

\mbox{
\includegraphics[width=2.2in]{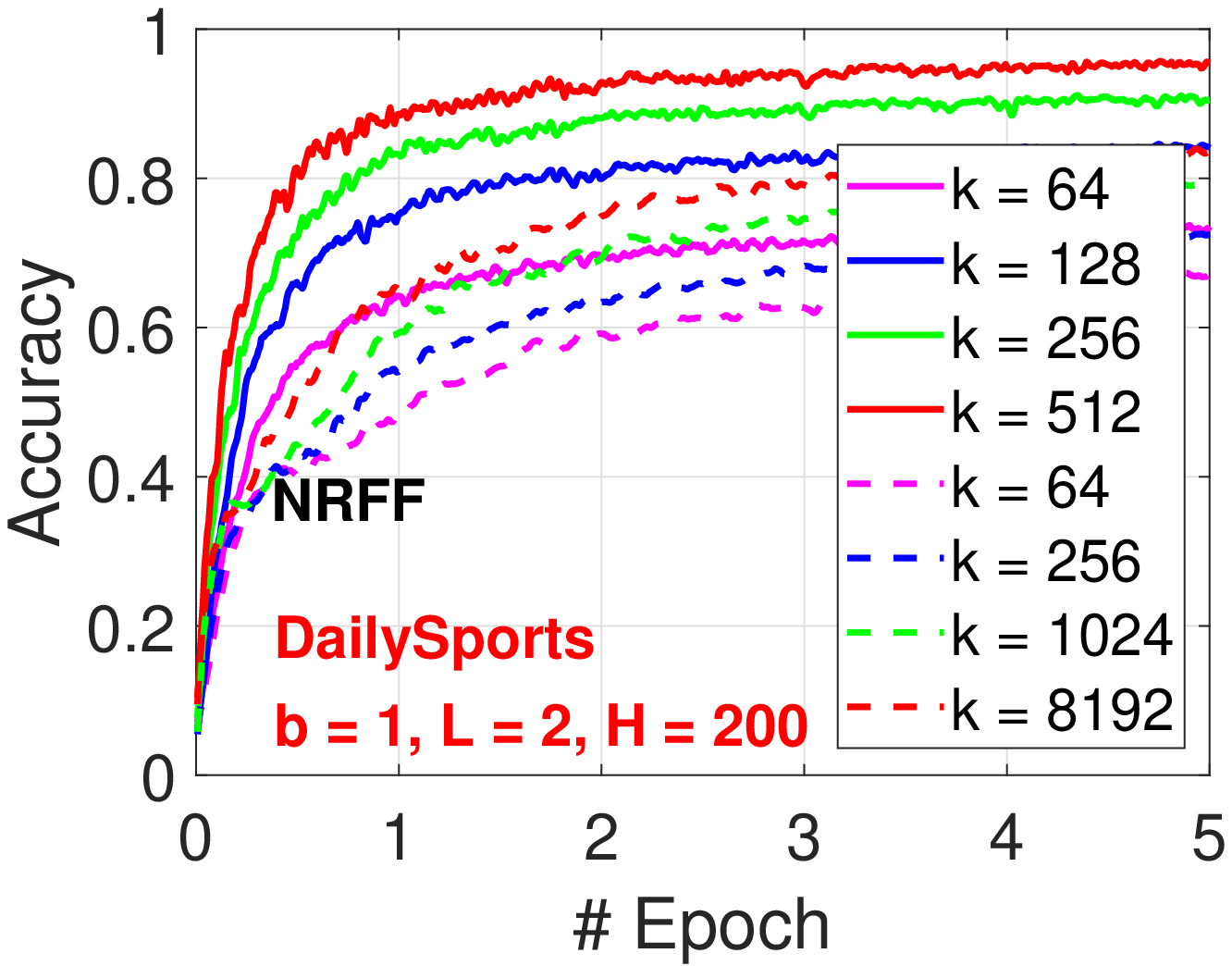}
\includegraphics[width=2.2in]{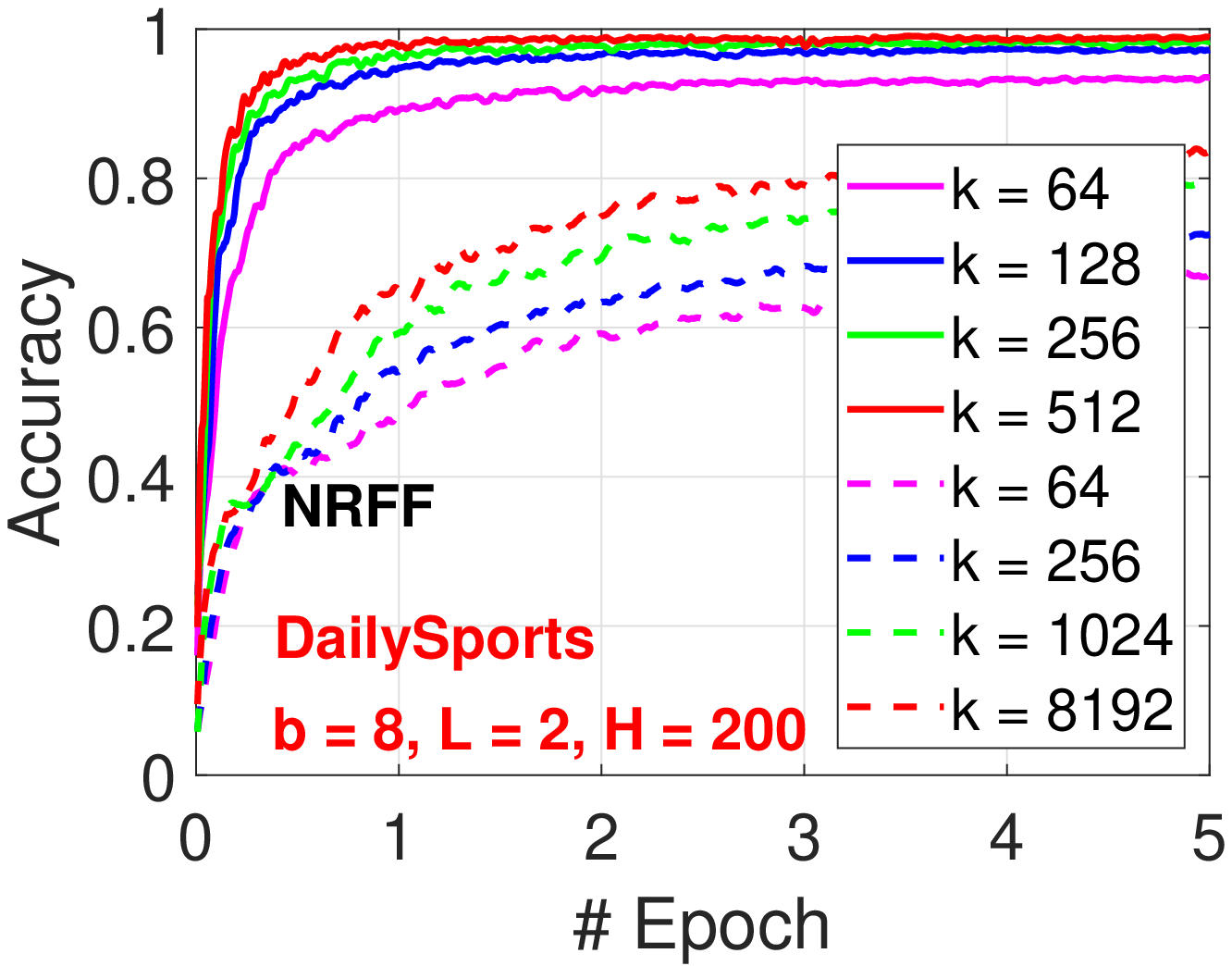}
\includegraphics[width=2.2in]{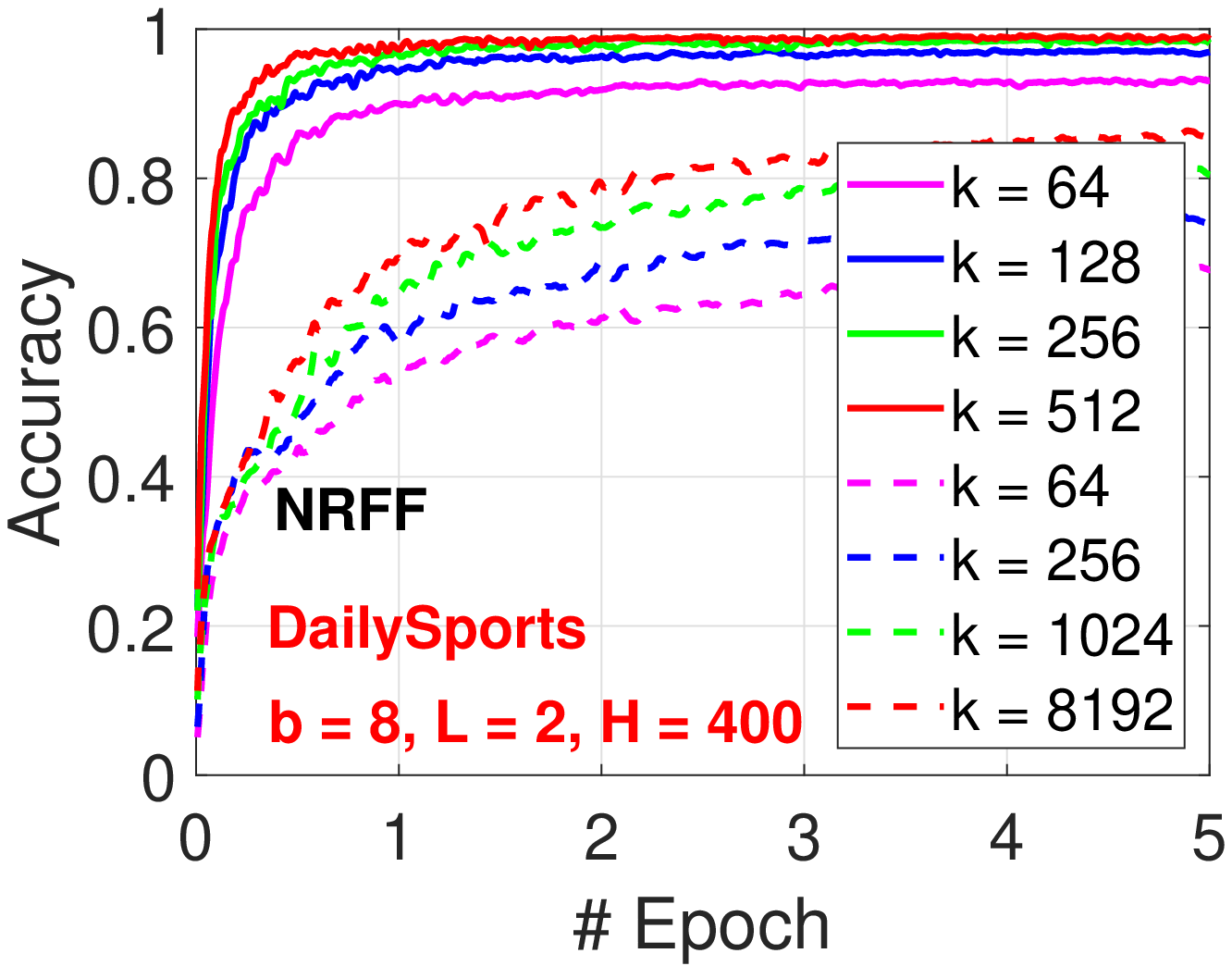}
}

\end{center}

\vspace{-0.2in}

\caption{NRFF (dashed curves) versus GCWS (solid curves), for two datasets: SEMG and DailySports and $b\in\{1,8\}$. We use much larger $k$ values for NRFF, as large as $8192$. }\label{fig:nrff}
\end{figure}

Figure~\ref{fig:nrff} provides an experimental study to compare NRFF with GCWS on two datasets. While we still use $k\in\{64, 128, 256, 512\}$ for GCWS (solid curves), we have to present for $k$ as large as 8192 for NRFF (dashed curves) because it is well-known that RFF needs a large number of samples in order to obtain reasonable results. It is quite obvious from Figure~\ref{fig:nrff} that, at least on these two datasets, GCWS performs considerably better than NRFF at the same $k$ (even if we just use $b=1$ bit for GCWS). Also, we can see that GCWS converges much faster.

\vspace{0.1in}

The method of random Fourier features is very popular in academic research. In comparison, research activities on consistent weighted sampling and variants are sparse. We hope this study might generate more interest in GCWS and motivate researchers as well as practitioners to try this interesting method.

\section{Concluding Remarks}

In this paper, we propose the pGMM kernel with a single tuning parameter $p$ and use GCWS (generalized consistent weighted sampling) to generate hash values, producing sparse binary data vectors of size $2^b\times k$, where $k$ is the number of hash samples and $b$ is the number of bits  to store each hash value. These binary vectors are fed to neural networks for subsequent tasks such as regression or classification. Many interesting  findings can be summarized  which might benefit future research or practice. 

\vspace{0.1in}

\noindent Experiments show that GCWS converges fast and typically reaches a reasonable accuracy at (less than) one epoch. This characteristic of GCWS could be highly beneficial in practice because many important applications such as data streams or CTR predictions in commercial search engines train only one epoch.  

\vspace{0.1in}
\noindent GCWS with a single tuning parameter $p$ provides a beneficial and  robust power transformation on the original data. By adding this tuning parameter, the performance can often be improved, in some cases considerably so. Inside GCWS, this tuning parameter acts on the (nonzero) data entry as $p\log u$ (which is typically robust). The nature of GCWS is that the coordinates of entries with larger values have a higher chance to be picked as output (in a probabilistic manner). In other words, both the absolute values and relative orders matter, and the power transformation does impact the performance unlike trees which are invariant to monotone transformations. Experiments show that GCWSNet with a tuning parameter $p$ produces more accurate results compared with two obvious strategies: 1) feeding power-transformed data ($u^p$) directly to neural nets; 2) feeding log-power-transformed data ($p\log u$, with zeros handled separately) directly to neural nets. 

\vspace{0.1in}
\noindent GCWSNet can be combined with count-sketch to reduce the model size of GCWSNet, which is proportional to $2^b\times k$ and can be pretty large if $b$ is large such as 16 or 24. Our theoretical analysis for the impact of count-sketch on the estimation variance provides the explanation for the effectiveness of count-sketch demonstrated in the experiments. We recommend the ``8-bit'' practice in that we always use $2^8$ bins for count-sketch if GCWS uses $b\geq 8$. With this strategy, even when $k=2^{10} = 1024$, the model size is only proportional to $2^{8} \times 2^{10} =2^{18}$, which is not a large number. Note that the outputs of applying count-sketch on top of GCWS remain to be integers, meaning that a lot of multiplications can be still be avoided.

\vspace{0.1in}
\noindent There are other ways to take advantage of GCWS. For example, one can always apply GCWS on the embeddings of trained neural nets to hopefully boost the performance of existing network models. We also hope the comparison of GCWS with random Fourier features would promote interest in future research. 

\vspace{0.1in}
\noindent Finally, we should mention again that hashing pGMM can be conducted efficiently. For dense data, we recommend algorithms based on rejection sampling~\citep{Proc:Shrivastava_NIPS16,Proc:Li_Li_AAAAI21}. For sparse data, we suggest GCWS. For relatively high-dimensional data, we always recommend the bin-wise algorithm in~\cite{Proc:Li_NIPS19_BCWS} combined with rejection sampling or GCWS, depending on data sparsity.

\vspace{0.1in}
\noindent {All reported  experiments were conducted on the PaddlePaddle \url{https://www.paddlepaddle.org.cn} deep learning platform.}

\bibliographystyle{plainnat}
\bibliography{standard}
\end{document}